**A HYBRID TABU/SCATTER SEARCH ALGORITHM FOR SIMULATION-BASED**

**OPTIMIZATION OF MULTI-OBJECTIVE RUNWAY OPERATIONS SCHEDULING**

by

Bulent Soykan
B.S. August 1997, Turkish Military Academy (TMA)
M.S. March 2007, Defense Sciences Institute, TMA

A Dissertation Submitted to the Faculty of
Old Dominion University in Partial Fulfillment of the
Requirements for the Degree of

DOCTOR OF PHILOSOPHY

ENGINEERING MANAGEMENT

OLD DOMINION UNIVERSITY
August 2016

Approved by:

Ghaith Rabadi (Director)

Resit Unal (Member)

Mecit Cetin (Member)

# ABSTRACT

A HYBRID TABU/SCATTER SEARCH ALGORITHM FOR SIMULATION-BASED
OPTIMIZATION OF MULTI-OBJECTIVE RUNWAY OPERATIONS SCHEDULING

Bulent Soykan
Old Dominion University, 2016
Director: Dr. Ghaith Rabadi

As air traffic continues to increase, air traffic flow management is becoming more
challenging to effectively and efficiently utilize airport capacity without compromising
safety, environmental and economic requirements. Since runways are often the primary
limiting factor in airport capacity, runway operations scheduling emerge as an important
problem to be solved to alleviate flight delays and air traffic congestion while reducing
unnecessary fuel consumption and negative environmental impacts. However, even a
moderately sized real-life runway operations scheduling problem tends to be too complex
to be solved by analytical methods, where all mathematical models for this problem belong
to the complexity class of *NP*-Hard in a strong sense due to combinatorial nature of the
problem. Therefore, it is only possible to solve practical runway operations scheduling
problem by making a large number of simplifications and assumptions in a deterministic
context. As a result, most analytical models proposed in the literature suffer from too much
abstraction, avoid uncertainties and, in turn, have little applicability in practice. On the
other hand, simulation-based methods have the capability to characterize complex and
stochastic real-life runway operations in detail, and to cope with several constraints and
stakeholders' preferences, which are commonly considered as important factors in practice.

This dissertation proposes a simulation-based optimization (SbO) approach for multi-
objective runway operations scheduling problem. The SbO approach utilizes a discrete-
event simulation model for accounting for uncertain conditions, and an optimization
component for finding the best known Pareto set of solutions. This approach explicitly
considers uncertainty to decrease the real operational cost of the runway operations as well as fairness among aircraft as part of the optimization process. Due to the problem's large, complex and unstructured search space, a hybrid Tabu/Scatter Search algorithm is developed to find solutions by using an elitist strategy to preserve non-dominated solutions, a dynamic update mechanism to produce high-quality solutions and a rebuilding strategy to promote solution diversity. The proposed algorithm is applied to bi-objective (i.e., maximizing runway utilization and fairness) runway operations schedule optimization as the optimization component of the SbO framework, where the developed simulation model acts as an external function evaluator. To the best of our knowledge, this is the first SbO approach that explicitly considers uncertainties in the development of schedules for runway operations as well as considers fairness as a secondary objective.

In addition, computational experiments are conducted using real-life datasets for a major US airport to demonstrate that the proposed approach is effective and computationally tractable in a practical sense. In the experimental design, statistical design of experiments method is employed to analyze the impacts of parameters on the simulation as well as on the optimization component's performance, and to identify the appropriate parameter levels. The results show that the implementation of the proposed SbO approach provides operational benefits when compared to First-Come-First-Served (FCFS) and deterministic approaches without compromising schedule fairness. It is also shown that proposed algorithm is capable of generating a set of solutions that represent the inherent trade-offs between the objectives that are considered. The proposed decision-making algorithm might be used as part of decision support tools to aid air traffic controllers in solving the real-life runway operations scheduling problem.

This thesis is dedicated to my beloved family,
for their unwavering love and support.

# ACKNOWLEDGMENTS

There are many people who have contributed to the successful completion of this dissertation. However, there are certain people that I have to acknowledge and appreciate their help and support.

First and foremost, my deepest and most sincere appreciation goes to my advisor, Dr. Ghaith Rabadi, especially for introducing me to runway operations scheduling and metaheuristic optimization methods, and for his constant encouragement, support and, most of all, friendship. His professional guidance and belief in me always gave me the confidence to continue researching. I feel very lucky to have the chance of working with him. I hope all Ph.D. candidates find such an advisor.

I would also like to express my deepest gratitude to my other two dissertation committee members, Dr. Resit Unal, and Dr. Mecit Cetin, for their time, interest, valuable insights and suggestions. They monitored this research and took effort in reading and providing valuable comments on this dissertation.

Last, but certainly not least, my love and deepest gratitude is reserved for my family. Without their continuous support, understanding, and motivation throughout my Ph.D. work, this would not have happened. The completion of this dissertation came at the expense of my hours of absence from enjoying life with them.

# NOMENCLATURE

| | |
|---|---|
| *ANSP* | Air Navigation Service Provider |
| *ARTCC* | Air Route Traffic Control Center |
| *ASDEM-X* | Airport Surface Detection Equipment Model X |
| *ASPM* | Aviation System Performance Metrics |
| *ASQP* | Airline Service Quality Performance |
| *ATC* | Air Traffic Control |
| *ATCSCC* | Air Traffic Control System Command Center |
| *ATCT* | Airport Traffic Control Tower |
| *ATM* | Air Traffic Management |
| *ATFM* | Air Traffic Flow Management |
| *ATO* | Air Traffic Organization |
| *B&B* | Branch-and-Bound |
| *B&C* | Branch-and-Cut |
| *B&P* | Branch-and-Price |
| *BTS* | Bureau of Transportation Statistics |
| *CAA* | Civil Aviation Authority |
| *CCD* | Central Composite Design |
| *CDM* | Collaborative Decision Making |
| *CPS* | Constrained Position Shifting |
| *DoE* | Design of Experiments |
| *DoSE* | Design of Simulation Experiments |
| *DP* | Dynamic Programming |
| *ETL* | Estimated Time of Landing |
| *ETT* | Estimated Time of Take-off |
| *Eurocontrol* | European Organization for the Safety of Air Navigation |
| *FAA* | Federal Aviation Administration |
| *FAF* | Final Approach Fix |
| *FAST* | Final Approach Spacing Tool |
| *FCFS* | First-Come-First-Served |
| *GA* | Genetic Algorithm |
| *GDP* | Ground Delay Program |
| *IAD* | Washington Dulles International Airport |
| *IAF* | Initial Approach Fix |
| *ICAO* | International Civil Aviation Organization |
| *IFR* | Instrument Flight Rules |
| *IMC* | Instrument Meteorological Conditions |
| *LHS* | Latin Hypercube Sampling |
| *MEANS* | MIT Extensible Air Network Simulation |
| *Meta-RaPS* | Meta-heuristic for Randomized Priority Search |
| *MINIT* | Minutes in Trail |
| *MIP* | Mixed Integer Programming |
| *MMC* | Marginal Meteorological Conditions |
| *MOEA* | Multi-Objective Evolutionary Algorithm |
| *MOGA* | Multi-Objective Genetic Algorithm |
| *MOO* | Multi-Objective Optimization |
| *MPS* | Maximum Position Shifting |
| *NAS* | National Airspace System |
| *NextGen* | Next Generation Air Transportation System |
| *NM* | Nautical Mile |
| *NP* | Non-deterministic Polynomial |
| *NSGA* | Non-dominated Sorting Genetic Algorithm |
| *OAG* | Official Airline Guides |
| *OOA* | Object-Oriented Architecture |
| *OOM* | Object-Oriented Modeling |
| *OOOI* | Out of the gate, Off the ground, On the ground and Into the Gate |
| *OOP* | Object-Oriented Programming |
| *PESA* | Pareto Enveloped Based Selection Algorithm |
| *PR* | Path Relinking |
| *RHA* | Rolling Horizon Approach |
| *RNAV* | Area Navigation |
| *ROT* | Runway Occupancy Time |
| *RSM* | Response Surface Methodology |
| *SbO* | Simulation-based Optimization |
| *SESAR* | Single European Sky ATM Research |
| *SPEA* | Strength Pareto Evolutionary Algorithm |
| *SS* | Scatter Search |
| *TAAM* | Total Airspace and Airport Modeler |
| *TFDM* | Tower Flight Data Manager |
| *TFMSC* | Traffic Flow Management System Counts |
| *TMA* | Terminal Maneuvering Area |
| *TRACON* | Terminal Radar Approach Control |
| *TS* | Tabu Search |
| *TSP-TW* | Traveling Salesman Problem with Time Windows |
| *UML* | Unified Modeling Language |
| *VFR* | Visual Flight Rules |
| *VMC* | Visual Meteorological Conditions |
| *V&V* | Verification and Validation |

# TABLE OF CONTENTS

# LIST OF TABLES

Table                                                                                              Page

# LIST OF FIGURES

# CHAPTER 1

# INTRODUCTION

Economic growth has been broadly influencing air transportation demand and pushing the industry's infrastructure and resource capacities to its limits. However, in terms of infrastructure, it is one of the most neglected industries (Wensveen, 2015). The essential physical infrastructure of air transportation are the airports where a modal transfer is carried out from the air mode to land mode and vice versa. As demand for air transportation continues to increase throughout the world, air traffic volume in major airports approaches airport infrastructure capacity. Consequently, delays are becoming inevitable in air transportation facilities as the demand for services exceeds its capacity. The resulting high-volume air traffic typically leads to airport congestion and long queues for both arrivals and departures, and in turn, results in additional fuel costs, passenger dissatisfaction, as well as environmental pollution.

In airport infrastructure, runways have been typically identified as the primary limiting factor (bottleneck) that causes congestion and delays. Hence, the capacity of an airport heavily depends on the runways in use. Air traffic congestion and delays stem from the scarcity of runways pose safety risks, and also increase operational and environmental costs. Although one may think that investing in airport infrastructure can solve the problem, most of the time it is not practical or feasible. Since majority of the busy airports are constrained by the lack of physical space for new runways and newly promulgated environmental restrictions, this prevents adding more runways as a way to increase capacity. As a result, it is significantly important to utilize terminal maneuvering area (TMA) effectively to increase the overall capacity of the airports and to smooth the flow of air traffic.

To improve TMA utilization and ensure air traffic flow safety, runway operations need to be scheduled effectively and efficiently. This real-life combinatorial problem is commonly referred to as runway operations scheduling problem. Over the past several decades, researchers and practitioners have developed various models and tools for this real-life problem. However, for major airports, it is still a challenge to schedule runway operations considering the complexity and uncertainty inherent to these operations. In 2015, air traffic volume accounted for 34.58 percent of all aircraft delays in the United States (US) (FAA, 2016a); therefore, the problem is motivated by a clear evidence of a steady increase in air traffic congestion and delays at the major airports.

One of the main reasons for this inefficiency is that currently available operational planning models and decision support tools used by air traffic controllers do not consider explicitly the stochastic nature of runway operations and the interests of different stakeholders, particularly airlines. However, uncertainty is an integral part of runway operations, which usually renders the deterministic models sub-optimal or even infeasible in practice, and also in real-life there are several stakeholders related to runway systems where each may have separate and possibly conflicting interest. Hence, there is an apparent need for improving these models and tools for planning and controlling the air traffic flow in runway operations to a level applicable for practical use.

This dissertation investigates an effective way for optimizing the multi-objective runway operations scheduling while explicitly considering the uncertainties inherent in runway operations and fairness among all aircraft in a natural way by utilizing a simulation-based optimization (SbO) approach. Furthermore, due to the problem's large, complex and unstructured search space, a hybrid Tabu/Scatter Search algorithm is designed and implemented as the optimization engine of this SbO framework, which is the main focus of this dissertation. This chapter is dedicated to present the background of the study focusing specifically on the problem definition and fundamental characteristics to set up the foundation, followed by the research philosophy and methodology. A summary of the contributions and outline of the dissertation are also provided.

## 1.1 Motivation

According to a Federal Aviation Administration (FAA) aerospace forecast, US carrier combined domestic and international passenger growth is estimated to be an average of 2.2 percent per year over the next two decades, and system capacity in available seat miles (both domestic and international) is estimated to increase by 2.6 percent in 2016 and by 2.5 percent of the average annual rate through 2036. By 2036, US commercial air carriers are projected to transport 1.24 billion enplaned passengers, as shown in Figure 1 (FAA, 2016b). According to a similar forecast study done by Eurocontrol (European Organization for the Safety of Air Navigation) for Europe, average annual air traffic growth rate is projected to be 3 percent through 2021 (Eurocontrol, 2015).

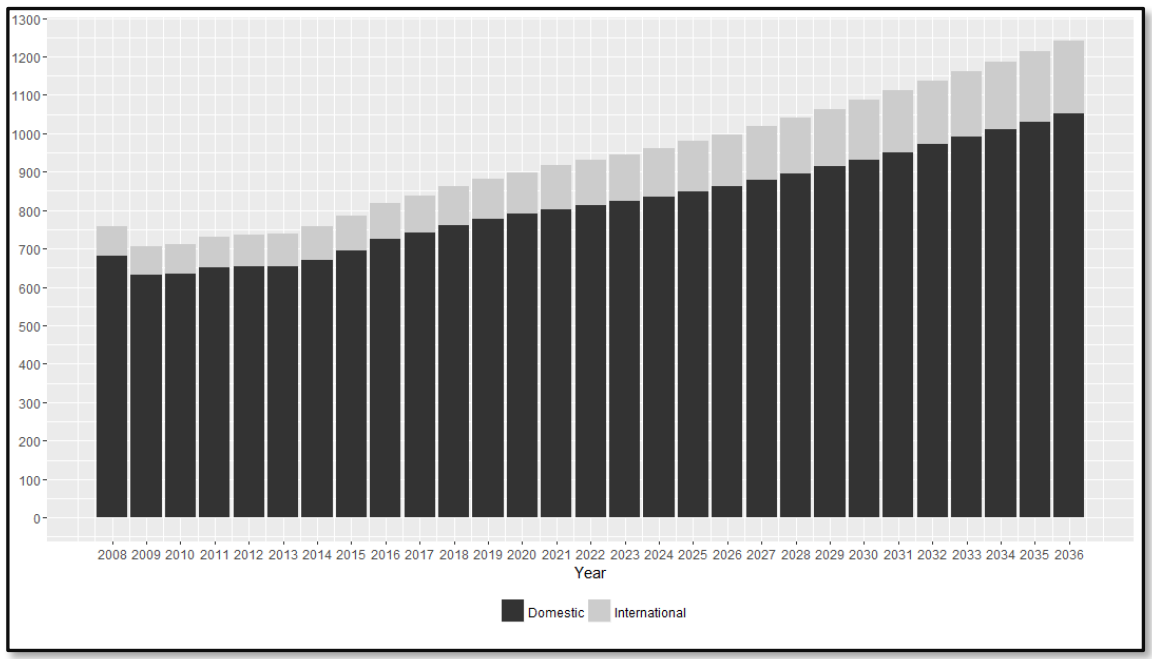

**Figure 1:** US Commercial Air Carriers Passenger Enplanement (2008-2036)
(Source: https://www.faa.gov/data_research/aviation/aerospace_forecasts)

As the upward trend in air traffic is expected to continue and outpace the capacity, the possibility of delays is likely to increase. According to FAA supported study on the total delay impact in the US, the total cost of all US air transportation delays in 2007 was estimated to be $31.2 billion. In addition to these direct costs imposed on the airline industry and passengers, aircraft delays have indirect effects on the US economy. It was also estimated that air transportation delays reduced the US GDP for that year by $4 billion (Ball et al., 2010). In another study, the FAA estimated that increasing congestion in the US air transportation system, if not addressed, would cost the economy $22 billion annually in lost economic activity by 2022 (FAA, 2007). Hence, flight delays are a growing challenge not only for air transportation industry but also for the environment and the whole economy.

The Bureau of Transportation Statistics (BTS) categorizes causes of air traffic delay into five broad groups: extreme weather, air traffic volume in the TMA, equipment problems, runway closure, and other. Figure 2 shows the causes of flight delays in the National Aviation System (NAS) between 2010 and 2015, where the largest contributor was weather, causing more than half of delays. The next major factor is the air traffic volume, which accounts for more than 31 percent, and this leaves a large room for improvement by increasing the effectiveness of TMA utilization.

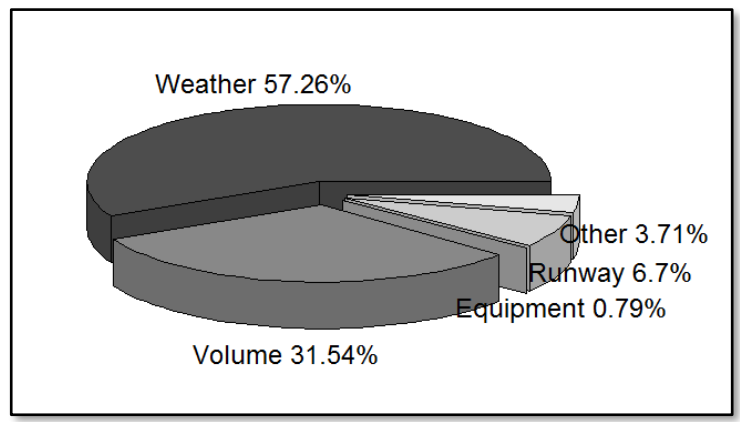

**Figure 2:** Causes of National Aviation System Delays (2010-2015)
(Source: http://www.transtats.bts.gov/)

The negative impact of high air traffic volume in TMA is exacerbated mainly by two factors. The first factor is the hub-and-spoke flight network structure widely adopted by airlines to enable more efficient use of their resources and improve service frequency. However, to offer a large variety of possible connections for passengers and limit waiting times at a hub airport, it is necessary for airlines to schedule as many runway operations (landings and take-offs) as possible during a short time frame, which are not equally spread throughout the day. This situation results in high traffic peaks during these times and often causes delays due to the scarcity of airport resources. The second factor is the unforeseen disruptions that cause sudden capacity drop. Due to the network structure of airport operations, delays propagate throughout the entire network and cause a knock-on effect of system-wide delays especially at major airports. Hence, these factors highlight the importance of finding robust solutions for better practical resource utilization.

Out of various resources in TMA, runways have been the main constrained resource that requires special attention. Hansman and Idris (2001) investigated the underlying dynamics of the aircraft departure process based on field observations and data obtained from a major US airport. They concluded that the largest delays and queues are mainly manifested in runways. In a more recent study, Mehta et al. (2013) analyzed the runway queues and sequences with both human-in-the-loop simulations and during operational tests at a major US airport. They concluded that given a significant diversity of aircraft; delay savings can range approximately up to four hours a day by optimizing landing and take-off sequences (Mehta et al., 2013).

It is primarily air traffic controllers' responsibility to ensure the safe and efficient flow of air traffic on the ground and in the close vicinity of airports. This task requires air traffic controllers to consider three dimensions of space and maintain a safe flow of operations and at the same time airport capacity has to be utilized efficiently with reduced fuel consumption. In order to accomplish this challenging task, they partially utilize automation and decision support tools to avoid human errors and achieve better resource utilization. Most of these tools require incorporation of operational data manually or cognitively by air traffic controllers. One of the significant tasks air traffic controllers perform is to schedule mixed departures and arrivals, due to the cognitive complexity in considering various operational constraints, such as wake turbulence separation requirements, time constraints of landing aircraft, etc.

The real-life combinatorial problem of scheduling runway operations (landing and take-off) is commonly referred to as runway (airport) operations scheduling problem. This problem includes determining the optimal landing/take-off sequence and start times over each arrival/departure runway in order to improve runway utilization, decrease delays, etc. This scheduling process is significantly important for achieving efficiency and effectiveness in runway operations, ensuring safety, improving passenger satisfaction, reducing air traffic delays, fuel burn and negative environmental impacts, etc.

Despite the vast body of knowledge related to runway operations scheduling, there are still literature gaps that need to be addressed. In particular, there is a clear knowledge gap on developing a methodology that explicitly take into consideration the inherent uncertainties, reflecting the stochastic and complex nature of runway operations. The quasi-optimal schedules that are obtained in the planning stage without considering uncertainties become far from optimal or even infeasible in practice. Usually, external factors, such as inclement weather or equipment failures, are held responsible for delays. However, certain delays are predictable and avoidable uncertainties are considered in the schedules at the planning stage. Mehta et al. (2013) provided clear evidence that there is a need for effective and efficient algorithms for recommending runway operations schedules that are robust to uncertainties.

Due to the complexity of this real-life problem, and in order to apply analytical methods, it has been inevitable to make numerous assumptions to reduce the complexity. As a consequence, the solutions acquired from the analytical models are far from practical. On the other hand, considering less simplifying assumptions for the sake of more realistic modeling usually lead to intractable analytical models, where results can only be estimated via some heuristic or approximation techniques. Therefore, appropriate modeling techniques and solution algorithms need to be employed. The simplifying assumptions, which had been made in many previous researches, often render the underlying model analytically tractable. These analytical models are quite often not validated because it is assumed that it is intrinsically valid since it is analytical. Therefore, the applicability of the analytical models seems quite narrow because of the assumptions made.

In addition, analytical models tend to oversimplify the dynamic and stochastic nature of the complex and nonlinear real-life problem; otherwise they will be computationally intractable. Also, it is a challenging task to develop a stochastic optimization model since even for a small-scale problem a large number of scenarios exist. On the other hand, simulation-based methods have proved to be a powerful tool to account for complex and stochastic nature of scheduling problems and have become an area of extensive investigation over the past decades. Hence, a simulation-based approach has the potential to overcome difficulties related to stochastic nature of the problem to address air traffic growth and reduce delays.

In comparison with a large number of publications on applying deterministic approaches to the runway operations scheduling problem, it seems that the exploration of using a SbO approach, especially within the context of multi-objective optimization (MOO), is promising. Therefore, the core motivation for this research is the need to improve decision-making in the TMA by developing a practical and operationally feasible optimization approach for scheduling runway operations. To this end, a SbO approach is proposed to be utilized as part of operational planning models and decision support systems used by air traffic controllers in order to find solutions to real-life multi-objective runway operations scheduling problem. Since several stakeholders are involved, decisions require considering interests of these stakeholders. Therefore, two potentially conflicting objective functions are considered simultaneously, namely maximizing runway throughput by exploiting separation requirements between aircraft, and maximizing fairness among airlines. These objectives are commonly considered as important in providing an operationally useful capability.

It is evident that the development of a SbO approach for runway operations requires an understanding of the air traffic management architecture and air traffic control practices. Hence, a brief overview of the air traffic management architecture and current air traffic control procedures are presented in the next section due to their high relevance, and also, inefficiencies are highlighted as a research motivation.

**1.2 Air Traffic Management**

Air traffic management (ATM) system is an essential component of airport operations, and it provides a set of services to guarantee the safety and efficiency of air traffic flows (De Neufville & Odoni, 2013). The smooth functioning of the ATM system is the primary task provided by the air navigation service providers (ANSPs). In the US, the FAA is the primary ANSP, whereas the Eurocontrol assumes this responsibility for the Europe. The operational side that directly interacts with the aircraft crew is commonly referred to as air traffic control (ATC). The specific purpose of ATC is to ensure the safety of aircraft by guaranteeing conformance to minimum separation requirements and maximize efficiency by increasing resource utilization.

In the US, FAA utilizes a hierarchically organized structure to implement ATC system, which is a large scale and multi-layered system with a single air traffic control system command center (ATCSCC) supervising the overall air traffic flow. This system has three components where these components interact with each other constantly. These components of the ATC system are listed below and illustrated in Figure 3:

(a) *Air route traffic control centers* (ARTCC) control the en-route airspace with generally low traffic density away from airports.

(b) The high traffic density region around an urban airport within a radius of 5 to 40 nautical miles (NMs) or below an altitude of 10,000 feet is called the *terminal radar approach control* (TRACON) area. The TRACON handles departing and approaching aircraft within its space.

(c)  *Air traffic control tower* (ATCT) is located at every airport that has regularly scheduled flights. ATCT controls aircraft during takeoff, landing and ground traffic, and in the air within 5 NMs.

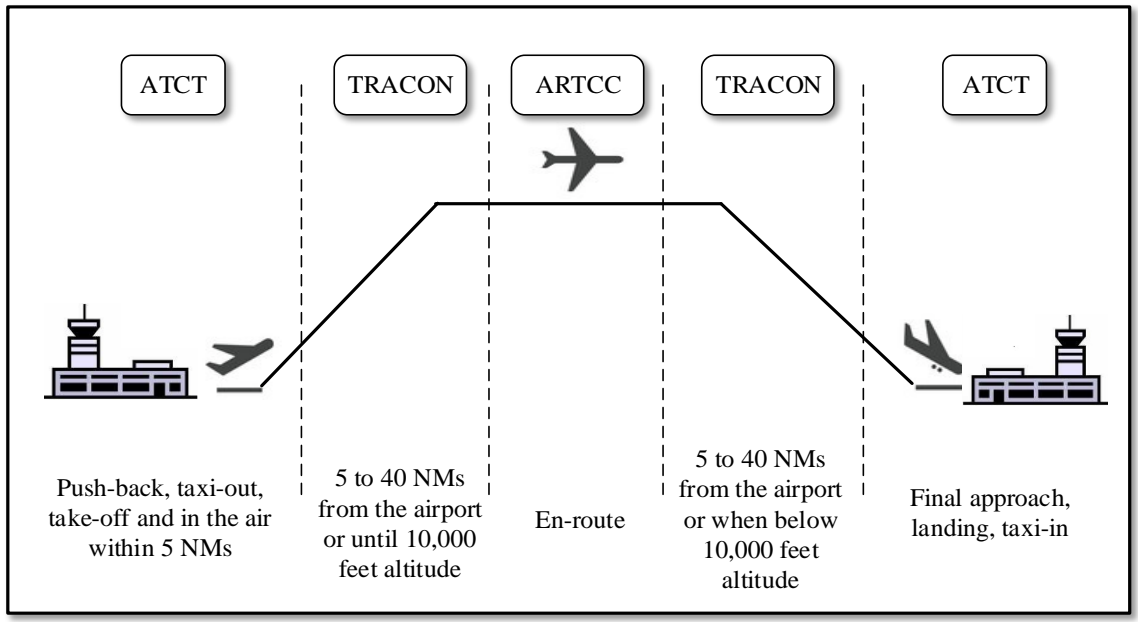

**Figure 3:** Components of Air Traffic Control System

Airport operations are traditionally divided into landside and airside operations. The landside operations comprise of operations that are directly related to passenger access to aircraft. On the other hand, the airside operations consist of aircraft operations in the runways, taxiways, spots, ramp areas, and gates. ATCT is responsible for the airside operations, i.e. safe and efficient handling of aircraft on the ground and in the close vicinity of an airport for both departing and arriving air traffic.

The airside operations are very complex and difficult to handle as a whole service; hence, it is partitioned into several services. The primary objectives of these services are to ensure the safety of the operations and to maximize the effectiveness of the system. In major airports, the air traffic controllers commonly consist of four groups of controllers: (1) gate controller (assigns gate to aircraft and grants pushback clearance), (2) ramp controller (provides clearance for ramp and sequences aircraft at the ramp), (3) ground controller (issues taxi clearances and arranges departure/arrival taxi queue), and (4) local (tower) controller (assigns runway and start times for landing/take-off of arriving/departing aircraft). The arrival and departure air traffic is controlled by the gate controller in the gate area, by the ramp controller in the ramp area, by the ground controller between the spot and the holding area/runway exit, and by the local controller between the holding area/runway exit and the runway. Responsibilities of these air traffic controllers are illustrated in Figure 4.

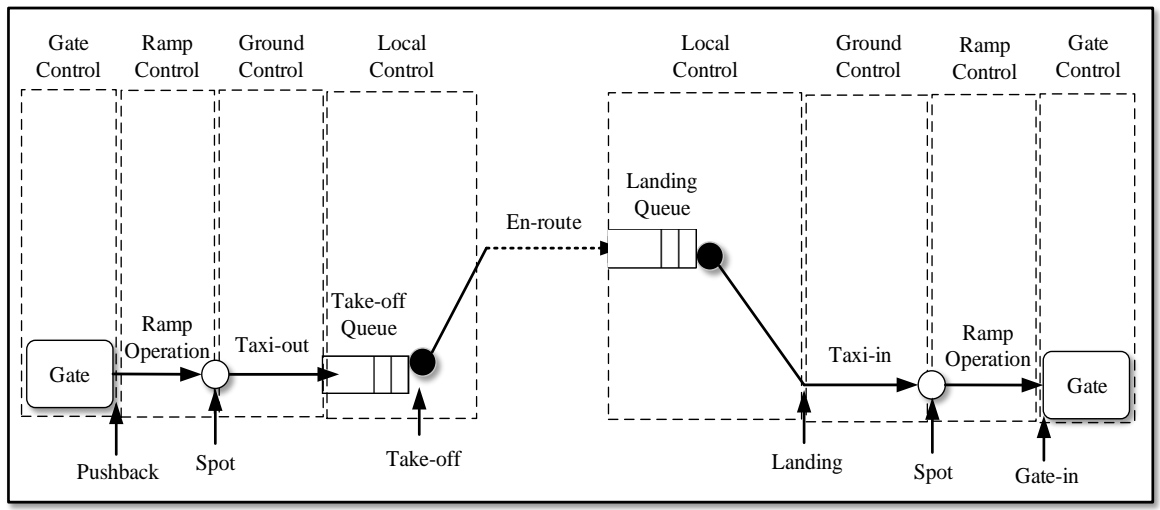

**Figure 4:** Responsibilities of Air Traffic Controllers

In ATCT, local controllers are the ones who are responsible for ensuring the safe and smooth flow of air traffic, while trying to minimize the delays and congestion on runways. The volume of runway operations, especially during peak hours, combined with some disruptions stem from uncertainties result in a very complex and dynamic operational environment. However, local controllers commonly utilize low-level of automation in this critical scheduling process. In an ideal situation, the local controllers should conduct this scheduling job interactively via a decision support system that is linked to a wider information system and databases. The development of efficient algorithms to be used in these decision support systems for runway operations scheduling is a significant and active area of research.

Furthermore, currently in practice, local controllers deal with uncertainties via a reactive approach by updating the initially planned aircraft schedules with unplanned event occurrence. However, the TMA is an extremely busy environment and rescheduling aircraft highly increases the workload of local controllers, and also, every intervention to the initial plans impacts the entire air traffic flow one way or another. One may think that adding buffer (slack) times to all aircraft separations may produce solutions robust to uncertainties; however, such schedules will lead to suboptimal solutions, and in turn, underutilization of runways and degrade in airport capacity. Therefore, uncertainty needs to be explicitly incorporated into the models to obtain robust schedules and realize better utilization of runways.

Existing air traffic flow operations heavily rely on local controllers to sequence aircraft, schedule the landing/take-off times and issue clearance instructions to each aircraft, and adjust the schedules when necessary to maintain minimum separations between aircraft. However, a paradigm shift has been pursued to change this reliance on local controllers as part of complete Air Traffic Management (ATM) system transformation both in the US and in Europe. FAA's Next Generation Air Transportation System (NextGen) program in the US and Eurocontrol's corresponding Single European Sky ATM Research (SESAR) program in Europe are still under development. These new systems adopt new technologies, such as 4-dimensional ATM (trajectory control), performance-based ATM, satellite-based navigation as well as collaborative decision-making (CDM) concepts.

The primary benefit of NextGen and SESAR technologies and concepts is that aircraft are expected to fly more exact routes in a more automated manner, thus improving system predictability and reliability, and increasing runway throughput and efficiency under varying demand and weather conditions. Therefore, the new concept of operations envisaged by NextGen and SESAR for the near future presents new opportunities for dealing with air traffic flow management on the current operational environment.

In NextGen and SESAR, it is envisaged that in major airports runway operations will be conducted using pre-determined surface trajectories for all aircraft. These pre-determined trajectories are planned to be a coordinated effort among all stakeholders especially ANSPs and airlines. Also, it is aimed to improve information sharing between the trajectory automation tools and airlines' automation systems. However, it is evident that substantial uncertainty in airport and runway operations stems from various sources as well as unexpected events will still be existent as they exist currently. Therefore, in the current as well as in NextGen's operational environment, it is highly important to take into account uncertainties during scheduling runway operations.

Recently, several air traffic tower automation tools have been developed to assist air traffic controllers in managing air traffic flow. The Airport Surface Detection Equipment-Model X (ASDE-X), which is a surface surveillance system, is used in major airports to help air traffic controllers maintain safe separation of aircraft and vehicles on the airport surface. Local controllers mainly use this surveillance system (if available in that airport) and voice communication systems to issue control clearances to aircraft. Another tool that is developed for scheduling arrivals is the Center TRACON Automation System, Traffic Management Advisor (CTAS/TMA) (this tool was developed by NASA), which is already in-service at many US major airports. As part of FAA NextGen efforts, the CTAS/TMA is planned to be replaced by an advanced decision support tools suite called the Tower Flight Data Manager (TFDM). The primary aim of TFDM is to serve as a platform for air traffic controllers to manage aircraft operations on the airport surface and in the TMA. There are some runway sequencing and scheduling related decision support tools considered under TFDM. The capabilities of ASDE-X and TFDM should be integrated to schedule runway operations better in the near future.

In Mehta et al. (2013), a recent study performed by the Massachusetts Institute of Technology (MIT) Lincoln Laboratory and sponsored by the FAA, the decision support functions of the latest TFDM prototype is analyzed and evaluated, and several gaps in performance are identified. This study indicates that the greatest potential operational benefits would come from decision support tools that facilitate managing runway queues and sequences. In addition, the same study provides evidence that outputs of the optimization approaches proposed in the literature are not operationally feasible, and in practice air traffic controllers still rely on first-come-first-served (FCFS) strategy, which usually yields suboptimal aircraft schedules and fails to utilize the available runway capacity completely. In the same study, two major issues have been identified to be addressed to make the proposed approaches applicable to practical runway operations, which include: (1) computational time for finding a feasible schedule, and (2) the impact of uncertainties on the resulting optimization algorithms (Mehta et al., 2013).

## 1.3 Problem Statement

### 1.3.1 Formal Problem Definition

The prominent problem in operations research literature which deals with scheduling aircraft for landing and take-off is referred to as runway operations scheduling problem. This real-life problem can be briefly defined as follows. Given a set of departing aircraft and another set of arriving aircraft, where each aircraft belongs to a weight class, we need to assign each aircraft to a runway, and then determine the start time for each runway operation (landing or take-off) on the assigned runway, while considering operational constraints, such as minimum separation times, time windows etc. This large-scale scheduling problem consists of a three-step process. The first step involves allocating aircraft to different runways; the second step is sequencing the aircraft allocated to each runway, and the third step is determining the operation start times for each aircraft. This problem usually arises at busy airports where runway utilization needs to be optimized to prevent delay-related costs.

The runway operations scheduling problem has a short planning horizon, which is usually 20-30 minutes. For arriving aircraft, each aircraft is considered as soon as it arrives at the extended TMA, which is about 30-40 minutes before its target landing time, and scheduled landing time is assigned before it reaches the final approach path, (about 20-30 minutes in advance of landing). For departing aircraft, each aircraft is considered as soon as it enters the holding area, and this time period varies among airports depending on the characteristics of taxiway, holding area, and runways. But, most of the time, take-off sequence for a runway should be determined before an aircraft enters the taxiway, since it is not possible to change the sequence during taxiing or at the holding area. Hence, take-off operations are scheduled approximately 20 minutes before target take-off time (Bennell et al., 2013). Therefore, from a practical standpoint, a solution method to the problem should generate a runway operations schedule within the mentioned short planning horizon.

Since the 1960s, developing efficient methods for tackling runway operations scheduling problem has been of great interest for both academic researchers and practitioners. Several techniques have been formulated to solve various forms of the problem. Bennell et al. (2013) identified the primary modeling approach as mapping the problem to a machine scheduling (job shop or parallel machine scheduling) problem, and the main solution techniques as dynamic programming (DP), branch-and-bound (B&B), and heuristics. Nearly the entire literature addresses the problem as a deterministic problem with the assumption that all the input data is known with certainty. However, there are numerous sources of uncertainty that need to be considered during scheduling, such as adverse weather, ground speed variations caused by the wind, piloting indecisions, unexpected delays, etc. In such cases, the quasi-optimal schedules become far from optimal in practice because of challenges posed by uncertainty. Hence, there is still an apparent shortage of study that takes uncertainty into account with a proactive approach.

From a local controller's point of view, the easiest scheme to use for scheduling runway operations is through the FCFS order. At the same time, FCFS is perceived as reasonably fair by most airlines. For the landing operation, this order is based on the order the aircraft enter the radar range, and for the take-off operation, it is based on the order of the aircraft queuing at the holding area. Although FCFS is an efficient strategy in terms of implementation, it does not typically produce the best schedule for runway utilization (Caprì & Ignaccolo, 2004). For major airports, low level of runway utilization leads to traffic congestion and delays, which, in turn, leads to inefficiency and waste of resources as well as environmental pollution.

The runway operations scheduling problem is usually modeled as a mixed integer program (MIP), a set partitioning problem or an asymmetric traveling salesman problem with time windows, and solved by utilizing mathematical programming methods where the obtained solutions are guaranteed to be optimal. However, since the minimum separation requirements are dependent on the aircraft order in terms of their weight class and runway operations type, the problem is combinatorial in nature. Therefore, for the practical large-scale problem sizes, the solution time is often prohibitive for these mathematical programming methods.

The computational complexity of the single runway operations scheduling problem has been shown to be Non-Deterministic Polynomial-time Hard (*NP*-Hard), which means that there is no known algorithm for efficiently finding optimal solutions to real-life problem sizes in polynomial time (Garey & Johnson, 1979). The multiple runways case of the problem is also *NP*-hard because it is a generalization of the single runway scheduling problem. In addition to the problem's inherent computational complexity, the magnitude of the problem's difficulty is exacerbated by considering uncertainties and multiple conflicting objectives. Therefore, exact (optimal) solution methods are not capable of solving practical problem sizes, and one of the main alternative solution methods is to use heuristic or metaheuristic algorithms.

1.3.2 Problem Characteristics

There are several dimension of classifying the runway operations scheduling problem, and most prominent ones are listed below:

*Static vs. dynamic*: This characteristic of the problem relates to time. The static (offline) case is solved before actual operations with known or predicted information while the dynamic (online) case is solved to generate schedules in real time as aircraft arrive in real time. The most common researched case in the literature is the static case where it is assumed that all data including ready times, target times and due times is known upfront and can be taken into account in the process. For this case, it is commonly assumed that a target time for landing or take-off is provided by another tool, such as Runway Scheduler (Balakrishnan & Jung, 2007). On the other hand, in dynamic case, all these data become known only when an aircraft is ready to land or take-off. Modeling approaches for dynamic case tend to be quite different from the ones for the static case, which requires some additional considerations about which aircraft to reschedule and when to reschedule the sequence of the aircraft. Therefore, in the dynamic case, fast computation times are necessary. In this research, the static case of the problem is considered.

*Runway configurations and procedures*: These characteristics are related to the number of runways that are considered (single or multiple), the interaction between the runways (interacting or independent), and the mode in which each runway is operated (mixed-mode or segregated-mode). In interacting runways, the separation requirements on one runway are affected by the operation at the other runway, and in independent runways this is not the case. In mixed-mode, each runway is utilized for arrival and departure at the same time. On the other hand, in segregated-mode, each runway is utilized only for either arrival or departure. Airport capacity can potentially be increased in mixed-mode, especially when there exist multiple runways, the opportunity is even greater. Arrival and departure aircraft inevitably interact through the common use of taxiways and runways. Thus, managing the air traffic flow successfully requires considering both arriving and departing aircraft in the TMA.

The runway operations scheduling problem is a challenging task both from a technical and an implementation point of view like most of the real-life scheduling problems. The major challenges on the technical part are considering uncertainties and multiple conflicting objectives in modeling the actual scheduling problem. On the other hand, implementation challenges are typically related to the fidelity and accuracy of the model developed for the analysis of the actual scheduling problem. In this research, uncertainties that are inherent to practical operations and multi-objective nature of the problem is taken into consideration, and detailed in the rest of this section.

1.3.3 Operational Constraints and Typical Objectives

The commonly considered operational constraints and typical objectives for the runway operations scheduling problem are briefly presented below:

*Time windows*: Once an aircraft enters the radar range for landing or pushbacks from the gate for take-off, air traffic (local) controllers assign a runway to it and a start time for landing/take-off. The start time has to be between the predetermined earliest and latest land/take-off time, so-called "time windows", which is a hard constraint. Also, there is a target time to land/take-off within this time window, which is the time that aircraft can land if it flies at its cruise speed for landing, and the most probable time for take-off considering the taxi-out and holding times for take-off.

*Minimum separation requirements*: This is the principal safety constraint that needs to be taken into account in a runway sequence, which is the spacing (time interval) between successive aircraft and it has to be equal or greater than the minimum requirement stated by the FAA. This spacing requirement is required for the wake vortices to dissipate. Wake vortices are turbulences of air which are caused by a leading aircraft as a result of its lift force. The FAA and other Civil Aviation Authorities around the world specify a set of minimum separation requirements in units of distance or time. The FAA enforced minimum separation requirements are largely determined by the type of operation and weight class of the leading and trailing aircraft. In the presence of multiple interdependent runways, these separation requirements are asymmetric (non-triangular), where the sequence of operations determines the actual separation time. Therefore, generating efficient aircraft schedules by exploiting the asymmetric separation times has the potential to increase runway utilization and delay reduction. However, the existence of asymmetric separation times between aircraft makes this scheduling problem a non-trivial one.

*Limited flexibility in deviating from the FCFS order*: In practice, local controllers often simply depend on FCFS strategy, which is the most straightforward and widely used approach. Although FCFS order eases local controllers' workload, maintains a sense of fairness among airlines and is easy to implement, usually it is not capable of providing the best schedule in terms of runway utilization. In practice, deviating from FCFS order is not common.

*Typical Objectives*: Different objectives are utilized in the literature considering various stakeholders' point of view. It is not practical to address the interests of all the stakeholders at the same time. Hence, the most commonly used ones are related to ANSPs' interests, such as minimizing the total delay (tardiness), minimizing the total deviation from the target time (earliness and tardiness), minimizing the average delay per aircraft, maximizing the throughput (makespan which is the landing or take-off time of the last aircraft), and minimizing the maximum delay. Total weighted tardiness measures the cost of delay that is a function of the length of delay multiplied by a weight (penalty) value related to each aircraft, and it is capable of addressing different stakeholders' needs. This objective is also very important for airline companies since every second the aircraft waits to land or take-off increases operating costs. From airlines' point of view, schedules for runway operations should ensure some degree of fairness, where FCFS sequence is typically perceived as relatively fair by airlines.

## 1.3.4 Consideration of Practical Aspects

A comprehensive literature survey, presented in Chapter 2, shows that there is a considerable gap between practitioners and academic researchers in the field of runway operations scheduling. Academic researchers are often not aware of the real-world complexities encountered by the industry practitioners, namely local controllers, and, in turn, they usually do not take into account most of the practical aspects of the problem. Most of the academic research conducted have the following assumptions:

(a)      They assume that all aircraft are already present on the final approach and holding area for landing and take-off, respectively. They also assume to have precise and reliable data on the aircraft and operating environment.

(b)      They consider mostly the deterministic problem in which the presence of uncertainties in actual runway operations is ignored to limit the computational complexity of the problem.

(c)      They highly focus on considering a single objective for optimizing the runway operation schedules, and they commonly do not consider Collaborative Decision Making (CDM) aspects and interests of different stakeholders at the same time.

However, local controllers are faced with daily challenges where uncertainties are real, and a trivial change in environmental conditions or small variations in implementation can be critical to operational safety and performance. In addition, finding the trade-offs between interests of different stakeholders is a key characteristic of the real-life problem. For example, concentrating only on runway utilization can cause unacceptable delays for individual aircraft, and in turn, this can impair the operational efficiency of the airline to which the aircraft belongs. Hence, the real-life problem involves several contradicting objectives that need to be satisfied simultaneously, and the most important ones are maximizing runway utilization, and maximizing fairness among all aircraft.

Essentially, runway operations scheduling problem is an applied area of research, and its benefits are ultimately derived from the results it achieves. However, by no means does this imply that the theoretical elements of this problem are not worthy of rigorous and careful treatment. Therefore, both theoretical and practical aspects of the problem are attempted in this research. To this end, the following practical elements have been taken into account to bring the problem closer to the practical real-life applications: (1) uncertainties inherent to runway operations, (2) multi-objective nature of the problem, and (3) ensuring fairness among aircraft.

*Uncertainties*: In practice, there are numerous sources of uncertainty that need to be considered during scheduling runway operations, such as inclement weather, airport congestion, equipment failure, unexpected delays in pushback or taxiing and so on; however, the main sources of uncertainty are push-back times, taxi times and wheels-on times for arrivals. These uncertainties are often very difficult, if not impossible, to avoid in practice. In such cases, the quasi-optimal schedules become far from optimal in practice because of challenges posed by uncertainty impacts. Uncertainty in runway operations usually manifests in the TMA in the form of traffic queues, and such queues typically result in operational inefficiencies, additional costs, such as fuel costs and environmental consequences. As a result, the methods to solve this scheduling problem should be robust enough to consider uncertainties, and efficient enough to produce solutions in a reasonable time.

*Stakeholders and Their Desired Interests*: There are various stakeholders in scheduling the aircraft landings and take-off on runways, and each has different interests. The most important of these stakeholders include ANSPs, airlines, airport managements and government agencies. The viewpoints of the various stakeholders who affect or be affected by the scheduling of aircraft over runways differ substantially. Moreover, each stakeholder is usually concerned with multiple performance measures. For example, ANSPs are primarily focused on the safe flow of air traffic and runway utilization. On the other hand, airlines are mainly concerned with resource utilization, punctuality, air traffic and on-time performance, etc.

As a relatively recent concept, airport CDM has been proposed as a means to deal with challenges at major airports, which has a potential to improve runway operations. CDM allows airlines to participate in air traffic decision-making that affects them. It also enhances collaboration between stakeholders, especially between ANSPs and airlines, to improve the operational efficiency of air traffic flow and satisfaction of the airlines by taking both stakeholder's interests into account.

In order to integrate CDM concept into models, the objective functions for the practical runway operations scheduling problem should reflect the preferences of interested stakeholders as precisely as possible, in particular, the interests of airlines. Provided that runway operations scheduling problem are multi-objective in nature as a result of CDM concept, the problem should be handled in an MOO context that considers more than one objective simultaneously to avoid imbalances among them. Therefore, there is an increasing need for considering multiple objectives as a way of taking into account various stakeholders' interests. One of the prominent objectives that is an integral part of CDM is ensuring fairness among airlines.

In the presence of multiple conflicting objectives, the resulting MOO problem gives rise to a number of optimal solutions, known as Pareto-optimal solutions. Since these are all trade-off solutions, the initial task for solving the MOO problem is to find as many such Pareto-optimal solutions as possible. There exist several traditional methods that convert the MOO problem into single optimization problem by utilizing some user-defined parameters. However, the key challenge for this is the fact that most of these objectives are non-commensurable, and it is hard to aggregate them into one synthetic objective. Therefore, finding near Pareto-optimal solutions is important in terms of facilitating effective trade-off decision-making.

The presence of uncertainty and multi-objectives introduces further challenges and additional computational complexity to the modeling and solution process. The difficulty is two-fold: (1) how to model the uncertainty, and (2) how to deal with computational complexity resulting from the existence of uncertainty and multiple objectives. Stochastic and dynamic nature of the runway operations renders simulation modeling as the only viable alternative for modeling uncertainty explicitly with a computationally tractable manner. The SbO approach is a widely utilized method for certain settings where analytical methods are not capable of optimizing complex models. The major downside of this method is the required computation time for simulation runs. Therefore, metaheuristic algorithms are commonly incorporated into the optimization component to deal better and efficiently with real-life, large-scale problems, such as multi-objective runway operations scheduling problem.

## 1.4 Research Philosophy and Methodology

The motivation and problem statement discussed in the previous sections have laid a foundation from which this dissertation research is formulated. This section describes the research philosophy, and then, outlines the aim, objectives, and scope of the research, guided by the forming of research questions.

The philosophy of science underlying this dissertation research, which pertains to the knowledge development and assumptions with regards to how the reality and knowledge are perceived, guided the research from problem definition to the conclusion in order to minimize researcher biases. In ontological consideration, critical realism is preferred due to our belief that even though there exist some form of reality that is external to the observer, it is yet to be perfectly understood. In epistemological consideration, post-positivism is embraced, since we believe that reality can be agreed upon by independent observers whereas the idea of relativity is respected. In methodological consideration, largely quantitative methods are utilized, and the following research methods are employed to achieve the research objectives: literature review, prototyping, and a case study. During the research, we tried to put emphasis on the validation of the proposed models and the whole SbO framework to address the epistemological question of "how we can know something to be true."

### 1.4.1 Research Questions

The runway operations scheduling problem is an important and challenging real-life problem; yet it has significant merit to be addressed efficiently as better utilization of the runways typically the only option left in response to the increasing demands for air traffic as most major airports are located close to residential areas, and in turn, there are concerns related to environmental issues, such as noise and pollution. A substantial number of models and approaches have been developed over the years. However, as the conducted literature review indicates, there is still room for improvement, especially regarding closing the gap between academic research and practice. The research in this dissertation is an effort to partially fill this gap.

The research described in this dissertation, which is based on the aforementioned research philosophy, is guided by the main and sub-research questions that can be summarized as follows:

Main research question:

"How to design and implement a hybrid Tabu/Search Scatter algorithm that is capable of generating best Pareto-optimal solutions for multi-objective runway operations scheduling problem within a SbO framework while considering the requirement of reasonable solution time enforced by the practical problem's planning horizon?"

Sub-research questions:

(a)     How to help local controllers to alleviate delays and increase runway utilization for future runway operations by explicitly incorporating uncertainties inherent to runway operations such that schedules are robust with respect to changes in the input data? In addition, how to formulate and incorporate fairness among aircraft into the scheduling process, which is perceived as an important factor by airlines?

(b)     Considering that multi-runway operations scheduling problem under uncertainty is a difficult problem to solve, does SbO approach generate robust solutions fast enough to be used in a real-life situation? How to develop a SbO framework for solving more realistic runway operations scheduling problem?

(c)     How to design and implement a discrete-event simulation model that simulates runway operations with an appropriate level of fidelity? How to develop methods to provide an effective initial solution? How to deal with the challenges posed by the simulation noise?

(d)    To what extent the SbO approach can contribute to the practical scheduling process where there are multiple conflicting objectives? What are the benefits that this approach can provide in comparison to deterministic and FCFS approaches in solving real-life runway operations scheduling problem?

## 1.4.2 Research Scope

The formulated research questions provide the conceptual framework that guides the scope of this research. The scope is delineated by forming a set of guidelines which ultimately result in achieving the practical application of the proposed methodology. The scope of the research in this dissertation is detailed as follows:

(a)    The static case of the runway operations scheduling problem with multiple interacting runways operating in mixed-mode with non-triangular separation times is considered. This is a more constrained problem than the single runway case or multiple runways operating in segregated-mode case. This version of the problem is addressed because it offers a better opportunity to generate operational benefits through effective runway operations scheduling.

(b)    In order to model the uncertainty in the problem, some of the input data are assumed to have a random element. Although taxi times are one of the main sources of uncertainty, the taxi routing problem is not integrated with the runway operations schedules. Therefore, the information related to the aircraft on the taxiways is accepted as inputs to the system, rather than as a part of the problem to be solved.

(c)    The simulation methodology used for the SbO framework is based on discrete-event simulation, which is an approach that models the runway operations as a network of activities, and where the system state is changed at the discrete point of time. In this approach, each entity, i.e. aircraft, is individually represented and specific attributes are attached to these entities in order to follow and collect some performance measures during the simulation.

(d)      In terms of multi-objective optimization (MOO) component of the SbO framework, best Pareto (trade-off) solutions are considered instead of aggregating the objectives into a single objective as trade-off solutions allow decision-makers to compromise between multiple objectives and make decisions that consider different stakeholders' interests simultaneously.

### 1.4.3 Research Aim and Objectives

Given the complexity and practical requirements of the runway operations scheduling problem, a SbO approach seems to be one of the best suitable methods for solving it. The main advantage of integrating simulation into optimization is that it can include less modeling assumptions, resulting in a more realistic and valid model and, in turn, leading to a better decision-making process. In SbO approach, a simulation model is commonly utilized for evaluating the performance of a solution which provides a convenient means to capture more realistic aspects of runway operations. However, this approach still faces challenges in terms of optimization especially when there exist multiple and conflicting objectives as well as the burden of computational time resulting from the simulation runs. Due to the complexity of this MOO, metaheuristic algorithms are practical and suitable techniques to find optimal or near-optimal solutions without much computational intractability issues.

The overall aim of this research effort is to fill a portion of the knowledge gap between theory and practice in runway operations scheduling by accounting for some practical complexities while keeping the computational tractability at a reasonable level, and to develop a hybrid Tabu/Scatter Search algorithm based on for approximating the Pareto-optimal solutions of the multi-objective runway operations scheduling problem in a SbO framework. The algorithm tries to evolve the reference set of solutions towards the Pareto-frontier in each iteration and distribute it over the Pareto-frontier to maintain a diverse set of solutions. To the best of our knowledge, this is the first attempt in the literature to employ a SbO approach for solving this real-life scheduling problem that considers uncertainties as well as fairness among aircraft as a secondary objective.

The proposed algorithm, which is the optimization component of the SbO framework, is a novel algorithm that takes advantage of the structural details of the problem, and can be distinguished from the current algorithms in the way it uses an elitist strategy to preserve non-dominated solutions, a dynamic update mechanism to produce high-quality solutions and a rebuilding strategy to promote diversity across the Pareto-frontier. Furthermore, the proposed SbO approach designed to be utilized as part of decision support tools used by local controllers, capable of finding reasonably good quality solutions in a relatively considerable time.

The thesis statement is defined as follows: develop a hybrid Tabu/Scatter Search algorithm, which is capable of handling multiple conflicting objectives and stochastic noise, can efficiently and effectively tackle the multiple runway operations scheduling problem within a simulation-based optimization (SbO) framework.

The following key research objectives have been identified as the main steps in addressing the aforementioned aim of the research (the chapter that describes the related activities of each particular objective is given in parentheses):

(a)     Since the motivation of this dissertation is to propose a more practical and efficient problem solving approach, a comprehensive literature review is performed to identify the knowledge gaps, inefficiencies in existing models, and their causes to mitigate them to a practical level. (Chapter 2)

(b)     Design a SbO framework to tackle the challenges posed by stochastic and multi-objective nature of runway operations. First, related concepts and methods are reviewed for coping with SbO as well as MOO. Compared to published literature, some of the simplifying assumptions related to runway operations are reduced, and additional practical considerations are included. (Chapter 3)

(c)     Design and implement a discrete-event simulation model with an object-oriented architecture to replicate uncertainties to a realistic extent so as to evaluate and integrate it into a SbO framework. (Chapter 4)

(d)     Design and implement a new hybrid Tabu/Scatter Search algorithm to search for best known Pareto-optimal solutions within reasonable computational times considering the planning horizon of the practical runway operations scheduling problem. (Chapter 5)

(e)     Evaluate the robustness and effectiveness of the proposed optimization and simulation components as well as the whole SbO approach as a proof-of-concept by conducting computational experiments using real-life data from a major US airport. Analyze the experimental results in terms of potential operational benefits compared to deterministic and FCFS approaches. Derive conclusions based on the solutions obtained through statistical analysis of the outputs. (Chapter 6)

(f)     Analyze the whole problem-solving methodology, and the overall approach on the basis of the computational results, and propose extensions that can be done as a future research. (Chapter 7)

## 1.5 Summary of Contributions

The proposed approach in this research expands the current models by integrating a simulation model to consider explicitly the uncertainties related to some variables that have been treated as deterministic in order to limit the computational complexity thus far. It also contributes to filling the literature gap by taking into account two conflicting objectives simultaneously such that fairness among aircraft is considered as a secondary objective. This approach provides more realistic and robust solutions that can be applied to practical runway operations scheduling.

Therefore, this research effort provides novel contributions originate from the development and integration of optimization and simulation models. The main contributions (theoretical and practical) and scientific novelty of this dissertation, and how these relate to the research questions can be summarized as follows:

(a)    *A hybrid Tabu/Scatter Search algorithm is designed and developed that consider multiple conflicting objectives*. The primary contribution of this dissertation research lies in a novel hybrid Tabu/Scatter Search algorithm. The proposed algorithm utilizes a Pareto approach to deal with multiple conflicting objectives where a set of best trade-off solutions is searched for and presented to the local controllers. The algorithm is based on a population-based metaheuristic in order to capture multiple trade-off solutions in a single run since they are capable of maintaining a set of solutions.

(b)    *A modeling framework based on SbO is proposed for the real-life runway operations scheduling problem, which addresses the identified knowledge gap*. A SbO framework is developed with a potential to generate better runway utilization than deterministic optimization procedures under certain conditions. Although both the scheduling problem and incorporating the uncertainty is complex, the SbO approach is simplified by using an appropriate decomposition approach, thereby enabling the design of an effective solution system. Therefore, computational efficiency is achieved through the utilization of metaheuristic algorithm.

(c)    *A simulation component consists of a discrete-event simulation model developed based on an object-oriented architecture*. Unlike many works in the literature that focus mainly on developing simulation models with the help of existing commercial or open-source simulation software packages, a modular and flexible discrete-event simulation model is designed and implemented by an object-oriented programming paradigm.

(d)    *Computational experiments are conducted using data from a major US airport, and the obtained results are presented as a proof-of-concept*. Furthermore, in order to demonstrate that the proposed approach can lead to higher quality runway schedules over current methodologies, the computational evidence is provided with the help of computational experiments. Also, it is shown that the proposed approach is computationally feasible for large-scale real-life problem instances. The results show that the proposed approach improves the overall performance of the runway operations scheduling for better runway utilization, low-level of delays and fairness among airlines. Along with these results, a discussion on how characteristics of an airport can affect the strategy is provided.

(e)     *A comprehensive literature review of academic research and practical applications in runway operations scheduling and solution approaches for the problem are performed*. This literature review provides an insight into the needs for further development of modeling and solution approaches for runway operations scheduling as well as SbO and MOO methods. This review is not by any means as a comprehensive treatment to literature in SbO and MOO methods; its only purpose is to identify the knowledge gaps in the literature.

## 1.6 Outline of the Dissertation

The structure of this dissertation is organized as follows. The first (this) chapter has briefly introduced the motivation behind this research, the basics of the air traffic management and the problem statement, followed by the research scope, aim, objectives, and a summary of the contributions of the dissertation.

Chapter 2 provides general background information and presents the review of the extensive literature on current research pertaining to the runway operations scheduling problem including approaches that explicitly consider uncertainties as well as the concept of CDM. An additional literature review is given on machine scheduling under uncertainty since it relates to the area of research. Also, existing mathematical programming models related to the problem are presented along with the notation. Furthermore, alternative approaches for optimization under uncertainty are reviewed and compared. Finally, based on the information gleaned from the literature review, knowledge gaps are identified, which eventually motivated this research.

Chapter 3 outlines the important theoretical and practical underpinnings of the SbO framework. Also, the specific factors that shaped the development of this framework are provided. Basic concepts, principles, and terminology of SbO are also given for the sake of completeness. In addition, an overview of the definition and characteristics of MOO are described, and an overview of different MOO techniques are given. Next, simulation-based multi-objective optimization is discussed, and the currently utilized methods for dealing with noise in the simulation are outlined. It is noteworthy to mention that the primary aim of this chapter is not an in-depth analysis and treatment of SbO and MOO but to give a basic understanding of the related fields' principles. Then, this chapter sets out the methodology on which the proposed approach depends, and addresses the considerations and rationale for the overall SbO approach, which comprises an optimization and a simulation model as well as a greedy heuristic algorithm to generate an initial solution.

Chapter 4 explains the discrete-event simulation model in detail, setting out the key elements of the model. The main focus is to outline the design and implementation of the simulation model including assumptions. First, an overview of the state-of-the-art airport and runway simulation models is provided. Then, the purpose and a high-level framework of the model are presented along with a conceptual model. Also, the quantitative modeling process, object-oriented design architecture, and implementation specifics are explained. Finally, how the verification and validation study conducted before the experimental stage is described.

Chapter 5 details the proposed hybrid Tabu/Scatter Search algorithm. First, metaheuristic algorithms and their foundational concepts are provided because the proposed algorithm is based on these concepts. In particular, basic design elements including initialization, representation, search operators, fitness function, and search strategies are presented. The multi-objective search components, namely fitness assignment, diversity preservation and elitism, are discussed. This chapter is concluded with the object-oriented design and implementation specifics of the proposed algorithm.

Chapter 6 presents conducted computational experiments as well as their detailed results. Computational experiments are reported in two separate parts as it is conducted: first, multi-objective optimization, and then, simulation-based optimization (SbO) experiments. The SbO experiments are based on actual historical operational data from a major US airport. Prior to presenting the experimental results, the way the experiments are designed and setup are also provided. The results demonstrate both that the proposed approach can provide significant benefits in practice compared to FCFS sequence and deterministic schedule, and also it is computationally tractable.

Finally, Chapter 7 provides a summary of the research and derives several conclusions from the proposed approach. Also, several areas deserving further investigation are discussed, and a few potential directions for future research are pointed out.

The dissertation flow and each component's corresponding chapters are shown in Figure 5, which better illustrates the outline of the dissertation.

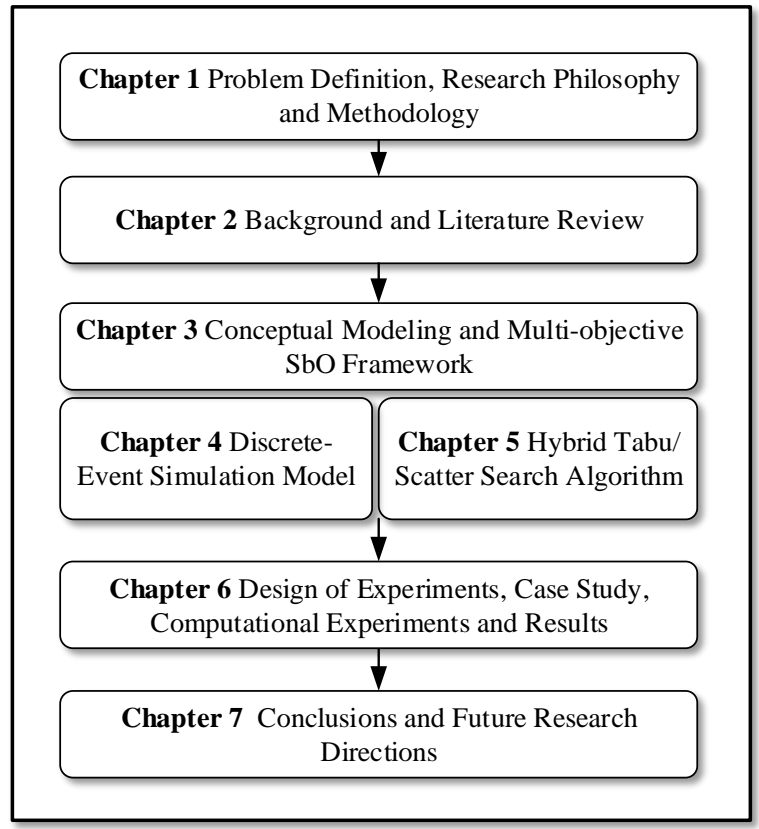

**Figure 5:** Dissertation Flow and Chapters

Throughout the dissertation, scalar quantities are denoted as lowercase, non-bold face symbols (e.g., $x \in \mathbb{R}$), vector quantities are denoted as lowercase, boldface symbols (e.g., $\boldsymbol{x} \in \mathbb{R}^n$, $n > 1$), and matrices are denoted as uppercase, boldface symbols (e.g., $\mathbf{A} \in \mathbb{R}^{n \times n}$, $n > 1$).

# CHAPTER 2

# BACKGROUND AND LITERATURE REVIEW

This chapter is dedicated to present the background information to answer the research questions that are formulated in the previous chapter. The outcome of a wide but non-exhaustive literature review that focused specifically on the recent publications is also provided. The presented literature review in this chapter mainly contributes to defining the scope and focus of this research. More specific literature reviews are provided in the context of the different subsequent chapters when necessary.

The first section presents the review of the existing literature to have a better understanding in various aspects of the runway operations scheduling problem. Also, the existing robust and stochastic approaches to the problem, and proposed approaches that consider CDM aspects and fairness are presented concisely. Then, the literature review on machine scheduling under uncertainty is provided briefly, since it relates to the area of research. In addition, mathematical modeling formulations along with the notation are presented. The full details of the existing mathematical models are relegated to references cited to avoid getting immersed in unnecessary detail. Also, alternative modeling approaches for optimization under uncertainty are identified to consider necessary practical aspects of the problem that supports the assumed methodology. The chapter concludes by providing a summary of the findings obtained from the literature review, discussing the knowledge gaps in the existing literature, and justifying the need of employing a SbO approach for the problem.

## 2.1 Runway Operations Scheduling Problem

Since the 1960s, developing efficient methods for tackling runway operations scheduling problem has been of great interest to both academic researchers and industry practitioners. It is a large-scale scheduling problem consists of a three-step process. The first step involves allocating aircraft to different runways; the second step includes sequencing the aircraft allocated to each runway, and the third step consists of determining the start times for each aircraft. This problem usually arises at busy airports where runway utilization needs to be optimized to reduce air traffic congestion and minimize delay, which would eventually provide the benefit of increasing flight safety and decreasing delay-related costs. For most of the major airports that operate with multiple concurrently active runways, scheduling multiple runway operations is a difficult task that local (air traffic tower) controllers face in a daily basis. Carefully scheduling runway operations has a potential to result in substantial improvements in runway utilization and safety. Therefore, reducing the high costs of aircraft delays, pollution (both regarding fuel consumption and environmental effects) and passenger dissatisfaction is a key motivator for this area of research.

The previous research identified on runway operations scheduling problem differ in terms of both scope and modeling perspective, but they share some foundational facts and assumptions. The identified research trends for the problem include the following: (1) developing models for various aspects of the problem, (2) developing assessment tools and performance measures (particularly simulation models for evaluating different scheduling approaches), and (3) conducting empirical studies and experiments to solve a particular airport's runway operations scheduling problem. In this dissertation research, all three areas for the problem are taken into consideration, since a SbO approach necessitates all these trends to be addressed.

Bennell et al. (2011); (2013) have provided a comprehensive review of the runway operations scheduling problem. The solution methods for the problem can be classified as exact and heuristic algorithms. Exact algorithms, such as branch-and-bound (B&B), dynamic programming (DP), etc., guarantee optimal solutions, but they are extremely computationally intensive for large-scale problem instances. On the other hand, heuristic algorithms generate solutions which are not guaranteed to be close to the optimum and the performance of heuristics is often evaluated empirically. These algorithms are often more time efficient. The heuristic algorithms are usually classified as constructive and improvement heuristics. The composite dispatching rules, such as Apparent Tardiness Cost with Setups (ATCS) rule, are examples of constructive heuristics whereas the local search methods are examples of improvement heuristics.

Although aircraft take-off problem is regarded as more difficult to solve compare to the landing problem (De Maere & Atkin, 2015), aircraft landing problem has received greater interest than take-off problem in the literature (Bennell et al., 2013). Therefore, the literature review for aircraft landing problem is presented first.

### 2.1.1 Aircraft Landing Problem

Bianco et al. (1987) and Bianco et al. (1997) mapped aircraft landing problem to a machine scheduling problem (job shop scheduling problem) on a single machine formulation with earliest release time for the jobs, and sequence-dependent setup times. They solved the single runway problem with a mixed integer linear programming (MILP) model.

C. R. Brinton (1992) utilized a B&B tree search algorithm to find the optimum landing sequence which minimizes the total delay assigned to a group of aircraft. The proposed algorithm first obtains an upper bound on the delay by computing the delay associated with an FCFS sequence. Then, it simply expands the tree, whose nodes correspond to aircraft within the arrival sequence, while keeping track of the total incremental delay assigned until the last expansion.

Abela et al. (1993) studied the aircraft landing problem with a single runway and proposed a B&B algorithm based upon a 0–1 MIP formulation, where a cost component for each aircraft is included in the objective function that is related to either speeding up or holding. Also, they proposed a heuristic based on genetic algorithm, and reported computational results for problem instances with up to 20 aircraft.

NASA Ames Research Center developed the Final Approach Spacing Tool (FAST) for assisting air traffic controllers in the management and control of landing aircraft. FAST includes a fuzzy decision logic that computes a new runway operations sequence, each time a new aircraft arrives the TMA at a given entry fix (point). The first aircraft in the outermost flight segments are sequenced, and then, sequences are merged repetitively at intermediate flight segments until it finally determines the aircraft sequence at the runway threshold. FAST consist of several major components: route analyzer and trajectory synthesizer, a sequencer and scheduler, a conflict resolver, a runway allocator, and a controller interface (Davis et al., 1997).

Ernst et al. (1999) also considered the single runway aircraft landing problem and pointed out that the single runway problem could be extended to multiple runways scheduling problem. They suggested a B&B algorithm and a genetic algorithm, where the objective function consists of penalty costs for landing before and after target times. Proposed algorithms are tested via problem instances available in the literature involving up to 50 aircraft on both a single runway and multiple runways.

Beasley et al. (2000) built and analyzed a MIP model and solved it by a technique based on the relaxation of binary variables by adding additional constraints such as limits on the maximum number of position shifts. In addition, an effective heuristic algorithm is presented. Also, they reported computational results presenting that their formulation produces optimal results within reasonable time limits for the problem instances found in the OR-Library (Beasley, 1990) involving up to 50 aircraft and four runways. However, their MIP model is not capable of addressing all real-life instances.

Bäuerle et al. (2007) modeled the landing problem as a special queueing system and utilized M/SM/1 queues with dependent service times to model a single runway. They considered the case of two runways with some heuristic routing strategies. They also analyzed and compared these strategies numerically with respect to the average delay for assigning aircraft to two runways.

Artiouchine et al. (2008) proposed an approach based on a general hybrid branch-and-cut (B&C) framework, based on constraint programming and MIP, to solve the single runway problem with arbitrary time windows.

M. J. Soomer and Franx (2008) studied a collaborative strategy where airlines assign a cost function for each of their flights, and these cost functions are scaled per airline to achieve fairness between airlines. They also developed a local search heuristic to incorporate the fairness into the schedule.

Yu et al. (2011) proposed a two-stage algorithm based on cellular automata. In the first stage, a good sequence of aircraft is found using cellular automata with updating rules in order to enforce separation and reach a good objective value. In the second stage, a local search heuristic is utilized to determine the start times for landing.

Although the majority of the literature have focused on aircraft landing scheduling over a single runway, there are several published research for multiple runways. Ciesielski and Scerri (1998) suggested a genetic algorithm for scheduling aircraft on two runways. Cheng et al. (1999) developed a genetic algorithm for multiple runways.

Wen et al. (2005) addressed the aircraft landing problem and formulated it as a set partitioning problem with side constraints. They suggested a branch-and-price (B&P) algorithm, which is similar to the B&B, but column generation is applied at each node of B&B tree. A combination of a genetic algorithm and an ant colony optimization algorithm for multiple runways has been proposed by G Bencheikh et al. (2009).

Pinol and Beasley (2006) suggested two population-based metaheuristics for multiple runways: Scatter Search and Bionomic algorithm. Their objective was to achieve effective runway utilization, where two different objective functions (a non-linear and a linear) were used during the experiments.

Boeing developed the multiple runway planner (MRP), which is introduced by Berge et al. (2006), to address multiple runways landing problem including the features of both last phase of the en-route and runway operations. The MRP calculates runway assignments and schedules at the fixes and at the assigned runways for a given set of landing aircraft with pre-assigned meter fixes and estimated times of landing (ETL).

Liu (2011) presented a genetic local search algorithm, where a local search procedure is incorporated into a genetic algorithm framework, for solving the aircraft landing problem with runway dependent attributes. They conducted several numerical experiments to test the validity of their genetic algorithm based on test instances from the literature.

Ghizlane Bencheikh et al. (2011) studied the aircraft landing problem in the multiple runway case and developed an ant colony optimization algorithm. They also proposed a new heuristic algorithm for scheduling aircraft landing times on a single runway from an order determined by a priority rule, and they compared several priority rules to test their heuristic algorithm. They tested their ant colony optimization algorithm with the OR-Library problem instances involving 10 to 50 aircraft and 1 to 5 runways.

Xiao-rong et al. (2014) considered the multiple runways aircraft landing problem with the objective of minimizing the total deviation from the target time and suggested a hybrid bat algorithm, where several local search procedures are integrated into the framework.

Faye (2015) proposed a method based on an approximation of the separation time matrix by a rank two matrix and on the discretization of the planning horizon. They suggested an exact method based on a dynamic constraint generation algorithm and also a heuristic method used to solve the model.

Girish (2016) developed a hybrid particle swarm optimization algorithm in a rolling horizon approach as a solution method for single and multiple runways cases of the aircraft landing problem. The considered objective function is to minimize the total penalty cost due to deviation of landing times of aircrafts from the respective target landing times. They assessed the performance of the proposed algorithm using OR-Library benchmark instances involving up to 500 aircrafts and 5 runways, and concluded that their algorithm produce high-quality solutions in short computational times.

2.1.2 Aircraft Take-Off Problem

As previously mentioned, aircraft take-off problem has received less attention in the literature compare to aircraft landing problem due to the fact that this problem includes more operational constraints, and it is heavily related to taxi-out scheduling problem, which requires these two problems to be integrated. However, this integration commonly renders the problem complex and intractable.

Anagnostakis and Clarke (2003) introduced a two-stage optimization algorithm for solving the take-off problem. In the first stage, throughput maximization is addressed to determine the best take-off class sequence to be used in the second stage, while ignoring the operational constraints. In the second stage, an integer programming formulation is utilized that generates a solution representing the assignment of aircraft to class slots while considering the related constraints.

Atkin et al. (2007) and Atkin et al. (2008a) dealt with the take-off scheduling with the objective of maximizing the runway throughput. They proposed different metaheuristics (Steeper Descent, Tabu Search, and Simulated Annealing) and analyzed their performance. As a result, it is reported that Tabu Search outperformed the others but with a small margin.

Atkin et al. (2008b) addressed the dynamic aircraft take-off problem and proposed a scheduling algorithm and a decision support system where taxi times are considered as uncertain. The main drive of their research is to study how the uncertainty influences the proposed scheduling algorithm. Also, experimental results are presented where the effect of taxi times are measured explicitly, and real data is utilized from different times of the day, showing how the performance of the aircraft departure system differs according to the volume of traffic and the accurateness of the provided taxi time estimations.

Rathinam et al. (2009) proposed a generalized DP approach to solve the take-off problem optimally, which exploits the chain-like ordering of the aircraft and tries to minimize total delay. Also, they used simulation to evaluate if their approach is fast enough to be considered for implementation in a real-time decision support tool and the quality of the solutions compare to FCFS order. After the computational results, they concluded that their approach is fast for a real-life implementation and can be used to reduce the aircraft delays at an airport.

Stiverson (2009) developed a greedy algorithm and a *k*-interchange heuristic algorithm to find improved take-off sequences. They also provided lower bounds on an optimal solution with the help of a MIP model. The proposed heuristic algorithms were tested using randomly generated datasets, and it was reported that, in general, the heuristic solutions were within 10–15 percent of the optimal solution.

### 2.1.3 Integrated Aircraft Landing and Take-Off Models

Trivizas (1998) considered a DP approach based on the Constrained Position Shifting (CPS) concept for solving the static aircraft landing and take-off scheduling problem for multiple runways in mixed and segregated-mode. They also conducted computational experiments with real data and reported that even a modest maximum position shifting (MPS) value is capable of increasing the runway capacity up to 20 percent with respect to FCFS.

Bianco et al. (2006) carried out a study examining the incorporation of a practical consideration which consists of constraining the set of feasible positions in the sequence for the new aircraft to prevent too many perturbations to the schedule.

Hancerliogullari et al. (2013) proposed three greedy algorithms and two metaheuristics including Simulated Annealing and Meta-RaPS (Metaheuristic for Randomized Priority Search) to solve static case of the multiple runway operations scheduling problem.

Ravidas et al. (2012) considered the two-runway scheduling problem where total delay tried to be minimized subject to operational constraints such as timing, safety, and chain-type precedence restrictions. A solution approach based on generalized DP is developed to solve the problem optimally. The presented computational experiments illustrate that their algorithm is capable of generating solutions in a reasonable computational time.

Ghoniem et al. (2014) addressed aircraft scheduling problem over a mixed-mode single runway or close parallel runways for the static case. They proposed a MIP formulation based on asymmetric traveling salesman problem with time-windows. They further embedded this model within the framework of two heuristics. They also tested the proposed exact and heuristic solution methods with real data and simulated instances and over 50 percent computational time savings were reported.

A novel branch-and-price (B&P) algorithm has been recently introduced by Ghoniem et al. (2015), where the set partitioning formulation is used. The model decomposed into a master problem and a pricing sub-problem, and the pricing sub-problem is formulated as an elementary shortest path problem and solved with a specialized DP approach, which is identified as the main factor for accelerating the solution process substantially.

D'Ariano et al. (2015) extended the existing job shop scheduling models proposed in the literature for integrated aircraft landing and take-off problem by considering additional practical constraints: (1) holding circle constraints, (2) separation time interval constraints for air segments, (3) blocking constraints for run-ways, and (4) time windows constraints for the aircraft travel time in air segments. They proposed several exact and heuristic algorithms to handle the constraints of the specific formulation. After conducting computational experiments, they concluded that their optimization models and algorithms has a significant potential for enhancing the performance of the TMA and decreasing the workload of air traffic controllers.

Lieder and Stolletz (2015) presented a MIP model and a dynamic programming solution approach for integrated aircraft landing and take-off problem with interdependent runways. They also proposed a rolling planning horizon (RPH) heuristic for solving large-scale instances. They conducted numerical experiments to evaluate the performance of their both exact and heuristic approach, and reported that both approaches yield high computation performance. They concluded that additional runway capacity can be obtained from optimized runway schedules compared to FCFS schedules using realistic runway settings.

In the literature, there is very few research that considers the multi-objective version of the problem. Montoya et al. (2014) proposed a multi-objective DP algorithm that minimizes the total delay of aircraft and the makespan of a sequence to find a set of Pareto-optimal solutions that completely represent the non-dominated frontier. The simulation results to validate the proposed algorithm were also provided.

Related literature on deterministic runway operations scheduling problems is tabulated and presented in Table 1. The table is divided into five sections: "Source" column indicates the article's reference; "Research Scope" column displays the problem characteristics considered in the article; the column "Modeling Approach" shows the approach used to model the problem; "Solution Approach" column illustrates the solution technique employed; and "Objective(s)" column presents the objective or the various objectives studied in the surveyed article.

**Table 1:** Related Literature on Deterministic Runway Operations Scheduling Problem

| Source | Research Scope | Modeling Approach | Solution Approach | Objective Function(s) |
|---|---|---|---|---|
| (Beasley et al., 2000) | Single and multiple interdependent runways | MIP and time-indexed MIP | B&B | Minimize weighted earliness and tardiness |
| (Bianco et al., 2006) | Single and two interdependent runways | No-wait job-shop scheduling model with sequence dependent machine set-up times and job release dates | Local search heuristic | Minimize makespan and minimize average delay |
| (Pinol & Beasley, 2006) | Single and multiple interdependent runways | MIP | Scatter Search and Bionomic algorithms | Minimize weighted earliness and tardiness |
| (Balakrishnan & Chandran, 2010) | Single runway | Constrained position shifting network | Dynamic programming | Minimize makespan and minimize total delay |
| (Ghizlane Bencheikh et al., 2011) | Single and multiple interdependent runways | MIP | Ant Colony Optimization | Minimize weighted earliness and tardiness |
| (Hancerliogullari et al., 2013) | Multiple interdependent runways | MIP | Dispatching rules and Simulated Annealing | Minimize weighted total delay |
| (Salehipour et al., 2013) | Single and multiple interdependent runways | MIP | Simulated Annealing and Variable Neighborhood Search | Minimize weighted earliness and tardiness |
| (Ghoniem et al., 2014) | Single runway | Asymmetric TSP-TW | Dispatching rules | Minimize makespan |
| (Farhadi et al., 2014) | Multiple independent runways | MIP | Dispatching rules and MIP-based heuristics | Minimize weighted total delay |

| Source | Research Scope | Modeling Approach | Solution Approach | Objective Function(s) |
|---|---|---|---|---|
| (Ma et al., 2014) | Single runway | MIP | Ant Colony Optimization | Minimize makespan |
| (Samà et al., 2014) | Two interdependent runways | Decomposition approach | B&B and Rolling Horizon Approach (RHA) | Minimize maximum delay |
| (Furini et al., 2015) | Single runway | MIP position-based | RHA | Minimize weighted total delay |
| (Ghoniem & Farhadi, 2015) | Multiple independent runways | Asymmetric TSP-TW | Column generation | Minimize weighted earliness and tardiness |
| (Lieder et al., 2015) | Multiple independent runways | MIP | Dynamic Programming | Minimize weighted total delay |
| (Ghoniem et al., 2015) | Multiple independent runways | Asymmetric TSP-TW | Branch-and-price | Minimize weighted total delay |
| (Sabar & Kendall, 2015) | Single and multiple interdependent runways | MIP | Iterated local search algorithm | Minimize weighted earliness and tardiness |
| (Samà et al., 2015) | Multiple independent runways | MIP based on job shop scheduling | Mathematical programming | Minimize maximum tardiness and total travel time spent |
| (D'Ariano et al., 2015) | Two interdependent runways | Integrated modeling arrivals and en-route traffic | B&B and greedy heuristic algorithms | Minimize maximum delay |
| (Lieder & Stolletz, 2015) | Single and multiple interdependent runways | MIP | Dynamic Programming | Minimize weighted total delay |
| (Girish, 2016) | Single and multiple independent runways | MIP | Hybrid Particle Swarm Optimization-local search algorithm in an RHA | Minimize weighted earliness and tardiness |

2.1.4 Approaches for Considering Uncertainties

There is substantial uncertainty in airport and runway operations stem from ground speed variations, piloting indecisions, delays in pushback or taxiing, arrival prediction error, airport congestion, flight cancellations, etc. Also, unexpected events, such as safety incidents, equipment failure, inclement weather, increase the uncertainty. As a result, most of the time these uncertainties render the schedules suboptimal or even infeasible that are found with a deterministic approach.

Even though most of the previous research  have been focused on deterministic runway operations scheduling, there are two models in the literature that consider explicitly the uncertainties inherent to runway operations: (1) robust model of NASA Ames Research Center (Chandran & Balakrishnan, 2007; Gupta et al., 2011), and (2) stochastic model of Georgia Institute of Technology (Solveling & Clarke, 2014; Solveling et al., 2011).

NASA Ames Research Center researchers considered a runway schedule as "robust" if there is a high probability that an air traffic controller does not have to interfere once the schedule has been determined. They considered two conflicting objectives: maximizing runway throughput (or minimizing makespan) and maximizing reliability. They only considered the landing problem on a single runway and, therefore, assumed that the separation times satisfy the triangle inequality for all aircraft types. As a solution algorithm, they proposed a DP approach which is computationally efficient enough for a real-time application, which schedules aircraft while limiting the number of positions an aircraft can move from its FCFS position. Their algorithm calculates a trade-off curve between throughput and reliability, which is defined as the probability that random deviations of aircraft from the scheduled landing times that violate operational constraints.

Georgia Institute of Technology researchers addressed the stochastic airport runway scheduling problem in which a set of aircraft are to be scheduled on a single or multiple dependent runways. They developed a two-stage stochastic integer program and a solution method using scenario decomposition based on Lagrangian relaxation. Also, a stochastic B&B algorithm is proposed, which is a sampling-based approach in which the stochastic upper and lower bounds are generated. The proposed models of the stochastic runway scheduling problem correspond to a single machine scheduling problem with probabilistic release times (and due dates) and sequence-dependent setup times.

In both of the robust and stochastic approaches, instead of actual operational distributions, representative probability distributions were utilized for computational experiments. In addition, these approaches were not applied to real-life, large-scale problem instances. The fundamental consequence of such assumptions is that these approaches are not yet mature enough for operational deployment (Mehta et al., 2013). In particular, the effect of uncertainties related to push-back times, wheels-on times and taxi predictions are not taken into account explicitly by the robust approach (NASA Ames Research Center).

On the other hand, the main issue with the two-stage stochastic integer programming (Georgia Institute of Technology) model is that it is not feasible to develop such a model for a problem instance in a practical size since numerous scenarios exist even for a small number of aircraft. Also, it is assumed in this model that the deviation from earliest runway time is independent between aircraft, whereas delay is usually dependent between aircraft in practice. Furthermore, this model assumes that the probability distributions and realizations of runway landing/take-off times are independent; specifically, there is no correlation between aircraft. However, in practice when the arrival and departure rates are high, deviations from scheduled landing/take-off times will presumably be correlated between aircraft, which makes this assumption invalid in actual runway operations.

Consequently, considering the shortfalls of the robust and stochastic optimization approaches, a simulation-based approach seems to be a promising methodology in terms of dealing with real-life, large-scale problem sizes as well as with the impact of dependent runway operations.

2.1.5 Collaborative Decision Making and Fairness

Collaborative Decision Making (CDM) concept is a means to collaborate and share real-time operational information to improve situational awareness and decision-making. It has the potential to improve TMA operations by allowing airlines to participate in air traffic decision-making that affects them. In the implementation of the NextGen, CDM concept is considered important for enhancing operational effectiveness through increased information exchange among stakeholders and consideration of desired intents. Even though final decision-making authority is the Air Navigation Service Provider (ANSP), the involvement of other stakeholders, in particular, airlines have the potential for considerable benefits.

In the CDM context, the air traffic decision-making responsibilities are shared mainly among four distinct stakeholders: the ANSPs, the airport operators, airlines, and government authorities. Each of these stakeholders has different interests. In order to integrate CDM concept into models, interests of these stakeholders have to be taken into consideration, and the key challenge for this is the fact that most of these interests are conflicting and non-commensurable, and it is often hard to aggregate them into one synthetic objective. As an example, for ANSPs, safe flow of air traffic and runway utilization are the primary concerns. On the other hand, airlines are mainly concerned with resource utilization, punctuality, operational costs and on-time performance, etc.

One of the main considerations within CDM concept is that ANSPs need to ensure that the outcome of runway operations scheduling is perceived by all airlines as fair. However, fairness is a significant challenge in terms of clearly defining what is considered as fair by airlines. Roger George Dear (1976) developed a heuristic methodology referred as Constrained Position Shifting (CPS), which limits the number of positions an aircraft can be moved from its FCFS ordered position, to make the scheduling scheme fair. The maximum allowable number of position shifts is determined through a parameter called maximum position shifting (MPS). They examined and tested its effectiveness for several objective functions and concluded that by limiting the MPS to a small number, typically 2 or 3, it is possible to achieve most of the potential benefits.

Roger G Dear and Sherif (1991) and Venkatakrishnan et al. (1993) presented DP algorithms based on CPS. Recently, Balakrishnan and Chandran (2010) developed a shortest path algorithm based on CPS using a discretized network. M. Soomer and Koole (2008) demonstrated various definitions of fairness by using aircraft landing problem, which includes absolute fairness, relative fairness, and fairness measured by the delay. The proposed MIP formulations that include fairness is solved by local search heuristics. Also, computational experiments are conducted to assess how the fairness definitions and solution heuristics behave with real-life problems. The results of these experiments demonstrate that it is possible to attain more fairness while still obtaining considerable cost compared to the FCFS schedule.

Bertsimas and Gupta (2009) formulate three integer programming models that accommodate fairness considerations and demonstrate challenges corresponding to fairness considerations in the solutions obtained from these models. The first model tries to control the total number of pairwise reversals in the resulting order of aircraft landings. The next model attempts to control the difference between airline flight delays. The last model incorporates both notions of fairness proposed in the other two models. After computational experiments where national-scale, real-life datasets are used, it is concluded that last model is capable of fulfilling both notions of fairness at a less than 10 percent increase in total delay costs. Y. Wang et al. (2012) proposed a fairness definition which considers the historical fairness information for aircraft landing problem. They consider fairness as the average affected additional cost of an airline.

Although CDM concept's main focus is on information exchange among stakeholders of the air traffic flow management to enhance shared situational awareness, fairness is an integral part of CDM processes for all stakeholders. However, recent studies are short of fairness considerations, especially in scheduling runway operations. Although there exist several different fairness definitions and notions, the most commonly embraced definition of fairness in terms of runway operations scheduling is the "sequence equity" among aircraft, which limits position shifts from the FCFS sequence.

## 2.2 Machine Scheduling under Uncertainty

Although separation requirements for take-off are substantially more complex than landing problem in practice, both segregated or mixed-mode operation problems can be modeled as a machine scheduling problem with asymmetric sequence-dependent setup times. Commonly used objective functions for this problem are minimizing the makespan (maximizing throughput) and minimizing the total weighted tardiness. It is worth to mention that the single machine scheduling problem can be transformed into a traveling salesman problem, if the objective function is makespan minimization.

Deterministic machine scheduling problems with sequence-dependent setup times are studied extensively in the literature. However, the same problem under uncertainty has not received much attention. Skutella and Uetz (2005) considered identical parallel machine scheduling problems, where the processing times of jobs are ruled by independent probability distributions. Their model's objective function is to minimize the expected value of the total weighted completion time.

Cai and Zhou (2005) studied a single machine scheduling problem, where each job has a random processing time, a general stochastic cost function, a random due date, and a weight value. The processing times are assumed to be exponentially distributed, whereas the stochastic cost functions and the due dates are assumed to follow any distribution. The objective is to minimize the expected sum of the cost functions.

Anglani et al. (2005) proposed a robust approach for solving the parallel machine scheduling problem with sequence-dependent set-up costs. They formulated a fuzzy mathematical programming model by considering the uncertainty in processing times to provide the optimal solution as a trade-off between total set-up cost and robustness in demand satisfaction.

Wu and Zhou (2008) addressed single machine scheduling problem with random due dates to minimize the expected maximum lateness. They first developed a deterministic equivalent to the expected maximum lateness, and then, proposed a DP algorithm to obtain the optimal solutions. However, sequence-dependent setup times are not included in the model.

Contrary to the literature on machine scheduling under uncertainty, the modeling structure in runway operations scheduling problem relates to machine scheduling models with probabilistic release times with sequence-dependent setup times, which have received very limited attention in the literature.

## 2.3 Mathematical Modelling Approaches

Several methods proposed in the literature for modeling the runway operations scheduling problem, which can be considered as a three-stage process. First, aircraft landings and take-offs have to be allocated to the available runways; next, the sequence of aircraft for each runway has to be determined; and then, start times of runway operations have to be determined. Depending on the objective(s) of the problem one of these stages are more important. For instance, only the allocation stage is important for the makespan minimization objective (completion time of the last aircraft take-off or landing which implies the runway utilization). The principal mathematical programming formulations and objective functions are identified below.

Considering its similarities with production scheduling problems, multiple runway operations scheduling problems can be viewed as an identical parallel machine scheduling problem where "aircraft" and "runways" represent "jobs" and "machines," respectively. Classical parallel machine scheduling problem consists of assigning a number of jobs on a set of parallel machines where the release time, start time and latest finish time for each job and sequence-dependent setup times are given. These setup times include the activities depending on both the job to be processed and the immediately preceding job. The mapping of the identical parallel machine scheduling to runway operations scheduling problem relies on the following assumptions: (1) if an aircraft begins to land or take-off, it cannot be interrupted by another aircraft, (2) at most one aircraft is allowed to land on or take-off from each runway at any time, (3) runways are available and reliable at all times, (4) any aircraft can land on or take-off from at most one runway at any time, and (5) all input data are known with certainty.

In the literature, the three-term notation, $\alpha \mid \beta \mid \gamma$, is commonly adopted, which is proposed by Graham et al. (1979), as the classification scheme for scheduling problems. In three-term notation, $\alpha$ indicates the machine environment; $\beta$ describes the job and the resources characteristics, and $\gamma$ defines the objective function to be minimized. As a result, the runway operations scheduling problem is denoted by $P_m/s_{ij},tw/\Sigma w_j T_j$ where $P_m$ denotes the parallel machine scenario; $s_{ij}$, denotes the sequence-dependent times between aircraft $i$ and $j$, respectively; $tw$ denotes the time windows, and the objective is to minimize total weighted tardiness costs.

Total weighted tardiness is a widely used performance measure in parallel machine scheduling problems. Tardiness is commonly defined as the amount of time by which a job's completion time exceeds its target time. In the tardiness problem, there is no benefit from completing jobs early, and the delay penalty is proportional to the length of the delay and the weight associated with each job, which refers to a late penalty for an individual job in case this job is tardy.

Even the problem of single machine scheduling with total tardiness objective function is proved to be *NP*-Hard, i.e., it is unlikely that there can be developed a polynomial-time algorithm for finding an optimal schedule. The computational complexity of the identical parallel machine scheduling problem with total weighted tardiness objective function is still *NP*-Hard due to its combinatorial nature. Since exact algorithms require long computation times, different heuristics and metaheuristics are commonly employed to find near optimal values in a shorter amount of times. Therefore, this justifies the use of heuristic and metaheuristic methods over exact methods for solving the runway operations scheduling problem.

The literature presents three mathematical programming formulations for the problem, which include a 0-1 MIP formulation, a set partitioning formulation and an asymmetric traveling salesman problem with time windows (TSP-TW) formulation. Before presenting these formulations, the notation used throughout the formulations is given below.

Notation:

| | | |
|---|---|---|
| $M$ | : | set of $m$ independent runways, $M=\{1,2, ..., m\}$ |
| $N$ | : | set of $n$ aircraft, $N=\{1,2, ..., n\}$ |
| $P$ | : | set of all feasible columns |
| $i, j$ | : | aircraft indices |
| $r$ | : | runway index |
| $p$ | : | column (sequence of aircraft) index |
| $r_j$ | : | ready time for aircraft $j$ |
| $\delta_j$ | : | target time for aircraft $j$ |
| $d_j$ | : | due time for aircraft $j$ |
| $O_j$ | : | operation type of aircraft $j$ |
| $C_j$ | : | class of aircraft $j$ |
| $w_j$ | : | weight value assigned to aircraft $j$ based on its operation type and class |
| $s_{ij}$ | : | separation time between aircraft $i$ and $j$ |
| $a_j^p$ | : | 1 if aircraft $j$ is covered by column $p$, 0 otherwise |

Decision Variables:

| | | |
|---|---|---|
| $t_j$ | : | start time of aircraft $j$ |
| $T_j$ | : | piecewise tardiness of aircraft $j$ with respect to $\delta_j$ |
| $z_{jr}$ | : | 1 if aircraft $j$ is assigned to runway $r$, 0 otherwise |
| $y_{ij}$ | : | 1 if aircraft $i$ and $j$ are assigned to the same runway and $t_j > t_i$ ($\forall i, j \in N, i \neq j$), 0 otherwise |
| $x_p$ | : | 1 if column $p$ is involved in the solution, 0 otherwise |
| $v_{ij}$ | : | 1 if aircraft $i$ directly precedes aircraft $j$ ($\forall i, j \in N, i \neq j$), 0 otherwise |
| $v_{0i}$ | : | 1 if aircraft $i$ is first in the sequence ($\forall i \in N$), 0 otherwise |
| $v_{i0}$ | : | 1 if aircraft $i$ is last in the sequence ($\forall i \in N$), 0 otherwise |

The 0-1 MIP formulation that is given below for the runway operations scheduling problem, which is commonly referred as Beasley model in the literature, is based on the formulation presented in Beasley et al. (2000). The main difference with the original formulation is that two auxiliary binary variables, one related to precedence on the same runway and the other related to whether aircraft pair assigned to the same runway, are merged into one.

$$min. \quad \sum_{j \in N} w_j T_j \tag{2.1a}$$

$$s.t. \quad \sum_{r \in M} z_{jr} = 1 \qquad \forall \, j \in N \tag{2.1b}$$

$$1 \leq \sum_{j \in N} z_{jr} \leq \left\lceil \frac{n}{m} \right\rceil \qquad \forall r \in M \tag{2.1c}$$

$$r_j \leq t_j \leq d_j \qquad \forall j \in N \tag{2.1d}$$

$$t_j \geq t_i + s_{ij} - (1 - y_{ij})(d_i - r_j + s_{ij}) \qquad \forall i, j \in N, i \neq j \tag{2.1e}$$

$$y_{ij} + y_{ji} \geq z_{ir} + z_{jr} - 1 \qquad \forall r \in M, \forall i, j \in N, i \neq j \tag{2.1f}$$

$$T_j \geq t_j - \delta_j \qquad \forall \, j \in N \tag{2.1g}$$

$$0 \leq T_j \leq d_j - \delta_j \qquad \forall \, j \in N \tag{2.1h}$$

$$z_{jr}, y_{ij} \in \{0,1\} \qquad \forall r \in M, \forall i, j \in N \tag{2.1i}$$

where the objective function (Eq. 2.1a) is to minimize the total weighted tardiness. Constraints in Eq. 2.1b ensure that each aircraft land on or take-off from exactly one runway. Constraints in Eq. 2.1c are load balancing constraints that enforce lower and upper bounds on the number of aircraft. Constraints (Eq. 2.1d) guarantee that each aircraft land or take-off within its time windows. Constraints (Eq. 2.1e) ensure required separation times between any pair of aircraft. Constraints in Eq. 2.1f actuate the sequencing variables between any pair of aircraft that are assigned to the same runway. With the help of constraints in Eq. 2.1f, constraints in Eq. 2.1e enforce separation only between aircraft that are assigned to the same runway. Constraints (Eq. 2.1g) specify aircraft tardiness, with respect to target times. Constraints in Eq. 2.1h enforce non-negativity restrictions and upper bounds on aircraft tardiness. Constraints (Eq. 2.1i) define binary decision variables.

The alternative mathematical programming formulation for the problem is a set partitioning model. The set partitioning model aims to partition all elements into a number of subsets, and each binary variable (column) represents a subset of elements defined by the coefficients. In this formulation, each column $p$ represents a feasible sequence of aircraft with an aggregated cost. The set partitioning formulation can be stated as follows:

$$min. \quad \sum_{p \in P} \left( \sum_{j \in N} w_j t_j a_j^p \right) x_p \tag{2.2a}$$

$$s.t. \quad \sum_{p \in P} a_j^p x_p = 1 \quad \forall j \in N \tag{2.2b}$$

$$\sum_{p \in P} x_p = m \tag{2.2c}$$

$$x_p \in \{0,1\} \tag{2.2d}$$

where the objective function (Eq. 2.2a) minimizes the total weighted tardiness. Constraints in Eq. 2.2b, which are the set partitioning constraints, ensure that each aircraft is assigned to exactly one runway. The constraint in Eq. 2.2c guarantees the limit on the number of the runways and constraints in Eq. 2.2d are the integrality constraints on the decision variables $x_p$. Set partitioning problem is one of the first problems shown to be *NP*-Hard; therefore, no polynomial time solution algorithm is likely to exist also for this formulation.

There are two major challenges related to the set partitioning model which are outlined below:

(a)     The number of binary variables corresponding to feasible sequences of aircraft usually reaches into millions for most real-life applications, which eventually renders the model computationally intractable.

(b)     Each column $p$ does not give the information related to the order of aircraft in that sequence. For this reason, it is computationally impractical to enumerate all the columns for large-scale problems.

The third formulation, which is an asymmetric TSP-TW formulation is based on Ghoniem et al. (2014) and given below. The major difference from the 0-1 MIP formulation (Beasley model) is the introduction of the binary variable $v$ to build the tours. This decision variable enforces the sequential precedence order of the aircraft on each runway.

$$min. \quad \sum_{j \in N} w_j T_j \tag{2.3a}$$

$$s.t. \quad \sum_{r \in M} z_{jr} = 1 \qquad \forall \, j \in N \tag{2.3b}$$

$$1 \leq \sum_{j \in N} z_{jr} \leq \left\lceil \frac{n}{m} \right\rceil \qquad \forall r \in M \tag{2.3c}$$

$$r_j \leq t_j \leq d_j \qquad \forall j \in N \tag{2.3d}$$

$$\sum_{j \in N-\{i\}} v_{ij} = 1 \qquad \forall i \in N \tag{2.3e}$$

$$\sum_{i \in N-\{j\}} v_{ij} = 1 \qquad \forall \, j \in N \tag{2.3f}$$

$$\sum_{i \in N} v_{i0} = m \tag{2.3g}$$

$$\sum_{i \in N} v_{0i} = m \tag{2.3h}$$

$$t_j \geq t_i + s_{ij} - (1 - v_{ij})(d_i - r_j + s_{ij}) \qquad \forall i,j \in N, i \neq j \tag{2.3i}$$

$$t_j \geq t_i + s_{ij} - (1 - y_{ij})(d_i - r_j + s_{ij}) \qquad \forall i,j \in N, i \neq j \tag{2.3j}$$

$$y_{ij} + y_{ji} \geq z_{ir} + z_{jr} - 1 \qquad \forall r \in M, \forall i,j \in N, i \neq j \tag{2.3k}$$

$$T_j \geq t_j - \delta_j \qquad \forall \, j \in N \tag{2.3l}$$

$$0 \leq T_j \leq d_j - \delta_j \qquad \forall \, j \in N \tag{2.3m}$$

$$z_{jr}, \, y_{ij}, \, v_{ij} \in \{0,1\} \qquad \forall r \in M, \forall i,j \in N \tag{2.3n}$$

where the objective function (Eq. 2.3a), the constraints Eq. 2.3b-3d and Eq. 2.3j-3n are the same as the 0-1 MIP formulation. The assignment constraints in Eq. 2.3e-3f guarantee that each aircraft to be followed by, and to follow, exactly one aircraft. Constraints in Eq. 2.3g and 3h ensure that exactly one aircraft will be assigned to the first place in the sequence, and one to the last place for all runways, respectively. Constraints in Eq. 2.3i actuate the sequencing variables between any pair of aircraft that are assigned to the same runway.

The triangle inequalities are not violated by the separation requirements when only landing or take-off operations are considered. The main advantage of this is that all separation requirements can be met by only taking into account the separation times between consecutive operations. The triangle inequality for given $i$, $j$, and $k$ weight classes is shown below:

$$s_{ik} \leq s_{ij} + s_{jk} \qquad (2.4)$$

The key issue with the triangle inequalities that needs to be addressed is that in mixed-mode operations on the same runway or close parallel runways the triangle inequality does not always hold.

*Objective functions*: There are various stakeholders in scheduling the aircraft landings and take-offs on runways, and each has different objectives. The most important stakeholders and their desirable objectives are listed below, and these desirable objectives are given in a mathematical form in Appendix B:

(a)     *Air navigation service providers (ANSPs) or air traffic controllers*: They are mainly responsible for flight safety and runway utilization, and they execute the actual scheduling process. Their key objectives are maximizing runway throughput as possible, minimizing landing and take-off delay.

(b)     *Airlines*: They are typically concerned with operational costs, on-time performance and reputation (goodwill). They have a preference on minimizing delay and fuel cost, minimizing total passenger delays, maximizing fairness.

(c)    *Airport management*: They are commonly focused on the smooth flow of traffic both in air-side and land-side of the airport. Their main interests are maximizing punctuality relative to the operating schedule and minimizing the need for gate changes due to delays.

(d)    *Government agencies*: They primarily deal with reducing environmental effects, such as noise disturbance, air pollution, etc. Therefore, their main objective is minimizing environmental effects.

It is noteworthy to mention that delay is typically considered as the deviation of actual landing/take-off time from the estimated landing/take-off time calculated by the FCFS principle instead of the planned aircraft schedule. From air traffic controllers' point of view, throughput and average delay are important objective functions, while from airlines' point of view, the operating costs, especially fuel costs, are important (Hanbong & Balakrishnan, 2008). Runway throughput is commonly defined as the number of aircraft landings or take-offs during a specific time (usually an hour) that an airport's runways able to sustain during periods of high demand.

## 2.4 Alternative Approaches for Optimization under Uncertainty

By and large, there are two primary approaches to deal with uncertainty in a scheduling environment: proactive and reactive scheduling. Proactive scheduling involves predictive schedules that account for statistical knowledge of uncertainty. On the other hand, reactive scheduling comprises of rescheduling or recovering the schedule when an unpredicted events occur (Figure 6).

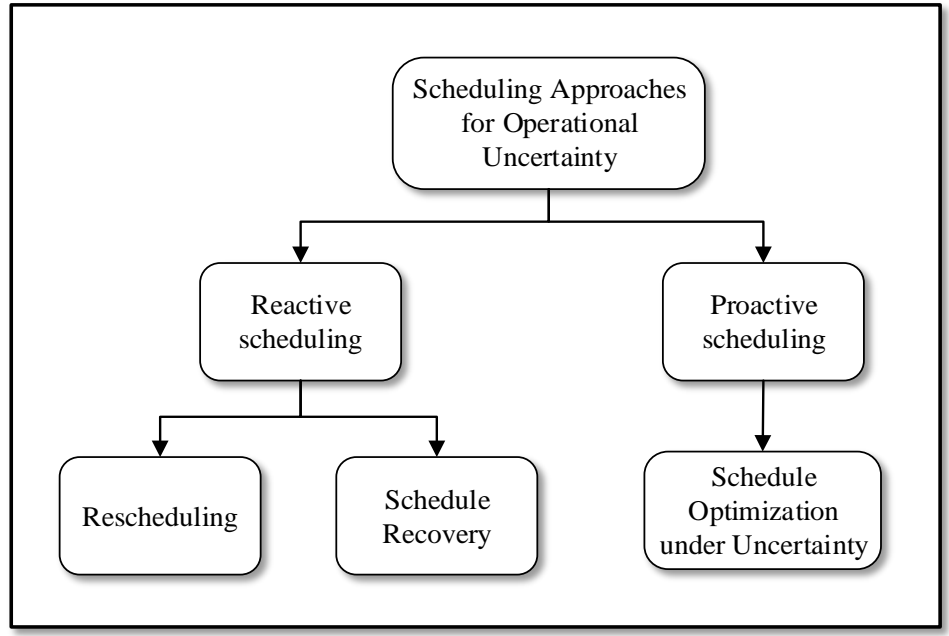

**Figure 6:** Scheduling Approaches for Operational Uncertainty

In reactive scheduling the schedules are re-optimized after disruptions occur as a result of uncertainties by rescheduling from scratch or recovering the schedule as possible. The main drawback of this approach is the fact that a solution is expected immediately. In addition, they require considering several complex constraints at the same time, which typically lead to intractability issues. Because of the need of an immediate response, practitioners currently employ manual or heuristic methods that yield sub-optimal solutions. On the other hand, in proactive scheduling the uncertainty is integrated into the schedules during schedule optimization stage so that schedule adjustments can be made more easily and more conveniently. While reactive scheduling tries to find the most appropriate way to react to disruptions on the day of operations, the proactive approach seeks to consider the effects of uncertainties in advance. Thus, schedules that are robust and less susceptible to effects of uncertainties could be generated with optimization.

Regarding runway operations scheduling in practice, existing operational planning models and decision support tools depend on frequent rescheduling to deal with uncertainty with the expectation that updating the schedules repeatedly with the latest data will mitigate the negative impacts of uncertainty. However, this approach is not proved to provide considerably high-quality and computationally efficient solutions. Therefore, uncertainty should be considered explicitly during optimization with one of the alternative approaches for optimization under uncertainty.

Three main alternative modeling approaches currently exist that consider uncertainty in schedule optimization, which are briefly explained below:

*Stochastic optimization*: The most widely applied stochastic programming model to address schedule optimization under uncertainty is the two-stage linear stochastic programming, where the objective functions and the constraints are both assumed to be linear. The basic idea of this model is that decisions should be based on data available at the time the decisions are made, and these decisions should not depend on future observations. In this model, the decision-maker takes some action in the first stage, after which a random event occurs affecting the outcome of the first stage decision. Then, a recourse decision can be made in the second stage that compensates for any bad effects that might have been experienced as a result of the first stage decision. This method necessitates probability distributions of random variables or sample approximate methods that need to be employed to account for the randomness (Shapiro & Philpott, 2007).

To utilize stochastic optimization approach, uncertain data needs to be considered as a second stage data that can be modelled as a random (not just uncertain) variable with a known probability distribution. The main drawback of this approach is the difficulty in measuring the quality of obtained solutions and finding an optimal solution in a reasonable time for large-scale real-life problems.

*Robust optimization*: The basic idea of robust optimization is based on the fact that while individual uncertainties may be unpredictable as the number of random variables grows large, uncertainties tend to cancel out, and the averages can be predicted fairly well (Ben-Tal & Nemirovski, 2000). In this approach, there is no need to know the exact probability distributions of the uncertain data. This situation will potentially result in increased tractability properties, and also eliminate the need to assign such distributions artificially to the random quantities (Bertsimas & Sim, 2004). The main challenge associated with achieving robust solutions is deciding what constitutes robustness and how to capture these in the optimization model. One possible approach may be to make an aircraft schedule robust by using a metric, such as total expected delay, etc. This metric can be calculated based on historical data, and this will also allow uncertainty to be modeled in the objective function.

*Simulation-based optimization (SbO)*: The SbO approach is a form of optimization in which simulation experiments are used to find the optimal values for the decision variables (Fu, 2015). In this approach, the simulation model functions as the external evaluator of the objective functions that are to be optimized by the optimization algorithm. This approach is detailed in Chapter 3.

Each of these alternative approaches has strengths and weaknesses, and each has its own assumptions. The primary strength of stochastic optimization is the explicit incorporation of uncertainty into the optimization model. The main assumptions of this approach are that the underlying probability distributions of the uncertain parameters have to be known, and these distributions will not change over the considered planning horizon. If these assumptions are met and the stochastic model is tractable, then stochastic optimization is the right optimization approach for applying.

Although robust optimization is computationally more tractable than stochastic optimization, it is more conservative, since it is worst-case-oriented. The main assumptions of this approach are that the constraints of a given problem must be satisfied for all realizations of the uncertain parameters in a so-called uncertainty set, and these uncertainty sets are readily available, which are typically derived by expert opinion or using historical data.

In the context of multi-objective runway operations scheduling problem, the most promising methodology seems to be the SbO approach. Stochastic optimization is not a suitable methodology since it introduces many additional variables which eventually renders the model computationally intractable for real-life problem sizes. Robust optimization is also not an appropriate methodology to pursue because it generates too conservative solutions which are usually too far from Pareto-optimal solutions. Because of its flexibility and effectiveness, a SbO approach is applied to the multi-objective runway operations scheduling problem in this research.

## 2.5 Summary of Findings and the Knowledge Gaps

From the comprehensive literature survey presented above, the following findings may be asserted:

(a)     Recently scheduling researchers and practitioners have been devoted more attention to the runway operations scheduling problem. Most of the studies in the literature consider only a single runway or a single type of operation; however, it is usually not possible to extend them to apply multi-runway and mixed-mode situations. Although multiple runways and mixed-mode operations introduce additional complexity into the problem, there is a potential to obtain operational benefits, such as increasing the efficiency of the runway operations and decreasing the delays and, in turn, reducing delay related operational and environmental costs.

(b)     The literature review demonstrates that from both a mathematical and practical perspective, multiple runway operations scheduling is a challenging problem even in the deterministic context. The computational complexity of the problem is classified as *NP*-Hard for almost all of its configurations and instances. Therefore, a polynomial time exact algorithm for this scheduling problem is very unlikely to exist. As a result, most of the research conducted on the problem concentrated on using heuristics and metaheuristic algorithms to obtain acceptable solutions to the real-life instances of the problem promptly. In the literature, the exact methodologies have been used as a reference to benchmark the performance of heuristic methodologies rather than as a practical method applicable to real-life problem instances.

(c)     There have been numerous publications based on research completed in this particular field; however, very little research has been undertaken to solve the problem under uncertainty conditions in a timely manner. Nearly all of the methods related to aircraft scheduling on runway proposed in the literature assume that all the input data are known with certainty. In addition, there are existing models that consider uncertainty for machine scheduling problems; however, they also do not take into account all the problem characteristics involved in runway operations scheduling problem.

(d)     Most of the proposed algorithms in the literature have not been tested and validated using real-life, large-scale datasets as a proof-of-concept. They mostly tried to be validated against benchmark problem instances that are available in the literature, which are not representative of practical problem instances.

(e)     Most of the published research have considered the problem with respect to a single performance measure (objective function). Commonly employed ones include total delay and runway throughput. However, the focus should be on finding the trade-off solutions between conflicting objectives that reflect various stakeholders' interests.

(f)     A number of analytical and simulation models have been developed for modeling runway operations in the TMA. The analytical models are typically macroscopic in nature while most of the simulation models are tend to be microscopic. Cost-effective solutions can be obtained promptly by utilizing analytical models. However, when model details influence solutions, simulation models provide a more suitable mechanism for handling system complexity. Also, simulation models are capable of generating various scenarios related to uncertainty and examining the impacts of delays. Therefore, a simulation-based methodology is much more appropriate in addressing the complexity of the runway operations, which is not adequately captured when using analytical models.

(g)     Stochastic optimization approaches are only suitable to solve problems where the scale of the problem is relatively small compared to problems emerge in real-life. Although a two-stage stochastic programming approach for addressing runway operations scheduling problems have been proposed, the main drawback of this approach is that it is very computationally expensive such that it is not applicable to real-life problem sizes. Another drawback of this approach is the fact that it is not capable of taking into account the dynamic nature of scheduling environment, and for the sake of tractability some limiting assumptions are required related to probability distributions in the model, which might not be true in actual runway operations. By the same token, robust optimization approaches generate too conservative solutions considering the worst-case possibilities of the uncertain data, which result in solutions far from optimal. On the other hand, SbO approaches are more computationally efficient in handling large-scale problems that explicitly consider uncertainties inherent to practical runway operations.

Given the above findings gleaned from the literature review, there is a requirement for a model that brings the runway operations scheduling problem closer to practice. Previously proposed deterministic models are so simplified that they are not capable of capturing all relevant aspect of the actual problem. The complexity of the problem increases with the extra restrictions and conditions added to the problem. Two aspects that increase the complexity of the problem significantly is the consideration of uncertainty and multiple conflicting objectives to be optimized, which is essential for closing the gap between academic research and practice related to the problem.

Uncertainty is inevitable in runway operations schedule optimization and can significantly degrade the performance of an optimized solution or even render it infeasible. There are various factors that cause uncertainty in runway operations, the most notable of these factors are ground speed variations caused by the wind, piloting indecisions, delays in pushback or taxiing, arrival prediction error, airport congestion, flight cancellations, etc. Also, unexpected events such as safety incidents, equipment failure, inclement weather, etc. also contribute to uncertainty. All these sources of uncertainty can result in variability in landing/take-off sequence and/or time windows (earliest, latest possible target landing/take-off times or target times) (C. Brinton & Atkins, 2009). Therefore, due to these sources of uncertainty, deviations from the estimated input parameters are unavoidable.

As a conclusion, based on the literature review and to the best of our knowledge, there is no work reported that deals with the runway operations scheduling problem under uncertainty with utilizing a simulation-based approach. Also, fairness is not taken into consideration during the optimization process as a second objective along with runway utilization, which converts the problem to a bi-objective optimization problem. Although there exist several approaches for modeling this real-life scheduling problem, it is evident that these different modeling approaches force different solution methods and resulting in inefficiencies in the schedules and computational challenges. However, a simulation-based optimization (SbO) approach seems to be the most promising one because of its capability in dealing with the stochastic and dynamic nature of this scheduling problem. It also considers uncertainty explicitly such a way that it is capable of reducing the negative impact of randomness on the optimized solutions and increasing the reliability of these solutions in practice.

# CHAPTER 3

## SIMULATION-BASED OPTIMIZATION FRAMEWORK

Simulation studies are conducted to determine an estimate of the system output from a set of system input configuration. However, the simulation does not include the capability to search for a set of system inputs that can produce the optimal or near-optimal system output. Hence, an optimization procedure needs to be incorporated into simulation models to empower it with an optimization capability. Numerous cases have been reported in the literature in which simulation and optimization methods were combined successfully, and these efforts extended the growth of research in the field of simulation-based optimization (SbO).

SbO approaches try to determine the exact combination of system parameters which produce the optimal or near-optimal performance measures. The main strength of SbO approaches is that they can consider the dynamic and stochastic nature of the real-life problem while optimal or near-optimal solutions can be obtained without much of computational intractability. However, these methods still face challenges especially when there exist multiple and conflicting objectives, and they typically require costly development and challenging verification & validation process. Due to the fact that simulation is not an optimization tool, in essence, simulation experiments require to be designed in a systematic way for analysts to understand the simulation model's behavior.

This chapter presents what is to the best of our knowledge the first SbO approach to solve runway operations scheduling problem under uncertainty. First of all, basic concepts, terminology and a concise classification of the SbO are presented for use throughout the rest of the dissertation. Also, such concepts, terminology and taxonomy help to define and clarify the framework for the SbO approach we have proposed and justify the decision to move into that direction. Afterwards, the principles of multi-objective optimization (MOO) are outlined and basic concepts are formally defined. This is followed by a discussion on simulation-based multi-objective optimization. Finally, an overview of the SbO framework is presented, and each component is outlined to provide a high-level understanding of the whole methodology.

## 3.1 Simulation-based Optimization

Simulation can be defined as the process of designing an abstract model of a real-life system and conducting experiments with this model for either understanding system behavior or evaluating various strategies within the limits imposed by a set of criteria for the operation of the system (Shannon, 1975). Simulation is usually considered as an effective performance evaluation tool especially when the real system is stochastic and dynamic in nature, and too complex to be presented by mathematical terms. However, despite its capabilities, simulation retains its downsides when it is used individually to deliver the optimal schedule. In order to mitigate this shortcoming, simulation and optimization need to be integrated. Integration of simulation models with optimization procedures, which is commonly referred as a simulation-based optimization (SbO), has the potential to enable more realistic modeling and produce more robust solutions. From an optimization point of view, SbO tries to find a set of decision variable values that optimize (minimize or maximize) an objective function that is estimated by a simulation component.

In recent years, SbO has been an active area of research with many important applications, such as planning, scheduling, etc. Also, it has become the method of choice for optimizing complex models due to the following advances in simulation technologies: (1) latest developments in computational capabilities, modeling paradigms, software and techniques for developing simulation models, and (2) latest improvements in the methods and the software for statistical design and analysis. As a result, it has been increasingly utilized in practice and incorporated into commercial and non-commercial simulation packages. For example, in Arena simulation software package, optimization of simulation is carried out by using OptQuest package, which uses Scatter Search, Tabu Search, and Neural Networks (F. Glover et al., 1999).

### 3.1.1 Basic Concepts of Simulation-based Optimization

Many real-life problems, including runway operations scheduling problem, involve uncertainty and their solutions highly depend on these uncertain input parameters. The constructed models of such problems are stochastic in nature. Simulation is an indispensable operations research method for analyzing such problems. One of the limitations of simulation models, in general, is that they act as a black-box system such that they can only evaluate the model for the decision variables that are pre-specified. Therefore, to use a simulation model for evaluating the performance of a process, the values of decision variables need to be set, and then, a simulation run needs to be conducted to forecast the performance of that particular input parameter configuration. Manually adjusting these input parameter configuration values is not practical due to the combinatorial nature of the process.

Moreover, it is often not clear how to adjust the decision variables from one simulation run to the next. In such cases, finding an optimal solution for a simulation model requires searching in a heuristic or ad hoc fashion. This situation usually involves running a simulation for an initial set of decision variables, analyzing the results, changing one or more variables, running the simulation again, and repeating this process until a satisfactory solution is obtained. As mentioned above, the simulation itself can not automatically adjust the decision variables so as to reach an optimum solution. This issue was one of the main problems of simulation which left large-scale models unresolved in the past (Law, 2014).

Until the last two decades, the two operations research methods - optimization and simulation - were kept largely separated in practice, even though there was a large body of research literature relevant to combining them (Fu, 2002). However, recent developments in both disciplines already herald a commonality between these two distinct disciplines. Moreover, in time, it already became a necessity to integrate optimization techniques into simulation practice. In the last couple of decades, such cooperation appeared between well-known optimization routines and simulation software packages. SbO is an invaluable tool when there is a need for a systematic and efficient method to find which of a large number of system configurations leads to an optimal or near-optimal value for an output performance measure (Law, 2014).

### 3.1.2 Simulation-based Optimization Terminology

Simulation-based Optimization (SbO) is the process of combining different input parameter values that can be controlled to find the combination that provides the most desirable output from the simulation model. In SbO terminology, different keywords for the terms related to inputs and outputs are used. They all express the same meaning either intentionally or inadvertently. The terms related to the inputs and outputs of a SbO problem, which are defined in Fu (2002), are as follows:

(a)     Inputs are referred as (controllable) parameter settings, values, variables, (proposed) solutions, designs, or configurations.

(b)     Outputs are referred as performance measures, or criteria (see Figure 7).

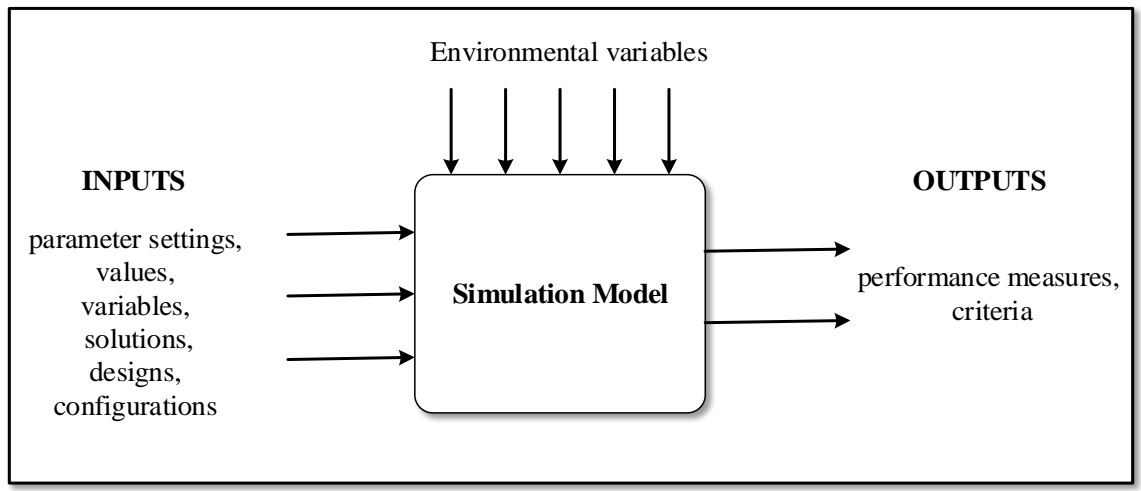

**Figure 7:** Simulation-based Optimization Terminology

In statistical terms, the input parameters typically refer to "factors", and the output performance measures refer to "responses". In the area of optimization, the factors turn into "decision variables" and the responses are used to model an "objective function" and "constraints". The SbO process seeks to find the combination of factor levels that minimizes or maximizes a response subject to constraints imposed on factors and/or responses.

The input parameters of the real system are set to the "optimal" parameter values determined by the SbO process, rather than in an ad hoc manner based on qualitative insights gained from exercising the simulation model. A very general formulation of the SbO problem is to minimize the expected value of the objective function with respect to its constraint set. Therefore, a SbO problem can be formulated as a classical mathematical optimization model as follows:

$$min \ z = \quad \mathbb{E} < f(x) > \qquad \text{(Objective function)} \qquad (3.1a)$$

$$s.t. \quad \boldsymbol{A}x \leq \boldsymbol{b} \qquad \text{(Constraints on input variables)} \qquad (3.1b)$$

$$g_l \leq G(x) \leq g_u \qquad \text{(Constraints on output measures)} \qquad (3.1c)$$

$$l \leq x \leq u \qquad \text{(Upper and lower bounds)} \qquad (3.1d)$$

where the objective function (Eq. 3.1a) represents the expected value of a key output performance measure obtained from the simulation model, and it is a mapping from a vector $\boldsymbol{x}$ of decision variables to a real value. The inequality in Eq. 3.1b represents the constraints, where both the coefficient matrix $\boldsymbol{A}$ and the right-hand-side values corresponding to vector $\boldsymbol{b}$ are known. The inequality in Eq. 3.1c represents the constraints enforce simple upper and/or lower bound requirements on an output function $G(\boldsymbol{x})$, where the values of the bounds are known constants. As imposed by Eq. 3.1d, all decision variables ($\boldsymbol{x}$) are bounded, and some may be restricted to be discrete. Each evaluation of the objective function and $G(\boldsymbol{x})$ requires an execution of a simulation of the system.

### 3.1.3 Simulation-based Optimization Methods

There are several optimization methods to be used in a SbO framework; hereafter these methods will be referred to as SbO methods. Depending on their various criteria SbO methods are classified in a number of different ways in the literature (Carson & Maria, 1997), (Azadivar, 1999), (Fu, 2002), (L.-F. Wang & Shi, 2013). There is a gap for a classification which covers the full spectrum of the approaches and launches the discussion on the different strategies. Since the possibilities of linking an optimization method with a simulation model are so vast, it is very essential to have a good overview of the different approaches. A possible classification for SbO methods, based on the structure of the problem and search scheme is shown in Figure 8. These methods are first partitioned depending on the criteria type of the variables (discrete or continuous).

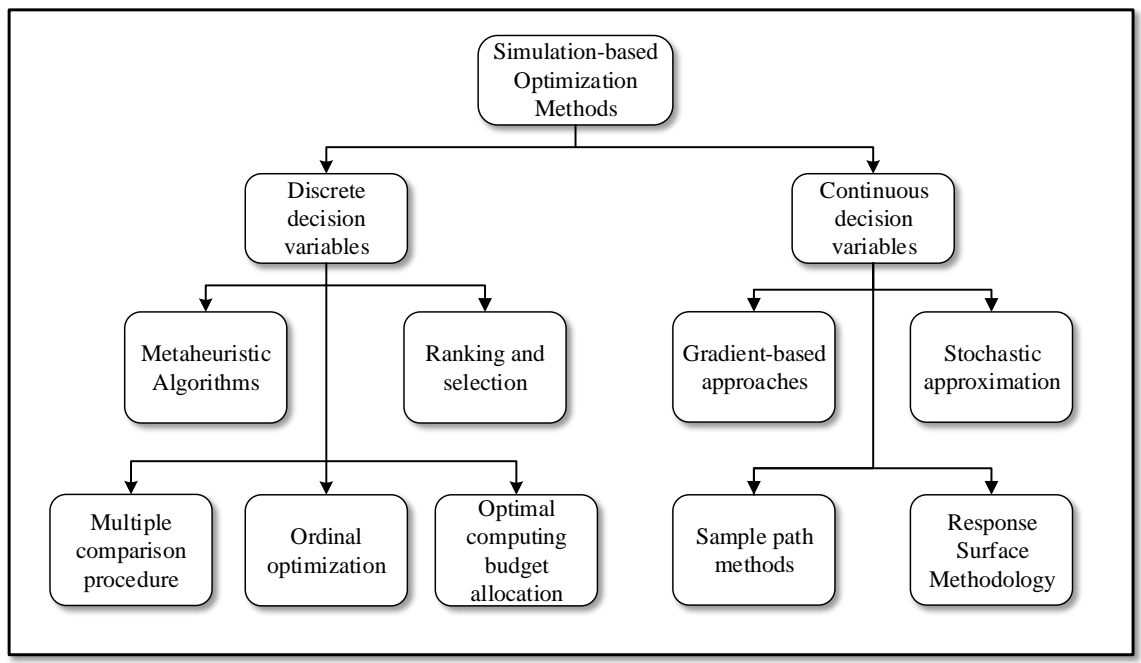

**Figure 8:** Classification of Simulation-based Optimization Methods

If the decision variable of the optimization problem is continuous, then it may be appropriate to utilize a method based on a metamodel, or a gradient method to search the solution space in an iterative way. If the decision variable is discrete and the solution space is large, then metaheuristic algorithms are the most suitable ones. Otherwise, if the solution space is small, statistical selection methods, such as ranking & selection method, are the best choice; hence, parameter inputs to be comprehensively examined and what-if questions answered. Due to their common utilization in the literature, some of the SbO methods, in particular, response surface methodology, metaheuristic algorithms, and stochastic approximation, are briefly outlined below:

*Response Surface Methodology (RSM)*: The basic idea of RSM is to construct a metamodel to approximate the objective function based on a set of input decision variables. A metamodel, also referred to as a surrogate model, is a mathematical model that approximates the response space (solution space) of a simulation model and is used to provide information concerning the system without the need for costly simulation runs. By utilizing a metamodel, the objective function can be evaluated efficiently for each parameter setting, since optimization procedures are executed over the metamodel instead of the objective function that is expensive to evaluate. The two most widely used techniques for obtaining the metamodel are regression and neural networks. Regression, as a statistical technique, summarize how the simulation model's output reacts to changes in the model's input. On the other hand, neural networks act as a screening device to eliminate points where the objective function value is predicted to be low-quality by the neural networks model without actually searching additional iterations.

*Metaheuristic Algorithms*: The main advantages of metaheuristic algorithms are that they do not require any gradient information, and they are not problem dependent. Despite these advantages, these algorithms still possess drawbacks. These methods typically suffer from local optimality, and they pose several challenges in tuning parameters and dealing with multiple objectives. Chapter 5 presents a detailed treatment of these algorithms along with a brief explanation of the most popular ones.

*Stochastic Approximation*: These methods are gradient-based, and they typically divided into two categories: (1) techniques that are based on direct gradient estimation techniques, and (2) techniques that are based on indirect gradient estimation. The main challenge in these methods is that a large number of iterations is required before obtaining the optimum value.

Although early research on the SbO field is more focused on exact methods, such as gradient search methods or statistical inference techniques, more recent research has been more focused on metaheuristic algorithms. The main reason for this is that recent real-life simulation models become so complex to be optimized with exact methods, where metaheuristic algorithms are capable of solving this type of problems more conveniently. The metaheuristic algorithms are preferred to exact methods mainly for their efficiency in terms of computation time; however, they do not guarantee optimal solutions. So, the quality of the result needs to be balanced with the time spent on computation. Also, it is worth to mention that while it is rare, on certain instances, metaheuristic algorithms methods fail to find any result, particularly in the presence of multiple objectives.

Unlike other optimization methods such as linear programming or MIP, the main difficulties of SbO method include, but not limited to, the following:

(a)     There may not exist an analytical expression of the objective function and in some cases, even the feasible region may not be explicitly described.

(b)     The stochastic nature of the simulation model makes it difficult to estimate the objective function values of solution points. Therefore, there exist various levels of noise. Usually, multiple simulation replications are needed to ensure the estimation accuracy and to handle simulation noise.

(c)     Most of the time, simulation runs are very expensive in terms of overall computation time, which may result in longer solution times.

### 3.2 Multi-Objective Optimization

Multi-objective optimization (MOO) problems require the simultaneous optimization of more than a single objective function, which is intrinsic to many real-life problems. In MOO, there is no accepted definition of optimum as it is in the single-objective case; hence, the notion of optimality is different where there is a set of optimal solutions instead of a single optimal solution. One of the primary challenges regarding MOO problems is that since two or more conflicting objectives are optimized simultaneously, the search space often becomes partially ordered, which requires a special treatment. Before delving into the details of MOO, the principles of MOO are outlined and basic concepts are formally defined in the following sub-section.

### 3.2.1 Basic Concepts of Multi-Objective Optimization

Several basic principles and formal definitions are required for proper analysis of MOO structures and related evaluation of these structures. These principles and definitions are introduced below.

Let's consider an MOO problem with $k$ objectives as an example:

$$min\ z = \begin{cases} f_1(x) = c_1 x \\ f_2(x) = c_2 x \\ \quad\vdots \\ f_k(x) = c_k x \end{cases}$$

$$s.t. \qquad Ax \geq b$$

$$x \geq 0$$

(3.2)

where $A$ is an $m \times n$ matrix, $b$ is an $m$-vector, $c_1$, $c_2$, ..., $c_k$ are $n$-vectors and $x$ is a $n$-vector of the decision variables of the problem. The feasible set of the above problem is $X = \{x \mid Ax \geq b,\ x \geq 0\}$. Here $k$ objective functions $z_i\colon X \longrightarrow \mathbb{R}$, $1 \leq i \leq k$, mapping a solution $x$ in decision space $X$ to its objective vector $f(x) = (f_1(x),\ ...,\ f_k(x))$ in the objective space $\mathbb{R}^k$, have to be minimized concurrently. Contrary to single-objective optimization, in most MOO problems, no solution exists that optimizes all objective functions simultaneously.

As opposed to finding or approximating the optimal objective function value, MOO is mainly concerned with finding or approximating the set of so-called *Pareto-optimal solutions* representing the best trade-offs between the objectives. For that purpose, the Pareto dominance relation is defined. A solution $x^*$ is called *Pareto-optimal (vector maxima* or *efficient)*, if there is no other solution which is at least as good as $x^*$ on all objectives and strictly better with respect to at least one. The concepts of dominance and Pareto-optimality are fundamental in MOO due to the fact they constitute the foundation of solution quality. Definition of the Pareto-optimal solution in a more formal way is given below, and illustrated in Figure 9.

**Definition 3.2.1 (Pareto-optimal solution)** A feasible solution $x^*$ is called Pareto-optimal or efficient, if there is no other $x \in X$ such that $f_i(x) \leq f_i(x^*)$, $\forall i \in \{1, 2, 3, \ldots, k\}$ and $f_i(x) \neq f_i(x^*)$ (Pareto-optimal solutions are also referred as non-dominated solutions.) (Ehrgott, 2006).

A solution $x \in X$ is said to dominate another solution $y \in X$ if and only if (iff) $\forall 1 \leq i \leq k$ : $f_i(x) \leq f_i(y)$ and $\exists 1 \leq i \leq k : f_i(x) < f_i(y)$. This can also be written as $x \prec y$. A solution $x^* \in X$ is then called Pareto-optimal iff there is no other solution in $X$ that dominates $x^*$. By the same token, the *weak dominance relation* is defined below.

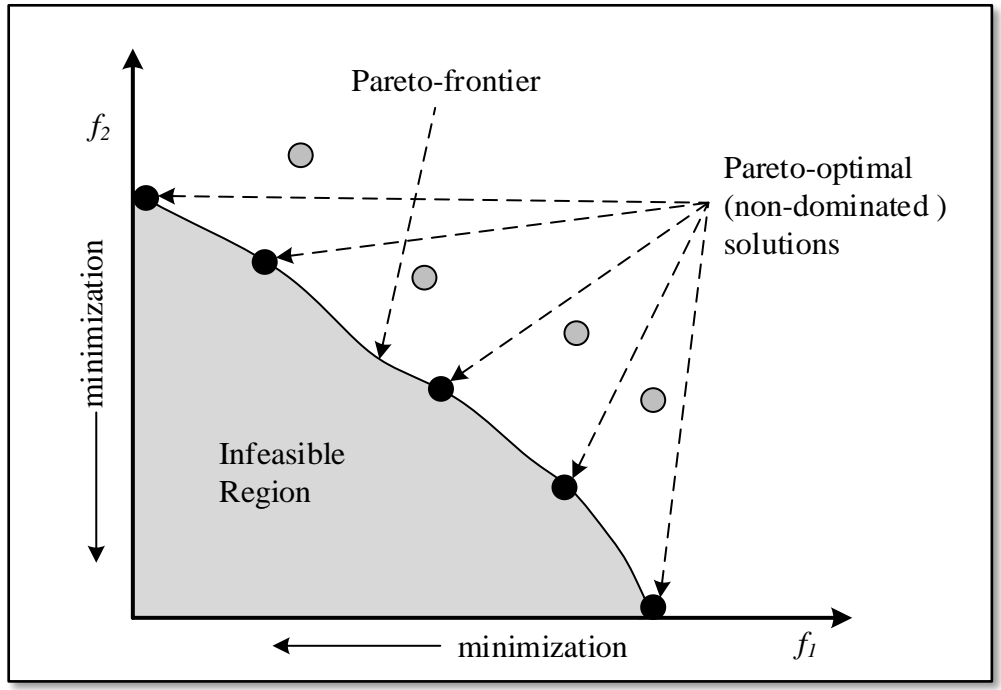

**Figure 9:** Illustration of Pareto-optimal Solutions and Pareto-frontier

**Definition 3.2.2 (Weak Dominance Relation)** A solution $x \in X$ *weakly dominates* a solution $y \in X$ $(x \preccurlyeq y)$ iff $\forall 1 \leq i \leq k : f_i(x) \leq f_i(y)$. Two solutions that are mutually weakly dominating each other are called indifferent whereas they are called *non-dominated* iff none is weakly dominating the other.

The definitions above can be used to define two different non-dominated sets: *strongly* and *weakly non-dominated set*. Strongly non-dominated set can be defined as follows: Among a set of solutions $P$, the strongly non-dominated set of solutions $P'$ are those that are not weakly dominated by any member of set $P$. Likewise, weakly non-dominated set can be defined as follows: Among a set of solutions $P$, the weakly non-dominated set of solutions $P'$ are those that are not strongly dominated by any member of set $P$.

These dominance relations can be generalized to relations between sets of solutions. For instance, a solution set $A \subseteq X$ weakly dominates a solution set $B \subseteq X$ iff $\forall b \in B \; \exists a \in A : a$

≼ *b*. Specific sets of pairwise non-dominated solutions are called *Pareto set approximations* (Ehrgott, 2006).

From scheduling perspective, a multi-objective scheduling procedure generates a solution set in the objective space in which some of the solutions are optimal trade-off solutions among the optimization objectives. These optimal solutions are known as *Pareto-optimal schedules,* which together constitute the so-called *Pareto-frontier*. A formal definition of Pareto-optimal schedule is given below.

**Definition 3.2.3 (Pareto-optimal schedule)** A schedule is called Pareto-optimal if it is not possible to improve the value of one objective without deteriorating the value of the other (Pinedo, 2016).

In Figure 10, the search space (solution or decision space), and the objective space are illustrated for bi-objective optimization, where four solutions are shown in decision space with their corresponding location at the objective space after applying to them the function *f(x)*. As shown in Figure 10, the location of the solutions in the search space does not hold any correspondence with their location at the objective space.

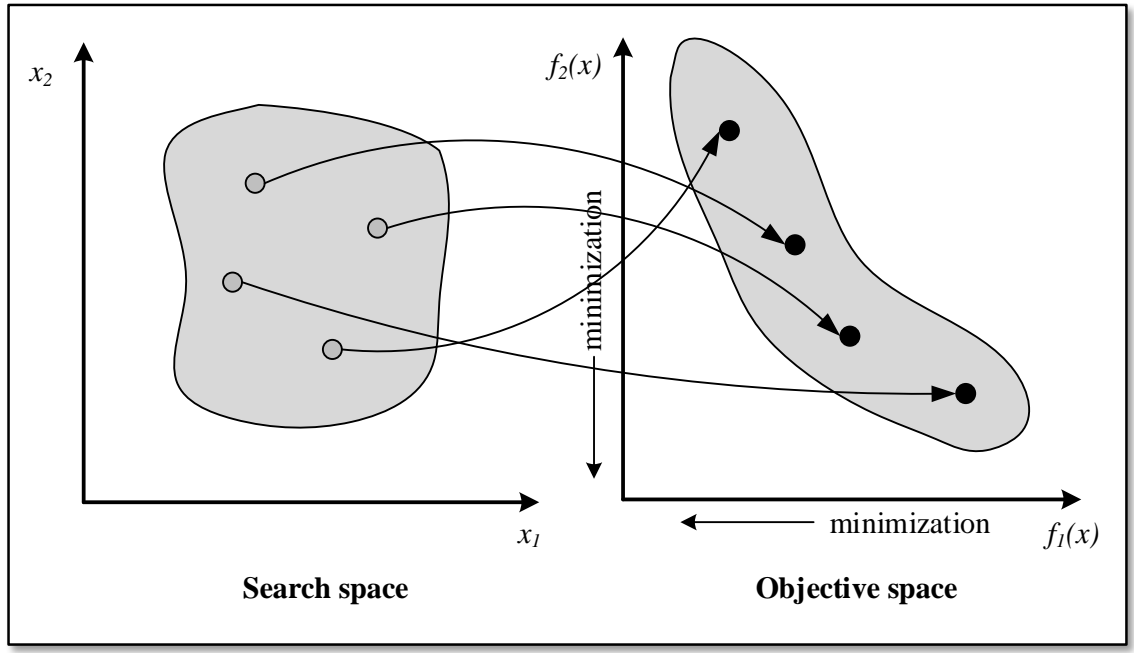

**Figure 10:** Illustration of Search and Objective Space in MOO

Coello et al. (2007) identified four essential goals for optimizing multiple objectives: (1) to find a set of solutions as close as possible to the true Pareto-optimal solutions and preserve progress towards the Pareto-frontier in objective space, (2) to find well-distributed Pareto-optimal solutions that cover the entire Pareto-optimal region in order to ensure a good set of trade-off solutions and retain diversity of Pareto-frontier in objective space and diversity of Pareto-optimal solutions in decision space, (3) to maintain non-dominated points in objective space and accompanying solution points in decision space, and (4) to provide the decision-maker "enough" but limited number of Pareto points for selecting the resulting decision variable values (Coello et al., 2007).

Effective design of MOO algorithms requires considering both of these goals since the fulfillment of one goal does not necessarily guarantee the other goal. Thus, not only it is important for mechanisms guarantee convergence to the Pareto-optimal region but they should also maintain a diverse set of solutions. This requirement renders the MOO as more challenging than the single-objective optimization.

MOO problems can be tackled by either exact or metaheuristic methods where both have different strengths and weaknesses. Selecting the most suitable method heavily depends on the characteristics of the problem at hand. However, in either case, the concept of Pareto-optimality is commonly embraced.

3.2.2 Classification of Multi-Objective Optimization Methods

Even though there exist several different classifications of MOO methods, the one based on the extent of preference information proposed by Hwang and Masud (1979) and later modified by Miettinen (1999) is the most accepted classification in the literature. In this classification, MOO methods are categorized into four classes according to the participation of the decision-maker in the solution process: (1) *no-preference* methods (no articulation of preference information is used), (2) *a posteriori* methods (a posteriori articulation of preference information is used), (3) *a priori* methods (a priori articulation of preference information is used), (4) *interactive* methods (progressive articulation of preference information is used).

To better explain the differences among these methods, the MOO process can be decomposed into three phases as follows: model building, optimization, and decision-making. In *no-preference* methods, as the name indicates, without any preference information, all the non-dominated solutions are enumerated with optimization and presented to the decision-maker. In *a posteriori* methods, the decision-making stage comes after optimization. In *a priori* (*scalarization*) methods, first the decision-maker articulates preference information, and then, the optimization phase tries to find the optimal solution by converting the MOO problem into a single-objective optimization problem. Lastly, in *interactive* methods, decision-making phase is interlaced with optimization in an iterative manner.

The most widely used MOO approach in the literature is the a priori methods where the multiple objectives are aggregated into a single objective. However, in the decision-making phase, the possibilities and limitations of the problem may not necessarily be known in advance, and this may result in too optimistic or pessimistic expectations. Also, this approach does not account for identifying the best trade-offs between contradicting objectives that can provide useful insights to the decision-maker. The primary task in MOO is to identify a set of Pareto-optimal solutions that helps to understand the problem structure better and provides a basis for decision-makers in picking the best trade-off solution.

On the other hand, a posteriori methods produce a set of Pareto-optimal solutions from which the decision-maker selects a solution based on post-preference assessment. These methods often lead to a more conscious and better choice. Moreover, resulting Pareto-optimal solutions can be analyzed to identify interdependencies among decision variables, objectives, and constraints. The focus of this dissertation is on a posteriori methods, in particular, metaheuristic algorithms where it is tried to generate an approximation of the Pareto-optimal set. However, first, a priori methods are explained briefly in order to give a basic understanding of the MOO field's principles.

### 3.2.3 Scalarization (A Priori) Methods

Scalarization (also referred as a priori or traditional) methods update a single solution in each iteration, and that utilizes a deterministic transition rule for generating the Pareto-optimal set. These methods convert a MOO problem into a single-objective optimization problem by scalarizing the objective vector into a single composite objective function, and thus, a single trade-off optimal solution can be sought effectively (Deb, 2001). The most frequently used a priori methods for handling MOO problems are the weighted sum method, the $\varepsilon$-constraint method, goal programming, Tchebycheff methods, and the minimax approach. Due to their wide utilization in the literature, the weighted sum and the $\varepsilon$-constraint methods are briefly discussed below.

*Weighted Sum Method*: This method might be the most intuitive approach for solving MOO problems. The basic idea of weighted sum method is to combine all of the objective functions into a single functional form with a weighted linear sum of the objectives. In this method, scalar weights are specified for each objective to be optimized, and then, they are combined into a single function that can be solved by any single-objective optimizer.

Clearly, the solution obtained depends on the values of the weights specified. A weighted sum version of the multi-objective linear programming problem with *k* objectives can be formulated as follows:

$$min \ z = \lambda_1 c_1 x + \lambda_2 c_2 x + \cdots + \lambda_k c_k x$$

$$s.t. \quad \boldsymbol{A}x \geq b$$

$$\lambda_1 + \lambda_2 + \cdots + \lambda_k = 1$$

$$\lambda_1, \lambda_2, \ldots, \lambda_k, x \geq 0$$

$$(3.3)$$

where $\lambda_1, \lambda_2, \ldots, \lambda_k$ are the weights. The main drawback of weighted sum method is the fact that it is essentially subjective, where decision-maker needs to provide the weights, and in order to accomplish this, different weight vectors have to be evaluated perpetually. The process of specifying and fine-tuning the weight vector is usually a tedious and inefficient task. Also, this approach cannot identify all efficient solutions. These drawbacks render this method very inefficient to be used in a SbO framework because running a simulation to evaluate the objective functions takes significantly long computation time. Therefore, setting the appropriate weight vector until finding an acceptable solution may take longer than a considerable solution time.

*ε-Constraint Method*: In this method, one of the objective functions is optimized using the other objective functions as constraints. Incorporating objective functions in the constraint part of the linear programming problem is shown below:

$$min \ z = c_1 x$$

$$s.t. \quad \boldsymbol{A}x \geq b$$

$$c_2 x \leq \varepsilon_2$$

$$c_3 x \leq \varepsilon_3$$

$$\vdots$$

$$c_k x \leq \varepsilon_k$$

$$(3.4)$$

$$x \geq 0$$

The parameters $\varepsilon_2$, $\varepsilon_3$, …, $\varepsilon_k$ are used to represent the upper bound values for the corresponding objective function values. The efficient solutions of the problem can be obtained by parametric variation of the right-hand-side of the constrained objective functions.

The main strengths of traditional (scalarization) methods include the proof of convergence to the Pareto-optimal set, simplicity, and easy implementation. Despite these advantages, they all suffer from a number of difficulties in common when finding multiple Pareto-optimal solutions. First of all, they necessitate several runs to find an approximation of the Pareto-optimal set. Second, because the optimization runs are done independently from each other, interdependencies typically cannot be exploited. Finally, they require some problem knowledge, such as weights or $\varepsilon$ values, etc. (Deb, 2001).

### 3.2.4 Approximate Methods

Recently, approximate algorithms have become established as an alternative to traditional methods for several reasons, including the following: (1) large search spaces can be handled, and (2) multiple alternative trade-off solutions can be generated in a single run. Due to the complexity of the MOO problems, approximate algorithms are a very practical and suitable approach. The main drawback of these algorithms is that even if the solutions found are non-dominated in the recent population, they may not necessarily be the actual Pareto-optimal set of solutions. Since these objectives are conflicting, trade-offs need to be found between objectives to obtain satisfactory results.

Among approximate methods, the multi-objective evolutionary algorithms (MOEA) are the most popular metaheuristic algorithms, which have been increasingly dominating the literature for solving practical multiple objective problem optimization in recent years. This is mainly due to their capability of dealing with multi-objectives in a natural way and capturing multiple trade-off solutions in a single optimization run since multiple solutions are simultaneously generated in each iteration, and a population of solutions is maintained.

Applying the principle of "survival-of-the-fittest" in natural selection, MOEAs have the unique feature of sampling multiple solutions concurrently.

In a study done by Jones et al. (2002), it is concluded that majority of the MOO approaches proposed in the literature to approximate the true Pareto-frontier were utilizing a metaheuristic algorithm, and 70 percent of all metaheuristic approaches are based on MOEAs. Even though a large variety of MOO algorithms have been suggested in the literature, there is still room for improvement to develop computationally efficient algorithms with an enhanced capability to converge non-dominated solutions along the Pareto-frontier for complex and large-scale MOO problems.

## 3.3 Simulation-based Multi-Objective Optimization

Simulation-based multi-objective optimization is an evolving area of research that integrates optimization techniques into simulation modeling and analysis, where optimization deals with multiple conflicting objectives simultaneously. Numerous real-life applications in different fields demonstrate the potential of the SbO with a single objective; however, multiple conflicting objectives are usually not taken into account. It is only recently that the research community has initiated experimentation with MOO in combination with simulation. This situation is because of the fact that in addition to the challenges inherent to the SbO models (high computational cost and stochastic noise), considering the additional complexity of multiple and possibly conflicting objectives makes the solution process more challenging and tedious.

Because solution evaluations are expensive in terms of time, traditional (a priori) methods may not be feasible to solve simulation-based multi-objective optimization problems. The classic MOO methods avoid the complexity originated from the existence of multiple conflicting objectives by utilizing traditional techniques by converting the MOO problem into a single-objective optimization problem. In general, the optimal solution to this single-objective problem is expected to be a Pareto-optimal solution. However, such a solution highly depends on the parameters used in the conversion method. In order to find different

Pareto-optimal solutions, the parameters should be changed, and the resulting single-objective problem should be solved again. Therefore, traditional MOO methods require solving single-objective optimization problems multiple times to find the Pareto-frontier, which is not analytically tractable in many real-life settings. On the other hand, MOEAs only require a limited number of fitness evaluations, which usually do not suffer from the computational tractability issues.

The objective function of a simulation-based multi-objective optimization problem with noise can be defined as follows:

$$min \ z = \ \mathbb{E} < \boldsymbol{f}(x) > = \ \boldsymbol{f}(x) + \boldsymbol{\delta} \ ,$$

$$\boldsymbol{f}(x) = \begin{bmatrix} f_1(x) \\ \vdots \\ f_n(x) \end{bmatrix}, \qquad \boldsymbol{\delta} = \begin{bmatrix} \delta_1 \\ \vdots \\ \delta_n \end{bmatrix}. \tag{3.5}$$

where objective function value is the expected value of sampled fitness value ($E<f(x)>$), which consists of true fitness value ($\boldsymbol{f}(x)$) and noise value ($\boldsymbol{\delta}$).

The primary issues with considering a simulation-based multi-objective optimization approach are individual objective ranges and noise involved in the objective functions. First difficulty can be mitigated conveniently by normalizing the objective functions. However, several issues stem from the noise include the following: (1) fitness evaluation component of the optimization component often becomes unstable, and in turn, it results in dismissal of high-quality solutions and premature convergence, (2) the diversity of solutions becomes biased because of the inappropriate allocation of the diversity measure, and (3) the risk of treating a dominated solution as a non-dominated solution due to the dominance comparison based on the sampled fitness value. These issues, if not addressed appropriately, easily decrease the performance of the approach, and lead to weak Pareto-frontier (Figure 11).

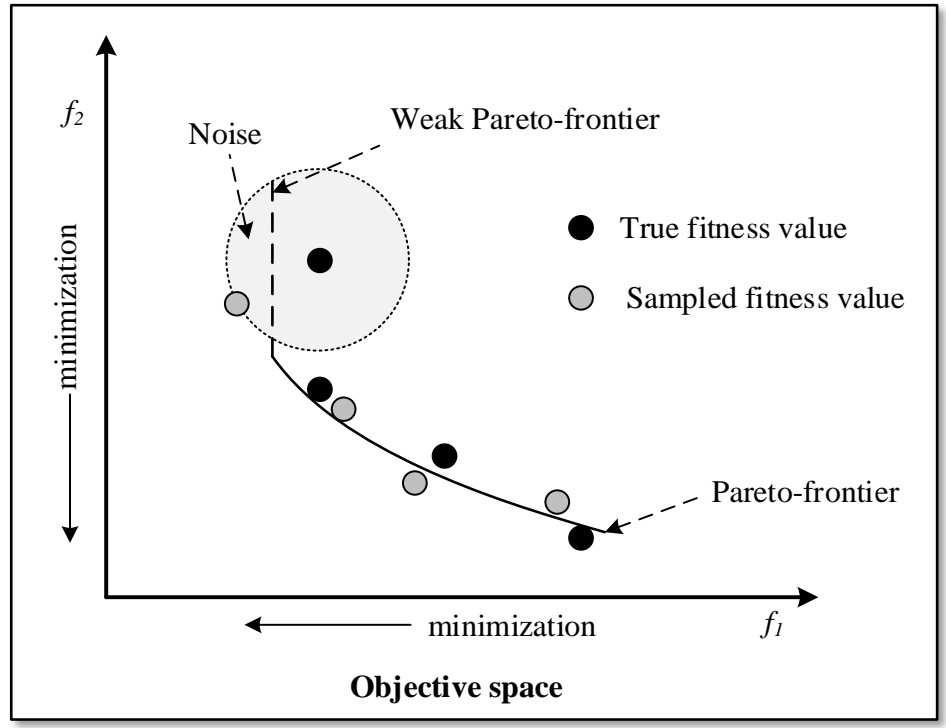

**Figure 11:** Illustration of Weak Pareto-frontier Resulting from Noise

In order to overcome the aforementioned issues inherent to simulation-based multi-objective optimization, several methods have been suggested. Joines et al. (2002) proposed a GA-based approach by adapting the enhanced (elitist) version of Non-Dominated Sorting Genetic Algorithm (NSGA-II) and applied it to a real-life supply chain optimization problem with two objectives. Eskandari et al. (2005) incorporated a simulation model with a stochastic non-domination-based MOO method and GA. New operators, which include elitism, dynamic expansion, importation operators, for GA are introduced to improve the performance of the algorithm in terms of both effectiveness and efficiency.

As a conclusion; although there exist several difficulties inherent to simulation-based multi-objective optimization, this field has become a significant research field with its specific theoretical foundations, which leads to a better understanding of the fundamental principles, and in turn, development of practical algorithms. Considering these difficulties in the context of the runway operations scheduling problem, it is evident that a framework that structures effective design and implementation of simulation-based multi-objective optimization is necessary.

## 3.4 Overall Simulation-based Optimization Framework

The overall problem-solving approach is based on a SbO framework, as presented in Figure 12. SbO is utilized for obtaining optimal system settings from sets of decision variables, i.e., input parameters, where the objective functions and performance of the system are evaluated through the output results of the simulation model over the system. The multi-objective metaheuristic algorithm and the discrete-event simulation model are separated into individual modules.

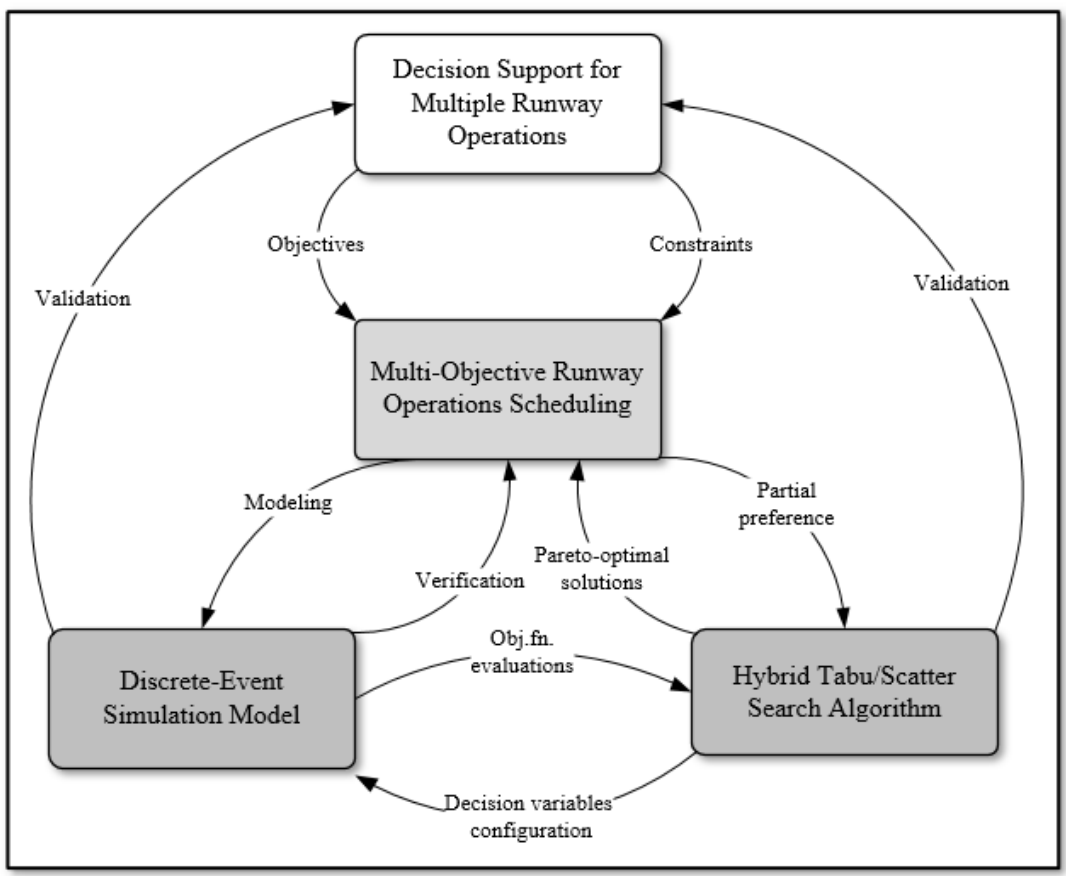

**Figure 12:** The Overall Problem Solving Approach

The flow-chart diagram of the SbO framework is given in Figure 13. The SbO framework starts with a greedy heuristic algorithm to produce a relatively good initial solution compared to a random one. The simulation model outputs the performance measures from a single simulation run where it is treated as a black-box model that evaluates the performance of a particular configuration of system parameters and provides these performance measures as bi-objectives. The optimization component employs a metaheuristic algorithm, which is discussed later, to search for the values of system parameters. It is an iterative process initiated by the optimization algorithm starting with inputting the initial solution and generating a set of candidate solutions which act as input values for the simulation model. After receiving the input values from the optimization algorithm, the simulation model is executed to compute the performance measures (including optimization objectives and other output parameters of interest) which are then fed back into the optimization algorithm.

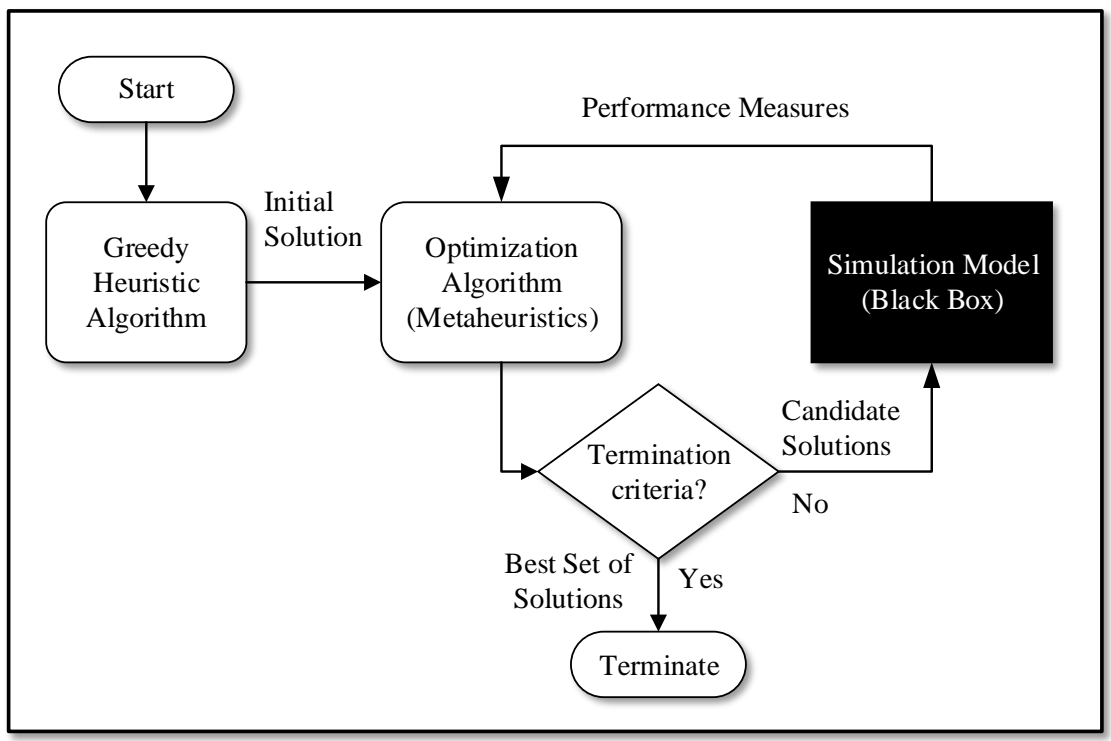

**Figure 13:** Schematic Representation of the SbO Framework

The proposed SbO framework is comprised of two main components: an optimization model for managing the search process, and a simulation model for evaluating the performance of candidate solutions. The results of the performance evaluation are used to refine the optimization process and return the best set of solutions at the end. In this method, a simulation model is utilized as a replacement for an analytical fitness function to better mimic the behavior of the real-life runway system as well as to account for uncertainty explicitly.

The whole SbO process treats the simulation model as a black-box where the optimization algorithm feeds candidate solutions to the simulation model which then generates the performance measures back to the optimization model. Since metaheuristic algorithms do not make explicit assumptions about the underlying structure of the objective function, the black-box nature of the simulation model does not create any difficulty.

### 3.4.1 Main Steps of the Simulation-based Optimization Framework

The proposed simulation-based optimization framework includes the following main steps:

**Step 1**: Build an initial feasible solution $S_0$, with the greedy heuristic algorithm.

**Step 2**: Improve the solution $S_k$, with the hybrid Tabu/Scatter Search algorithm.

**Step 3**: Given $S_k$ invoke a simulation procedure to generate random variable realizations and compute an estimate of the corresponding performance measure.

**Step 4**: Determine if any termination criterion is satisfied. If yes, stop the algorithm and output the best set of solutions found so far; otherwise, set $k \leftarrow k+1$ and go to step 2.

In the first step, the initial feasible solution is obtained from the greedy heuristic algorithm, and then, sent to the metaheuristic optimization algorithm as its starting point. In the next step, a metaheuristic algorithm generates a neighborhood of solutions and selects a candidate solution from them with a number of techniques and this candidate solution is sent to the simulation model to be evaluated. Subsequently, the simulation model estimates candidate solution's performance measure by running several replications the result is sent back to the optimization component. Then, optimization component continues improving the solution through information that is obtained from the simulation model. These steps are repeated iteratively until the termination condition is met.

The simulation model, which is a discrete-event simulation, further complicates the optimization process because information needs to be sent between the simulation and optimization algorithms in every iteration.

The main objective of the SbO framework is to determine the runway assignments, and find the best sequence of aircraft in each runway and the landing/take-off times for each aircraft. For the SbO framework, the following information is considered as given:

(a)     A set of arriving aircraft with pre-determined meter fix assignments and estimated time of landing (ETL). Meter fixes are the points along the established route from over which aircraft is metered prior to entering TMA.

(b)     A set of departing aircraft with estimated time of take-off (ETT).

(c)     A set of attribute for each arriving and departing aircraft, which include aircraft identification number, operation type, weight class, a maximum delay time for each aircraft as a hard constraint.

(d)     The minimum separation times between aircraft weight classes and runway occupancy times for each aircraft weight class.

(e)     Arrival times to entry points and holding area for arriving and departing aircraft, respectively.

In practice, runways used for mixed operations are operated in one of the following three ways: (1) arrival priority, (2) alternating runway operations, and (3) departure priority. In arrival priority, landing operation has higher priority over take-off operation, where take-offs are fitted in between landings whenever enough time space is available. In alternating runway operations, additional spacing is used among landing aircraft to allow at least one take-off between consecutive landings. In departure priority, take-offs are given priority over landings. This mode of operation is typically used only when there are many take-offs in the holding area queued for take-off. The SbO framework is designed to work in all configurations.

In the SbO framework, the following two primary objectives are considered:

(a)     *ATC-driven objective*: Minimizing makespan, which means maximizing runway utilization (throughput), where landings are weighted by importance.

(b)     *Airline-driven objective*: Minimizing unfairness, which corresponds to minimizing aircraft's position shift from their FCFS position. This is an important interest for airlines because FCFS order considered a fair rule among airlines. Also, this objective usually helps airlines to reduce operational costs and improve their on-time performance.

The decision variable configurations are evaluated in the simulation model, which can be seen as a black-box; the simulation model is fed with the decision variables and the objective function values are provided by running the simulation model. This iterative process is demonstrated in Figure 14. The main objective of this SbO process is to iteratively refine the objective values by adjusting the decision variables.

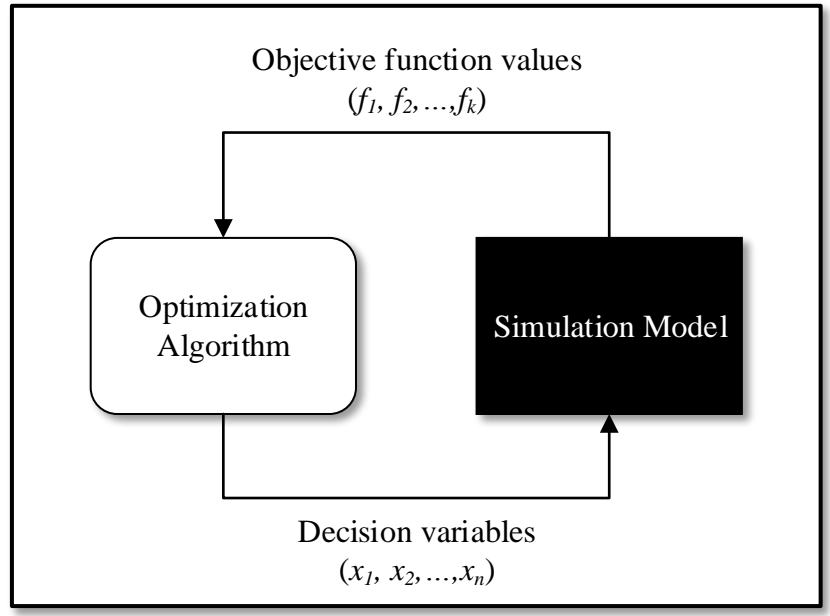

**Figure 14:** Simulation-based Multi-Objective Optimization Iterations

The resulting multi-objective problem is harder to deal with because a significantly larger portion of the search space needs to be explored to obtain a set of Pareto-optimal solution. The proposed multi-objective approach tries to deliver a Pareto-frontier (a set of optimal solutions), where decision-makers can use them to choose the appropriate solution based on different conditions and priorities. Therefore, a hybrid Tabu/Scatter Search algorithm is embedded in the optimization component to find decision variables configuration points in the decision space for which the corresponding points in the objective space reside on or close to the Pareto-frontier. The primary requirement for the metaheuristic algorithm is the ability to generate diverse solutions, which means solutions that cover a large area in the objective space, and to converge effectively towards the Pareto-frontier, i.e. solutions evolve iteratively.

The ultimate aim of the SbO framework is to generate a robust solution, which can be defined as the solution that has a low probability of violating the constraints in actual operations while being reasonably close to optimal. Robustness (reliability) of a schedule is measured in terms of the probability that none of the minimum separation times constraints will be violated, or in other words, the probability that an air traffic controller intervention will not be required. SbO approach allows evaluating the robustness of the solutions provided by the optimization method under near-real conditions. If the solution is not robust, its objective function value is used as a lower bound and the simulation optimization iteration will be repeated.

3.4.2 Design and Implementation Methodology

Before designing and developing the aforementioned SbO framework, the following activities are concluded.

(a)     Various sources of uncertainty in runway operations that influence the runway utilization performance are identified.

(b)     These uncertainty factors are embedded into the simulation model.

(c)     The impacts of these uncertainties are investigated by running simulations.

(d)     Depending on the degree of uncertainty, the landing/take-off delay changes are quantified through simulations for each uncertainty factor.

The design and implementation methodology of the SbO framework is shown as follows (depicted schematically in Figure 15):

Step 1: Understand the stochastic nature of the runway operations scheduling problem that the air traffic (local) controllers face on a daily basis. Define the problem and delineate the abstraction level for modeling.

Step 2: Analyze the properties and fundamental characteristics of the practical problem with consideration of the uncertainty systematically.

Step 3: Design and develop a rigorous, computationally tractable SbO approach.

Step 4: Implement and test an initial version of the SbO framework and test it.

**Step 5:** Develop an improved version of the solution approach, which can produce a robust and efficient solution.

**Step 6:** Conduct computational experiments, analyze the results and perform sensitivity analysis.

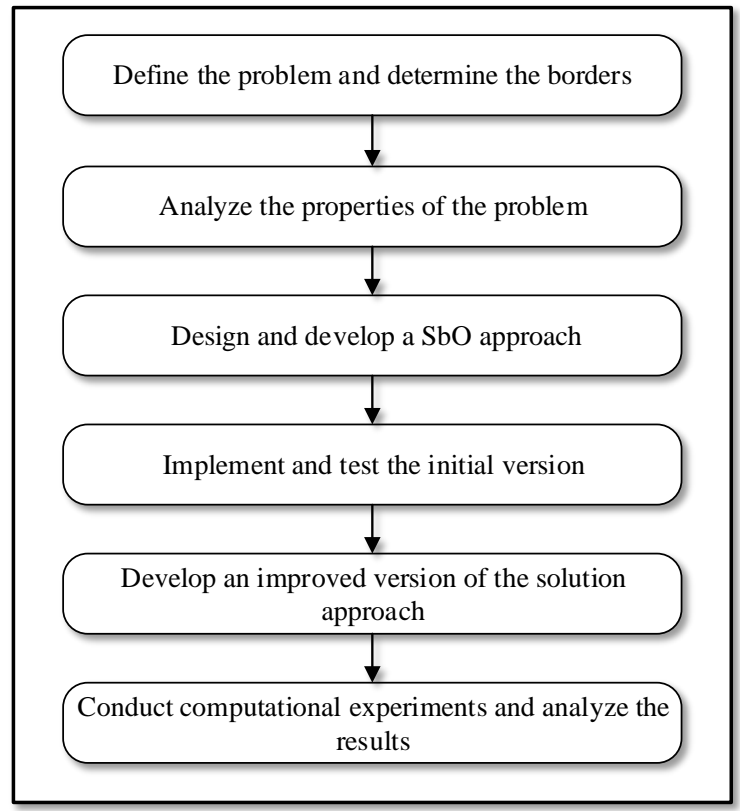

**Figure 15:** Design and Implementation Methodology

## 3.5 Components of the Simulation-based Optimization Framework

### 3.5.1 Initial Solution Generation Component

Initial solution generation component consists of a greedy heuristic algorithm based on a composite dispatching (priority) rule to efficiently construct a starting solution for the SbO framework. A heuristic algorithm is a rule of thumb used for finding a solution to a mathematical problem where it is usually based on a simple logical idea that does not need a detailed mathematical foundation to explain. The main advantages of a heuristic algorithm are relative simplicity, fast solution times and easy implementation. Also, it does not need significant time to adjust parameters to fit a problem instance. On the other hand, the main inherited shortcomings are that in most cases they do no guarantee optimal solution, and they may generate solutions with poor quality. Solutions obtained from a heuristic algorithm are typically used as an initial solution for a more advanced solution method, such as a metaheuristic algorithm.

In the past several decades, heuristic algorithms based on a dispatching (priority) rule not only have been studied extensively in the operations research literature but have also been applied to different scheduling problems successfully in practice. Dispatching rules can be classified into two main categories: static and dynamic rules. As it is evident from the term, dynamic rules are time dependent, static rules are not. A combination of elementary dispatching rules is typically called as composite dispatching rule (Pinedo, 2016). Typically, a look-ahead parameter scales the contribution of each part of the composite dispatching rule relative to the total, which has to be suitable for the problem instance at hand to get good quality solutions, look-ahead (scaling) parameters are often determined with an empirical study.

The composite dispatching rule utilized in the greedy heuristic algorithm exploits the structure of the problem. Basically, it is a function of aircraft ($i, j$) attributes, such as earliest time ($r_j$), latest time ($d_j$) and separation times ($s_{ij}$) as parameters. The overall priority of an aircraft is influenced by an attribute of the aircraft that is mainly determined by a look-ahead parameter. This parameter is determined empirically and validated in terms of its suitability for practical problem instances to get high-quality solutions (Hancerliogullari et al., 2013). In their composite dispatching rule, aircraft are scheduled one at a time, i.e. when a runway becomes free, a priority index is computed for each remaining aircraft and the aircraft with the highest priority index is then selected to be scheduled next. The priority index ($\eta_{jr}$), that is a function of the time $t$, is defined as follows:

$$\eta_{jr} = w_j \exp\left(\frac{-\max(d_j - t, 0)}{k_1}\right) \exp\left(\frac{-s_{ij}}{k_2 \bar{s}}\right) \exp\left(\frac{-\max(r_j - t, 0)}{k_3}\right) \tag{3.6}$$

where $j$ is the aircraft to be scheduled; $i$ is the previous aircraft; $r$ is the runway; $w_j$ is the weight parameter for $j$, and $k_1$, $k_2$ and $k_3$ are scaling parameters. As a result of an empirical study, the parameters $k_1$, $k_2$, and $k_3$ are determined to be 2, 0.75, 1.7, respectively.

The main idea in the greedy heuristic algorithm is to generate a better quality initial solution in which the aircraft are sorted by the priority index and are considered to be landing or take-off on the runway at its best available time one after another. The pseudo-code of the greedy heuristic algorithm is given below.

---

**Algorithm 1** Greedy Heuristic Algorithm for Initial Solution Generation (Hancerliogullari et al., 2013)

---

**Input**: List of aircraft and number of runways, $M$

1:     **begin**
2:         sort all aircraft according to the look-ahead priority index (1 to $N$)
3:         **for** $i = 1$ to $N$
4:             **for** $r = 1$ to $M$
5:                 calculate $E_{ir}$ (Earliest feasible time that aircraft $i$ can land or take-off from runway $r$)
6:             **end for**
7:         calculate start time for aircraft $i$, $s_i = min \{E_{ir} \mid r$ in $M\}$
8:         assign aircraft $i$ to the runway related to calculated $s_i$
9:         **end for**
10:     calculate the fitness value (obj.fn.value)
11:    **end**

**Output**: A feasible solution consists of runway operations schedule with a fitness value and start time for each aircraft

---

The greedy heuristic algorithm generates the initial solution by assigning aircraft to runways, determining the aircraft sequence in each runway and start time of each runway operation, considering the minimum separation requirements and time windows for landing and take-off. The runway assignment step tries to balance the air traffic on multiple runways, and the aircraft sequencing and scheduling steps attempt to generate a solution accounting for minimum separation times and time windows.

In addition, a fitness value is calculated based on the start time of the last runway operation, which is the objective function value corresponding to maximizing throughput (runway utilization). This simple heuristic algorithm provides a reasonably good solution in a relatively short computational time without guaranteeing an optimal solution.

### 3.5.2 Simulation Component

It is easy to measure the performance of a runway operations schedule once it has been executed, by measuring the associated performance measure, such as the runway utilization, or the total delay occurred during the runway operations. However, it is difficult to estimate these performance measure related to the particular schedule in advance at the planning stage. Therefore, simulation is commonly considered as a powerful method for dealing with these difficulties especially for scheduling problems which are dynamic and stochastic in nature.

In general, simulation is considered as an algorithmic technique for conducting experiments on dynamic numerical models. The internal representation of the simulation model utilizes a number of state variables which are employed to define the system state and mimic the dynamic behavior of the real runway system by changing these variables accordingly.

The simulation component of the proposed SbO framework is a discrete-event simulation model based on an aircraft trajectory model, and runway operations are approximated to fit the model. It mimics the movement of individual aircraft with an acceptable level of accuracy, and simulates the stochastic processes to reflect the uncertainty in runway operations. The data that the entities (aircraft) need as they move through the model include release times into the system, separation times, runway occupancy times (ROTs), probability distributions associated with release times and ROTs. Entities carry some of the data with them as they move, and some of the information are shared across entities, which belong to the system as a whole.

The simulation model simulates the arrival and departure segments through reasonable trajectory states at key points along the path of the aircraft, and they are represented by a network of nodes and arcs. Aircraft move on this network along prescribed trajectories that are made up of strings of nodes and arcs, where each arc can be occupied by a single aircraft at a time. The landing and take-off operations are broken into segments, and each segment forms a node of the simplified runway operations.

In the simulation model, these stochastic processes are simulated by utilizing random variables to reflect the stochastic behavior observed in actual runway operations. Given a solution (a runway operations schedule), the performance measures are evaluated stochastically with simulation runs by using particular values (realizations) of these random variables (sources of randomness). A detailed treatment of the simulation component is given in Chapter 4.

### 3.5.3 Optimization Component

The optimization component consists of a hybrid Tabu/Scatter Search algorithm. The main reasons for selecting a metaheuristic algorithm include the following:

(a)     Metaheuristics are capable of generating high-quality near-optimal solutions to multi-objective problems in a relatively short computational times.

(b)     Metaheuristics are more flexible to be adapted to different multi-objective optimization problem structures.

(c)     Metaheuristics do not require gradient or derivative information which is not easy to find for real-life multi-objective problems.

(d)     The decision-maker does not have to make beforehand preferences regarding the objectives until the alternative solutions are presented (Marler & Arora, 2004).

Although metaheuristic algorithms offer these advantages, their application is not completely straightforward. The dynamic balance of intensification and diversification has to be considered, where intensification refers to the exploitation of the best solutions found and convergence to the Pareto-optimal solution sets while diversification refers to the exploration of the search space and diversity of the obtained solutions around the optimal set.

The other aspect that needs to be considered is that in MOO problems a significantly larger portion of the search space needs to be explored to obtain a set of Pareto-optimal solutions. Considering that even a single simulation run may take considerable amount of computation time (ranging from a couple of minutes to hours), the solution process is computationally expensive when a large number of simulation evaluations are needed.

The optimization component generates a set of Pareto-optimal approximations, which are superior to all dominated solutions in the objective space and inferior to other Pareto-optimal approximations in at least one objective; thus, it provides the decision-maker with a whole set of alternative solutions that represent the trade-offs to choose from. The design and implementation details of the optimization component are given in Chapter 5.

### 3.5.4 Interfacing Optimization and Simulation Components

One way to interface optimization and simulation components is to provide a specific interface layer for taking care of the communication between two components. Most of the existing simulation environments provide some ways to integrate the simulation model with other applications with the help of different technologies, such as specialized application programming interfaces (API), component object model (COM) interfaces for inter-processing, or web services. The primary challenge in these specialized interfaces is that they have to be developed for each simulation environment and underlying technology.

In order to avoid the burden of developing a specialized interface, both optimization and simulation components are designed with an object-oriented architecture, and implemented with a common development environment and programming language. Therefore, interfacing optimization and simulation components are accomplished using direct calls. When an evaluation of a candidate solution is needed, a simulation model is called directly. Then, the simulation model runs with the parameter values supplied by the optimization algorithm, calculates the quality of the candidate solution and returns it to the optimization algorithm.

The main advantages of this interfacing scheme are: (1) the effort for interfacing the components is minimum, (2) it allows tight coupling of optimization and simulation components, (3) it is more efficient compared to specialized interfaces by reducing inter-processing overhead, and (4) complex interactions between components are applied conveniently.

# CHAPTER 4

# DISCRETE-EVENT SIMULATION MODEL

Simulation modeling is a powerful and widely accepted method for analyzing complex and stochastic systems. Hence, in the literature, there exist several simulation models and tools that simulate airport and runway operations, and these models and tools are often utilized for evaluating some performance measures, such as airport capacity, resource utilization, etc. However, these existing simulation models for quantifying runway operations' performance necessitate a vast amount of detailed data. Also, they are not open source and cannot be used as a simulation component in a SbO framework. Therefore, a new discrete-event simulation model is developed that is capable of capturing the essential interactions of key system components, representing the system with sufficient detail, and reflecting the uncertainties associated with the runway operations.

In order to realistically represent the components of the real runway system and their complex interactions, system approach is adopted in developing the simulation model. This approach allows decomposition of the real system into functional components and application of an object-oriented architecture. Furthermore, applying this type of architecture yields advantages in the design as well as in the implementation of the simulation model. This discrete-event simulation model is designed to simulate the flow of air traffic for both arrivals and departures, where aircraft are generated as objects that move through the airspace segments and the runways. It is built to compute the performance measures, which are the inputs for the optimization component, by tracking the flow of aircraft in specific points.

In this chapter, first, the most widely used airport and runway simulation models are reviewed. Then, the purpose and high-level framework of the simulation model, which is a fast-time simulation model that quantify performance measures of runway operations for different runway operations schedules, are presented. Next, an abstract representation of the TMA, which describes the elements, relationships, borders and assumptions without reference to the specific implementation details, is described. Afterwards, the data sources and input analysis for the simulation model are given. In addition, the object-oriented design, and implementation specifics are provided. Finally, details of the verification & validation process is presented. The verification process is necessary to show that the simulation model operates as expected and provides an accurate, logical representation of the conceptual model. On the other hand, the validation process is conducted to determine if the model's behavior validly represents the real runway system being simulated.

## 4.1 Airport and Runway Simulation Models

During the past few decades, many analytical, as well as macroscopic and microscopic simulation models, have been developed for modeling the airside and/or landside operations of an airport, and analyzing the related statistics. The models developed so far span from basic queuing models and Monte Carlo simulations to sophisticated, comprehensive computer simulation models, and they differ broadly in their objectives, scope, level of detail, fidelity and complexity. Some of these models are publicly available and open source while others are proprietary. The primary objectives of these existing models include, but not limited to the following: (1) providing performance data of airport capacity evaluation, (2) bottleneck analysis of critical airport resources, (3) estimating the capacities of the runways, and (4) analyzing various strategies in air traffic flow management (ATFM) and airline operations.

*FAA Airport Capacity Models*: These models are developed for airport and runway systems that focus on strategic aspects of ATFM, namely airport and runway capacity. They estimate arrival and departure capacity by utilizing various parameters, such as minimum separation standards, runway occupancy times, fleet mix, availability of exit taxiways, etc. These models consist of the following two models to conduct specific and focused airport capacity studies: (1) Airfield Delay Simulation Model (ADSIM) is designed to calculate travel time, delay, and flow rate data to analyze airport airfield components, airport aircraft operations, and operations in the immediate terminal airspace for measuring service delays. (2) Runway Delay Simulation Model (RDSIM) is primarily developed to provide runway capacity/delay analysis and models the final approach, runway threshold, and runway exits. This tool simulates runway operations and provides both capacity and delay information. It is simply a critical-event stochastic model that employs Monte Carlo sampling techniques (Odoni et al., 1997b).

*Airport and Airspace Delay Simulation Model (SIMMOD)*: SIMMOD is a microscopic, discrete-event simulation model that tracks the movement of individual aircraft as they travel through the airspace and on the ground. SIMMOD consists of a network of nodes and links to represent the aircraft paths throughout the area being simulated. The nodes represent the decision points, physical locations or logic changes along the path, whereas the links define the route of travel. Each aircraft is allowed transit from one link to another depending on the link's attributes and rules related to the analysis. In SIMMOD, there are five input categories: airspace definition, airfield (airport) definition, schedule events, flight banks (to model the dependency of arriving and departing flights in the hub), and aircraft definition. SIMMOD generates highly detailed output including individual flight level data. It has been widely employed in air traffic surface operations and capacity studies (Kleinman et al., 1997).

*Total Airspace and Airport Modeler (TAAM):* TAAM is a large-scale detailed fast-time simulation tool that models the layout of an airport, the operating rules for every aircraft type, and the dynamics of every gate, taxiway, and runway with high fidelity. TAAM is widely considered as a flexible tool that can assist airport operators to estimate and analyze the impact of present and future airspace and runway operations precisely as well as enhance the safety and efficiency of these operations. It also includes an interactive user interface that provides a 2D or 3D view of the airspace or airport; a real-time air traffic monitoring tool with simulation capability; and a reporting tool which can be used to create graphs and tables from data produced by the simulation model (Bazargan et al., 2002).

*MIT Extensible Air Network Simulation (MEANS)*: MEANS is a macroscopic, discrete-event simulation framework for analyzing various strategies in air traffic flow management and airline operations. It is also capable of capturing the effects of uncertainty in the operating environment. MEANS has seven modules, where each module simulates a particular section of the air transportation network. Four of these modules (en-route, tower, taxi, and gate) are state modules since they deal with the movement of entities (aircraft, crews, and passengers) through several states. The en-route (airspace) module simulates the national airspace as a whole. Two other modules (Air Traffic Control System Command Center (ATCSCC) and airline) are the decision-making modules since they control the desired changes to the flight schedule that are then performed by the state modules. The last module, which is the weather module, provides weather and weather prediction information to the other modules. (Clarke et al., 2007).

*The MITRE Corporation runwaySimulator*: This simulation tool is developed by the MITRE Corporation, and it is employed by various organizations which have an interest in understanding the capacity of airport runways, such as air traffic service providers, airlines, airport operators, etc. The main functionality of this tool is to estimate the capacity of the runways, and it enables rapid analysis of airport capacity by combining a package of different methodologies including analytical and simulation methods. It randomly generates flights according to a specified fleet mix and combines a trajectory model, airport and fleet characteristics, and separation rules to estimate hourly capacity. The main output of the tool is a capacity curve showing the efficient capacity as a Pareto-frontier of arrival-departure throughput. The runwaySimulator is coded in Java and shared publicly for analysis of only US airports (Kuzminski, 2013).

Existing microscopic simulation models, such as TAAM and SIMMOD, can simulate detailed airport and runway operations; however, these models require extensive adaptation of both the airport layout and the traffic scenarios to produce statistically significant results. Therefore, it is difficult to use these microscopic models to as part of our SbO framework since they require a simulation study over long periods of time and extensive data (Odoni et al., 1997a).

On the other hand, the existing macroscopic models, such as airport capacity models, have generally been utilized for the purpose of supporting policy decisions regarding the best runway utilization, assisting the design of airports, and simulating air traffic with significant fidelity. Although these simulation models are very useful for studying system behavior as well as capable of considering stochastic airport processes including runway operations, they have several drawbacks: (1) these models require long computation times to produce results, (2) they are not open source, and (3) they are not flexible enough to be used as a simulation component in a SbO framework.

## 4.2 Objective and Simulation Modeling Approach

The previous section provided the evidence that there is no one-size-fits-all simulation model for different decision levels (strategic, tactical and operational) in air traffic flow management, and the existing simulation models cannot be utilized as the simulation component in our SbO framework. Therefore, a simulation model is developed that is capable of capturing the essential interactions of key system components, representing the system with sufficient detail, and reflecting the uncertainties associated with the runway operations.

The main purpose of the simulation model is to evaluate the given runway operations schedule in terms of runway utilization and fairness among aircraft. Several other performance measures can also be estimated by utilizing the simulation model, such as hourly delays, travel times, and queueing data, but the optimization component in our SbO framework only requires these two measures. Since a relatively small number of aircraft needs to be simulated, a discrete-event, microscopic and stochastic simulation model is considered which is suitable for our SbO framework. It is noteworthy to mention that discrete-event simulation, where system state changes happen discretely at isolated times so-called events, is the most widely used kind of simulation for design and analysis of runway operations.

In developing the discrete-event simulation model, system approach is adopted to represent the components of the real runway system and their complex interactions accurately. This approach allows decomposition of the real system into the functional components and application of an object-oriented architecture, which is a powerful approach for coping with the development of complex computer simulations. Furthermore, employing an object-oriented architecture to the simulation of real runway system yields advantages in the design as well as in the implementation of the simulation model. In this architecture, the simulation model is seen as a collection of objects with attributes that interact with each other via messages. The primary benefit of this architecture is that it partially mitigates the challenges presented by large computational time and memory requirements for the simulation of runway operations. Based on the object-oriented architecture, the simulation model developed according to the following steps:

**Step 1**: Identification of objects and collections of objects in the problem domain.

**Step 2**: System evaluation and the identification of the problem in a clear statement.

**Step 3**: Problem analysis, formulation, and identification of variable relationships.

**Step 4**: Determining the appropriate level of modeling sophistication, model building, and definitions of performance measures.

**Step 5**: Data acquisition and abstraction.

**Step 6**: Model translation into computer code (implementation).

**Step 7**: Model verification (whether the model does what it is intended for) and validation (whether the model is an accurate representation of the real system).

## 4.3 Structural Modeling

The most important part of simulation studies is considered as finding the right abstraction of the real-life system for the simulation model. This section lays out the abstraction and fundamental logic of the simulation model, and presents the conceptual model, which is developed to display the structure and the abstract behavior of the runway operations consistent with the purpose of the simulation model outlined in the previous section. Limitations and assumptions of the structural modeling are also discussed.

4.3.1 Outline and Conceptual Design

The simulation model simulates the arrival and departure segments i.e. both landing and take-off modes of runway operations are simulated, and they are represented by a network of nodes and arcs. Aircraft move on this network along prescribed trajectories that are made up of strings of nodes and arcs, where each arc can be occupied by a single aircraft at a time. Accordingly, whenever an aircraft tries to use an arc that is already occupied by another aircraft, delay takes place.

In actual runway operations, both arrival and departure aircraft, which fly under instrument flight rules (IFR) or visual flight rules (VFR) conditions follow a filed flight plan while it is in the terminal maneuvering area (TMA). This flight route is represented by a sequence of fixes (waypoints) in the airspace, and they have to attain approval and clearances from air traffic controllers throughout the flight, in practice. Based on the actual runway operations the abstraction and fundamental logic of runway operations in the simulation model is outlined below.

As shown in Figure 16, the arrival segment begins when aircraft enters the TMA (freezing horizon) through entry points. After entering the TMA, it remains in it until it arrives one of the available meter fixes. For each of these fixes an initial approach segment is defined, starting from the initial approach fix. In practice the length of the initial approach segment differs depending on the approach conditions (IFR or VFR); however, in the simulation model only IFR conditions are considered because these conditions depend on air traffic controllers to maintain adequate separation. Then, the initial approach segments lead aircraft into final approach segments. These segments differ in length according to the direction from which the aircraft approaches to runway. After landing aircraft exits from the runway, aircraft leaves the system and arrival segment concludes. Briefly, arrival aircraft starts from the entry points and it follows the sequence of fixes including meter fix, initial approach fix, final approach fix and stabilized approach fix through the runway, as given in a flow chart diagram in Appendix D.

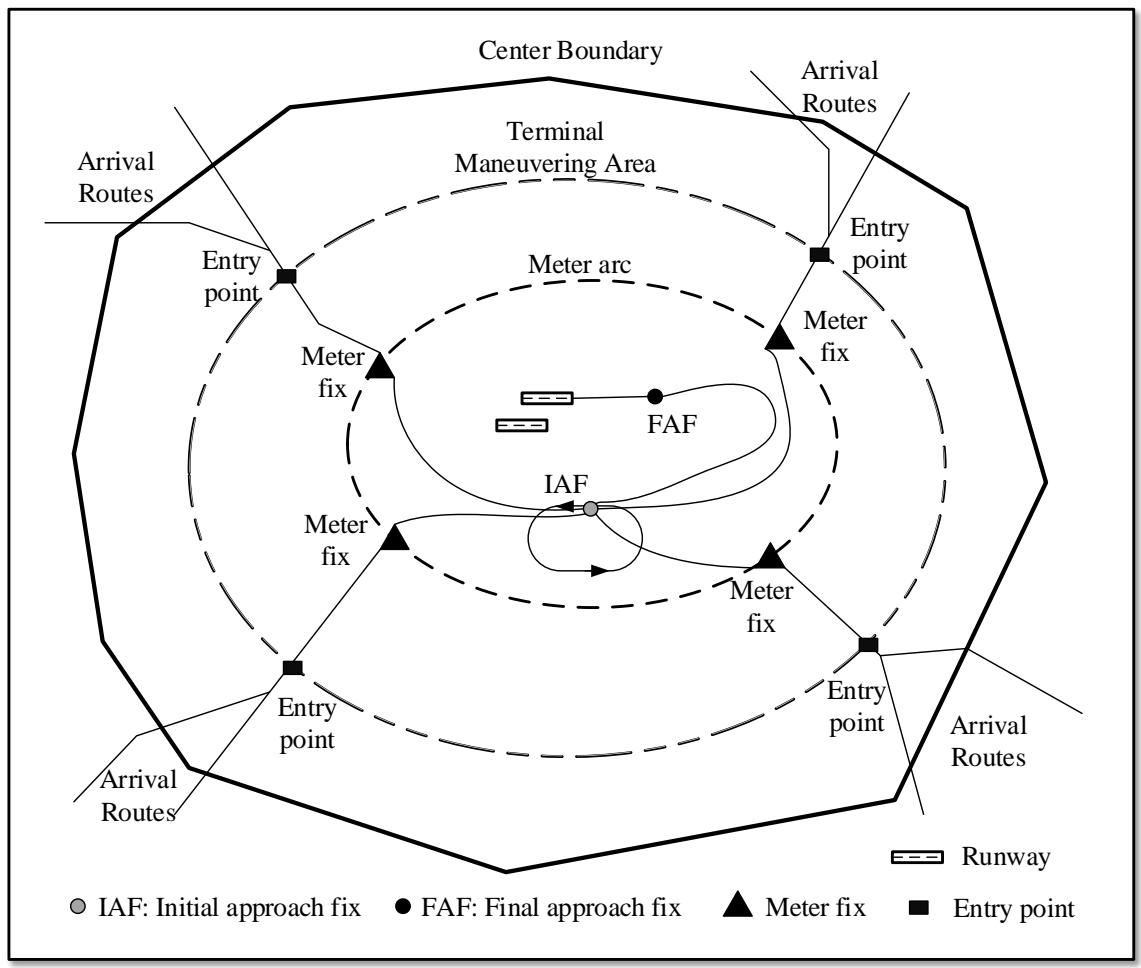

**Figure 16:** Sequence of Fixes in Airspace for Arrival Aircraft

As for departing aircraft, the departure segment begins when the aircraft enters to the holding area and ends when aircraft leaves TMA in a departure fix. After receiving runway clearance, aircraft starts moving with the intent to take off. As soon as it receives take-off clearance, aircraft wheels-off the runway, where take-off time corresponds to the wheels-off time for the aircraft. Then, it enters the initial climb segment, where at the end it reaches enough level of speed, and next, it enters the en-route climb segment, where aircraft maintain en-route speed. The departure segment concludes when an aircraft arrives departure fix and leaves the TMA. Similar to the arrival segment, a flow chart diagram of the departure segment given in Appendix D.

For arrivals, the primary nodes that are located in the network representation are shown in Figure 17 in which the flow of air traffic for arrival is represented as a network based on the arrival procedures. The primary nodes that are located in this network representation are listed below:

(a)    *Aircraft generation node* is the entry point that represents the point where aircraft are created and fed into the simulation.

(b)    *Meter (merge) fix node* represents the fix along an established route over which aircraft is metered prior to entering the TMA, where it is established at a distance from the airport which facilitate a profile descent.

(c)    *Initial approach fix (IAF) node* represents the fix that identifies the beginning of the initial approach segment, where aircraft flows merge and enter the holding pattern.

(d)    *Final approach fix (FAF) node* represents the fix from which the final approach to an airport is executed and which identifies the beginning of the final approach segment. The glide slope/path starts at this node, which is the location where all landing aircraft are required to pass before landing. The airspace between FAF and the runway is referred as final approach.

(e)    *Stabilized approach fix (SAF) node* represents the fix that starts maintaining constant flight conditions, such as speed, for the approach phase.

(f)    *Runway threshold node* represents the point across the runway that denotes the beginning and end of the designated space for landing.

(g)    *Touchdown node* represents the point at which the nominal glide path intercepts the runway and aircraft first make contact with the landing surface of the runway. (The landing time corresponds to the touchdown time for the aircraft.)

(h)    *Runway exit node* represents the point where aircraft depart the runway, and simulation model stops tracking them.

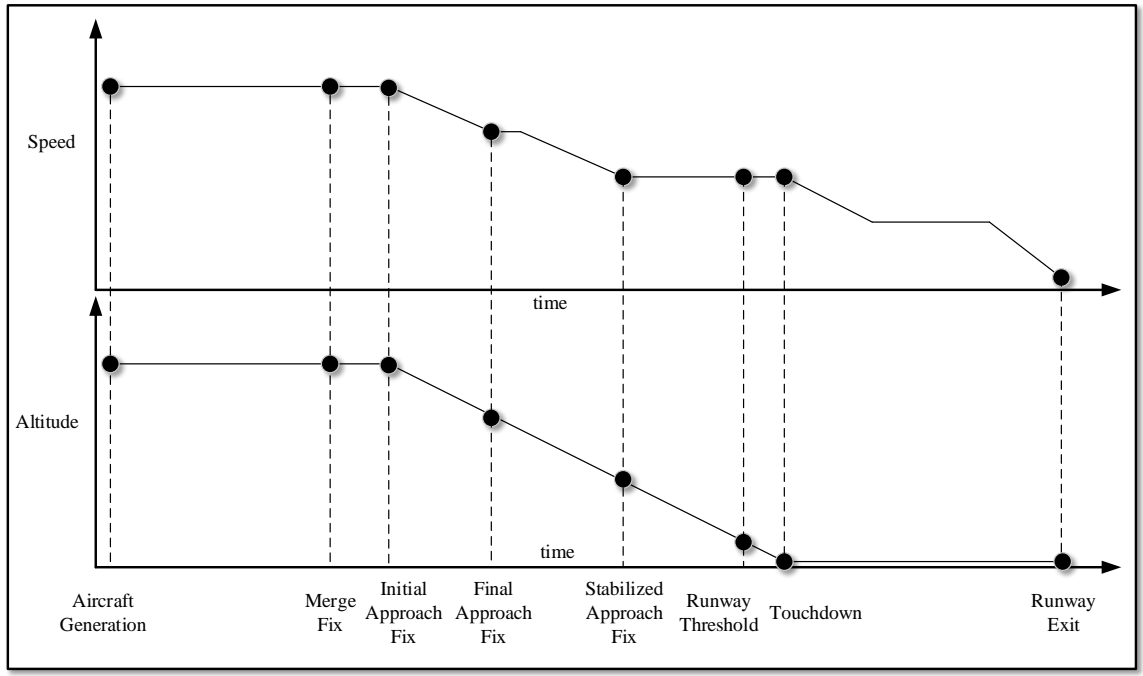

**Figure 17:** Primary Nodes for Arrival

For departures, the primary nodes that are located in the network representation are shown in Figure 18 in which similar to arrivals, the flow of air traffic for departures are represented as a network based on the take-off procedures. The primary nodes that are located in this network representation are listed below:

(a)    *Aircraft generation node* represents the point where aircraft are created in the holding area.

(b)    *Start of roll node* represents the point where an aircraft is aligned with the runway centerline and the aircraft starts to move with the intent to take off.

(c)    *Take-off node* represents the point at which aircraft wheels-off the runway. (The take-off time corresponds to the wheels-off time for the aircraft.)

(d)    *Initial climb node* represents the point that aircraft reach enough level of speed at the beginning of the initial climb segment.

(e)      *En-route climb node* represents the point at the beginning of the en-route climb segment where aircraft maintain en-route speed.

(f)      *Departure fix node* represents the point where aircraft leaves the TMA, and the simulation model stops tracking the aircraft.

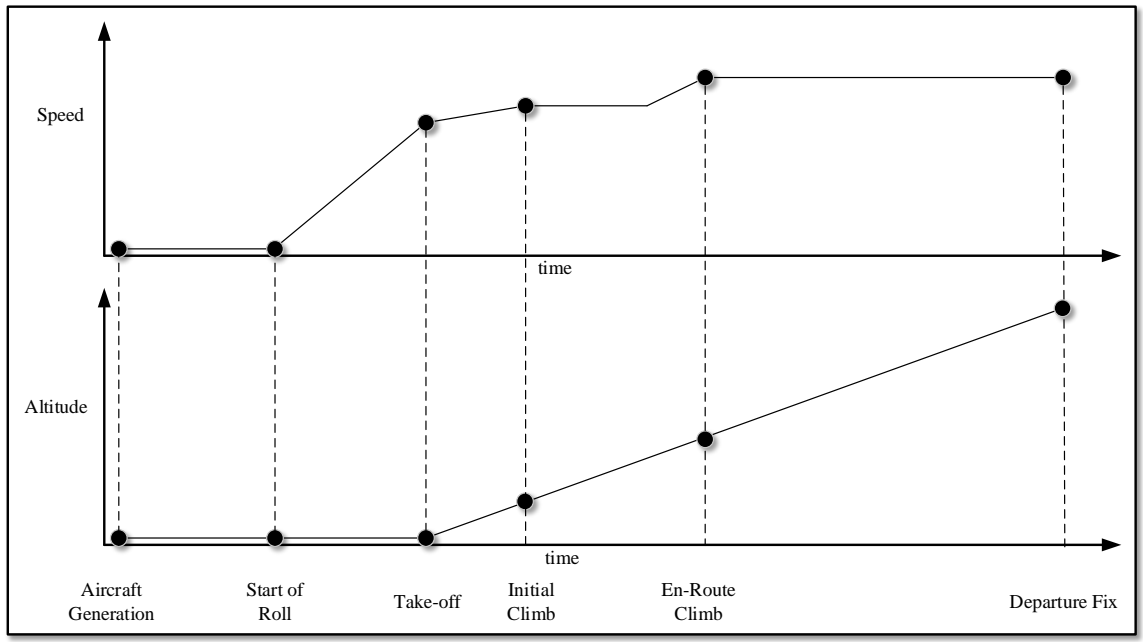

**Figure 18:** Primary Nodes for Departure

The primary advantage of this network structure is the detailed representation of the runway operations in the TMA and convenience in collecting various statistics. Essentially, the state of the simulation model changes only in discrete points in times, which is typically referred to as event times.

In the simulation model, the runway system is recognized as a terminating system and, hence, it is modeled as a terminating simulation where there is a specific starting and stopping conditions to reflect the actual operation of the real runway system. Aircraft are considered as dynamic objects (entities) that flow through the simulation model, and runways are considered as the main resource.

In arrival procedure, the scope of analysis includes the landing procedure starting from the arrival to the entry points, up to aircraft exit from the runway. Likewise, in departure procedure, the scope of analysis includes the take-off procedure starting from the arrival to the holding area, up to the departure fix where aircraft exit from the TMA. In each node that is located in this network representation, the separation time between the aircraft is checked based on time-based distance. If the separation time is not obeyed, then the following aircraft is delayed until the separation time is obeyed.

For arrivals, the holding pattern, which is an airspace section for aircraft waiting to continue the final approach to the runway, is modeled as the queue for the arrival aircraft where aircraft fly at a certain speed. In the simulation model, the holding pattern is used as an indicator for infeasibility. If more than a specific number of aircraft exist in the holding pattern at the same time or if any aircraft spends more than a threshold of time in the holding pattern that schedule is considered infeasible. Likewise, for departures, the holding area, which is an area close to the runway for aircraft waiting to start to roll, is modeled as the queue for the departure aircraft.

4.3.2 Assumptions and Parameters

Several assumptions are made to simulate the dynamic nature of the arrival and departure operations outlined in the previous sub-section. The primary assumptions are listed below:

(a)     The number of runways is assumed to be more than one, and in the runway configuration, it is assumed that there are no crossing active runways or closely spaced parallel runways.

(b)     A constant nominal approach and climb speed assumed depending on aircraft weight class. To simulate the practical variations in the system arrival times and introduce related practical uncertainty into the model, perturbations are imposed on the system arrival times for both arrivals and departures.

(c)      Weather restrictions, especially effect of wind is not considered because the data of wind grid is not available. Therefore, aircraft airspeed is assumed to be equal to ground speed. Also, change in runway configuration because of the crosswind is not taken into consideration.

(d)      It is assumed that there are four weight classes of aircraft and each aircraft belongs to one of these class. These four classes include the following: "Heavy", "B757", "Large", and "Small". In a recent change, FAA added a new weight class, designated as "Super" (FAA, 2014). However, this weight class is not considered in the simulation model, because this weight class of aircraft is not included in the fleet mix data that we obtained from FAA databases.

(e)      All arrival and departure aircraft assume to follow pre-defined trajectories through entry points to runway exit for arrivals, and through holding area to TMA exit for departures. Also, aircraft are not allowed to change their trajectory during the arrival or departure procedure once they are assigned a trajectory.

(f)      Air traffic (local) controllers' workload is not taken into account. Therefore, it is assumed that complexity of the air traffic control environment does not increase the controllers' workload.

(g)      Air traffic (local) controllers do not add buffer times to minimum separations requirements. The buffer is the additional spacing applied by air traffic controllers to account for the variability inherent in controlling aircraft. It is noteworthy to mention that even though buffer increases safety margins, it reduces runway throughput.

(h)      Minimum separation times and runway occupancy times are dependent on the type of aircraft weight, and the runway is equipped with a dedicated rapid exit for each type of aircraft.

### 4.3.3 Performance Measures

As in many real-life problems, the performance measures are not easy to capture in runway operations scheduling, but it is key for obtaining practically applicable results to the problem. The main performance measures to be collected via the simulation model are runway utilization and fairness, which are detailed below.

*Runway utilization*: Although most of the previous research on runway operations scheduling problem have been focused on minimizing delay, maximizing runway utilization (throughput) is not the same as minimizing delay. In practice, air traffic (local) controllers try to maximize runway utilization instead of to minimize total delay directly. Minimizing the delay will maximize the throughput over the long term but the opposite is not valid. Hence, maximizing runway utilization (minimizing makespan) is selected as one of the performance measures.

The runway utilization is defined in terms of the period of time that can accommodate a given number of runway operations, and it is calculated as the actual landing or take-off time of the last runway operation.

*Fairness*: Several models for Collaborative Decision Making (CDM) aspects of air traffic flow management have been proposed in the literature where information related to stakeholders (primarily airlines) is integrated to enhance decision-making process. Among the CDM aspects, fairness is typically considered as an important factor for all the stakeholders and a runway schedule that preserves the FCFS order has been agreed upon as fair.

Therefore, the second performance measure is motivated by social justice, where scheduling using the arrival order is considered fair. This metric attempts to measure the deviation from the FCFS order, in particular, fairness is calculated in the simulation model through the concept of "position shift" compared to FCFS sequence and the value of the difference between target landing/take-off time and start time. For the position shift concept, suppose a landing aircraft sequenced in the 4th position according to FCFS order. If it gets scheduled in the 7th position, then its position shift is +3 (behind), and likewise, if it is scheduled in the 2nd position, then its position shift is -2 (ahead) (Figure 19).

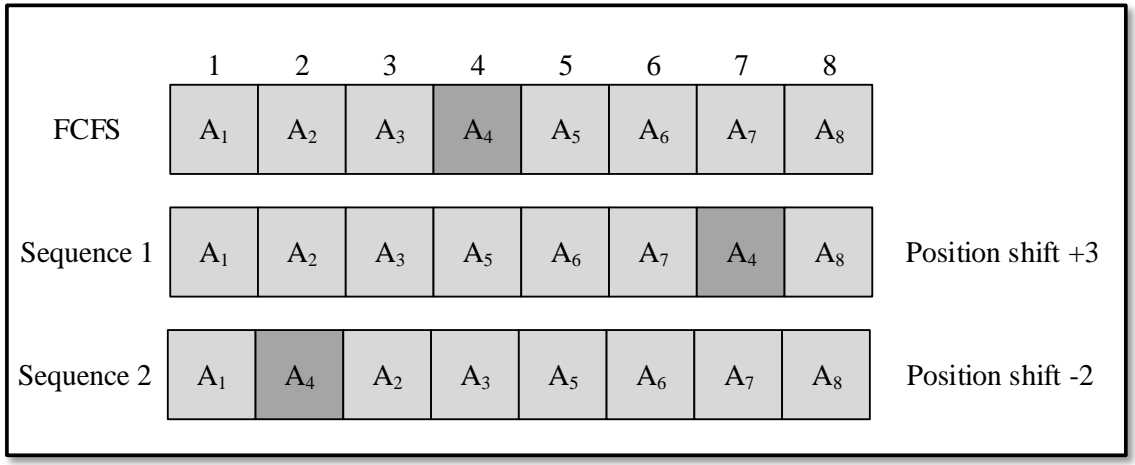

**Figure 19:** Representation of Position Shift Concept

The fairness is modelled by the square deviation from the target times with absolute values of aircraft $j$ position shifts as the weights as shown in Eq. 4.1, where $\delta_j$ represents the target landing/take-off time of aircraft $j$, $t_j$ represents the start time of aircraft $j$, and $w_j$ represents the position shift value of aircraft $j$.

$$\min \sum_{j \in N} \left| w_j \right| \left( t_j - \delta_j \right)^2 \qquad (4.1)$$

Since FCFS order generates aircraft groups that are closely sequenced with larger gaps between individual groups, it deteriorates runway utilization. On the other hand, in order to schedule aircraft for maximum runway utilization, total position shift from the FCFS order has to be increased. Therefore, there is a requirement to find trade-off solutions.

## 4.4 Quantitative Modeling (Input Data Analysis)

### 4.4.1 Data Sources and Design Parameters

Multiple types of data sources are utilized for Input Data Analysis of the simulation model. These data sources are listed below:

(a)      FAA Operations & Performance database (provides publicly available real-world historical data, and available online at *https://aspm.faa.gov*).

(b)      Bureau of Transportation Statistics (BTS) database (provides data reported by airlines, and available online at *www.rita.dot.gov*).

(c)      Flightstats (available online at *www.flightstats.com*) database.

(d)      Data obtained as a result of simulation runs with a validated and FAA approved simulation tool, namely the MITRE Corporation runwaySimulator.

FAA Operations & Performance database provides access to historical traffic counts, forecasts of aviation activity, and delay statistics, and this database consists of several FAA-based core databases. For input data analysis, only Aviation System Performance Metrics (ASPM) and the Airline Service Quality Performance (ASQP) core databases are used. The ASPM database includes information on individual flight performance, airport efficiency, and runway configurations for every quarter hour. The ASPM database is structured into two groups of data: (1) the Out of the gate, Off the ground, On the ground and Into the Gate (OOOI) flight data, as reported by the airlines, and (2) non-OOOI flight data.

As illustrated in Figure 20, OOOI data consist of the times of aircraft pushback from the gates, their take-off and landing times, and the gate-in times. Most of the airlines report OOOI data for the majority of their flights. OOOI data is commonly considered as reliable because the data is collected automatically using input from Aircraft Communications Addressing and Reporting System (ACARS) sensors mounted on the aircraft. However, OOOI data is only reported to FAA by the airlines that have the aircraft with ACARS sensors, which are commonly referred as OOOI airlines.

On the other hand, the ASPM database estimates OOOI data for flights of non-OOOI airlines and for OOOI airlines where OOOI data are not available, which include the non-OOOI data. Considering that ASPM database estimation may not be accurate, non-OOOI data is detected and corrected by using the Flightstats database when it is available.

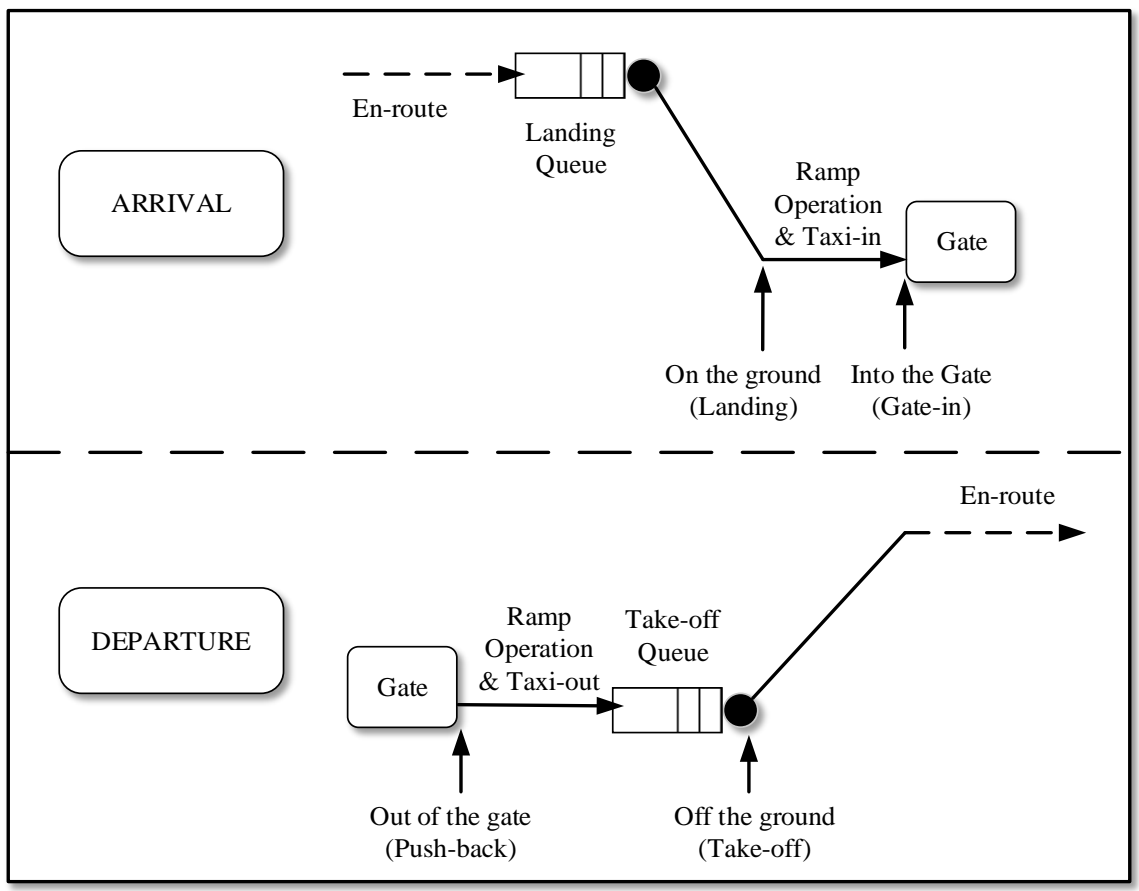

**Figure 20:** Illustration of OOOI Data

In addition to the aforementioned OOOI and non-OOOI data, ASPM database provides airport-level aggregate data, which enumerates the total number of arrivals and departures in 15-minute increments. This data is used only for validating the simulation model. The transit times between nodes in the simulation are obtained by analyzing the navigation information obtained by experimenting with the MITRE Corporation runwaySimulator.

The design parameters for the discrete-event simulation model are listed in Table 2. Among these design parameters of the simulation model, runway layout, procedures and initial and final approach segment length and speed are airport specific parameters. On the other hand, minimum separation requirements and runway occupancy times are not airport specific parameters, and the details of these two parameters and how they are determined are outlined in the following sub-sections. The remaining design parameters, fleet mix and operating sequence, is given as part of input data.

**Table 2:** Simulation Model Design Parameters

| Design Parameter | Description |
|---|---|
| Runway layout | Number of runways and configuration |
| Runway procedures | The way runways are operated |
| Fleet mix and operating sequence | The percentage of operations among all aircraft based on weight classes and their arrival/departure sequence |
| Minimum separation requirements | The required minimum distance/time between leading and trailing aircraft |
| Runway occupancy times | The time difference between when an aircraft crosses the runway threshold, and when it clears the runway |
| Initial and final approach segment length and speed | The distance of the initial and final approach segments and the speed of the aircraft |

The following are given as an input to the simulation model: (1) number of aircraft and a set of attributes for each aircraft including aircraft identification number, operation type, weight class, a maximum delay time for each aircraft as a hard constraint, arrival times to entry points and holding area for arriving and departing aircraft, respectively, (2) aircraft sequence for each runway (runway assignments), and (3) estimated landing and take-off times for each aircraft. And the output of the simulation model is the performance measures associated with the given schedule, which are runway utilization and fairness.

These input data are used in the simulation model as they are specified in the schedule generated by the optimization model since there is no internal mechanism in the simulation model logic that can change these attributes during simulation run. For instance, runway operations schedule and runway assignments cannot be changed to maximize utilization or to increase fairness since the input schedule is utilized as a base for evaluating its performance. The overall working mechanism of the simulation model is explained below.

A set of arrival and departure aircraft is generated at each iteration of the simulation, where each aircraft is defined by two features: a scheduled runway operations time (landing or take-off) and a weight class from among the four FAA weight class based on the provided fleet mix. The scheduled runway operations time corresponds to the time at which the aircraft can cross the runway threshold for arrival aircraft. By the same token, this time, corresponds to the time at which the flight can begin its take-off roll for departure aircraft.

4.4.2 Minimum Separation Requirements

Minimum separations are required by FAA to maintain the safety of runway operations, and they are intended to prevent aircraft collisions as well as eliminate the hazard to aircraft that subject to wake turbulence of a leading aircraft. According to FAA safety regulations, there exist two types of minimum separation requirements enforced by the FAA: (1) an airborne minimum longitudinal separation, which is mostly given as distance, and (2) a minimum separation at the runway threshold, which is typically given as time. The second type of minimum separation requirements are especially important for controlling the risk of simultaneous runway occupancy and collision.

The most important factor that needs to be ensured during the simulation is that the aircraft are separated by at least the minimum distance from the previous ones on the same runway and all other dependent parallel runways. In the simulation model, separation requirements are utilized in accordance with the FAA Aircraft Wake Turbulence Advisory Circular 90-23G as a default. The FAA wake separation standards are given in Table 3. The FAA uses five aircraft weight classes for wake turbulence separation minima: "Super", "Heavy", "B757", "Large", and "Small". "Super" is a relatively new class added to the classification that has been approved on an interim basis for aircraft type, such as the Airbus A380. Also, the Boeing "B757" was previously classified as a "Large", but because special wake turbulence separation criteria apply to this type of aircraft, it is classified as a separate weight class.

**Table 3:** FAA Minimum Separation Standards in NMs
(Source: FAA (2014))

| Leader/Follower | Super | Heavy | B757 | Large | Small |
|---|---|---|---|---|---|
| Super | MRS | 6 | 7 | 7 | 8 |
| Heavy | MRS | 4 | 5 | 5 | 6 |
| B757 | MRS | 4 | 4 | 4 | 5 |
| Large | MRS | MRS | MRS | MRS | 4 |
| Small | MRS | MRS | MRS | MRS | MRS |

MRS: Minimum Radar Separation

The FAA minimum separation standards depend on runway operations type (landing or take-off), the weight class sequence, the runway configuration and runway assignments, and the flight rules in use (IFR or VFR). For runway operations under VFR rules, the FAA does not enforce numerical minimum separation, where pilots are responsible for maintaining the separation visually. In the simulation model, only runway operations under IFR conditions are simulated rules since these conditions depend on air traffic controllers to maintain adequate separation.

Most of the FAA minimum separation standards are described distance-based instead of a time-based separation. Since time-based separation requirements are expected to be implemented, distance-based separations are converted to time-based separations. This conversion is done by assuming a 5 NMs final approach path and a nominal approach speed, i.e. each aircraft flies at a constant speed over the final approach path. This assumption is commonly used in the literature and apparently adequate for converting the distance-based separation into time-based one.

The ground speed through the final approach depends on the headwind; as the headwind is higher, aircraft needs less thrust to maintain the necessary lift. Table 4 is the summary of the observed samples of average ground speed for different types of aircraft from the FAF to the runway threshold when instrumental landing system is in use. "Super" weight class is excluded from the minimum separation requirements table, because this weight class of aircraft is not included in the fleet mix data that we obtained from FAA databases.

**Table 4:** Nominal Approach Speeds in knots

| Runway Operation Speed | Heavy | B757 | Large | Small |
|---|---|---|---|---|
| Arrival Speed | 150 | 130 | 130 | 90 |
| Departure Speed | 170 | 150 | 150 | 100 |

The calculated minimum separation requirements for runway operations on the same runway are given in seconds in Table 5a, where the leading aircraft is given by the rows, and the trailing aircraft is given by columns. In Table 5b the separation requirements for parallel runways depending on their spacing are shown.

**Table 5:** Minimum Separation Times in seconds

| Departure → Departure | | | | | Departure → Arrival | | | | |
|---|---|---|---|---|---|---|---|---|---|
| Leader / Follower | Heavy | B757 | Large | Small | Leader / Follower | Heavy | B757 | Large | Small |
| Heavy | 60 | 90 | 120 | 120 | Heavy | 50 | 53 | 55 | 65 |
| B757 | 60 | 60 | 90 | 90 | B757 | 50 | 53 | 55 | 65 |
| Large | 60 | 60 | 60 | 90 | Large | 50 | 53 | 55 | 65 |
| Small | 60 | 60 | 60 | 60 | Small | 50 | 53 | 55 | 65 |
| Arrival → Departure | | | | | Arrival → Arrival | | | | |
| Leader / Follower | Heavy | B757 | Large | Small | Leader / Follower | Heavy | B757 | Large | Small |
| Heavy | 75 | 75 | 75 | 75 | Heavy | 96 | 133 | 157 | 196 |
| B757 | 65 | 65 | 65 | 65 | B757 | 74 | 107 | 133 | 157 |
| Large | 55 | 55 | 55 | 55 | Large | 60 | 65 | 69 | 131 |
| Small | 40 | 40 | 40 | 40 | Small | 60 | 65 | 69 | 82 |

(a) Minimum separation times for operations on the same runway

| Runway spacing | Departure → Departure | Departure → Arrival | Arrival → Departure | Arrival → Arrival |
|---|---|---|---|---|
| up to 2500 ft (up to 760 m) | As on single runway | As on single runway | Independent | As on single runway |
| 2500 ft – 4300 ft (760 m – 1310 m) | Independent | Independent | Independent | 40 |
| more than 4300 ft (more than 1310 m) | Independent | Independent | Independent | Independent |

(b) Minimum separation times for operations on parallel runways

In the simulation model, arrival and departing aircraft are generated at discrete times as specified by the given runway operations schedule. Whenever an aircraft's scheduled time of landing does not respect the minimum separation requirements, this aircraft is delayed. Likewise, if a departing aircraft's scheduled time of take-off violates the minimum separation times, it is delayed at the holding area.

### 4.4.3 Runway Occupancy Times

Although in the literature few research considered runway occupancy in scheduling runway operations, runway occupancy has a strong impact on runway throughput. In the simulation model, a couple of logical requirements are taken into account related to runway occupancy. First of all, aircraft are not allowed to occupy the same runway at the same time. Second, in order for an aircraft to execute a runway operation (landing or take-off), the previous aircraft must have exited the runway. This required time period that is the time an aircraft occupies the runway is commonly referred as runway occupancy time (ROT). For arrival aircraft, ROT starts when aircraft passes runway threshold and ends when aircraft exits the runway. For departure aircraft, ROT starts when aircraft enters the runway and ends when the aircraft passes the departure end of the runway.

Several researches have been conducted recently on estimating and measuring ROT. In earlier researches, observational data occasionally was not sufficient to conclude statistically significant results, and there was considerable uncertainty in measurements. Lee et al. (1999) analyzed ROTs at a busy US airport using the NASA Dynamic Runway Occupancy Measurement System (DROMS), which is an automated tool that collected ROTs for over 3000 arriving aircraft. They concluded that ROTs are dependent only on aircraft weight and speed, and there is no significant difference between airlines or head-wind/tail-wind conditions. Therefore, in the simulation model ROTs are considered as dependent only on aircraft weight class and aircraft speed.

The missed approach procedure (procedure to be followed if an approach cannot be completed to a full-stop landing) and Land and Hold Short Operations (LAHSO) (procedure that requires pilot participation for landing and holding short of an intersecting runway or point on a runway to balance airport capacity and system efficiency with safety) are not modeled in the simulation model, since these procedures are rarely utilized in practical runway operations, and they require control mechanisms which further complicates the model unnecessarily.

4.4.4 Random Variables

There are various factors that cause uncertainty in runway operations, such as ground speed variations caused by the wind, piloting indecisions, delays in pushback or taxiing, airport congestion, etc. Also, unexpected events such as safety incidents, equipment failure, inclement weather, etc. also contribute to uncertainty. All these factors can result in variability in target landing/take-off times, runway occupancy times, etc. (C. Brinton & Atkins, 2009). In the simulation model, these stochastic processes are simulated by utilizing random variables to reflect the stochastic behavior observed in actual runway operations. Given a solution (a runway operations schedule), the performance measures are evaluated stochastically with simulation runs by using particular values (realizations) of the following random variables (three sources of randomness).

(a)     *Arrival times to entry points and holding area:* In order to simulate the practical variations in the system arrival times and introduce related practical uncertainty into the model, perturbations are imposed on the system arrival times, i.e. arrival times to holding area for take-offs and arrival times to entry point for landings. In each simulation run, the input schedule generated by the optimization component is used to control the insertion of aircraft into the simulation model. In each simulation replication, the system arrival times for both arrival and departure aircraft are perturbed by the addition of a random lateness distribution, which may include negative values. In both current and previous works, the perturbations to both arrival and departure aircraft are assumed to follow a truncated normal distribution, which is confined between an upper and a lower bound. Departing aircraft have a mean of -30 seconds and a standard deviation of 1.5 minutes; arriving aircraft have a zero mean with a standard deviation of 30 seconds (Xue & Zelinski, 2014).

(b)     *Transit times between nodes*: In order to account for practical uncertainties that stem from ground speed variations caused by the wind, piloting indecisions and other unexpected events during transit times, an additive perturbation imposed to average transit times. The transit times between each node in the network are estimated as a function of the aircraft weight class. The values for these transit times are approximated by determining the vectoring the nodes and using approach profiles and standard fix speeds. The aircraft are assumed to fly directly from one node to another. The transit times are obtained by analyzing the navigation information obtained from *www.airnav.com*, and experimenting with the MITRE Corporation runwaySimulator.

As a result of the analysis, additive perturbations are assumed to be normal distributed with a zero mean and a standard deviation of 1.8 minutes. The statistical validity of this assumption is analyzed with a chi-square goodness of fit test. The chi-square test confirmed that perturbations are normal distributed at significance level of $\alpha = 0.05$.

*(c)*    *Runway occupancy times*: As mentioned previously, runway occupancy times (ROT) is the length of time required for a landing aircraft to proceed from the runway threshold to a point clear of the runway, and for a take-off aircraft to proceed from the runway and to a point when the aircraft passes the departure end of the runway. In order to account for uncertainties stemming from unexpected events during landing and take-off, perturbations are imposed to average ROTs calculated by the analysis of historical data for the airport under consideration (only for IFR conditions).

Ghalebsaz-Jeddi et al. (2009) provided statistical analysis of the ROT for a major US airport, and preferred the beta distribution for ROT because it has lower and upper bounds as in real situations for ROT. For both the early and late exits, the normal distribution is rejected at significance level of $\alpha = 0.05$. Distribution of ROT depends on the aircraft weight class as smaller aircraft exit earlier and larger ones later. The beta distribution shape parameters $(\beta, \alpha)$ of ROT for each aircraft weight class is given in Table 6 (Kolos-Lakatos, 2013; Kumar et al., 2009).

**Table 6:** Beta Distribution Shape Parameters for ROT

| Shape parameters | Heavy | B757 | Large | Small |
|---|---|---|---|---|
| Beta ($\beta$) | 12.03 | 12.03 | 12.42 | 12.42 |
| Alpha ($\alpha$) | 27.48 | 27.48 | 26.86 | 26.86 |

## 4.5 Object-Oriented Design

In general, developing a simulation model for runway operations requires analysis of many complex factors including airport's layout, runway operation procedures, fleet mix, the characteristics of the various aircraft using the runway, etc. However, the complexity of the simulation is minimized by employing an object-oriented design, which has been applied in a wide variety of domains to simulate real-life complex systems. The object-oriented design of the simulation model is based on Leathrum (2014).

The primary advantages of this kind of design are that it allows managing the inherent complexity by breaking the system into various objects, and it promotes reusability of existing objects. Objects are the data structures that encapsulate a state, which is the value of its attributes, and behavior, which constitutes its methods. This kind of design also provides a high-level of flexibility through its modular design, which can be employed to model any runway system, operating in any configuration. The general structure and design of the components and the whole simulation model is illustrated with block diagrams. In these diagrams, objects are represented by rounded boxes, while methods of individual objects are represented by ovals or circles.

The simulation model consists of two interconnecting and interrelated modules, namely "Simulation Executive" and "Simulation Application", which are detailed below:

*Simulation Executive*: This module acts as the main controller of the simulation model that manages the model when executing. It schedules the events and runs the simulation by providing a set of events as objects. Simulation Executive maintains an "Event List" where the future events are stored in a list ordered by their execution time. The time interval for advancing the simulation time is determined by the events in the Event List, which is the foundation of the event-driven simulation approach. Simulation time is incremented at the execution of the next event without considering the time interval between consecutive events. The execution time of a current event becomes the value of the simulation time. As soon as Event List becomes empty, simulation terminates. The architecture of the Simulation Executive is illustrated in Figure 21. As shown in this figure, the Simulation Executive has two main methods: Schedule Event and Run Simulation. Event List object maintains the event queue, updates the simulation time and schedules changes of states throughout the simulation period.

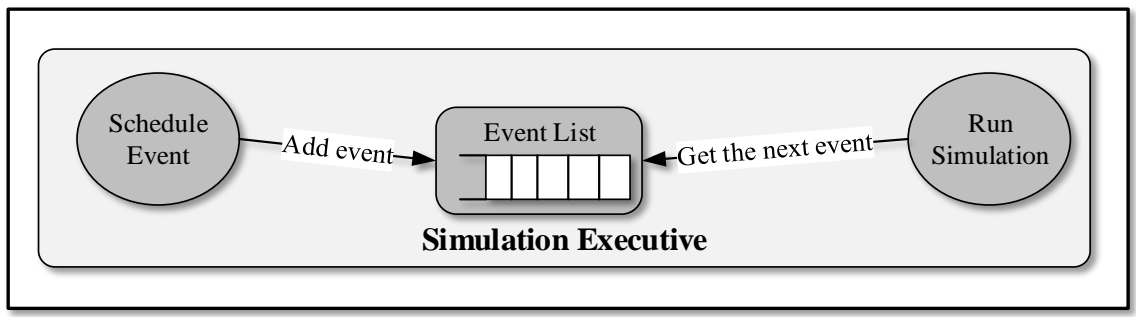

**Figure 21:** Architecture of the Simulation Executive

*Simulation Application*: This module includes all the application elements that are controlled by the Simulation Executive. Simulation Executive and Simulation Application components are interfaced by using objects named "Simulation Object", which provide connection to the Simulation Executive. This object is inherited by all application objects need access to the functionality of the Simulation Executive. Architecture of the simulation application is illustrated in Figure 22.

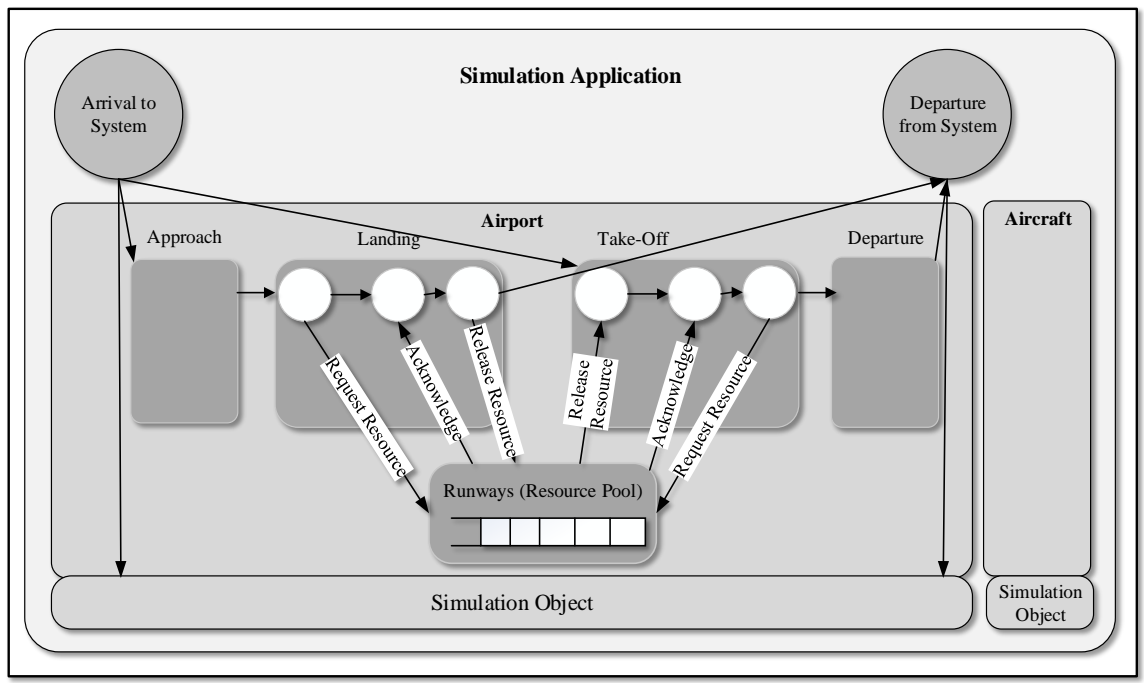

**Figure 22:** Architecture of the Simulation Application

The object-oriented design of the Simulation Application consists of the following object types:

(a)    *Aircraft object:* This object is the main active process object flows through the simulation.

(b)    *Airport object*: This object maintains the tasks (Approach, Landing or Take-off, Departure) for the aircraft objects.

(c)    *Resource pool object*: This object represents the runways. If any aircraft object needs a resource instance from a resource pool, it requests a resource from the resource pool object. If any resource is available, then resource pool acknowledges the aircraft object to acquire the resource; otherwise, this request will be put in the queue inside the resource pool object. As soon as the resource utilization is done by the aircraft object, it releases the acquired resource. Resources in the resource pool can only be acquired by an aircraft object at a time. A representation of the resource pool is given in Figure 23.

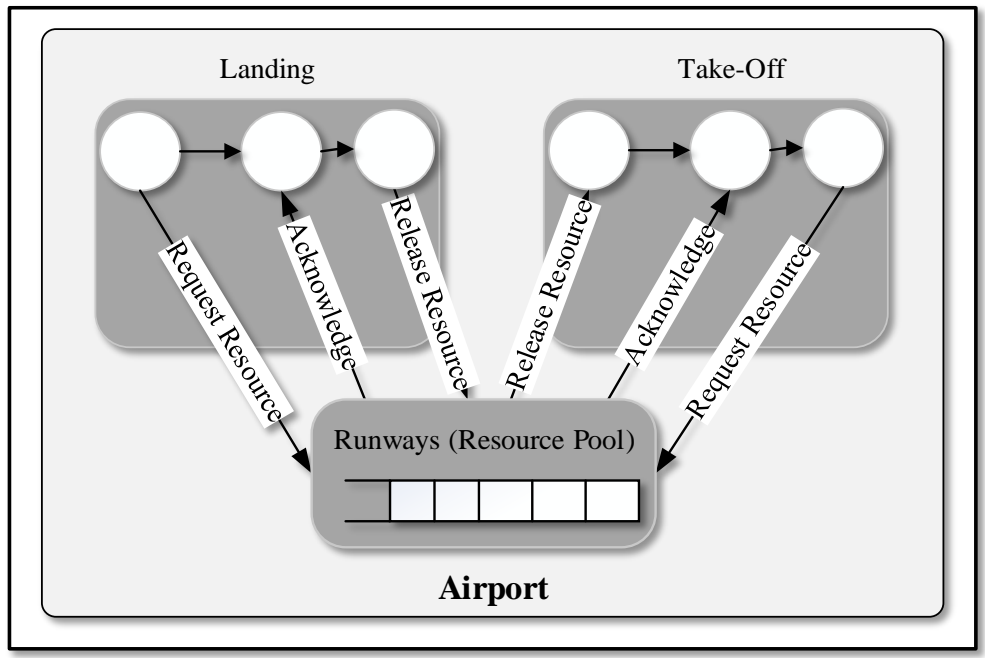

**Figure 23:** Representation of the Resource Pool

All objects have their data structures and methods that represent their attributes and procedures, respectively. Also, all objects interact with each other via messages during the simulation period.

The high-level block diagram (architecture) of the discrete-event simulation model is given in Appendix C.

## 4.6 Implementation Specifics

The object-oriented design extremely simplying the implementation, due to the fact that functional modules are self-sufficient and connected with each other through well-defined interfaces. The implementation is capable of controlling simulation model externally and passing data in real-time. Several implementation specifics are described below. Unified

Modeling Language (UML) is used to describe object-oriented elements, which includes a set of logical rules for representing the real system in a semi-graphical form.

The main challenges faced during the implementation phase are listed below:

(a)     How to instantiate the simulation executive once and provide a reference to the resulting object on each subsequent instantiation for scheduling events?

(b)     How to schedule execution of an event at some point in future which is the main feature of the Simulation Executive module?

(c)     How to assign unique identifiers to each entity (object) of a given class within the simulation?

The first two challenges mentioned above are dealt with two design patterns: "singleton" and "command design patterns", and the third challenge is overcome by "static attributes". Design patterns are abstract structures of classes and commonly utilized in object-oriented design and implementations. As explained below, these design patterns and static attributes facilitate corresponding implementation difficulties in a more systematic way.

*Singleton Design Pattern*: This design pattern involves a single class which is responsible for generating an object while ensuring that only a single object is generated. This design pattern allows creating a single simulation executive. In the first initiation of the singleton class, simulation executive is created. In the next initiations of the class, instead of creating an instance of the class it only provides a reference to the simulation executive. As illustrated by a UML diagram in Figure 24, design pattern does not allow access to the class constructor; rather it allows access to a static method, which creates a single object when it is called the first time and in the subsequent calls it provides a reference to that single object. In the implementation of the simulation model, the Simulation Executive component is created as soon as the simulation is initialized, and then, whenever a simulation object is created that needs access to the Simulation Executive for scheduling events, it receives a reference to the Simulation Executive.

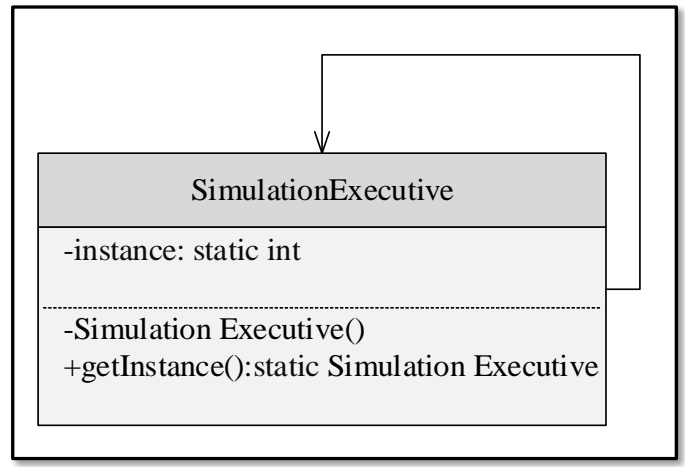

**Figure 24:** UML Representation of Singleton Design Pattern

*Command Design Pattern:* The main feature of the Simulation Executive component is that execution of an event has to be scheduled at some point in future, which requires separating the object that triggers the event from the object that executes the event. In order to fulfill this requirement, the Simulation Executive component has to possess the capability to encapsulate all information needed to execute an event in future and trigger an event at a later time. This information includes the object and method to be called as well as the values for method parameters. The command design pattern is employed to implement this requirement conveniently.

The command design pattern is a behavioral design pattern which is driven by data. It achieves the required separation in the Simulation Executive component by creating an abstract base class that maps a receiver (an object) with an action (a pointer to a member function). The base class contains an "Execute" method that simply calls the action on the receiver. As illustrated by a UML diagram in Figure 25, "Event" class is defined such that the class inheriting it defines the functionality of the "Execute" method.

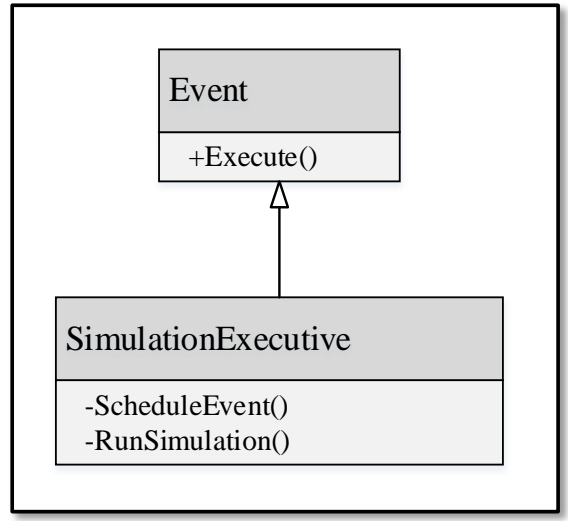

**Figure 25:** UML Representation of Command Design Pattern

*Static Attributes*: Static attributes are the ones that are associated with a class, not objects of that class, and these attributes are shared by all objects (instances) of that class. Static attributes fulfill the need for assigning unique identifiers to each entity (object) of a given class within the simulation. In particular, static attributes are utilized to identify an entity during collection of statistics related to this entity. Therefore, a static attribute is defined for the next available identifier to facilitate the access of all objects created from the class, which assigns a unique identifier to each object and increments this identifier as soon as the object is created.

All necessary input parameters as well as control parameters, such as random number seeds, are provided by the user through the parameter list section of the code. After simulation initiation, it executes the same model several times with different input and control parameter settings; thus, it provides an automated execution of multiple experiments. In addition, all necessary probability functions are implemented in order to simulate various statistical distributions, particularly the uniform, Poisson, normal, beta and exponential distributions.

The "Event List", which contains all events that are scheduled during the time of a simulation run, is implemented with a data structure based on a simple linked-list. Since events on the Event List need to be ordered depending on each event's scheduled execution time at each update, the linked-list implementation provides more efficient execution compare to other data structures.

A low-level programming language, namely C++, is selected as the implementation language for several reasons including the following: (1) it is an object-oriented programming language, (2) it has powerful features, such as pointers and Standard Template Library, and (3) it is capable of implementing system level designs such as the simulation executive. No graphic representation or Graphic User Interface (GUI) is implemented as part of the simulation model to avoid further complexity of the implementation.

**4.7 Verification and Validation Study**

As part of simulation model development, an iterative verification and validation process is used to determine whether the simulation model is valid to an acceptable level. Several versions of the simulation model were developed until a valid simulation model was obtained. The simulation model's validity was gradually improved through the process by increasingly building confidence in the accuracy of the model by applying verification and validation tests.

4.7.1 Verification Study

Verification study is done to ensure that computer implementation works as described in the conceptual model. Because a general-purpose programming language, i.e. C++, is utilized for implementation instead of a simulation software package, development time as well as verification time increased substantially. However, it reduced the simulation execution time significantly.

Since simulation model is designed and implemented using the object-oriented technique, verification study is mainly focused on ensuring that the simulation model functions (the time-flow mechanism, pseudo-random number generator, and random variate generators) and the computerized simulation model are implemented and coded correctly. Also, since implementation is done in Microsoft Visual Studio 2013 environment, debugging capabilities of this environment is used to check for logical errors at various stages of the implementation.

As part of verification study, each member function of the classes is debugged separately to ensure it works in accordance with the corresponding function in the conceptual model. In addition, simulation experiments are performed with the known deterministic data instead of stochastic data to verify that simulation model outputs plausible results. Finally, structured walk-throughs of outputs and deterministic run techniques are used to determine that the model is programmed accurately, and to test whether pseudo random number and random variate generators are implemented correctly. Furthermore, the outputs are compared with outputs of simulation runs performed by the MITRE Corporation runwaySimulator using the same airport data. Consequently, adequate evidence obtained from the verification study to conclude that computer implementation is an accurate representation of the logical behavior of the conceptual model.

4.7.2 Validation Study

Validation commonly regarded as a crucial step in simulation studies, since it tests simulation model predictions against reality and ensures that model is an accurate representation of the real system. Therefore, special emphasis is given to the validation study.

The validation study is conducted in two phases of validity analysis, including face validity and statistical validity checking. In face validity, an independent assessment of the appropriateness of the model structure and plausibility of the assumptions is conducted with the help of two subject matter experts. One of the experts is an aviation professional with more than 20 years of experience as an operator, and the other one is an experienced analyst in the industry. Both experts evaluated the model structure, where their judgment constituted a crucial component of the validation study. As a conclusion, the subject matter experts concluded that results produced by the simulation model are appropriate and reasonable, and model is a valid representation of the real-life runway system.

Statistical validity checking is conducted by comparing model outputs with the actual data obtained from the real system outputs. The actual data is obtained from FAA Aviation System Performance Metrics (ASPM) database, which is a part of FAA Operations & Performance Data. This database consists of 15-minute arrival and departure counts, weather conditions, and detailed information on individual flights based on runway operation times as provided by airlines through Airline Service Quality Performance (ASQP) data or Enhanced Traffic Management System (ETMS) messages. The data is obtained only for Washington Dulles International Airport (IAD) covering the year 2015. Given the complexity of runway operations, validating the discrete-event simulation model was a challenge, and statistical comparison of numerical values of the output performance measures to the real-life runway system was conducted to overcome this challenge.

Statistical validity checking divided into different steps. In the first step, the hours to be analyzed are determined by simply bundling 15-minute arrival and departure counts into hourly counts, and eliminating the non-busy hours according to certain criteria. These criteria include the following: (1) if the primarily used runway configuration was not used for the entire hour, and (2) if the demand for the hour is not the highest average demand for the day. In the second step, 20 hours are selected from the remaining set of hours randomly. In the third step, runway utilization estimates are obtained by running the simulation model. The actual data and simulation outputs are compared based on only runway utilization as an index for the simulation accuracy. Since the data related to the position shifts of aircraft compared to FCFS sequence is not available, this output measure of the simulation model is not evaluated.

After simulating 20-hour period in the simulation model, runway utilization values are collected (Let $Y_j$ be the random variable defined on the $j^{\text{th}}$ set of simulation model data).

Then, actual runway utilization values are extracted from the historical data (Let $X_j$ be the random variable defined on the $j^{th}$ set of actual data). In order to determine if simulation model is a valid representation of the real runway system, a confidence interval approach is employed to identify the statistical difference between two sets of data. Since a confidence interval approach provides more information, it is preferred to corresponding hypothesis test.

It is assumed that $Y_j$'s were generated by independent replications, and $X_j$'s are homogeneous with mean $\mu_y$ and $\mu_x$, respectively. We compared the simulation model with the real system by constructing a 95 percent confidence level for $\zeta = \mu_x - \mu_y$. The results of the paired-$t$ test are given in Table 7, where it is assumed that $W_j$ is the difference between actual value and simulation model value ($W_j = X_j - Y_j$).

**Table 7:** Results of the Paired-$t$ Test

| Statistics | Real Values ($X_j$) | Simulation Values ($Y_j$) |
|---|---|---|
| Mean | 3286 | 3192 |
| Variance | 241 | 213 |
| Number of observations | 20 | 20 |
| 95 percent confidence interval for $\zeta$ | -94 ± 107 | |

As a result, since the 95 percent confidence interval for $\zeta$ (-201, 13) contains 0, the observed difference between the mean runway utilization for the real system and the mean runway utilization for the simulation model is not statistically significant.

Similar to runway utilization, simulated flight delays are compared with the real system flight delays, which are calculated by the difference between the actual landing/take-off time and the corresponding scheduled time. Likewise, a confidence interval approach is employed to identify the statistical difference between two sets of delay data. After applying the paired-$t$ test, it is concluded that the difference between the mean delays for the real system and the mean delays for the simulation model is not statistically significant at a significance level of $\alpha = 0.05$.

Furthermore, the practical significance of the differences is evaluated by the two subject matter experts, who also supported the face validity, and they concluded that the differences are not practically significant as well.

# CHAPTER 5

# HYBRID TABU/SCATTER SEARCH ALGORITHM

Over the last several decades, interest in metaheuristic algorithms in solving multi-objective optimization (MOO) problems has risen considerably among researchers, and they have become more widely accepted as a viable alternative to exact methods. However, it is still a challenging task to develop an efficient metaheuristic algorithm for generating Pareto-optimal solutions, even for relatively easy bi-objective optimization problems. Furthermore, this difficulty is exacerbated in the context of simulation-based optimization (SbO) because of the noise stemming from the simulation component, which can easily render the optimization process unstable. This additional challenge can be compensated for by performing multiple simulation runs for each optimization iteration; however, this compensation will most probably result in long computational times. Therefore, developing an efficient metaheuristic algorithm for simulation-based multi-objective optimization requires finding a balance between intensification and diversification mechanisms in the design of such an algorithm.

This chapter presents the novel hybrid Tabu/Scatter Search algorithm, which is capable of finding a compromise between the quality of the obtained solution and the computational time requirements when used for simulation-based multi-objective optimization. The proposed algorithm generates solutions by using an elitist strategy to preserve non-dominated solutions, a dynamic update mechanism to produce high-quality solutions and a rebuilding strategy to promote solution diversity.

The first section presents a short introduction to the field of metaheuristics in order to provide a basis for terminology and a general classification, and also, foundational metaheuristic algorithms are discussed for the sake of completeness and better understanding the capabilities of these algorithms. Then, details of the Scatter Search (SS) algorithm template are outlined. In addition, salient features of Multi-Objective Evolutionary Algorithms (MOEAs) are briefly described, and a MOEA, namely the Elitist

Non-Dominated Sorting Genetic Algorithm (NSGA-II), is presented due to its wide utilization in the literature. Afterwards, mechanics of the proposed hybrid Tabu/Scatter Search algorithm along with its main methods as well as its differences from the traditional SS algorithm template are provided. Finally, implementation specifics of the algorithm are presented, concentrating on object-oriented design elements of the proposed algorithm. Validation of the algorithm is done in the context of computational experiments and provided in the next chapter.

## 5.1 Metaheuristic Algorithms

Metaheuristic algorithms are general-purpose heuristics that utilize more advanced intensification (i.e., procedures that exploit previously found solutions) and diversification (i.e., procedures to explore the search space) mechanisms to find near-optimal solutions with low computational effort. Over the last few decades, a wide variety of metaheuristics has been proposed, and this area of research has developed rapidly both from a theoretical and practical standpoint. These algorithms are commonly considered as flexible enough to tackle *NP*-Hard problems, and they can achieve good quality solutions promptly. Given that the computational complexity of the problem of optimizing multiple objectives in a SbO setting is *NP*-Hard, metaheuristic algorithms seem to be the most promising approach for finding the best trade-off solutions efficiently. Recently, metaheuristic algorithms have become an important and integral part of the state-of-the-art SbO tools especially when the use of exact algorithms is impractical, and they dominate the optimization routines of simulation software packages.

In general, metaheuristic algorithms can be categorized in a number of different ways depending on their various properties. The most commonly used categorization is based on the number of candidate solutions maintained and improved simultaneously: single solution-based, population-based and set-based (see Figure 26). In single solution-based (also referred as trajectory-based) algorithms, a single state is preserved during the optimization process, and a search procedure is utilized for local improvement. In population-based algorithms typically a population of candidate solutions is kept and updated at each iteration instead of following a single path in the search space. On the other hand, in set-based metaheuristics, a global sampling strategy is utilized that is continuously adapted using partitioning the search space into sets.

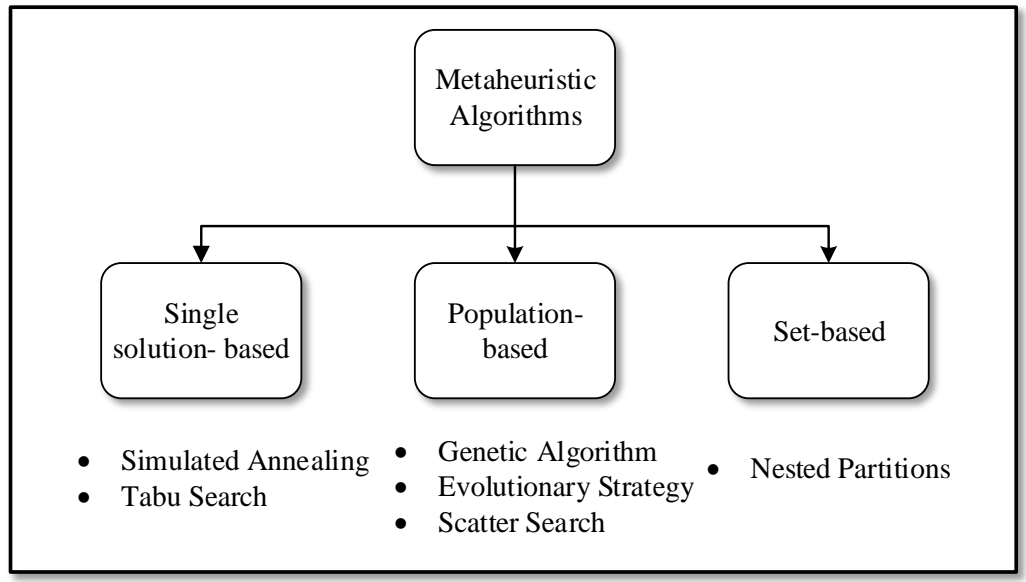

**Figure 26:** A Classification of Metaheuristic Algorithms

As No Free Lunch Theorems state, there is no single algorithm which is suitable for all optimization problems (Wolpert & Macready, 1997), which is also valid for SbO approaches. There are two major issues need to be addressed to employ metaheuristic algorithms in a SbO approach: (1) the necessity to consider the simulation noise in the implementation of metaheuristic method, and (2) difficulty in the analysis of convergence and diversity (Henderson & Nelson, 2006).

The metaheuristic methodology to SbO is grounded on treating the simulation model as a black-box function evaluator, which makes the search procedure problem independent. When combining the metaheuristics with simulation models some input parameters are given to the black-box, then the simulation models will give some feedbacks or responses, which can be used to guide the search process in metaheuristic algorithms. In general, metaheuristic methods require more simulation runs than metamodel methods to obtain near optimal solutions due to the simulation noise. Therefore, the main challenge is that a large number of simulation runs is required, which may result in long computation times. Also, failing to find a balance between intensification and diversification mechanisms will eventually result in an inefficient SbO, which may cause premature convergence and finally trapping in a local optimum (Fred Glover & Kochenberger, 2003) and (Michalewicz & Fogel, 2004).

The behavior of a metaheuristic algorithm is largely determined by the intensification and diversification mechanisms for the search. Intensification is the mechanism for exploring intensely the most promising search areas, and it is commonly implemented with local search techniques. On the other hand, diversification is the mechanism for diversifying the search process to move towards new areas of the search space, and it is commonly implemented with tracking the search history such as long-term memory utilization. In metaheuristic algorithm design, it is significantly important to find a good trade-off between these two mechanisms.

In recent years, hybrid metaheuristics have been widely used to solve large-scale real-world MOO problems because systematic combination of different metaheuristics has the potential to provide more efficient and flexible solutions (E. G. Talbi, 2015). One of the most widely used ways of hybridization is the utilization of single-solution based metaheuristic algorithms in population ones. The main strength of population-based metaheuristic algorithms is their capability to generate new solutions by recombining current ones, which enhance the convergence rate. On the other hand, single-solution based metaheuristic algorithms explore a promising area in the search space more systematically than population-based ones. Thus, hybrid metaheuristics combine the strength of population-based ones (the identification of promising areas) with the advantage of single-solution based ones (exploration of promising areas).

Scatter Search (SS) and its generalized version Path Relinking (PR) are one of the most promising population-based metaheuristics for SbO, and recent applications of both SS and PR utilize adaptive memory principles of Tabu Search (TS). Also, all three have a shared history, since their basic principles are suggested by Fred Glover (1977). In addition, SS has numerous similar attributes with Genetic Algorithms (GAs), where both of them are evolutionary methods and evolve over a set of solutions. Therefore, before delving into SS's details, brief descriptions of TS, PR and GAs are provided in the rest of this section.

### 5.1.1 Tabu Search

As previously mentioned, Tabu Search (TS) is a single solution-based metaheuristic proposed by Fred Glover (1989); (1990) and has been applied successfully to solve many combinatorial optimization problems. TS is an iterative improvement algorithm based both on neighborhood search methods and the use of diverse types of memories/strategies to guide the search. The idiosyncratic characteristic of TS is its utilization of memory to guide the local search to escape from the local optimum. When a local optimum is faced, a move to the best neighbor is done even if this move may cause to worsen the objective function value. In order to avoid cycling, a tabu list is utilized, which tracks attributes of recent moves and forbids any recurrence of such moves. Fundamental components of any basic TS algorithm are described below:

*Search Space*: Determining a search space along with a neighborhood structure is the most significant step of any TS implementation. The search space of TS is the space of all solutions that can be visited during the search. To allow the search to move infeasible solutions is usually desirable in order to escape from local optimum.

*Neighborhood Structures*: Considering that the quality of the final solution relative to global optimum heavily depends on the structure of the neighborhood, a problem specific neighborhood structure needs to be defined to cover all search space. There are several options for the neighborhood structures of the solution, such as adjacent pairwise interchange, swapping, insertion, etc. Adjacent pairwise interchange requires exchanging positions of two elements directly next to each other. Swapping or all pairwise interchange entails exchanging positions of two different elements. Insertion is related to removing an element from its original position and placing it immediately after another one. Previously done computational experiments indicate that the neighborhood insertion structure produces better quality solutions than the swapping neighborhood structure (Laguna & Glover, 1993). However, a hybrid neighborhood structure including both swapping and insertion has the potential to yield better solutions (Barnes & Laguna, 1991).

*Memory Structures*: Memory structures are the basic elements of TS and in general there are two types of memory, namely, explicit and attributive. Explicit memory keeps complete solutions, which is typically utilized for memorizing very good (elite) solutions encountered during the search. In contrast, attribute memory keeps the modifications that were done while proceeding from one solution to the next solution. Both explicit and attribute memory are used to build the short term and the long term memory of TS. For short term memory, a tabu list is retained in order to avoid cycling back to previously visited solutions. For long term memory, typically a frequency matrix is employed to detect more promising areas in the search space. It is noteworthy to mention that short term memory is used to store recency information; on the other hand, long-term memory is used to store frequency information. The number of iteration that an attribute remains in the tabu list, which is referred to as tabu tenure, is also an important search parameter for TS. If the tabu tenure is too small, preventing the cycling might not be achieved; on the other hand, too long tabu tenure might create so many restrictions.

*Aspiration Criteria*: Since a move or an attribute that is in the tabu list, may forbid moving to attractive unvisited solutions, it is necessary to overrule the tabu status of this move or attribute in certain situations, which is typically achieved by an aspiration criterion. The most commonly used aspiration criterion consists of releasing the restrictions on a move or an attribute, which is in the tabu list, if the current objective function value is better than the best objective function value found so far.

*Termination Criteria*: The most commonly used termination criteria in TS are as follows: (1) if the current iteration is equal to the maximum allowable iterations or the maximum allowable CPU time, (2) if the current iteration is equal to the maximum allowable iterations without an improvement in the fitness value, and (3) if the best fitness value found so far is equal to a pre-determined threshold value.

Main steps of a generic TS algorithm are given below (It is important to note that the term "solution" does not necessarily correspond to a final solution of the problem, it is just a component in the search space.):

**Step 1**: Generate all candidate solutions which are reachable by applying one move.

**Step 2**: Choose the best candidate solution based on tabu restrictions (which is not in the tabu list) and aspiration criteria.

**Step 3**: Update the current solution and the best solution found so far.

**Step 4**: Determine if any termination criterion is satisfied. If yes, stop the algorithm; otherwise, go to step 2.

The performance of the basic version of TS, which is explained above, often needs to be improved to tackle difficult problems, because it tends to get stuck in a local optimum in the end. In order to escape from local optimum, additional components for intensification and diversification have to be included in the search. Intensification is a myopic approach and it is done by implementing some strategies to explore more thoroughly the areas of the search space that seem promising. On the other hand, diversification is done by either performing several random restarts or implementing some strategies to penalize frequently performed move attributes.

It is crucial to find a balance between the diversification ability to move towards new areas of the solution space and the intensification ability to explore intensely the most promising areas. In TS, balancing the intensification and diversification mechanisms is usually done by controlling the length of the tabu list when fixed-length tabu lists are used or by controlling the tabu tenure. The diversification effect will be stronger if the tabu list is longer or the tabu tenure is, and intensification effect will be stronger if the tabu list is shorter or the tabu tenure is relatively small.

## 5.1.2 Path Relinking

Path Relinking (PR) was originally proposed as a strategy in TS to integrate intensification and diversification mechanisms (Fred Glover, 1994b). PR produces new solutions simply by identifying paths that link high-quality solutions. The PR procedure starts from one of the high-quality solutions, referred as "initiating solution", and identifies a trajectory in its neighborhood that guide through the other solutions, referred as "guiding solutions." This is commonly achieved by selecting moves that introduce attributes that are present in the guiding solutions.

After PR procedure initiated with high-quality solutions, these solutions are ordered with respect to their quality. Then, new solutions are created by exploring trajectories between and beyond the selected solutions in the neighborhood space. The characteristics of the guiding solution are progressively transferred to the intermediary solutions in order to ensure that these solutions include more characteristics from the guiding solution rather than the initial solution as search moves along the trajectory. At each step, procedure incorporates attributes of the guiding solutions as well as keeps track of the objective function values.

PR can be regarded as an extension of the Solution Combination method of SS. In PR, a trajectory is produced between and beyond the chosen solutions in the neighborhood, instead of generating a new solution by combining two or more original solutions as in the Solution Combination method of SS. The primary difference between SS and PR is that PR procedure typically starts from a set of given high-quality solutions rather than building a reference set as in SS. Hence, even though both algorithms operate on a set of reference solutions or a set of high-quality solutions, they simply differ in the way in which these set of solutions are created, maintained, updated and improved (Laguna & Marti, 2003).

5.1.3 Genetic Algorithms

Genetic algorithms (GAs) were introduced by Holland in the 1970s, and they belong to the class of evolutionary methods that adopt the evolution theory of genetic variation and natural selection (survival of the fittest), where successful individuals have a high probability to participate in reproduction for the next generation. Inspired by these principles, low-quality solutions are eliminated from the population, and fitter individuals reproduce to guarantee successful offspring. GAs are less susceptible to premature convergence to a local optimum compare to single solution-based metaheuristic algorithms, but they often tend to be computationally expensive.

In GAs, a solution to the problem at hand is often referred to as a "chromosome", which is analogous to the genetic material of an organism. They search the solution space first by generating a set of solutions called a "population." Then, they evolve this population over a number of iterations by using genetic operators such as selection, crossover, and mutation, where each iteration is called a "generation." The selection operator chooses parent solutions based on their fitness function. The crossover operator combines parent solutions to produce new trial solutions (offsprings). The mutation operator perturbs a solution to maintain diversity in the search, and to avoid premature convergence. There are various schemes for implementing these operators; the appropriate one should be chosen that best fit the problem at hand. The general framework of GAs is presented below:

**Step 1**: Generate an initial set of solutions (population).

**Step 2**: Select individuals from the population to be parents.

**Step 3**: Create offsprings (new individuals) as combinations of selected parents.

**Step 4**: Mutate some offsprings.

**Step 5**: Select the offsprings to insert into the population and the individuals to remove from the population.

**Step 6**: Update the best solution found so far.

**Step 7**: Determine if any termination criterion is satisfied. If yes, stop the algorithm; otherwise, go to step 2.

## 5.2 Scatter Search

Scatter Search (SS) was first introduced by Fred Glover (1977) as a heuristic for integer programming. SS has been commonly considered as a flexible and adaptable metaheuristic algorithm because it offers various implementation alternatives by exploiting its foundational strategies. Although SS shares some features with evolutionary approaches, its principles were established by concepts developed independently from the evolutionary paradigm. SS is based on the methodology of combining available solutions to generate new ones, which was originated from strategies for creating composite decision rules and surrogate constraints. SS has captured the attention of numerous researchers and practitioners. Recently, SS has been successfully applied to a wide range of real-life combinatorial optimization problems, such as vehicle scheduling, linear ordering, quadratic assignment, production scheduling problems, etc.

SS has several common features with Genetic Algorithms (GAs), even though it also has some differences. Similar to GAs, SS maintains a "reference set" derived from a population and new candidate solutions are generated by weighted linear combinations. As opposed to GAs where the population updating mechanism depends on random selection rules that select solutions with respect to their fitness value, in SS, the reference set update mechanism relies on adaptive memory structures where a balance between intensification and diversification tried to be maintained. Trial solutions are selected for reference set based on this memory structures. The number of solutions in the reference set is usually smaller than a "population" in GA, which is typically around 100. In general, the reference set has at most 20 solutions. The other fundamental difference between SS and GAs is the fact that SS has more deterministic rules about how to combine candidate solutions and how to improve them at each iteration (Deb et al., 2002).

In fact, SS is one of the most suitable metaheuristic algorithms for simulation-based multi-objective optimization. Since each simulation experiment is time-consuming, the chosen optimization algorithm should require as few simulation replications as possible for efficiency without compromising from effectiveness too much. Suitably, SS require too less fitness evaluation compare to other evolutionary methods such as GAs. The other important aspect is the existence of multiple conflicting objectives and the need for finding the Pareto-optimal solution set efficiently. Because SS maintains a reference set of solutions in each iteration, it is capable of dealing with multiple objectives. Therefore, SS has a high potential to address challenges stem from a large number of lengthy simulation runs as well as multiple objectives.

It is should be mentioned that SS has transformed since it was first introduced, and has been continuing to evolve over time. The first introduced version in 1977 is commonly referred as the "original SS algorithm" (Fred Glover, 1977). Fred Glover (1994b) extended the basic SS by combining it with adaptive memory structures of Tabu Search to balance search intensification and diversification, which is usually referred as the "hybrid Scatter/Tabu Search algorithm". Finally, Fred Glover (1998a) provided a simplified "SS algorithm template" that has been serving as the main reference for recent SS applications. These three versions of the SS algorithm and some prominent advanced design strategies are briefly presented in the following sub-sections.

5.2.1 Original Scatter Search Algorithm - 1977

In a nutshell, the working mechanism of the original SS algorithm is as follows: SS starts with generating an initial population of candidate solutions. The initial population of solutions is generated by considering features in different parts of the solution space without randomization. Then, it reduces this population to a reference set of solutions. In the next phase, it builds, maintains and evolves this reference set throughout the search where preferred subsets of solutions in the reference set are combined to generate new trial solutions. The reference set is updated by selecting the promising solutions from the trail solutions by finding a convex combination of the solutions in the reference set, which is referred as the central point.

The original SS algorithm relies on combining more than two candidate solutions to produce central points. Although the algorithm does not consider randomization, there is no guidance on how to select proper weights to generate biased central points. Also, there is no method mentioned for distribution of reference points relative to each other (Laguna & Armentano, 2005).

## 5.2.2 Hybrid Scatter Search Algorithm - 1994

Although the general description of the Scatter Search (SS) was first published in 1977, the original SS algorithm was not discussed or applied until the 1990s. In Fred Glover (1994a), the original proposal was extended by providing some implementation details. Also, in Fred Glover (1994b), nonlinear, binary and permutation problems are included as the application areas, and the algorithm is combined with Tabu Search by utilizing adaptive memory structures and aspiration criteria. This hybrid Scatter Search version gives emphasis to line searches and utilizing weighted combinations to create new solutions from the lines that connect reference points. The main advantage of integrating adaptive memory structures to the SS is that it provides a proper balance between diversity and quality.

This hybrid Scatter Search version of the SS served as the basis for many SS implementations proposed previously, and it is commonly considered as a hybrid evolutionary approach. In this version, the solutions are generated using combination strategies as opposed to probabilistic learning approaches, where these combination strategies facilitate the connection between diversification and intensification mechanisms.

## 5.2.3 Scatter Search Algorithm Template - 1998

A template algorithm for Scatter Search (SS) was provided by Fred Glover (1998a), which is a simplification of the hybrid Scatter Search version. The SS algorithm template has been commonly considered as the main reference for SS implementations so far.

As illustrated in Figure 26, SS algorithm template consists of two main phases (initial and SS) and five methods. In the initial phase, SS starts with generating a set of solutions (population) to ensure a high level of dispersion. Then, a subset of the best solutions is designated to be reference solutions. In the SS phase, new solutions are created by combining the subsets of the current reference solutions. The basic idea is to select the better quality and better dispersion solution from the reference set, using the reference set to produce the next generation of the solutions, in order to enhance the algorithm diversification capability. Typically, a new solution is formed by the combination of at least two reference solutions. Reference set evolves by deleting old solutions and adding new solutions. The theoretical underpinnings and basic principles of SS are based on the following five methods (Fred Glover, 1998b):

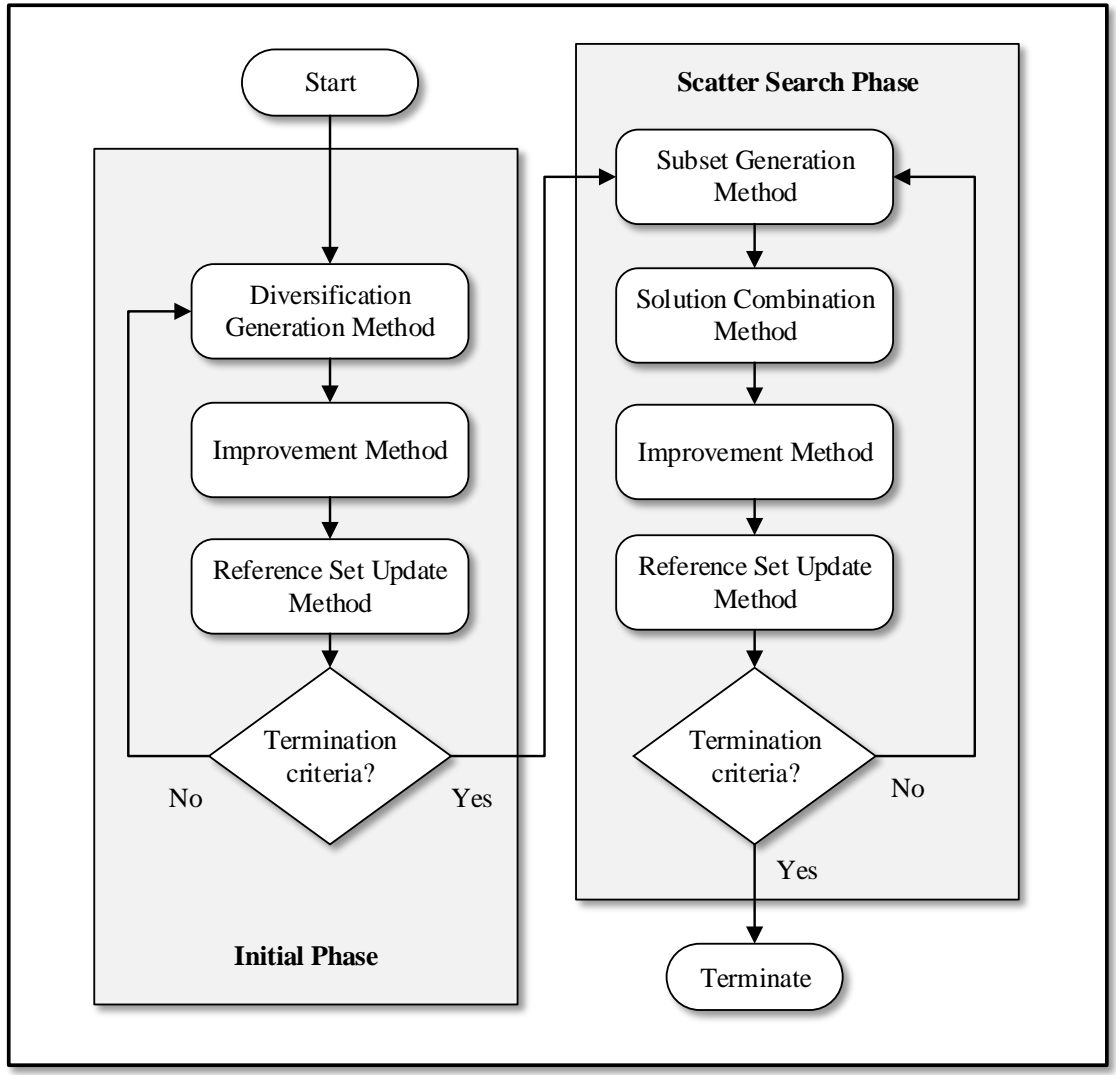

**Figure 27:** Illustration of Initial and Scatter Search Phases

*Diversification Generation Method*: The method is used to generate a set of diverse trial solutions using a seed solution or an arbitrary trial solution as an input. In general, the quality of the solutions is not important, instead main focus is to ensure a level of diversity. The method is often needed to be customized for the specific problem at hand and its effectiveness highly depends on the solution representation. The size of the set of diverse solutions generated by the diversification generation method is usually set to five to ten times the size of the reference set ($5 \times b$ to $10 \times b$), where $b$ refers to the size of the reference set. The number of solutions in the initial population needs to be large enough to guarantee that solution space is diversely covered.

*Improvement Method*: This method transforms an input trial solution into one or more improved output trial solutions. It must be able to handle both feasible and infeasible solutions. If the resulting output trial solution after applying this method is not improved, then the output trial solution is considered to be same as the input solution. This method is the only component that is not necessary to implement an SS algorithm.

*Reference Set Update Method*: This method tries to create a set of both high quality and diverse solutions by building and maintaining a reference set consist of $b$ solutions. The number of solutions contained within the reference set is usually not more than 20. Solutions are accepted to the reference set according to their quality or diversity.

*Subset Generation Method*: This method produces a subset of its solutions as a basis for creating combined solutions with the Solution Combination method by operating on the reference set. In general, subsets are constructed by including two solutions, although it is possible to include three, four or more solutions in the construction of subsets.

*Solution Combination Method*: This method is used to transform a given subset of solutions whose production is mentioned in the previous method into one or more combined solutions. This method is often problem-specific, and it can generate more than one solution at a time, where infeasible solutions can also be generated. Implementation of this method is systematic rather than probabilistic. For problems that have permutation type of representation, an adaptive structured combination based on the absolute position of the elements is presented as effective in Campos et al. (2000).

The pseudo-code and a detailed treatment of the SS algorithm template are given below, which is often implemented using a number of parameters. Before the pseudo-code and details of the algorithm, the necessary notation of these parameters and their definitions are given below:

| | |
|---|---|
| $P$ | the set of diverse solutions generated (population) |
| $P_{size}$ | the size of the set of diverse solutions generated |
| *RefSet* | the set of reference solutions |
| $b$ | the size of the reference set (*RefSet*) |
| $x^i$ | the $i^{th}$ solution in the *RefSet* |
| $b_1$ | the size of the high-quality solutions in *RefSet* |
| $b_2$ | the size of the diverse solutions in *RefSet* |
| *MaxIter* | maximum number of iterations |
| *NewSolution* | boolean variable that indicates whether or not a new solution has become a member of *RefSet* |
| *NewSubsets* | list of subsets of reference solutions that are subject to the Solution Combination Method |
| $s$ | subset of reference solutions |

---

**Algorithm 2** Scatter Search Algorithm Template

---

1:   **Initialization**
2:   $P=\{\}$
3:   **while** $|P|<Psize$ **do**
4:      generate a solution $x$ with **Diversification Generation Method**
5:      improve $x$ with **Improvement Method**
6:      **if** $x \notin P$ **then** $P=P \cup x$ **else** discard $x$
7:   **end while**
8:   build *RefSet* with $b$ ($b_1$ high-quality and $b_2$ diverse) solutions from $P$
9:   sort the solutions in *RefSet* according to their fitness in ascending order ($RefSet=\{x^1,...,x^b\}$)
10:  *NewSolution* = TRUE
11:  **while** (*NewSolution* or $iter<MaxIter$) **do**
12:     update the iteration counter, $iter=iter+1$
13:     generate *NewSubsets* with **Subset Generation Method**
14:     *NewSolution* = FALSE
15:     **while** (*NewSubsets* $\neq \{\}$) **do**
16:        select the next subset $s$ in *NewSubsets*
17:        apply **Solution Combination Method** to $s$ to obtain a new solution $x$
18:        improve the generated new solution $x$ with **Improvement Method**
19:        **if** ($x$ is not in *RefSet* and $f(x)<f(x^b)$) **then**
20:          insert $x$ into *RefSet* and reorder *RefSet* (**Reference Set Update Method**)
21:          *NewSolution* = TRUE
22:        **end if**
23:        delete $s$ from *NewSubsets*
24:     **end while**
25:  **end while**
26:  **return** best solution found so far

---

The SS template procedure starts with the generation of $P_{size}$ solutions with the Diversification Generation method. These solutions are originally generated to be diverse and subsequently improved by the application of the Improvement method. $P_{size}$ is usually five to ten times the size of *RefSet*. *RefSet* is constructed by Reference Set Update method with the first $b_1$ solutions in $P$ according to quality and $b_2$ solutions that are diverse with respect to the members in *RefSet*. Then, the value of True is assigned to the boolean variable *NewSolution*.

In the next step, the generation of the subsets occurs by applying the Subset Generation method, and the boolean variable *NewSolution* is switched to False. All subsets are subjected to Solution Combination method to generate new solutions. Then, these solutions are improved with the application of the Improvement method. If any of the improved solutions from the previous step is better (in terms of the objective function value) than the worst solution in *RefSet*, then the improved solution replaces the worst solution, and becomes a new element of *RefSet*. If any of the improved solutions is not admitted to the *RefSet* due to its quality, the solutions are tested for their diversity merits. If one of the solutions is diverse, then the solution is added to the *RefSet* and the less diverse solution is deleted.

The SS template procedure stops when a termination criterion met. The commonly used termination criteria include: (1) the maximum number of iterations, *MaxIter*, has reached, (2) the reference set does not change, or improvement does not warrant further iterations, and (3) the maximum allowed CPU time has passed.

Figure 28 presents a schematic representation of the SS algorithm template by illustrating the roles of five SS methods.

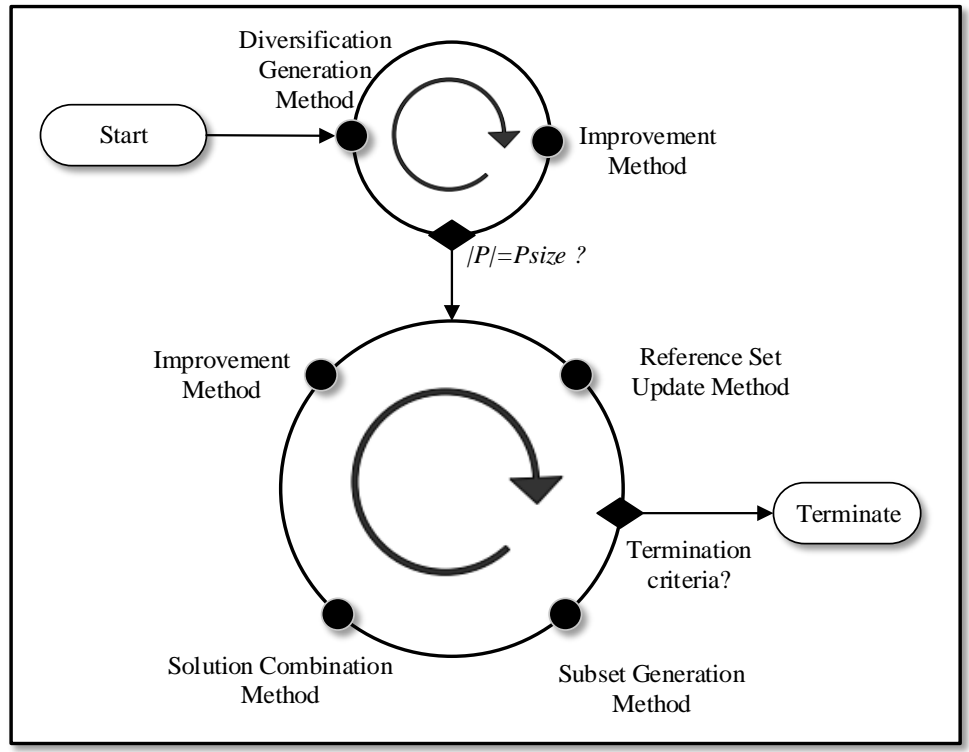

**Figure 28:** Schematic Representation of Scatter Search Template

Although previous SS applications often result in good performance and high-quality solutions, for complex problems, problem-specific considerations have to be integrated into the design of the algorithm. Out of these five components of the SS algorithm template, Reference Set Update method is not a problem-specific method, where it can be employed in different problem contexts. However, the remaining methods often have to be designed from scratch.

In basic SS algorithm template, the Reference Set Update method, which is based on improving the quality of the worst solution in the *RefSet,* and the Subset Generation Method, which consists of generating all pairs of solutions in *RefSet* that contain at least one new solution, are very simple mechanisms. More precisely, these procedures do not consider search diversity and do not allow for two solutions to be subjected to the Solution Combination method more than once. In the next section, several advanced SS design strategies that are proposed in the literature are presented.

5.2.4 Advanced Scatter Search Design Strategies

There are many advanced design strategies of SS, one differing from another in the way that how the five methods of SS are implemented, instead of including or excluding some of these methods.

*Reference Set Rebuilding Strategy*: The SS algorithm template terminates when there exists no new solution to be added to *RefSet*, which indicates convergence of the algorithm. This convergence might be premature, and search might have stuck in a local optimum. One of the possible ways to avoid such convergence situations is to enforce a form of diversity in *RefSet* by inserting a rebuilding step. This rebuilding step consists of creating a new population by applying Diversification Generation and Improvement methods again, and replacing half of the poor quality solutions in *RefSet* with the solutions from the newly generated population that increase the diversity in *RefSet*. When Solution Combination and Improvement methods are not able to generate solutions of adequate quality to enter the *RefSet,* this mechanism needs to be employed to rebuild the *RefSet* partially. The pseudo-code for the SS algorithm template with reference set rebuilding strategy is given below.

The working mechanism of reference set rebuilding strategy is as follows: After the final step is performed if *NewSolution* is False and iteration number has not reached maximum iteration number yet, the rebuilding step will be triggered. This step provides a seed for set $P$ by a new application of the Diversification Generation method. That is, a new set of diverse solutions $P$ is built by Diversification Generation method, and *RefSet* is reconstructed by the best solutions in the new set of diverse solutions $P$.

---

**Algorithm 3** Advanced Scatter Search with Reference Set Rebuilding Strategy

---

1:    **Initialization**
2:    $P=\{\}$
3:    **while** $|P|<Psize$
4:        generate a solution $x$ with **Diversification Generation Method**
5:        improve $x$ with **Improvement Method**
6:        **if** $x \notin P$ **then** $P=P \cup x$ **else** discard $x$
7:    **end while**
8:    build *RefSet* with $b$ ($b_1$ high-quality and $b_2$ diverse) solutions from $P$
9:    sort the solutions in *RefSet* according to their fitness in ascending order ($RefSet=\{x^1,....,x^b\}$)
10:   **while** the termination criterion is not met **do**
11:       update the iteration counter, $iter=iter+1$
12:       generate subsets with **Subset Generation Method**
13:       **while** no more new subsets **do**
14:           select the next subset
15:           combine the solutions immediately with **Solution Combination Method**
16:           improve the generated new solution with **Improvement Method**
17:           apply **Reference Set Update Method**
18:               **if** solution quality is not sufficient to displace current *RefSet*
19:               **then** apply **Rebuilding Mechanism**
20:       **end while**
21:   **end while**
22:   **return** best solution found so far

---

Figure 29 illustrates a schematic representation of the reference set rebuilding strategy.

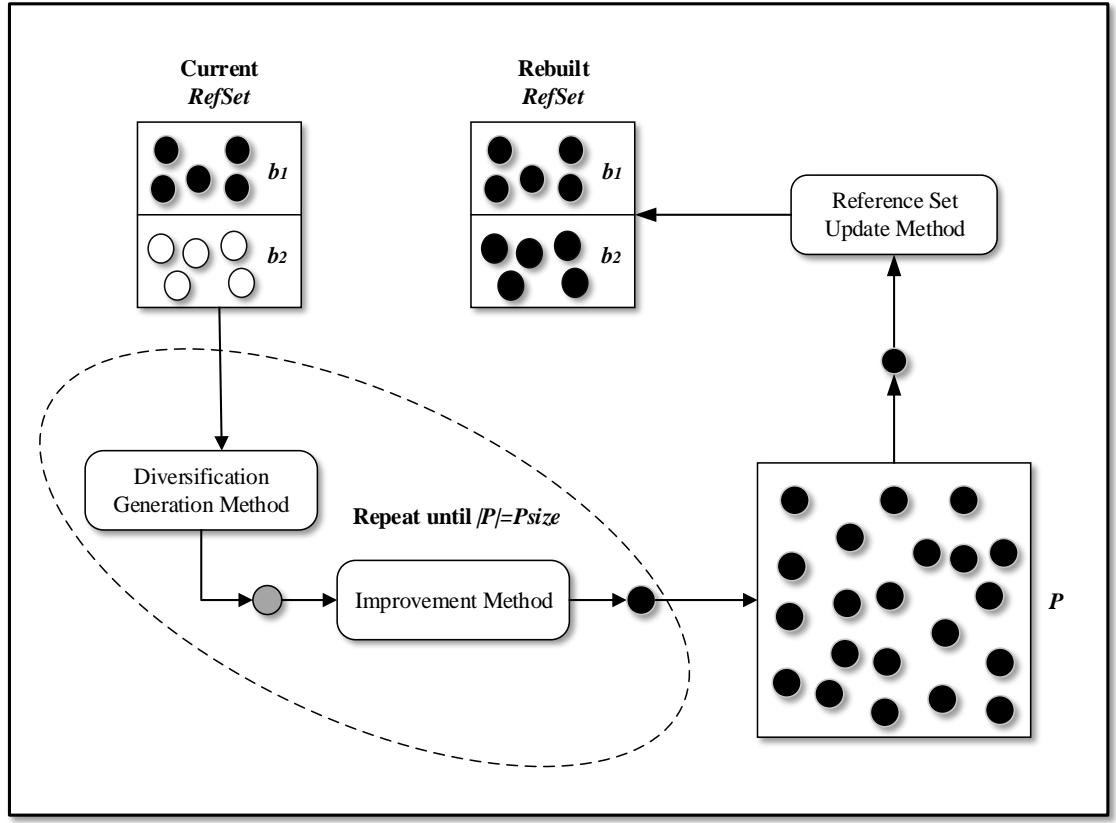

**Figure 29:** Schematic Representation of Rebuilding Strategy
(Adapted from Laguna and Marti (2003))

*Dynamic Reference Set Update Strategy*: In the SS algorithm template, new solutions that are selected as to become members of *RefSet* are not subject to Solution Combination method until the next iteration of the algorithm. This reference set update strategy is commonly referred as "static update" strategy. However, in dynamic update strategy, the primary objective is to apply Solution Combination method to new solutions that are accepted to the *RefSet* faster than the SS template. In other words, if a new solution is accepted to the *RefSet*, the aim is to allow this new solution to be subjected to Solution Combination method as quickly as possible. In order to accomplish this, the solution is immediately included in the *RefSet*, instead of waiting for the rest of parent solutions to be combined.

Figure 30 illustrates a schematic representation of the dynamic reference set update strategy with a hypothetical *RefSet*, which contains four solutions (*x1*, *x2*, *x3* and *x4*) ordered according to their objective function value *f(x)*. The figure shows the combination of the pair *x1* and *x2* in the current iteration where after applying the Improvement method, the solution *y* is generated. In the next iteration, the updated *RefSet*, which consist of solutions *x1*, *x2*, *y* and *x4*, will be used, and search will continue by combining the solutions *x1* and *y* instead of *x3* and *x,* which would have been made under the static reference set update method.

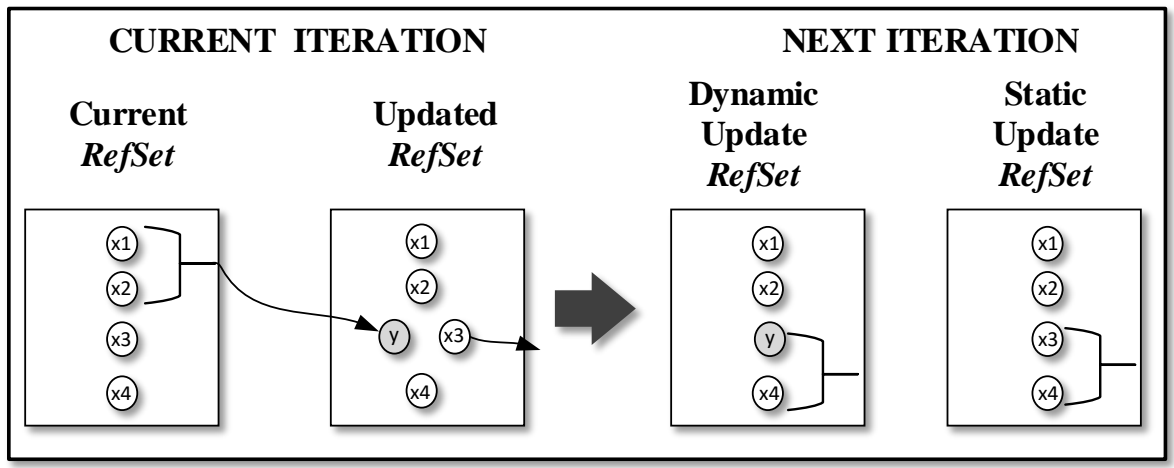

**Figure 30:** Schematic Representation of Dynamic Reference Set Update Strategy

The main benefit of the dynamic update strategy is that poor quality solutions in the *RefSet* are immediately replaced with high quality solutions and solution combinations are immediately done with these high quality solutions. On the other hand, this strategy has some drawbacks that need to be considered: (1) promising combinations may be ruled out, (2) implementation of this strategy is more complicated compared to static update strategy, and (3) it increases the computational complexity since it requires more *RefSet* ordering, which is important for determining the solutions to be replaced.

*Tiered Reference Set Update Strategy*: This strategy tries to maintain diversity in the *RefSet* in a dynamic way by avoiding admittance of only high-quality solutions. In order to accomplish this in "two-tier update" mechanism, *RefSet* is partitioned into two subsets and updated not only with high-quality solutions (*RefSet₁*) but also with diverse ones (*RefSet₂*), where the first tier is ordered with respect to their fitness value, and the second tier is ordered with respect to their diversity value.

$$RefSet_1 = \{x^1, x^2, \dots, x^a\}$$
$$RefSet_2 = \{x^{a+1}, x^{a+2}, \dots, x^b\}$$

(5.1)

The two-tier update mechanism can be combined with the *RefSet* rebuilding strategy simply by preserving *RefSet₁* and rebuilding only *RefSet₂* with solutions diverse with respect to the whole *RefSet*. Similar to two-tier update, three-tier update mechanism *RefSet* is divided into three sub-sets, where the first two subsets (*RefSet₁* and *RefSet₂*) are updated using the same rules as in the two-tier update. The third subset (*RefSet₃*) is updated by tracking a g-value (*g(x)*), which is the objective function value of the best solution ever generated from a combination of solutions from *RefSet₁* and any other solution in the *RefSet*. The *RefSet₃* is ordered with respect to g-value.

## 5.3 Multi-Objective Evolutionary Algorithms

Multi-objective evolutionary algorithms (MOEA) take a population-based approach for solving MOO problems. For more than three decades, MOEAs have been widely adopted for solving MOO mainly because of its capability to exploit the diversified and comprehensive set of Pareto-optimal solutions simultaneously in every iteration. MOEAs start with a set of candidate solutions and by applying stochastic operators (such as selection, crossover, mutation operators, etc.) try to reach the most accurate approximation of the Pareto-frontier.

5.3.1 Common Characteristics and Prominent Algorithms

The most widely-used MOEAs in the literature have three common features: fitness assignment based on Pareto-domination, elitism (archiving), and niching (Jensen, 2005).

The "fitness assignment" is related to assigning fitness value according to the incomparable objectives. The most widely used method is "non-dominated sorting", in which all the solutions from a population (set of solutions) are classified into different ranks or fronts. Each rank contains all the solutions that are dominated by at least one solution from the rank above it, but that dominate all the solutions in the ranks below it. "Crowding distance" is a parameter used by some MOEAs to promote the diversity within the population (set of solutions). It refers to a minimum distance that should be kept among all the solutions in the population.

The high-quality solutions found during the search process are often called as "elite" solutions. The "elitism" strategy is conceptualized by Dejong (1975), which involves preservation of good candidate solutions and utilization of the combined population rather than just replacing the old population with the new solution, which prevents the loss of promising solutions. Unlike single-objective optimization, where the elite solution is always copied into the next population, in MOEAs integration of elitism is more complicated since there exist a set of best solutions instead of a single best solution at each iteration. The main issues with integration of elitism into MOO include the following: (1) to determine the solutions to be kept in the elite set of solutions, and (2) to determine the elite solutions to be sent back to the population (Zitzler et al., 2000). To overcome these issues, elitism is commonly implemented by utilizing an archive, which is used to store identified non-dominated solutions and also interact with the population of individuals.

The concept of "niching" was originally proposed by Goldberg (1989) to promote population distribution to avoid genetic drift, and to search for potential multiple peaks. Niching tries to converge to more than one solution in a single iteration, which is accomplished by segmenting the population into disjoint sets so that at least one member in each region of the fitness function is covered more than one local optimum. Niching enforces each niche (neighborhood) to have no more than a specified number of solutions.

The main difficulties in developing MOEAs include the following: (1) quantifying the quality of a solution relative to other candidate solutions to update the population at each iteration, and (2) maintaining diversity among the non-dominated solution set in the Pareto-frontier. In order to address the first difficulty, various ranking procedures are proposed to ensure the most successful individuals are selected to reproduce. For the second difficulty, a niching method is commonly utilized to maintain a diverse set of non-dominated solutions.

Among the many MOEAs that have been introduced in the literature, the most popular and the ones that proved to be efficient in solving MOO problems are Non-dominated Sorting GA (NSGA-II), Strength Pareto Evolutionary Algorithm (SPEA, SPEA2), Pareto Archived Evolution Strategy (PAES), and Pareto Enveloped Based Selection Algorithm (PESA, PESA-II). Although the overall motivations of these algorithms are similar, they can be distinguished by the way in which the mechanisms of elitism and diversity preservation are implemented. These algorithms are discussed briefly in the rest of this sub-section except NSGA-II algorithm, which is presented in detail in the next sub-section.

Zitzler and Thiele (1999) introduced the SPEA, and Zitzler et al. (2001) proposed an enhanced version of it, referred as SPEA2. For fitness assignment, SPEA2 initially calculates the dominance count for each solution, where the dominance count corresponds to the number of solutions in the population that a given solution dominates. Then, fitness for a given solution is calculated by adding the dominance count of a solution to all dominating solutions. SPEA2 maintains an archive of non-dominated solutions explicitly. The archive is updated with new non-dominated solutions from both the recent population and archive at each iteration. In case that the archive's size exceeds a threshold, the solutions that have the poorest quality are deleted from the archive.

Knowles and Corne (1999, 2000) proposed the PAES, which is an elitist algorithm. In PAES, each generation comprises of a single parent and an offspring, and the selection between retaining the parent or the offspring as the parent solution is based on dominance. Also, a bounded archive of non-dominated solutions is maintained during the search. If the parent and the offspring are mutually non-dominating, and the offspring is not dominated by the archive, the one which resides in the less crowded region of objective space is chosen in order to maximize diversity.

The PESA and PESA-II were introduced by Corne et al. (2000) and Corne et al. (2001), respectively. PAES is an evolutionary strategy where the emphasis is placed on local rather than global search. The main operator in PAES is the mutation operator, and an archived list tracks the non-dominated solutions. PESA-II also employs an archive population to keep track of the current Pareto-optimal solutions. PESA-II utilizes a hyper-box scheme for determining the spacing of individuals on the Pareto-frontier, where these hyper-boxes divide the search space uniformly. Selection mechanism operates by hyper-box instead of by the individual, i.e. first a hyper-box is selected, and then one of the parents in that hyper-box is chosen at randomly. The main advantage of this selection mechanism is that it avoids search bias stem from having a high number of individuals in a given hyper-box.

5.3.2 The Elitist Non-Dominated Sorting Genetic Algorithm (NSGA-II)

Lately, the enhanced version of the NSGA-II has commonly been considered as one of the principal algorithms in the domain. Deb et al. (2002) proposed the NSGA-II, which implements a non-dominated sorting of a combined population with an elitist mechanism that helps to enhance the efficiency of the algorithm significantly. In addition to standard GA operators, NSGA-II algorithm uses a specialized non-dominated sorting operator that sorts and partitions the population into different Pareto-frontier approximations. Also, in addition to rank (fitness value), it utilizes crowding distance, which creates a fitness ranking for all of the individual solutions depending on each solution's closeness to its neighbors. For a population, large average crowding distance denotes that this population has a better diversity.

The time complexity of NSGA-II in generating Pareto-frontier in one generation for population size $n$ and $m$ objective functions is $O(mN^2)$. It has been empirically shown that NSGA-II achieves better convergence and diversity of solutions close to the true Pareto-frontier compared to PAES and SPEA (Deb, 2001). The distinct feature that characterizes NSGA-II is the incorporation of elitism. NGSA-II requires no parameters in addition to those required by a basic GA.

In NSGA-II, solutions in the current population are ranked into several classes at each generation. Then, two values are assigned to each solution. The first value relates to the "rank" the corresponding solution belongs to and represents the quality of the solution in terms of convergence. The second value is the crowding distance that refers to the density of solutions neighboring a particular solution in the population, and it is typically computed by the average distance between two points on either side of this solution along each of the objectives. A solution is said to be dominating another one if it has a better rank value, or, in the case of equality, if it has a better crowding distance. The deterministic tournament method is used as the selection operator between two randomly selected solutions. At the replacement step, only the best solutions survive with respect to a predefined population size.

The dominance process in NSGA-II operates as follows: All solutions in the population are searched to find non-dominated solutions, and these non-dominated solutions are all labeled with the front number "1" and not considered in the further iterations. Then, the remaining population is searched for non-dominated solutions again, and these solutions are labeled with the front number "2" and not taken into account in the further iterations. This process continues until all individuals in the population have been assigned a front number. Any solution with a lower front number is considered a fitter than a solution with a higher front number. Also, a hyper-boxed-based penalty function is utilized to reward solutions that are far apart so that solutions do not group in the same area on the Pareto-frontier.

In conclusion, NSGA-II owes its outstanding performance to three major features (Ding et al., 2008):

(a)     The non-dominated sorting approach, which reduces the O($mN^3$) complexity of Multi-Objective Genetic Algorithm (MOGA) to O($mN^2$).

(b)     The $\lambda + \mu$ elitism selection procedure, where binary tournament selection method is used for parent selection and survival selection was handled collectively on both the parents and offspring.

(c)     The crowding distance, as a measure for comparison and selection after the non-dominated sorting, to preserve the diversity of the solutions in the population.

### 5.3.3 Scatter Search-based Multi-Objective Algorithms

Although there has been a pervasive interest in applying GA to MOO problems in the literature, there has been some attempt to propose MOO algorithms based on Scatter Search (SS). Some of the main motivations for using SS to solve MOO problems include the following: (1) it has several powerful features that are desirable for MOO, such as maintaining diversity in the reference set in a natural way, and (2) it operates on a relatively small set of solutions compare to other evolutionary algorithms, which eventually contributes to efficiency. The most prominent SS-based multi-objective algorithms in the literature are detailed below briefly.

Rahimi-Vahed et al. (2007) suggested a non-dominated sorting procedure called MOSS (Multi-Objective Scatter Search), which ranks every solution of the reference set. In order to maintain non-dominated solutions uniformly dispersed along the Pareto frontier, an NSGA type of niching method is employed. In order to evaluate the generated solutions' quality, MOSS uses a weighted sum approach. This algorithm is compared against NSGA-II, SPEA-2, and PESA on a set of unconstrained benchmark test functions, and reported that MOSS outperforms the existing GAs, particularly in large-scale problems. Nebro et al. (2008) proposed a hybrid metaheuristic algorithm called AbYSS (Archive-based hYbrid

Scatter Search), which adapts the traditional SS algorithm template but utilizing mutation and crossover operators coming from the field of evolutionary algorithms. This algorithm is built on integrating the ideas of Pareto dominance, external archiving, and two different density estimators.

These previous research on SS-based multi-objective algorithms indicates that SS is a promising approach for MOO problems. Some of the open areas for research related to integrating SS to MOO problems are: (1) how to reuse the obtained search information that is available in the non-dominated solutions found by the SS, (2) different update mechanisms for Reference Set Update method to improve the diversity of the solutions, and (3) setting the parameters to enhance the convergence to the Pareto-frontier (El-Ghazali Talbi et al., 2012).

## 5.4 Mechanics of the Proposed Tabu/Scatter Search Algorithm

The proposed hybrid Tabu/Scatter Search algorithm is based on Scatter Search (SS) algorithm template and it makes use of the adaptive memory structures of Tabu Search (TS). SS algorithm template is chosen as basis of the optimization engine for the simulation-based multi-objective optimization framework for the following reasons:

(a)     It generates and maintains a reference set of solutions at each iteration rather than a single solution and this mechanism gives the ability to search for multiple Pareto-optimal solutions concurrently in a single run, without repeatedly finding each Pareto-optimal point one at a time.

(b)     It improves the solutions increasingly at each iteration, and this facilitates evaluating and improving the candidate policies through simulation.

(c)     It is capable of handling non-differentiability and discontinuity that often appear in simulation-based approaches.

The proposed hybrid Tabu/Scatter Search algorithm's general framework is similar to that of the traditional SS algorithm template. The primary additional procedure and mechanisms integrated into the SS template are listed below:

(a)     Adaptive memory structures are utilized explicitly to store complete solutions. After a new trial solution is created with the Solution Combination method, the memory structure ensures that this trial solution has not visited previously. Then, it is sent to the simulation model for performance evaluation. Since computational time is the limiting factor in any simulation-based approach, integration of adaptive memory structures is significantly important for efficiency.

(b)     A dynamic update procedure is employed in Reference Set Update method with the intention of producing high-quality solutions, where non-promising solutions are replaced immediately with more promising ones.

(c)     The fitness of each solution is computed with a non-dominating sorting approach, and a dominance procedure is utilized to classify solutions over the bi-objective domain, where both the objective value of the solution and its proximity to other solutions are considered.

(d)     A rebuilding mechanism is adopted to enhance and maintain the diversity of the Pareto-frontier approximations.

(e)     A two-step approach that includes a Tabu Search and a local search step is applied to improve solutions in the Improvement method.

(f)     An elitism mechanism is adopted where both dominated, and non-dominated solutions are stored in a fixed-size archive. Also, a truncation procedure is employed based on density assessment by measuring the Euclidean distance in order to restrict the number of stored solutions.

The high-level scheme of the proposed hybrid Tabu/Scatter Search algorithm is as follows: SS procedure is initiated by constructing a population of solutions ($P$) by using the initial solution obtained from the greedy heuristic algorithm as its starting point (seed) and, then a reference set (*RefSet*) is built from the population. This initial *RefSet* is selected by identifying non-dominated solutions consecutively from the $P$ with a dominance test procedure. The *RefSet* is then updated by applying the non-dominated criterion to the set of solutions that result from the union of the current *RefSet.*

During the procedure, *RefSet* is evolved through Subset Generation, Solution Combination, and Improvement methods. *RefSet* consists of two distinct subsets $H$ and $D$, representing the high-quality and diverse solution subsets, respectively (*RefSet* = $H \cup D$). *RefSet* is updated from iteration to iteration by Reference Set Update method. *RefSet* is always maintained in order, where $x^1$ is the best solution and $x^b$ is the worst one. Hence, in each iteration, *RefSet* is updated by assigning the incumbent trail solution to $x^b$ and reordering the *RefSet.* The proposed algorithm has two main loops: (1) a "while loop" that controls the generation of the $P$, and (2) a "while loop" in which *RefSet* is evolved until a termination criterion is met (when the current iteration is equal to the maximum allowable iterations or the maximum allowable CPU time.)

The mechanics of the proposed algorithm is illustrated in Figure 31. In the figure, circles represent solutions, where grey-colored ones represent solutions before Improvement method applied (trial solutions), and black-colored ones represent solutions after the application of Improvement method.

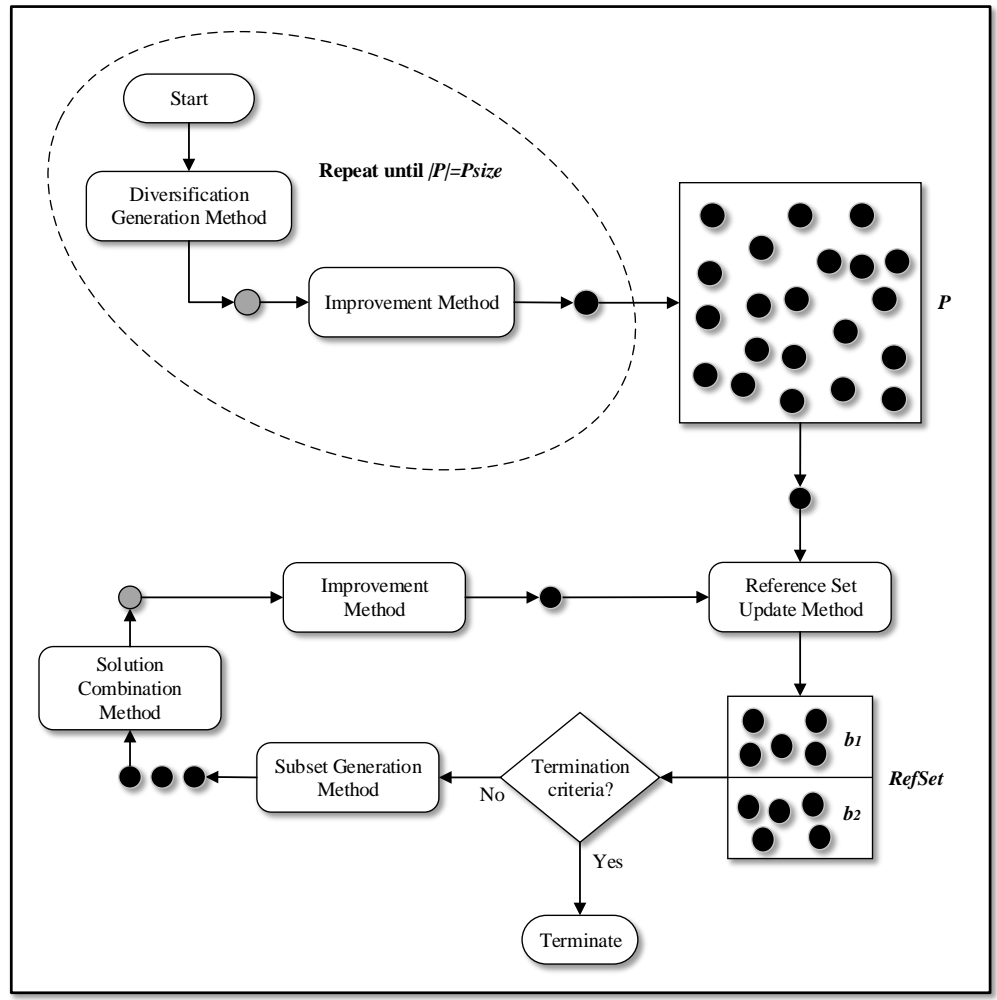

**Figure 31:** Schematic Representation of the Proposed Algorithm
(Source: Laguna and Marti (2003))

## 5.4.1 Design Elements

The four fundamental design elements that are incorporated into the proposed algorithm include the following: (1) the combination of representation and search operators, (2) the fitness function, (3) the initialization and the termination, and (4) the search strategy. Search strategies differ in the control of the intensification and diversification steps. These design elements typically depend on the problem context and search strategy which will be implemented. Each of these design elements is detailed below:

*Representation and Search Operators*: For the effective and efficient application of any metaheuristic algorithm, it is essential to find an appropriate representation for a candidate solution that largely depends on the nature of the problem at hand and search operators that conform well to the characteristics of the representation, and the proposed hybrid T/SS algorithm is no exception. The main requirement for the representation is the fact that it has to be capable of covering all candidate solutions in the search space since the design of appropriate search operators is closely related to representation. However, there are not many theoretical models available that explain how different types of representation impact algorithm's success and to what extent. The related properties of representation that impact solution quality, convergence time, and diversity for the multi-objective problems have to be identified. The most commonly accepted properties are redundancy, scaling, and locality (Franz, 2006).

A solution in the reference set corresponds to a set of decision variables for the optimization problem that is going to be simulated. Each iteration contains different input parameters that have to be experimented by the simulation model. Because generating and maintaining diversification effectively depends on the solution representation, a permutation encoding is employed where a solution is represented by a sequence of integers corresponding to the index of the aircraft, and each row corresponds to a runway and an aircraft sequence. In the literature, a number of representation types are proposed for encoding permutations, where integer numbers are utilized to represent a sequence directly. However, this representation type requires additional repair mechanisms in order to apply the Solution Combination method, which yields infeasible permutations with duplicate elements.

In Figure 32, the representation of a solution with $n$ aircraft is given, where $A_i$ is the index of the aircraft $i$, $R_i$ is the runway allocated to aircraft $i$, and $S_i$ is the sequence of aircraft $i$ in its allocated runway. This representation is selected mainly for two reasons: (1) to avoid the difficulty in maintaining feasibility after applying Solution Combination method, and (2) to facilitate the calculation of distance measures, which is the key element in Diversification Generation method, and mostly depend on representation.

| Aircraft | $A_1$ | $A_2$ | $A_3$ | $A_4$ | $A_5$ | ··· | $A_n$ |
|----------|-------|-------|-------|-------|-------|-----|-------|
| Runway | $R_1$ | $R_2$ | $R_3$ | $R_4$ | $R_5$ | ··· | $R_n$ |
| Sequence | $S_1$ | $S_2$ | $S_3$ | $S_4$ | $S_5$ | ··· | $S_n$ |

**Figure 32:** Representation of a Solution

*Fitness Function*: The fitness function essentially links the algorithm to the problem at hand, and typically dependent on the problem description and representation. In the proposed algorithm, the fitness function is estimated by running the simulation model, and it is calculated based on a Pareto-based fitness assignment method, which will be detailed under multi-objective components of the algorithm.

*Initialization and Termination*: The algorithm initiates by getting the initial solution provided by the greedy heuristic algorithm based on a composite dispatching rule. For the termination, various criteria have been used to terminate the optimization process of SS-based algorithms including criteria that take into account the landscape of the response surface, the convergence speed towards the Pareto-frontier, the desired quality of the solution found, the maximum number of solution evaluations, and the required computation time. The termination criteria for the proposed algorithm is chosen such that when the current iteration is equal to the maximum allowable iterations or the maximum allowable CPU time, the algorithm terminates and outputs the best Pareto-optimal solutions found so far.

*Search Strategy*: The search strategy is based on explicit memory approach of storing complete solutions. This approach is usually not preferred due to its time and memory consumption which usually renders algorithms intractable. However, since the evaluation of a solution by running the simulation model necessitate a large computational time, tracking every solution produced and evaluated during the optimization process is a viable approach in this simulation-based optimization setting.

The five SS methods of the proposed hybrid Tabu/Scatter Search algorithm are detailed below:

### 5.4.2 Diversification Generation Component

The Diversification Generation method is used both for initializing the reference set and also rebuilding the reference set during the search. The primary purpose of this method is to create a set of trial solutions systematically to guarantee a critical level of diversity both in initialization and rebuilding stages.

The algorithm starts with generating a set of trial solutions, which are required to be diverse. Hence, a systematic procedure is used to generate those trial solutions. When a termination criterion is met, the algorithm provides the best Pareto-optimal solutions found during any iteration. As previously mentioned, SS utilizes a reference set by combining the solutions in the reference set to generate new solutions, where the reference set is the core element. In a case such that all solutions in the reference set are similar, then the whole procedure will probably not be capable of improving the best solution found so far.

Modus operandi of Diversification Generation method is as follows: First, the output of the initial solution generation algorithm is used as a seed to generate subsequent trial solutions. Controlled randomization and frequency-based memory structures are employed to produce a collection of diverse solutions, where frequency-based memory structures are common Tabu Search mechanisms for implementing long-term memory strategies. Because too many high-quality solutions induce a premature convergence of the population into areas of the solution space containing only sub-optimal solutions, the main purpose of this method is to generate diverse solutions instead of high-quality solutions.

### 5.4.3 Improvement Component

Improvement method is an important intensification method to further transfer the incumbent solutions into a set of enhanced solutions of reasonable quality and diversity. This method is comprised of two steps: a simple Tabu Search, and a local search (neighborhood search) procedure. In the Tabu Search step, only the solutions that dominate the other solutions are considered. This threshold value is an assumed parameter for the algorithm. In local search step, all trail solutions are considered. In this step, "insertion" technique is used for moving from one solution to another. This procedure terminates when exploration of the neighborhood fails to find an improving move. This method is applied to all solutions present in the set $P$ initially, and then, to new solutions generated by the Solution Combination method.

### 5.4.4 Reference Set Update Component

Reference Set Update method is utilized to generate and maintain the *RefSet*. During the first application of this method (initial generation of *RefSet* from the population), a minimum diversity test is utilized, which operates as given in the following pseudo-code:

---

**Algorithm 4** Minimum Diversity Test for Initial *RefSet* Generation

---

**Input:** A population of improved trial solutions ($P$)

1:   **begin**
2:       find the best solution according to dominance test in $P$
3:       select this solution to become $x^l$ in the *RefSet*
4:       delete this solution, $x^l$, from the $P$
5:       **while** ( $\left| RefSet \right| < b$) **do**
6:           find the next best solution $x$ according to *Obj.Fn.* value in $P$
7:           select this solution, $x$, to be included in the *RefSet* only if
                      $distance_{min}(x) >= tresholdDistance$
8:           delete this solution, $x$, from the $P$
9:   **end**
10: **return** *RefSet*

---

The minimum diversity test procedure for the initial *RefSet* generation starts with choosing the best solution according to the dominance test in the population, selected as the non-dominated solution in the *RefSet*, and these solutions are extracted from the population. Then, at each iteration, the next non-dominated solutions in the population are selected only if the minimum distance between the selected solution $x$ and the solutions currently in *RefSet* ($distance_{min}(x)$) is at least as large as the threshold value (*tresholdDistance*).

In addition, an elitist sorting mechanism for the non-dominated solutions is utilized to sort the solutions in the *RefSet* according to the number of solutions they dominate. These solutions are then compared to each other to identify the distribution of solutions in the current Pareto-frontier. The decision for accepting a candidate solution to *RefSet* is made based on the dominance relation and the density of the *RefSet* (whether it improves the diversity of the set). The distance between solutions in *RefSet* is calculated based on the crowding distance from each member of *RefSet*.

Finally, a rebuilding mechanism is employed to rebuild the *RefSet* partially when the Solution Combination and Improvement methods are not able to provide solutions of sufficient quality to displace the current *RefSet*. This mechanism reinitializes the Diversification Generation method to generate diverse solutions with respect to high-quality solutions in the current *RefSet*. It consists of the $b_1$ best solutions from the preceding step (solution combination or diversification generation). It also consists of the $b_2$ solutions that have the largest Euclidian distance from the current solutions in the *RefSet*. At each iteration, a set of high-quality solutions replaces less promising solutions to improve the quality of the *RefSet*.

## 5.4.5 Subset Generation Component

Subset Generation method generates subsets from *RefSet* that will be used for creating new solutions where the subsets are constructed by including all pairs of *RefSet* solutions except the pairs that have already been included in previous iterations. Adaptive memory structures are utilized to exclude reference solutions during the application of this method, where the subsets that have already been processed by the Solution Combination method are recorded. Since execution time of the simulation model is the limiting factor for the whole computation time, these memory structures help to avoid unnecessary simulation runs. Nevertheless, it is also important to ensure that these memory structures do not grow to an unmanageable size during the execution. In order to fulfill this requirement, appropriate data structures are used in the implementation of these memory structures.

### 5.4.6 Solution Combination Component

Solution Combination method utilizes the generated subsets to combine the elements of each subset to create new trial solutions. The input for this method is not limited to *RefSet,* an intermediate pool of solutions is utilized in the implementation to enhance quality and diversity. Also, a dynamic update strategy is utilized where a new solution is included in the *RefSet* as quickly as possible before the next combination is performed. Furthermore, an intensification strategy is integrated into this method to improve the search towards the Pareto-frontier.

Solution Combination method also tracks the subsets of *RefSet* solutions that have already been exposed to this method in each iteration. Whenever a new trial solution is created with this method, by using memory structures, this trial solution is checked whether it has not been visited previously. Then, it is sent to the simulation model for performance evaluation. Since a ranking procedure is not appropriate to use due to small size of the reference set, which is typically 20, a dominance procedure is used to compare each candidate solution with the solutions in the *RefSet* and the pool. After an application of the Solution Combination method, the dominance test is applied to the solutions in the *RefSet* and the pool, and the reference set is updated with the solutions that have highest dominance value, where the two-tier *RefSet* prevent the optimization process from focusing on a given part of the Pareto-frontier.

### 5.4.7 Multi-Objective Search Components

The proposed Tabu/Scatter Search algorithm has three important multi-objective search components: (1) fitness assignment for better guiding the search towards the Pareto-frontier, (2) diversity preservation for maintaining well-spread non-dominated solution set and avoiding premature convergence, and (3) elitism for preserving high-quality solutions. Illustration of these multi-objective search components are shown in Figure 33.

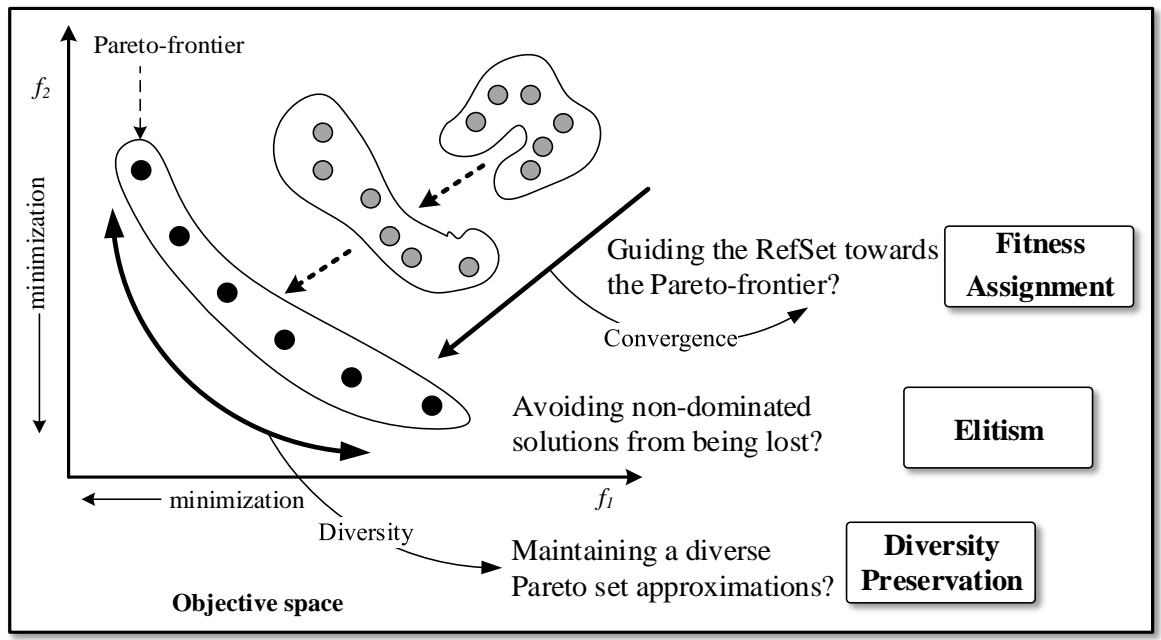

**Figure 33:** Illustration of Multi-Objective Search Components

*Fitness Assignment*: Due to the low-dimension (only two conflicting objectives) of the problem at hand, a Pareto-based fitness assignment method is employed to converge the solutions in a direction normal to the Pareto-optimal region and, at the same time, to promote diversity among solutions. This assignment method is applied together with a density measure, which is incorporated in such a way that adopts a two-stage process where first solutions are compared based on Pareto-fitness, then the density measure is applied. The main strength of this approach is that at the initial stages the force for diversity is higher, on the other hand, when the solutions begin to move to the Pareto-frontier, convergence force becomes dominant as most of the solutions that are equally fit.

*Diversity Preservation*: A diversity assessment scheme is adopted as the core element of diversity preservation component. Since the main goal of the proposed algorithm is to obtain a diverse Pareto-frontier, this diversity assessment scheme is applied in the objective space. And a distance-based assessment, in particular niching (niche sharing) is employed, which promote diversity in the reference set.

*Elitism*: Besides the fixed-size reference set, an archive employed to store non- dominated solutions along the evolution. The main function of the archive is to store a record of the non-dominated solutions identified during the optimization process and maintaining them to generate a diverse Pareto-frontier. The archive is updated at each iteration by adding a candidate solution to the archive if it is not dominated by any solution in the archive. Similarly, any solution in the archive dominated by this solution is removed from the archive. When the predetermined archive size is reached, a recurrent truncation process based on niche count is used to remove the most crowded solution in the archive. The crowding distance is an approximation of the density of solutions neighboring a specific solution in the archive, and it is calculated by averaging the distance of two points on either side of this point with respect to each of the objectives.

In the algorithm, elitism is applied by selecting solutions to a solution combination pool through a binary tournament selection of the combined archive and evolving reference set, where in case of a tie niche count is used. Similar to diversity preservation, niching (niche sharing) is employed in the tournament selection, where the crowding distance of NSGA-II is utilized as a niching measure.

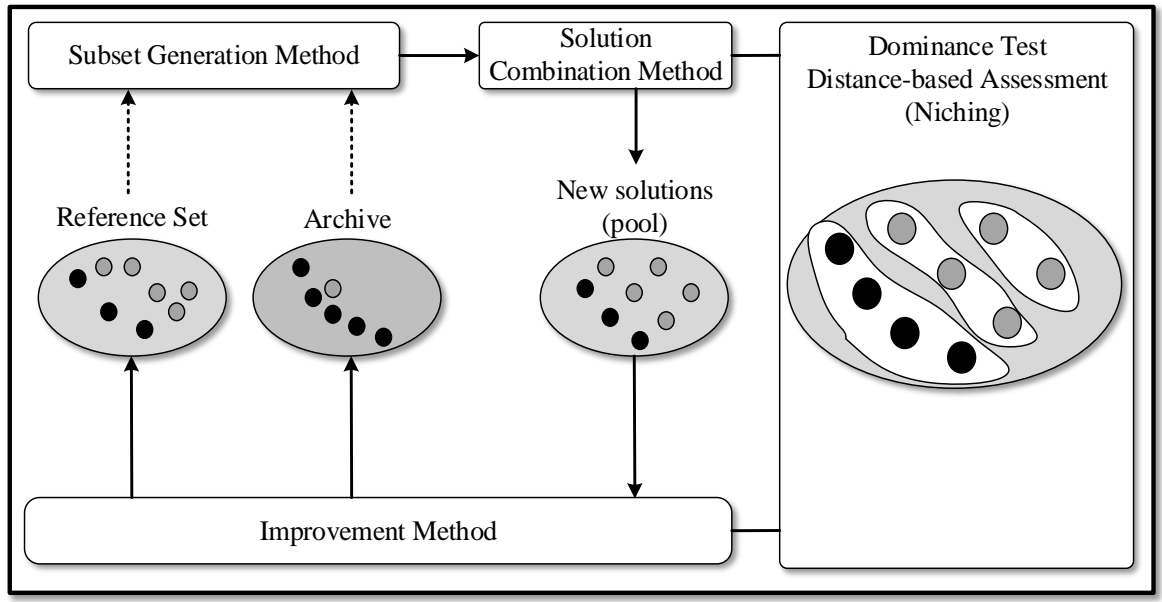

**Figure 34:** Representation of Elitism Mechanism

## 5.5 Object-Oriented Design and Implementation Specifics

Integrating advanced strategies in any metaheuristic algorithm, especially in multi-objective population-based algorithms, to improve the performance regarding effectiveness and computation time, typically comes with the burden of a design that has difficulties in implementation and parameter tuning. Also, these advanced designs more often add extra parameters and increase complexity. Therefore, a structured, two-stage approach is utilized for the design and implementation of the proposed hybrid Tabu/Scatter Search algorithm. First, the core data structures of the algorithm are created, and then, the algorithmic structure is built on top of them. Since adaptive memory structures heavily depend on data structures, object-oriented techniques are employed sensibly.

The main challenges faced during the design and implementation phases are listed below:

(a)     How to design and implement a multi-objective metaheuristic algorithm through incremental specification by first defining the skeleton of the algorithm in an abstract level and then building other aspects of it as soon as they become concrete?

(b)     How to reduce complexity by distributing the roles of algorithm and visualization components of the overall design, and simplify implementation?

In order to overcome these two challenges, template method design pattern and Model-View-Controller (MVC) architectural pattern are employed, which are presented briefly below:

*Template Method Design Pattern*: The general framework of the algorithm is based on template method design pattern, which allows redefining certain steps of an algorithm incrementally without changing the algorithm's overall structure. The template method design pattern has two components: (1) an abstract parent class, which is the template class used to define the algorithmic steps and preserve it across implementations, and (2) one or more concrete child classes, which extends the parent class and contains details of the abstract methods. In this way, the algorithm is defined as a skeleton of methods (operations) and leaving details to be implemented by the child classes, where the parent class preserves the overall structure and sequence of the algorithm. UML representation of an example of template method design pattern is given in Figure 35.

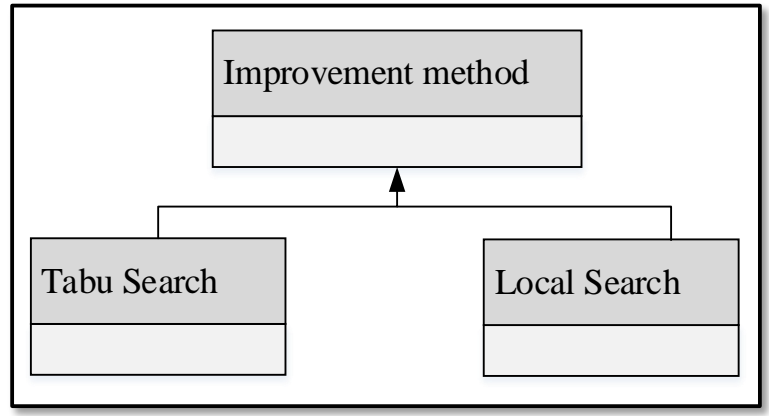

**Figure 35:** Example of Template Method Design Pattern

For implementing the template design pattern, the following steps are followed: (1) the main process (the template) is created by establishing a parent abstract class, (2) the sub-processes are created by defining abstract methods, (3) a special method is created that defines the sequence how the sub-process methods will be called, where child methods cannot override it, and (4) the child classes are created which can modify the abstract methods or sub-process to define a new implementation.

*Model-View-Controller (MVC) Architectural Pattern*: The user interface and visualization architecture is design and implemented based on Model-View-Controller (MVC) architectural pattern. This architectural pattern is used to separate user interface and visualization component from the metaheuristic algorithm component. In application of MVC architectural pattern the model established the metaheuristic algorithm, the view handled the visualization of the data that model contains, and the controller operated on both the model and the view by controlling the data flow into a model object and updating the view whenever data changes. Hence, this structure allowed keeping the view and the model separate. Also, setting the input parameters is handled within a function of the Algorithm Handler class. The representation of MVC architectural pattern is given in Figure 36.

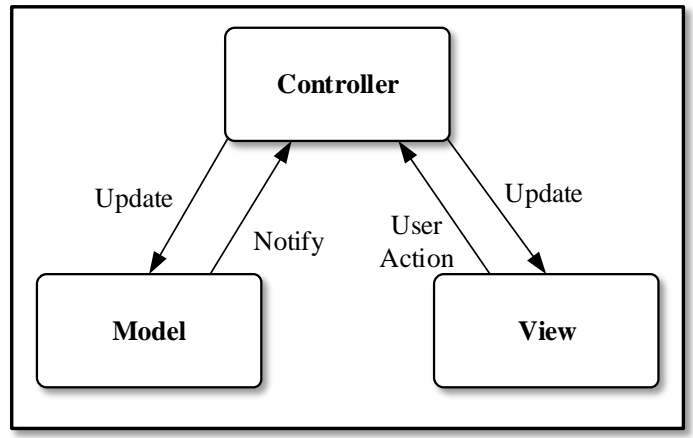

**Figure 36:** Representation of MVC Architectural Pattern

# CHAPTER 6

# COMPUTATIONAL EXPERIMENTS AND RESULTS

The main objective of the computational experiments is to study the quality and efficiency of the solutions generated by the proposed hybrid Tabu/Scatter Search algorithm and conduct a proof-of-concept (validation) of the whole simulation-based optimization (SbO) framework, in which the proposed algorithm is incorporated as an optimization engine. The experimental study is conducted in two phases. In the first phase, denoted as multi-objective optimization (MOO) experiments, the proposed algorithm's performance is evaluated based on multi-objective benchmark problems. In the second phase, denoted as SbO experiments, real-life historical datasets are utilized that belong to a major US airport. In experimental design, design of experiments approach is employed to analyze the impacts of parameters on the simulation as well as the optimization component's performance, and to identify the appropriate parameter levels. After the experiments are conducted and output data are collected, data analysis and visualization methods are utilized to identify patterns and draw conclusions.

This chapter initially discusses the experimental design including the objectives and the general framework of the computational study that is conducted. Next, exploratory MOO experiments are explained, which is conducted to locate algorithmic bottlenecks and guide parameter tuning efforts. Afterwards, exploratory simulation experiments are detailed, which investigate the most critical and sensitive parameters, and ensure the tolerance to which model outputs can be expected to alter with given input parameters. Then, experimental design and results of the MOO experiments are explained. Next, the performance metrics and experimental setup along with the considered scenarios for the SbO experiments are outlined. Then, the final SbO experiments are presented, and a statistical analysis of the results is provided. Next, key experimental results, as well as an analysis of these results, are presented. Finally, results of a safety risk assessment associated with our proposed approach is provided.

**6.1 Computational Framework and Experimental Design**

The objectives and general framework for designing computational experiments are presented in this section.

6.1.1 Objectives of the Computational Experiments

The primary objective of the computational experiments is to assess the effectiveness and performance of the proposed hybrid Tabu/Scatter Search algorithm as well as to validate the whole SbO framework in different current and future operational conditions. In particular, this computational study seeks to answer the following questions:

(a)     *Validation of models and the framework*: Is the proposed algorithm able to find the best known Pareto set of solutions? Is the proposed SbO framework truly capable of handling real-life applications? Are there any conditions that would make the key assumptions invalid?

(b)     *Effectiveness and computational tractability*: Is the proposed algorithm and SbO approach computationally tractable and effective (able to generate solutions within a reasonable computation time)?

(c)     *Practical contribution assessment*: Do the results contribute to the actual problem in practice significantly?

(d)     *Dealing with uncertainty and robustness*: Is the proposed approach consider uncertainty explicitly and generate robust solutions that are applicable in practice?

The proposed hybrid Tabu/Scatter Search algorithm, the simulation model and initial solution generation algorithm were all implemented in C++ and complied in a Microsoft Visual Studio 2013 Integrated Development Environment (IDE). All the experiments were performed on a standard PC machine with a 64-bit Intel(R) Core(TM) i5-3210M CPU 2.50 GHz processor and 8 GB of RAM running Microsoft Windows 10 operating system. The statistics were collected, analyzed and visualized by using R Statistical Software Packages and R Studio IDE.

6.1.2 Experimental Design Framework

Both the simulation and optimization (metaheuristic) components of the proposed SbO framework need some necessary parameter setting to adapt to the problem instances at hand, since the choice of parameter values has a significant effect on the quality of the solutions. Unfortunately, there is no one-size-fits-all parameter setting for any given simulation or optimization model. Because one-factor-at-a-time (OFAT) method does not consider the interactions between the parameters, which may significantly affect solution quality and performance, experimental designs are established according to formal procedures from the design of experiments (DoE) field. The DoE methods utilized for simulation experiments are commonly referred as "design of simulation experiments (DoSE)" methods; hence, these two terms are usually used interchangeably. Specifically, the DoE methods are employed for mainly two reasons: (1) to determine the various parameters' main and interaction effects on the solution quality and algorithm efficiency, and (2) to identify the optimal combination of parameter levels.

To this end, a DoE framework is developed which provides a step-by-step approach for formulating an experimental study and for evaluating the results to validate statistical significance. The main steps of the developed DoE framework are shown in Figure 37, and each step of this framework is described in further detail below:

> **Step 1**: *Determine objectives and identify characteristics to be observed*: In this step, objectives for the experimental design are determined and characteristics to be observed are identified. Also, the measurement methods are determined.

> **Step 2**: *Define responses/factors*: In this step, factors, factor constraints, and response(s) of interest are determined. Also, the factor settings (levels) that describe the experimental design space are identified. Since determination of the factors and their initial levels require a priori knowledge, an exploratory study is conducted. This exploratory study consists of several trials on a small subset of instances for preliminary analysis of the potential factors and their initial levels as a starting point.

**Step 3**: *Generate and evaluate the design*: In this step, design decisions, such as the number of experiments, and the number of replications, are specified. As a result, an experimental design is outlined, design matrix is constructed, and the order of experiments is determined. Also, estimation efficiency of the generated design and its power to detect effects are evaluated. Its prediction variance and the correlations between effects are also identified.

**Step 4**: *Conduct experiments*: In this step, experiments are conducted in the pre-determined order, and results are recorded.

**Step 5**: *Analyze the data*: In this step, (linear) regression analysis is applied to the results obtained from each experiment to find a (linear) approximation of the response surface. The factors that have an effect on the response are identified by utilizing response tables and graphs etc.

**Step 6**: *Select optimum levels and run a verification experiment*: In this step, optimum factor setting is selected, and a verification experiment is conducted. When verification experiment had failed, then the steps were repeated starting from Step 2.

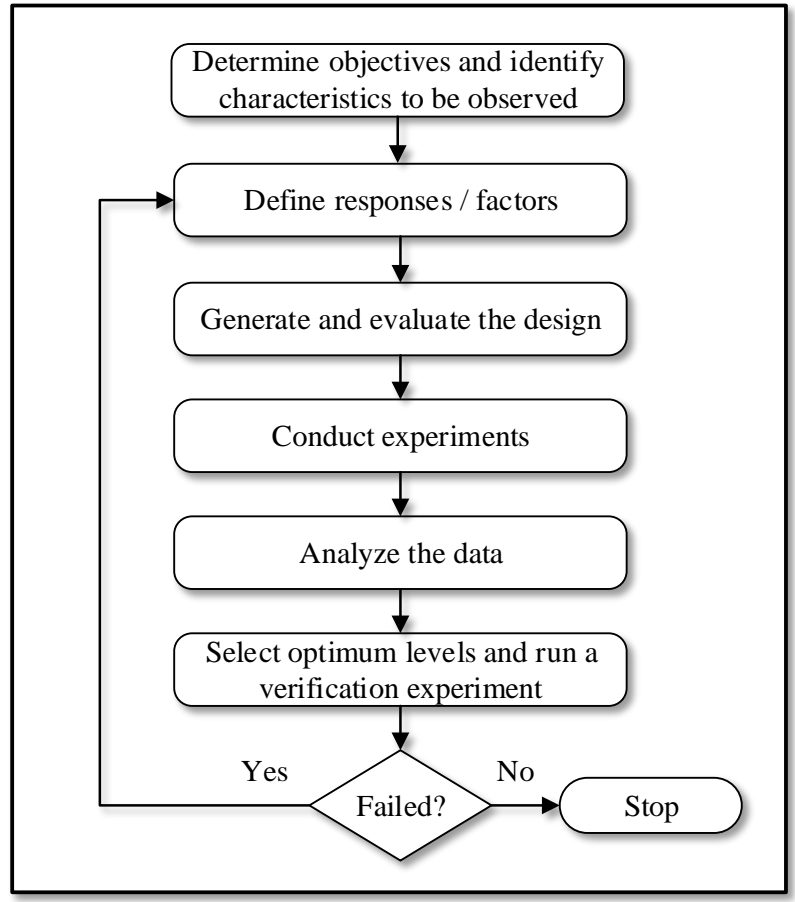

**Figure 37:** Design of Experiments Framework

### 6.1.3 Design of Optimization Experiments

Tuning an optimization (metaheuristic) algorithm to the specific problem being considered is significantly important for achieving high performance in terms of both solution quality and computational time. In this regard, Central Composite Design (CCD), which is a well-known and widely used DoE method, is used to determine the optimal values of the proposed hybrid Tabu/Scatter Search algorithm parameters. CCD allows to estimate all full second-order effects (i.e., main effects, two-way interactions, and quadratic effects) with a reasonable amount of experiments.

CCD typically consists of a full factorial or fractional factorial design ($2^k$ or $2^{k-p}$), a center point, and two points on axes for each factor at a particular distance from the design center which results $2^k+2k+1$ or $2^{k-p}+2k+1$ experiments in total. The experimental design for the optimization algorithm that we assumed (CCD) involves an "L16 Taguchi design" for factorial points ($2^k=16$ experiments), a center point (1 experiment) and axial points ($2k=10$ experiments), and in total 27 experiments. The experimental design that is used in optimization experiments is illustrated in Figure 38.

**Figure 38:** Assumed Central Composite Design

Since the determination of the factors and their initial levels require a priori knowledge of the behavior of the metaheuristic algorithm on the problem instances, a preliminary analysis is performed. This analysis consists of several trials on a small subset of the dataset obtained as part of Input Data Analysis, which is described in Chapter 4. A one-hour period of data is selected randomly from the dataset for determining the potential design factors and their initial levels. As a result of this preliminary analysis, factors that can influence the quality and computation time of the solutions, and their levels are determined as a starting point. Hence, the experimental ranges for each factor (parameter) are identified.

To design our optimization experiments, we identified five design factors ($k$=5) as potentially critical factors. These factors consist of five algorithm parameters that need to be tuned, where each of these design factors has two possible levels. These design factors and their possible levels (low and high levels are denoted as -1 and +1, respectively) are given in Table 8. The first and the second design factors (A and B) are size of the population and the reference set, respectively. The third design factor (C) is the threshold value for the Tabu Search step of the Improvement method, and the fourth design factor (D) is the threshold distance for minimum diversity test procedure in Reference Set Update method. Finally, the last factor (E) denotes the archive capacity.

**Table 8:** Optimization Algorithm Design Factors and their Possible Levels

| Design Factor | Low level (-1) | High level (+1) |
|---|---|---|
| A - Population size | 50 | 200 |
| B - Reference set size | 10 | 30 |
| C - Improvement threshold value | 2 | 10 |
| D - Threshold distance | 5 | 20 |
| E - Archive capacity | 30 | 80 |

Exploratory optimization experiments are conducted based on this experimental design to determine appropriate parameter values, which are presented in the next section.

6.1.4 Design of Simulation Experiments

Although discrete-event simulation is a widely accepted technique to estimate accurately the key output performance measures of complex systems, such as airport runways, the execution time of this kind of simulation is usually slow and can only evaluate one scenario at a time (except parallel simulations). Moreover, this kind of complex systems typically involves a large number of input parameters which potentially affect the system's output performance. In order to overcome these difficulties design of simulation experiments (DoSE) methods are utilized to reduce the number of input parameters with eliminating the unimportant ones. As a result, a smaller set of input parameters can be examined in a more efficient and effective way, and the interactions between these parameters can also be identified.

Since simulation responses typically have a random component, the input parameter strategy should have error control for misclassification of factors, which includes the probability of classifying a factor as important when it is not (Type I Error) and the probability of classifying a factor unimportant when it is important (Type II Error) (Law, 2014). The main objective of DoSE is to find which factors (input parameters) have the greatest effect on the response (output performance measure), where the effect of each factor can be formally estimated, and for a small number of factors the interactions between factors can also be identified (Kleijnen, 2007).

Space-filling designs are commonly considered as suitable for simulations that are complex and involve variables with complicated interrelationships. The main idea in this type of designs is to find a simpler empirical model that adequately predicts the behavior of the system over limited ranges of the factors. The most widely used space-filling design method for simulation experiments is the Latin Hypercube Sampling (LHS). Considering the challenges regarding simulation model's complexity and involvement of interrelated variables, LHS is chosen for exploring the interior of the parameter space and for determining the parameters of the simulation model.

LHS design maximize the minimum distance between design points but requires even spacing of the levels of each factor, and it provides an orthogonal array that randomly samples the entire design space partitioned into regions of equal probability. LHS can be considered as a stratified Monte Carlo sampling method, where the pairwise correlations can be minimized to a small value that is necessary for uncorrelated parameter estimates. LHS is useful especially for exploring the interior of the parameter space and for limiting the experiment to a fixed or a user-defined number of combinations. This technique ensures that the entire range of each parameter is sampled. LHS has good space filling properties, so they are efficient ways of exploring unknown, but potentially complicated response surfaces with many quantitative factors. Furthermore, LHS is flexible enough for exploring complex simulation models when information about the response surfaces is limited (Sanchez, 2005).

To design our simulation experiments, we listed five design factors as potentially critical factors. These design factors and assumed LHS design for simulation experiments are given in Table 9. These design factors include the following: seed for pseudo-random number stream (multiplied by 100) (A), minimum number of replications (B), percentage of relative error to achieve certain precision for performance measures (C), number of initial samples (D), and standard error threshold (E). It is worth mentioning that since in each optimization iteration multiple simulation runs are performed, seed for pseudo-random number stream is considered as an important factor.

It should also be stressed that the choice of LHS for the simulation model does not necessarily depend only on the power of the design and available resources, but also depends on the sample size (number of replicates), selection of a suitable run order for the experimental trials, and determination of whether or not randomization restrictions are involved.

Exploratory simulation experiments are conducted based on this experimental design to determine appropriate parameter values, which are presented in the following section.

**Table 9:** Assumed LHS Design

| Design Factor | A | B | C | D | E |
|---|---|---|---|---|---|
| **Low level (-1)** | 1 | 1 | 1 | 1 | 1 |
| **High level (+1)** | 100 | 10 | 20 | 15 | 25 |
| Experiment 1 | 32 | 10 | 16 | 6 | 7 |
| Experiment 2 | 7 | 3 | 18 | 9 | 1 |
| Experiment 3 | 13 | 5 | 2 | 5 | 16 |
| Experiment 4 | 20 | 7 | 7 | 15 | 15 |
| Experiment 5 | 75 | 9 | 9 | 3 | 9 |
| Experiment 6 | 100 | 4 | 8 | 12 | 3 |
| Experiment 7 | 63 | 3 | 20 | 5 | 22 |
| Experiment 8 | 57 | 9 | 15 | 14 | 21 |
| Experiment 9 | 51 | 6 | 11 | 8 | 13 |
| Experiment 10 | 69 | 1 | 5 | 10 | 19 |
| Experiment 11 | 94 | 8 | 3 | 7 | 25 |
| Experiment 12 | 88 | 6 | 19 | 12 | 10 |
| Experiment 13 | 81 | 4 | 14 | 1 | 12 |
| Experiment 14 | 26 | 2 | 12 | 13 | 18 |
| Experiment 15 | 1 | 7 | 13 | 4 | 24 |
| Experiment 16 | 38 | 8 | 1 | 11 | 4 |
| Experiment 17 | 44 | 2 | 6 | 2 | 6 |

## 6.2 Exploratory Optimization Experiments

Exploratory optimization experiments are conducted to locate algorithmic bottlenecks and guide parameter tuning efforts for the proposed hybrid Tabu/Scatter Search algorithm. For these experiments, small-scale problem instances are randomly generated to reflect realistic schedules of runway operations. The empirical validation with this synthetic dataset is only an initial step for algorithmic improvements and accompanying validation. This real-like synthetic problem instances are generated according to the following guidelines. Aircraft are generated assuming a Poisson process with respect to exponential inter-arrival times, because this assumption has been highly utilized and typically considered as acceptable in the literature (Balakrishnan & Chandran, 2010). The deviations from estimated landing/take-off times, the runway occupancy time distributions, and the transit time distributions are estimated by analyzing the FAA Operations & Performance database and by experimenting with the MITRE Corporation runwaySimulator.

In exploratory optimization experiments, to verify the statistical validity of the results and to ensure that the effects of the different levels of the factors are statistically significant, Main Effects Plot is used to determine the level of each factor, where the mean values of each level of a factor are shown graphically. Also, Interaction Plots are used to determine the mean values for each level of a factor with the level of a second factor held constant, which specifies that the effect of one factor is dependent on a second factor.

The parameter tuning is conducted in separate for MOO and SbO experiments in order to find the best parameter setting for each experimental setting. The primary difference between these two separate experiments is the termination criterion. In the MOO experiments, the algorithm terminates when 10000 function evaluations are computed. On the other hand, for the SbO experiments termination criterion is chosen in accordance with the planning horizon for the practical problem, which is 20 minutes.

## 6.2.1 Parameter Setting for MOO Experiments

According to the chosen experimental design for the optimization experiments, namely CCD, 27 experiments are performed and each of these experiments are replicated 30 times. These 30 replications are then averaged to obtain a response for each experiment. In order to statistically determine for each experimental condition if these design factors have a significant effect on the responses, an analysis-of-variance (ANOVA) test performed. Since an underlying assumption of the ANOVA test is that responses are samples from normally distributed populations, normal-scores plots are used to identify non-conformance to this assumption. Each experimental configuration is examined for only the high level and the low level, and separate normal-scores plot is constructed for each test.

As a result, it is concluded that the assumption that the responses of the experimental design are sampled from normally distributed populations is valid.

For the MOO experiments, the first design factors (A), i.e. size of the population (*Psize*) is set to 100. The second design factor (B), i.e. the size of the reference set (*b*) is set to 20, where half of the solutions in the reference set are selected according to their diversity (*b2*=10). The third design factor (C), i.e. the threshold value for the Tabu Search step of the Improvement method, is set to 7. The fourth design factor (D), i.e. the threshold distance for minimum diversity test procedure in Reference Set Update method, is set to 17. Finally, the last factor (E), which denotes the archive capacity is set to 55. Regression analysis conducted to determine if the fit was supported statistically, and results of the regression analysis yielded an adjusted $R^2$ value of 0.901, which indicates a strong relationship between the variables.

### 6.2.2 Parameter Setting for SbO Experiments

A subset of real-life data that was used for Input Data Analysis (presented in Chapter 4) is selected randomly for the exploratory experiments. According to the chosen experimental design explained previously, namely CCD, exploratory experiments are performed, and solutions are found.

For the SbO experiments, the first design factors (A), i.e. size of the population (*Psize*) is set to 120. The second design factor (B), i.e. the size of the reference set (*b*) is set to 22, where half of the solutions in the reference set are selected according to their diversity (*b2*). The third design factor (C), i.e. the threshold value for the Tabu Search step of the Improvement method, is set to 6. The fourth design factor (D), i.e. the threshold distance for minimum diversity test procedure in Reference Set Update method, is set to 14. Finally, the last factor (E), which denotes the archive capacity is set to 45. After completing regression analysis of the results, the adjusted $R^2$ values is found as 0.926, which indicates a strong relationship between the variables.

**6.3 Exploratory Simulation Experiments**

The main objective of the exploratory simulation experiments is to investigate the most critical and sensitive parameters, and to ensure the tolerance to which model outputs can be expected to alter with given input parameters, where this information also gives insight for determining the bounds beyond that application of the simulation model is not appropriate.

6.3.1 Variance Reduction Techniques

The simulation model is set up to implement two important variance reduction techniques: (1) common random numbers method is utilized to generate the sequence of pseudo-random number streams for uncontrollable factors in simulation experiments, and (2) antithetic variates is utilized to generate antithetic samples between successive pairs of replications. These variance reduction techniques are employed primarily to enhance the refinement of the simulation model.

*Common random numbers method*: The behavior of simulation model usually changes from one simulation run to next by simply changing the values utilized for the underlying pseudo-random number streams. Hence, to be confident that any observed differences in performance of alternative configurations are not due to fluctuations of the experimental conditions generated by pseudo-random numbers. We formed a 90 percent confidence interval for comparing the alternative configurations by using common random numbers method and observed 10.2 percent decrease in the variance.

*Antithetic variates*: The basic idea in this technique is that the variance of the simulation outputs might be reduced by using pseudo-random numbers that are negatively correlated in each pair of simulation runs. This is achieved by pairing the simulation runs, and if one of the pairs uses a stream of (0, 1) random variables $x(j)$, then the other pair should use stream of $y(j)$, where $y(j) = 1 - x(j)$. After applying this technique, the simulation outputs of each simulation run cannot be considered as independent; however, each pair of simulation run can be considered as independent. Therefore, the degrees of freedom in any average across $n$ simulation runs is taken as *(n/2)-1* instead of *n-1*, and as a result, any set of observations of an output across pairs of simulation runs is assumed to be normally distributed.

### 6.3.2 Dealing with Noise

One of the primary challenges in optimizing the simulation model is the fact that evaluation of each candidate solution is influenced by noise, where the source of this noise is the stochastic nature of the simulation model. The noise has a huge potential to undermine the performance of the proposed hybrid Tabu/Scatter Search algorithm by misleading the reference set to a local optimum and deteriorating the convergence rate. The proposed algorithm partially alleviates the effects of this noise by utilizing a set of solutions (reference set) and averaging these effects. Also, it does not require derivative and gradient information, which is a difficult task to approximate this information in the existence of noise.

The main effect of this simulation noise is that a high-quality solution might be evaluated lower than its true fitness value, and likewise, a low-quality (poor) solution might be evaluated higher that its true fitness value. This effect typically leads the search to a non-promising region in the search space, easily renders the optimization process unstable, and degrades the algorithm's performance. One way to deal with the noise is to increase the number of iterations and to utilize fitness averaging simply by evaluating each candidate solution several times and using the average fitness of these evaluations as the fitness of the candidate solution. However, this approach comes with the expense of high computational costs.

Several resampling schemes proposed in the literature that utilize resampling in order to reduce the noise of fitness evaluations in which fixed number of solution resampling simulation runs are distributed unevenly among the solutions. This unevenly distribution allow spending the biggest share of the computation time on the most promising solutions. This noise compensation technique is commonly referred as "dynamic resampling" (Syberfeldt et al., 2010). In this method, the most critical step is to find the best compromise between the number of solutions evaluated and the number of samplings of each solution. If more solutions are evaluated, the search space can be explored more widely, and in turn, a probability of finding Pareto-optimal solutions will increase.

To deal with the simulation noise, a dynamic resampling method, in particular Standard Error Dynamic Resampling (SEDR), is employed. SEDR is a sequential sampling method proposed by Pietro et al. (2004). It allocates sampling budget individually for each solution depending on the noise level (uncertainty) of the solution's fitness which is calculated by determining the standard sample deviation of the samples taken. The standard error of a solution $s$ decreases as the solution resampled (Eq. 6.1). The sample standard deviation is shown in Eq. 6.2. The solution is iteratively sampled until the standard error is below a threshold $SE_{threshold}$.

$$SE_{sk} = \frac{\sigma_{sk}}{\sqrt{n}}, \qquad k = 1,2. \tag{6.1}$$

$$\sigma_{sk} = \sqrt{\frac{1}{n-1}\sum_{i=1}^{n}(s_k - \mu_{sk})} \ , \qquad k = 1,2. \tag{6.2}$$

The pseudo-code for the utilized SEDR procedure is given below:

---

**Algorithm 5** Standard Error Dynamic Resampling (SEDR) (Pietro et al., 2004)

---

**Input**: Solution $s$ and parameters $t_{min}$ and $SE_{threshold}$

1:    **begin**
2:        perform $t_{min}$ initial samples of the fitness of $s$
3:        calculate mean of the available fitness samples for both objectives $\mu_{s1}$, $\mu_{s2}$
4:        calculate objective sample standard deviations with available fitness samples $\sigma_{s1}$, $\sigma_{s2}$
5:        calculate standard errors for both objectives $SE_{s1}$, $SE_{s2}$
6:        calculate average standard error $ASE_s$
7:        **if** average standard error ($ASE_s$) < standard error threshold ($SE_{threshold}$)
8:            **then** sample the fitness of $s$ one more time and goto step 2
9:    **end**

---

6.3.3 Parameter Setting

For setting the parameters, previously explained statistical experimental design (Latin Hypercube Sampling) is applied and experiments are performed. The performance characteristic of interest during parameter setting is the average CPU time. In order to achieve certain precision for performance measures, relative error is selected as $\gamma = 0.1$. Before simulation runs, to warm up to reach steady state the simulation model run for 10 minutes, and the system reached steady state after 8.5 minutes of simulation runs. Initially, the simulation model was run for each design factor configuration and the mean response for each configuration is estimated by the sample average of the output from the corresponding simulation run. The results are obtained from the simulation runs where all random number streams are seeded independently.

After all the experiments are completed and the responses are calculated for each experiment, linear regression analysis is conducted to determine if the fit was supported statistically, and results of the regression analysis yielded an adjusted $R^2$ value of 0.927, which is a measure of association between the variables with a value of zero indicating no correlation exists and a value of one representing the strongest correlation possible. In this case, the adjusted $R^2$ value indicates a strong relationship between the variables.

After finding a linear approximation of the response surface, the path of steepest descent on the response surface is calculated and small steps are made along this path by changing the parameter values. At each step, one trial is conducted and the process is continued until the limit of the experimental region is reached. The parameter vector associated with the best result found during this process is determined as the final parameter setting, where the most significant results are obtained with a medium level for A, C, D and E, and high level for B. As a result, the parameter values for the seed for pseudo-random number stream (multiplied by 100) (A) is set to 84, minimum number of replications (B) is set to 10, percentage of relative error to achieve certain precision for performance measures (C) is set to 17, number of initial samples (D) is set to 14, and standard error threshold (E) is set to 22.

## 6.4 Multi-Objective Optimization Experiments and Results

Since the main focus of the dissertation is on developing a multi-objective hybrid Tabu/Scatter Search algorithm, Multi-Objective Optimization (MOO) experiments are conducted separately to evaluate the proposed algorithm's performance based on multi-objective benchmark problems. As mentioned in Chapter 3, the three primary goals of a MOO problem are minimal distance to the Pareto-optimal front, good distribution, and maximum spread. Hence, the proposed hybrid Tabu/Scatter Search is evaluated based on all of these three goals. In this section, related performance metrics, experimental setup and results of the experiments are presented.

### 6.4.1 Performance Metrics for Multi-Objective Optimization

It is a challenging task to evaluate the performance of a MOO algorithm since the algorithm generates a set of solutions instead of a single value, but several performance measures exist to evaluate approximations of the Pareto-optimal set generated by the algorithm. These measures are usually based on the convergence rate of the optimization and diversity of the solutions, where all of MOO objectives are considered. However, most of these measures are only suitable for problems where the Pareto-frontier is known. For real-life optimization problems, the Pareto-frontier is typically unknown, and an appropriate performance measure is required that does not rely on this information to assess the convergence and spread of solutions in the algorithm.

As shown in Table 10, two primary performance metrics are selected for evaluating the performance of the proposed hybrid Tabu/Scatter Search algorithm, which are hyper-volume metric and CPU time. The hyper-volume metric (also referred as *S* metric) is chosen for the evaluation in terms of convergence rate and diversity of solutions that are generated, which is the most popular and significant performance metric for MOO algorithms in the literature. The second metric is related to computational performance, namely total CPU time, for finding near Pareto-optimal solutions.

**Table 10:** Multi-Objective Optimization Performance Metrics

| Category | Performance Metrics |
|---|---|
| Convergence rate and diversity of solutions | Hyper-volume metric |
| Computational performance | Total CPU time |

The hyper-volume measure is the area of the dominated region by a non-dominated solution set, and a reference point is needed for calculating hyper-volume measure, which is a point weakly dominated by all vectors in the Pareto-frontier (Deb, 2001). In the experiments, the origin of the objective space is used as the reference point. The hyper-volume metric provides a single measurement to assess both the convergence and spread of a Pareto-optimal set of solutions, and it does not rely on knowledge of the Pareto-frontier. The main strength of this measure is that it is strictly monotonic with respect to Pareto dominance. To avoid favoring for objectives with higher absolute value, the hyper-volume metric is typically calculated in a normalized objective space, where each objective function value is normalized to a common interval. The hyper-volume metric ($HVM$) can be formulated as follows:

$$HVM = \bigcup_{i=1}^{|P|} V_i^R \tag{6.3}$$

where $V_i$ is the volume of the objective space dominated by solution $i \in P$ with respect to the reference point $R$.

In the objective space with only two objectives, the hyper-volume metric measures the area of the objective space that is weakly dominated by the image of the solutions of a non-dominated set, where this area is bounded by the reference point *R* (Figure 39). The size of this area reflects the quality of the non-dominated set according to the hyper-volume metric. The larger this area, the greater the hyper-volume metric. An algorithm with greater hyper-volume metric is considered to be superior since it measures both the convergence and the spread of the solution to the Pareto-frontier. We applied the hyper-volume metric by utilizing normalized objective function values due to the possibility of arbitrary scaling of the objectives. It is worth to mention that the computational complexity of computing the hyper-volume metric for a set of *n* solutions with two objectives is *O(nlog(n))*.

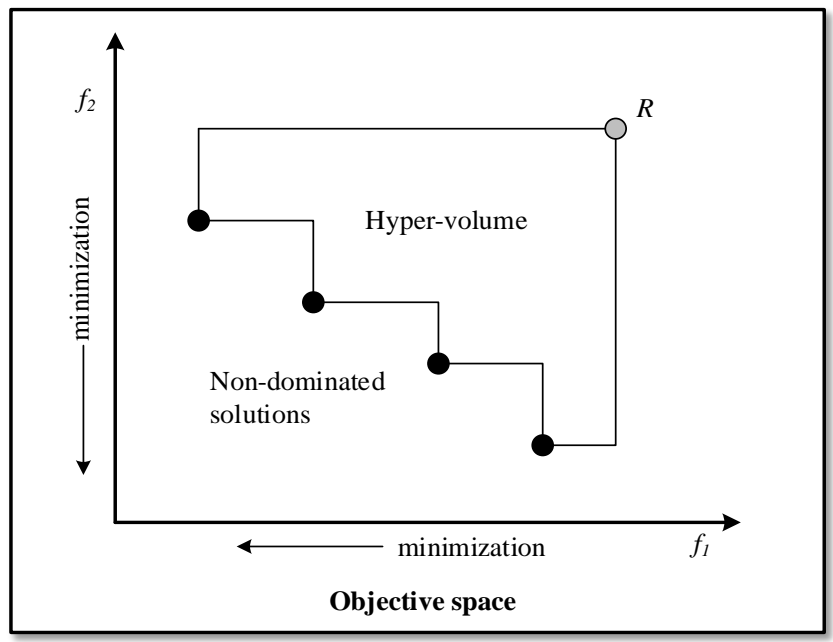

**Figure 39:** Illustration of Hyper-Volume Metric

The second metric, total CPU time, evaluates the computational performance of the algorithm by calculating total computation time spent for finding the approximate Pareto-frontier. This metric is significantly important for the solution time requirement based on the planning horizon of the practical runway operations scheduling problem.

### 6.4.2 Experimental Setup and Results

In MOEA literature, the performance of algorithms is widely evaluated by using standardized benchmark problems. Although computational time required to solve these benchmark problems is much lower than a real-life problem's solution time, they allow to compare different algorithms as well as to replicate the experiments. The two benchmark problems that are chosen for MOO experiments are: (1) Fonseca and Fleming's two-objective minimization problem, denoted as "F&F," and (2) Zitzler–Deb–Thiele's function number 3, denoted as "ZDT3." The main motivation behind utilizing these problems is that they both have challenging characteristics regarding convergence to actual Pareto-frontier and maintain diversity, and they represent features of real-life problems.

The first test function, Fonseca and Fleming's bi-objective minimization test problem, has been widely used as a benchmark problem for MOO and suggested by Fonseca and Fleming (1996). The solution to the problem has large and non-linear trade-off curve that challenge the algorithm's ability to find and maintain the entire Pareto-frontier uniformly. F&F benchmark function is shown below:

$$\min f_1(x_1, x_2) = 1 - \exp\left(-\sum_{i=1}^{2}\left(x_i - \frac{1}{\sqrt{2}}\right)^2\right)$$

$$\min f_2(x_1, x_2) = 1 - \exp\left(-\sum_{i=1}^{2}\left(x_i + \frac{1}{\sqrt{2}}\right)^2\right) \tag{6.4}$$

$$-4 \leq x_i \leq 4 \qquad \forall\, i = 1,2.$$

Based on the guidelines proposed by Deb (1999) in the development of benchmark problems for MOO, Zitzler et al. (2000) proposed six ZDT series benchmark problems. ZDT3 is one of these six problems, which has two objective functions and two decision variables. The Pareto-optimal set for this benchmark function comprises several discontinuous convex parts in the objective space. ZDT3 benchmark function is given below:

$$\min f_1(x) = x_1$$

$$\min f_2(x) = g(x)\, h(f_1(x), g(x))$$

$$g(x) = 1 + \frac{9}{29}\, x_2$$

$$h(f_1(x), g(x)) = 1 - \sqrt{\frac{f_1(x)}{g(x)}} - \left(\frac{f_1(x)}{g(x)}\right) \sin\left(10\pi f_1(x)\right)$$

$$0 \le x_i \le 1 \qquad \forall\, i = 1,2. \tag{6.5}$$

In order to evaluate performance under noisy conditions, a noise (random variation) element is integrated into the test problems by applying noise as an additive normal distributed perturbation with zero mean (Eq. 6.6). To mimic the simulation noise, the noise element is assumed to have a disrupting effect on the value of each solution in the objective space.

$$\overline{f(x)} = f(x) + Normal(0, \sigma^2) \tag{6.6}$$

where *Normal* denotes the normal distribution, and $\sigma^2$ represents the existing level of noise.

The non-elitist and elitist versions of the proposed hybrid Tabu/Scatter Search algorithm are compared with respect to the magnitude of noise present. Computational experiments are performed at noise levels of $\sigma^2 = \{0.01, 0.05, 0.1, 0.15, 0.2\}$ to assess the performance under the impact of noise.

The results from the noisy benchmark functions are shown in Table 11. The results are based on normalized objective values and constitute the average of 500 independent runs. Based on the reference point, the hyper-volume metric value is normalized between 0 and 1, where the optimization is performed for 10000 function evaluations.

**Table 11:** Benchmark Results

| Metric | Elitist HT/SS | | Non-Elitist HT/SS | |
|---|---|---|---|---|
| | HVM | CPU time (seconds) | HVM | CPU time (seconds) |
| F&F + 0.01 noise | 0.956 | 332 | 0.908 | 563 |
| F&F + 0.05 noise | 0.805 | 465 | 0.753 | 741 |
| F&F + 0.10 noise | 0.731 | 598 | 0.674 | 1102 |
| F&F + 0.15 noise | 0.657 | 846 | 0.442 | 1681 |
| F&F + 0.20 noise | 0.551 | 1278 | 0.367 | 1977 |
| ZDT3 + 0.01 noise | 0.978 | 231 | 0.921 | 367 |
| ZDT3 + 0.05 noise | 0.826 | 294 | 0.695 | 463 |
| ZDT3 + 0.10 noise | 0.721 | 583 | 0.553 | 896 |
| ZDT3 + 0.15 noise | 0.716 | 965 | 0.463 | 1430 |
| ZDT3 + 0.20 noise | 0.664 | 1164 | 0.457 | 1731 |

Average hyper-volume metrics and CPU times for (a) F&F and (b) ZDT3 attained by non-elitist and elitist versions of the proposed hybrid Tabu/Scatter Search algorithm (HT/SS) under the influence of different noise levels are shown in Figure 40 and 41, respectively.

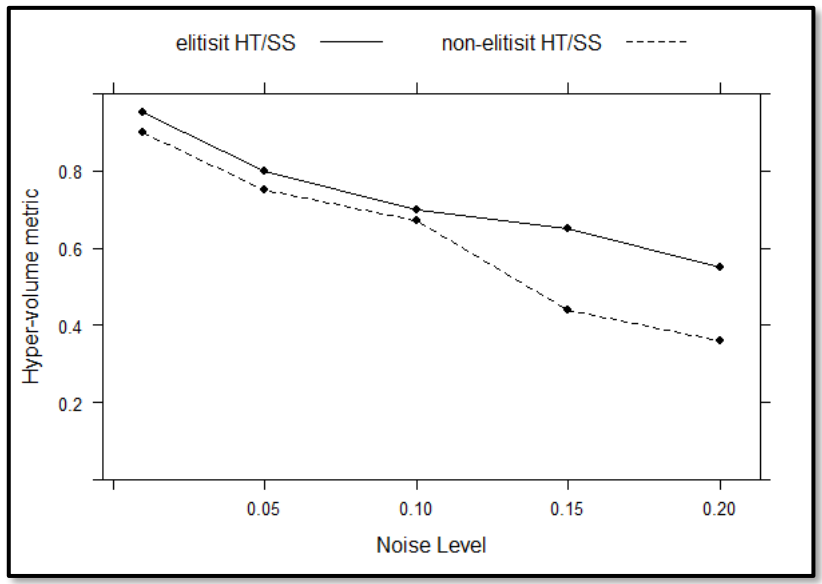

(a) F&F

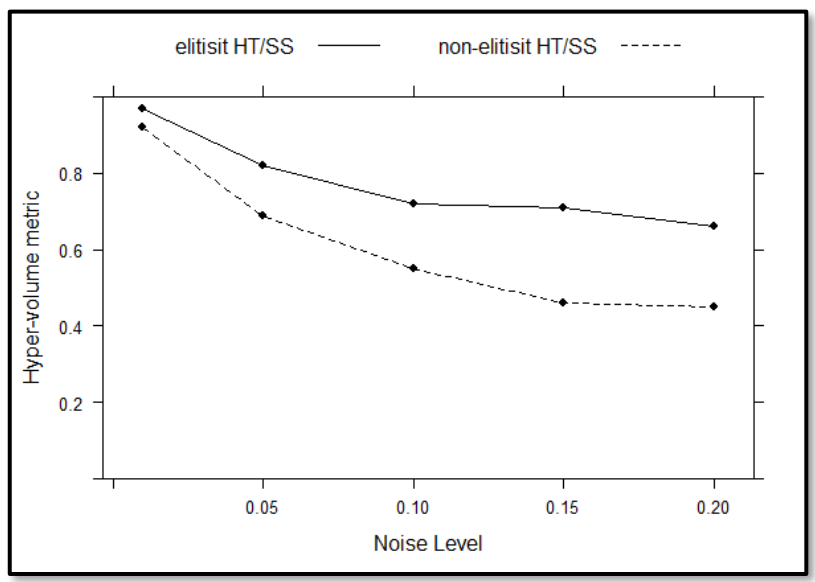

(b) ZDT3

**Figure 40:** Hyper-Volume Metric under Different Noise Levels

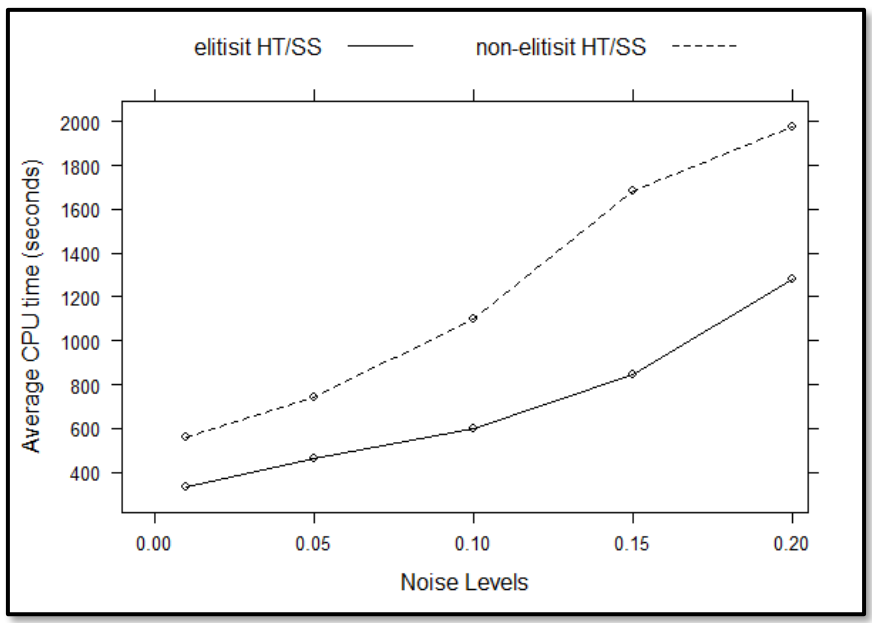

(a) F&F

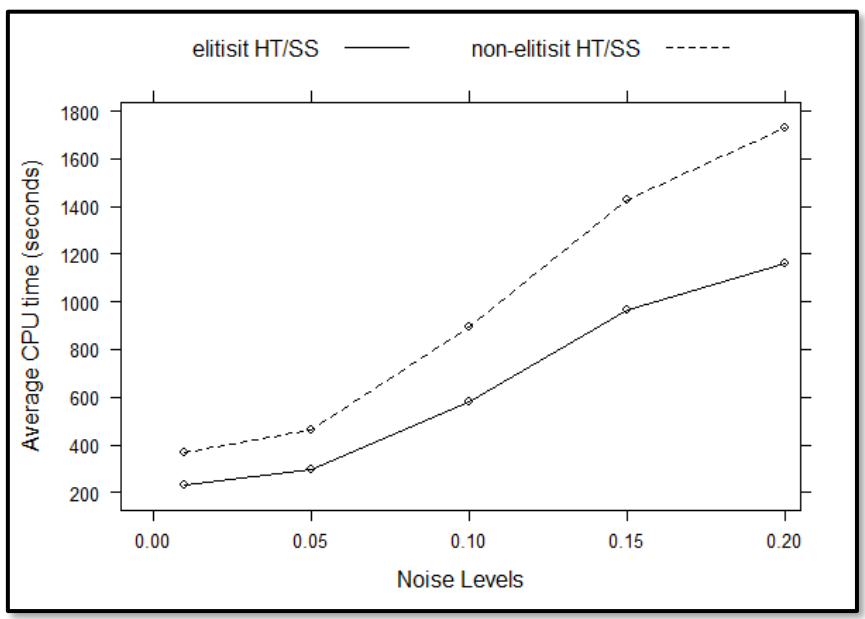

(b) ZDT3

**Figure 41:** Average CPU Time under Different Noise Levels

One of the primary requirements for an effective optimization component in a SbO framework is that it should be able to effectively handle different levels of noise. As Table11, Figure 36 and 37 illustrates, the elitist version of the proposed optimization algorithm outperforms the non-elitist version with respect to performance measures over several levels of noise. To compare the results whether the performance difference is statistically significant, we have applied a non-parametric Kruskal-Wallis test at a significance level $\alpha = 0.05$. The Kruskal-Wallis test indicated that the values have a statistical confidence in the sense that the differences are unlikely to have occurred by chance with a probability of 95 percent.

In addition, one-way ANOVA tests are performed to determine the effect of noise on the performance of both versions of the algorithm with respect to hyper-volume metric. The results of these one-way ANOVA tests revealed that the performance difference between versions of the algorithm is statistically significant with a significance level $\alpha = 0.05$ for all noise levels.

As a result, comparison of non-elitist and elitist HT/SS in terms of convergence rate and diversity of solutions as well as computational time shows that elitist version of the proposed algorithm yields greater values of hyper-volume metric and CPU times. It illustrates that solutions offered by the elitist version are closer to the Pareto-optimal front compared to the non-elitist version. Considering the efficiency of the algorithms, elitist HT/SS needs less time to complete all function evaluations. In summary, experiments provide evidence that elitist HT/SS presents better results than the non-elitist version in terms of both effectiveness and efficiency.

Finally, the effect of dynamic update mechanism in the Solution Combination method and the rebuilding strategy on the proposed algorithm's convergence rate is evaluated. To achieve this, a comparison is performed between the dynamic and the static update mechanisms in the Solution Combination method using the elitist version of the proposed hybrid Tabu/Scatter Search algorithm. Also, another comparison is conducted between the elitist version of the algorithm with and without the rebuilding strategy.

For these comparisons a commonly used measure for evaluating convergence, namely the $Y$ metric, is used. This metric measures the degree of convergence by calculating the average minimum Euclidean distances from each of the obtained non-dominated solutions to the closest solution in the true Pareto-frontier (Deb et al., 2002). It is worth to mention that the smaller the value of $Y$, the better the convergence rate of the algorithm. Figure 42 shows the comparison between the algorithm with the dynamic update mechanism and with the static mechanism in the Solution Combination under different noise levels, and Figure 43 illustrates the comparison between the algorithm with the rebuilding strategy and without the rebuilding strategy under different noise levels.

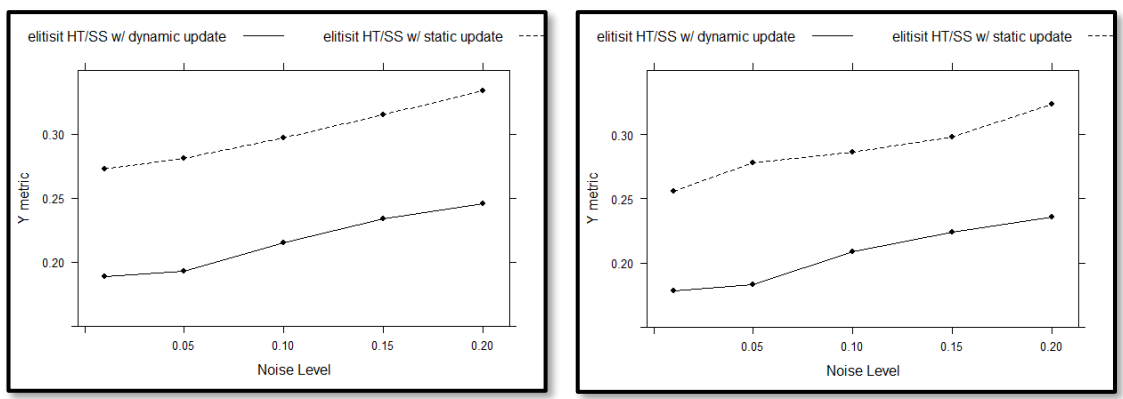

(a) F&F                                 (b) ZDT3

**Figure 42:** Comparison of the Proposed Algorithm with Different Update Mechanisms

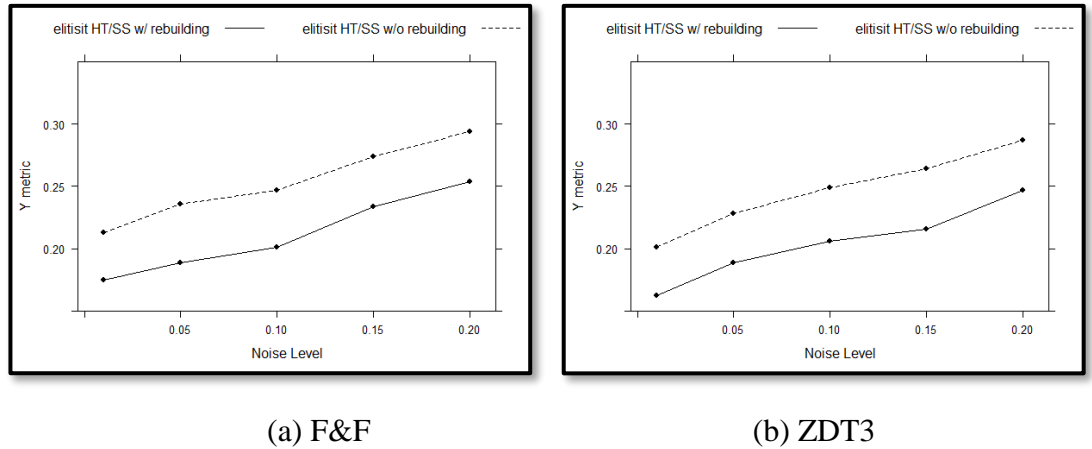

(a) F&F                    (b) ZDT3

**Figure 43:** Comparison of the Proposed Algorithm with and without Rebuilding Strategy

The results illustrate that the dynamic update mechanism in the Solution Combination method is an effective scheme in terms of convergence strength for generating new Pareto-optimal solutions compare to the static update mechanism. Also, results show that the rebuilding strategy, which partially rebuilds the reference set when the Solution Combination and Improvement methods do not provide diverse solutions, is an effective scheme in improving convergence towards the Pareto-frontier. We applied a non-parametric Kruskal-Wallis test to compare the results whether the performance difference is statistically significant at a significance level $\alpha = 0.05$, which indicated that the values have a statistical confidence in the sense that the differences are unlikely to have occurred by chance with a probability of 95 percent.

### 6.4.3 Key Findings from Multi-Objective Optimization Experiments

The above computational multi-objective optimization experiments led us to the following findings:

(a)     The computational results for two benchmark problems with different Pareto-optimality characteristics indicate that the proposed hybrid Tabu/Scatter Search algorithm is able converge to and provide diversity on the Pareto-frontiers of both benchmark problems. The benchmark problems, where a noise factor is added to these problems, provided necessary complexity to evaluate the proposed algorithm.

(b)      During the experiments the elitist version of the proposed hybrid Tabu/Scatter Search algorithm was compared with the non-elitist version in terms of both hyper-volume metric and computation time. The results concluded that the performance of the algorithm improves significantly when elitism strategy is employed. However, elitism strategy should be applied carefully in presence of noise since noise enhanced solutions in the archive might prevent true high-quality solutions out of the archive, and in turn, the search process might be biased towards less promising regions in the search space.

(c)      Also, the effect of the magnitude of the noise on both versions of the algorithms is evaluated. The results show that noise has a detrimental effect on the algorithm's performance in terms of convergence and diversity, which is observed as high. It is also observed that optimization process degrades as the level of noise increases. Simulating a solution multiple times reduces the noise by a factor of number of simulation runs, however, this comes at the expense of a higher computational time.

(d)      The dynamic update mechanism and the rebuilding strategy, which are employed in the proposed algorithm, significantly contribute to its convergence capability. The main strength of the dynamic update mechanism comes from its application in the Solution Combination method to new candidate solutions in such a way that it combines these solutions faster compare to a static update mechanism. On the other hand, the key strength of the rebuilding strategy stems from its capability of partially rebuilding the reference set when the Solution Combination and Improvement methods are not capable of generating solutions of satisfactory quality to dislocate current solutions in the reference set.

(e)      Utilizing adaptive memory structures is important for creating Scatter Search algorithms for solving practical MOO problem instances. This strategy observed to be systematic in the sense that it progresses towards to Pareto-optimal rather than revisiting the earlier developed solutions too many times unnecessarily.

(f)     The evidence obtained from the experiments show that the proposed algorithm can converge to multiple solutions simultaneously by encouraging competition between solutions within the same local optimum neighborhood. This is achieved mainly by maintaining a good balance between quality and diversity in the reference set. Also, diversity preservation and two-tier structure of the reference set prevent the optimization process from focusing on a specific part of the Pareto-frontier while neglecting the rest.

(g)     Although the SS algorithm template defines the generic strategies, to develop an effective SS algorithm still requires many design decisions to be made and a balance between diversification and intensification mechanisms to be adjusted. Experiments revealed that the dominance procedure increases the exploration capabilities of the optimization process.

(h)     Finally, employing Scatter Search's systematic and strategically designed mechanisms instead of probabilistic rules of evolutionary methods for solving multi-objective optimization problems provides a robust framework for developing gradually improved methods. In addition, connecting Scatter Search with the Tabu Search setting, where adaptive memory structures and responsive exploration mechanisms are used, makes it suitable for simulation-based optimization, which requires a capability of searching the solution space economically and effectively.

## 6.5 Experimental Setup for Simulation-based Optimization

### 6.5.1 Runway Operations Performance Metrics

As in any complex system, there is no single best performance metric that captures every aspect of runway operations. Therefore, different performance metrics are used for evaluation. Airport capacity and air traffic delays are two of the principal performance metrics for airports in practice. Airport capacity is a measure of the maximum number of runway operations (landing or take-off) that can be accommodated on an airport within a given period, which is usually an hour, with 95 percent confidence level (Odoni et al., 1997a). While estimating this measure, several assumptions required to be incorporated regarding minimum separation requirements, fleet mix, weather conditions and technological aides. A variety of tools and techniques are used in estimating airport capacity ranging from analytical models to simulation tools. However, it is commonly estimated with the help of simulation tools because certain aspects of the runway operations cannot be reasonably addressed by using existing analytical models.

The airport capacity is usually illustrated by a Pareto-frontier, which shows the maximum number of arrivals and departures that can be performed within one hour. As shown on Figure 44, all observed runway throughput values are within the capacity frontier. Expanding runway infrastructure may expand the feasible region, but most of the time it is not feasible or practical. Due to this fact, the better option is to increase airport capacity by sequencing aircraft such a way that total separation requirements are minimum.

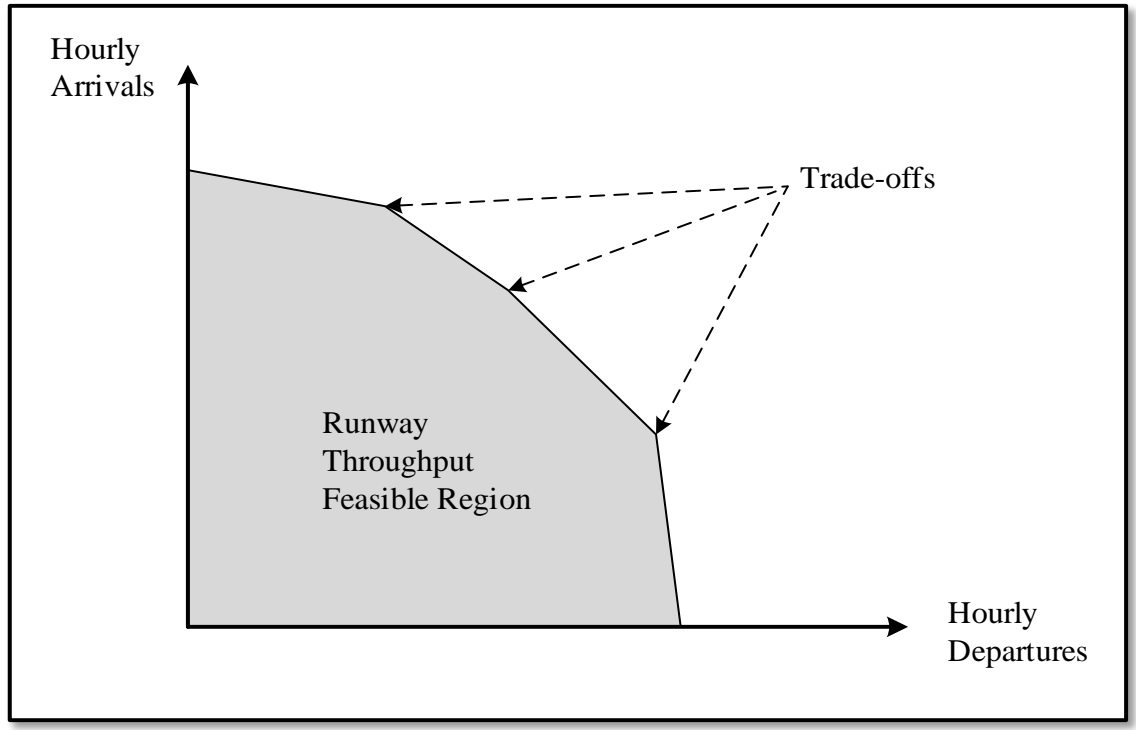

**Figure 44:** Typical Airport Capacity Curve

The most commonly used definition of airport capacity is practical runway throughput (utilization), which is the number of runway operations that can be accommodated. Aircraft delays are also an important performance metric in practice because of costs associated with them and their undesirable consequences, such as missed flight connections, cancellations, and diversions, etc. Aircraft sequence changes are also considered as an essential metric since these changes account for the workload of air traffic controllers, and also, this metric is a representation of the fairness among aircraft. Therefore, average runway utilization, average and longest runway operation delays, and average sum of sequence changes are determined as the metrics for evaluating the performance of runway operations, as given in Table 12.

**Table 12:** Runway Operations Performance Metrics

| Category | Performance Metrics |
|---|---|
| Capacity | Average runway utilization |
| Delays | Average landing delay |
| | Longest landing delay |
| | Average take-off delay |
| | Longest take-off delay |
| Position shifts | Average sequence change |

The runway utilization is calculated for every 5-minute interval as a percentage of time in each interval for which runways are being used for active runway operations. The types of active runway operations are listed as follows: (a) final approach, which is the time an aircraft enters to final approach fix to touchdown time, (2) runway occupancy for landing and take-off, which is the time between touchdown and leaving the runway for arriving aircraft, and the time between start of take-off roll to wheels-off for departing aircraft.

The average landing and take-off delays are calculated by aggregating all aircraft delays for each runway operations, and then, averaging over total aircraft for landing and take-off, respectively. The aircraft delay is calculated as the difference between the aircraft's actual runway operation time and the estimated runway operation time. Average landing and take-off sequence changes are calculated based on the number of position shifts compared to FCFS sequence for both runway operations.

6.5.2 Dataset

The set of instances most often used in the literature for aircraft landing problems (Airland 1-13) are certainly those in the OR-Library (Beasley, 1990). However, these instances do not reflect the actual real-life problem, and also, they are trivial for high-performance computers such that they can be easily handled in a reasonable time with state-of-the-art

MIP solvers. The second most commonly used benchmark problem instances are proposed by Ghoniem et al. (2015) for multiple runway aircraft scheduling problems. In these benchmark instances, each aircraft is characterized by its ready time, target time, due time, operation type (arrival or departure), weight class ("Heavy", "Large", and "Small"), priority (tardiness weight), and separation times with other aircraft. Every aircraft was set to a time window of 600 seconds. These instances are composed of $M = \{2, 3, 4, 5\}$ runways and $N = \{15, 20, 25\}$ aircraft. A set of 55 different instances is proposed, at the size ($N \times M$) and they are denoted using the pair ($n, m$), where $n$ is the number of aircraft and $m$ is the number of runways.

However, the benchmark problem instances of Ghoniem et al. (2015) are also not suitable for our experimental study for several reasons. First of all, these instances are structured in such a way that all data assumed to be deterministic. Second, aircraft weight classes were randomly generated without taking into account the fleet mix ratio in an airport, also without considering the weight classes "B757" and "Super" that are included in the official regulations issued by the FAA. Lastly, aircraft target times were calculated by adding 20 seconds to ready times which is an invalid assumption in practice. Consequently, both benchmark instances, i.e. OR Library (Beasley, 1990) and Ghoniem et al. (2015), do not represent the practical situation in a way that it can be used for our validation (proof-of-concept) study. Also, it is noteworthy to mention that although *www.SimOpt.org* website provides a testbed of simulation optimization problems and contains a variety of test problems for simulation optimization methods, aircraft or runway scheduling or any similar problem has not been included in the problem library yet.

For experimental study, actual operations dataset was utilized, which is obtained from a case study airport, namely Washington Dulles International (IAD) airport. The dataset is primarily used for determining the operational benefits that would be achieved by utilizing the proposed approach for practical runway operations scheduling. The dataset is collected through the following data sources:

(a)        *Aviation System Performance Metrics (ASPM) and Airline Service Quality Performance (ASQP) databases*: These core databases are part of FAA Operations & Performance database, which provides publicly available historical data, and they are available online at *https://aspm.faa.gov*. These databases provide flight-specific OOOI (Out of the gate, Off the ground, On the ground and Into the Gate) times and airport throughput in 15-minute interval, as reported by the airlines. The "Off the ground" times can be used to calculate the airport throughput in the same 15-minute interval. These databases also provide airport efficiency, runway configurations, and airport-level aggregate data, which enumerates the total number of arrivals and departures in 15-minute interval. Such data is commonly used to develop queuing models of airport operations or empirically estimate airport capacity envelopes. However, the level of detail is typically insufficient to investigate other factors that affect runway operations, such as interactions between landing/take-off aircraft, runway occupancy times, etc.

(b)        *The Operations Network (OPSNET)*: This database is the official source of historical air traffic activity provided as part of FAA Operations & Performance database (also available online at *https://aspm.faa.gov*). Monthly and annual counts of aircraft operations are available at the facility, state, regional, and national levels. Also, the number of runway operations (take-offs and landings) at major airports can be obtained from this database.

(c)        *Official airline guide (OAG)*: This database provides information only for scheduled flights, and also, fleet mix information is included, which is required for estimating the total runway activity by specific aircraft type, or aircraft grouping. This database is available online at *www.oag.com*, but most of the data is not publicly available.

(d)        *Bureau of Transportation Statistics (BTS) database:* This database provides T-100 Air Carrier Statistics, which is a monthly commercial aviation traffic data reported by airlines, and it includes not only scheduled passenger flights but also cargo and nonscheduled flights. This database is available online at *www.rita.dot.gov*.

(e)      *Flightstats database*: This database provides flight performance data based on operational data obtained from airline operational data feeds and aggregated data from its historical databases. This database is available online at *www.flightstats.com*.

(f)      Data obtained as a result of simulation runs with a validated and FAA approved simulation tool, namely the MITRE Corporation runwaySimulator.

We only considered scheduled flights in the dataset from IAD. Nonscheduled flights (general aviation and military) and other flights using IAD airport's TMA without landing are not considered due to the fact that all the required input data for the simulation model are available only for scheduled flights. However, according to ASPM data, these nonscheduled and other flights account for only a small fraction. Therefore, it is reasonable to assume that the extracted scheduled flight information reflects all the arrival and departure operations at IAD.

The scheduled arrival and departure flights data are mainly obtained from OAG database. Although OAG database includes the scheduled flights data, it does not include actual arrival and departure flight data and delay statistics. The limitations of OAG database is mitigated by obtaining actual arrival and departure flight data and delay statistics regarding individual flights from FAA Operations & Performance and BTS databases.

As explained in Chapter 4, ASPM database provides the OOOI data in where this data is estimated for flights of non-OOOI airlines and for OOOI airlines where OOOI data are not available. Since ASPM database estimation may not be accurate, non-OOOI data is detected and corrected by using the Flightstats database when it is available. In addition, Flightstats database is used for finding the missing data gathered from various databases, and ensuring accuracy. Additional information, such as information regarding merge fixes, initial approach fixes, stabilized approach fixes, final approach fixes and departure fixes, is obtained through the airport's website (*www.flydulles.com*) and through a commercial website *www.airnav.com*, which includes navigation and TMA related information.

After the analysis of the available data for IAD airport on its major runways, only peak demand time periods are considered. Total number of runway operation per hour in 2015 for IAD is given in Figure 45. The data between 7 am and 11 pm local time at the airport is taken into consideration to avoid periods of low activity since runway throughput is usually lower during such periods. Also, air traffic controllers' workload is usually high in such periods, and controllers can benefit the most from decision support tools in such periods. A time horizon of 20 minutes is considered for aircraft schedules mainly for two practical reasons: (1) for arriving aircraft, the scheduled landing time is assigned about 20-30 minutes in advance of landing, and (2) for departing aircraft, take-off is scheduled approximately 20 minutes before target take-off time.

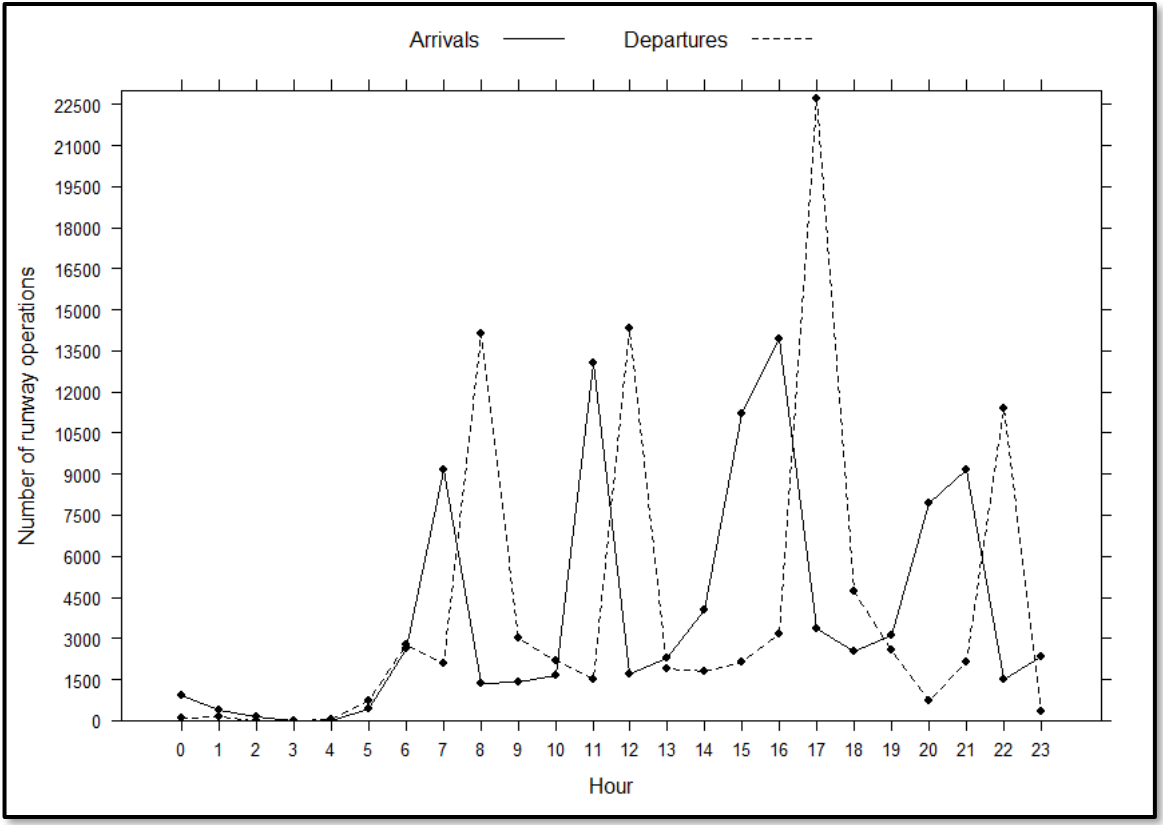

**Figure 45:** IAD Total Number of Runway Operation per Hour in 2015

IAD is operating under arrival or departure priority configuration with north or south flow operations both in visual flight rules (VFR) and instrument flight rules (IFR) conditions. Due to the fact that air traffic controllers enforce minimum separation requirements in only IFR conditions, we focused on only the periods in which IFR conditions are applied. Also, for the air traffic direction of the flow, based on the most recent (2014) FAA airport capacity profile, hourly runway operations rate under IFR conditions and arrival priority configuration for south flow (109) is greater than north flow (108) (FAA, 2015a). Therefore, data that include south flow operations are selected for experimental study. The data included in the problem instances, which are extracted from the dataset, are listed below:

(a)    Meter fix assignments and estimated time of landing (ETL) for arriving aircraft.

(b)    Estimated time of take-off (ETT) for departing aircraft.

(c)    Operation type and weight class of both arriving and departing aircraft.

(d)    Actual arrival times to entry points and holding area for arriving and departing aircraft, respectively.

Maximum delay time for both arriving and departing aircraft considered as 600 seconds, which is a hard constraint. Minimum separation times between aircraft weight classes are taken as calculated in Chapter 4.

### 6.5.3 Case Study Airport

As previously mentioned, historical data that belong to Washington Dulles International (IAD) airport is utilized in the experimental study. Historically, IAD has been ranked among the top 30 busiest airports in the US (FAA, 2015b). IAD has more than 150 runway operations per hour in VFR conditions and more than 100 runway operations per hour in IFR conditions. IAD has been suffering from a high level of delay that results from the scheduled demand exceeding the available capacity, which makes it an ideal case for experimental study.

IAD handles both domestic and international flights, and the traffic volume is relatively unstable throughout the day. IAD operates in either arrival or departure priority mode, as opposed to a single balanced operation between arrivals and departures to maximize capacity. Table 13 shows the capacity rates for arrivals and departures operations at IAD, presented as a range depending on the priority configuration mode (Jennifer Gentry et al., 2014).

**Table 13:** Runway Operations per hour in IAD

| Configuration | Weather Conditions | | |
|---|---|---|---|
| | **Visual** | **Marginal** | **Instrument** |
| Arrival Priority | 150-159 | 112-120 | 108-111 |
| Departure Priority | 156-164 | 136-145 | 125-132 |

Table 14 presents the annual fleet mix percentage for 2014 by weight class for IAD airport, where fleet mix does not change with the weather since IAD do not have substantial number of VFR operations (Jennifer Gentry et al., 2014).

**Table 14:** Annual Fleet Mix Percentage for 2014 by Aircraft Wake Class in Washington Dulles International Airport

| **Heavy** | **B757** | **Large** | **Small** |
|---|---|---|---|
| 10.1 | 3.8 | 74.3 | 11.8 |

IAD management has the authority and responsibility for controlling the IFR arrival, departure, and en-route aircraft within the IAD airspace. IAD operates in a north/south flow, and there exist four runways: (1) runway 01 Right (01R) – 19 Left (19L), (2) runway 01 Center (01C) – 19 Center (19C), (3) runway 01 Left (01L) – 19 Right (19R), and (4) runway 30 – 12. IAD's runway layout is illustrated in Figure 46.

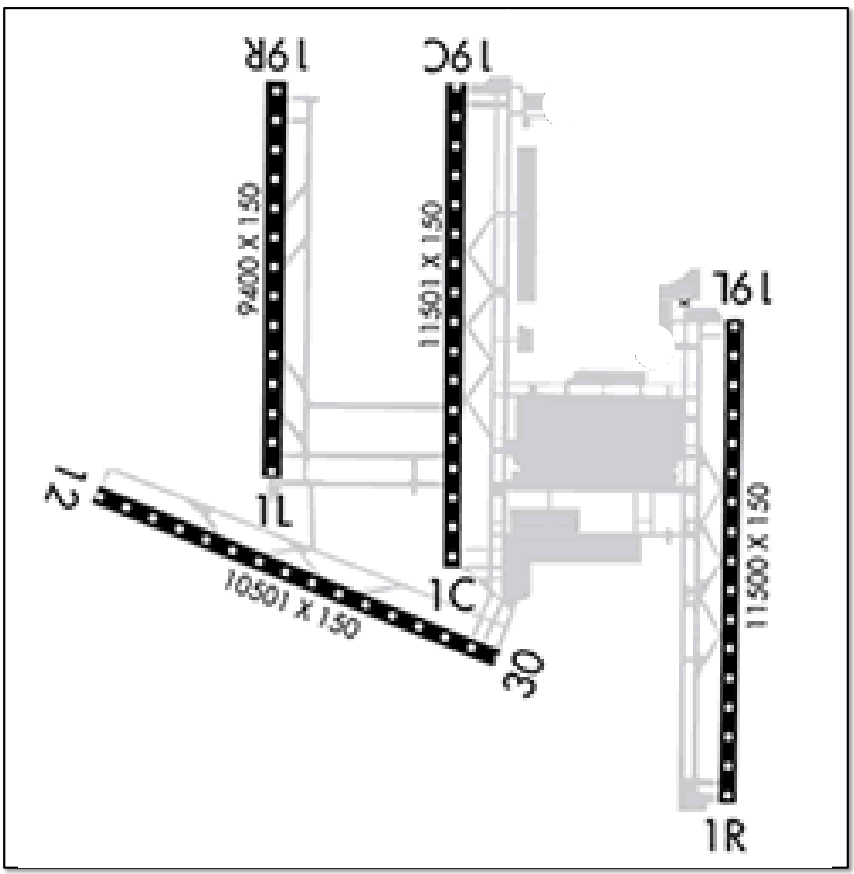

**Figure 46:** IAD Runway Layout

The runway configuration in IAD is often determined based on wind direction and weather conditions. Hence, depending on the following general rules, the most advantageous runway configuration is commonly selected to facilitate the air traffic. Active runways depending on the wind conditions in IAD are shown in Table 15, and detailed below:

(a)     Runway 12 can only be used for arrivals and can only be used when landing on the 19's.

(b)     Runway 30 can only be used for departures.

(c)     If using the 1's and winds are between 210 clockwise to 030, runway 30 will be active for departures only.

(d)     If using the 19's and winds are between 030 clockwise to 210, runway 12 will be active for arrivals only.

(e)     If using the 19's and winds are between 030 counterclockwise to 210, runway 30 will be active for departures only.

**Table 15:** Active Runways Depending on the Wind Conditions in IAD

| Wind | Runways |
|------|---------|
| 100 clockwise to 280 | 19R, 19L and 19C |
| 280 clockwise to 100 | 1R, 1L and 1C |
| Calm (up to 5 knots) | 19R, 19L, 19C and 12 |

As can be seen from the IAD airport layout in Figure 46, the 19R and 1L runways have four exits, two of which are before the halfway of the runway and two others are after that; 19L, 19C, 1R, and 1C runways have three exits, one in the middle, one before the halfway of the runway and the other is after that; 12 and 30 runways have only two exits, one of which is before the halfway of the runway and the other is after that. Average runway occupancy times are calculated by analysis of historical data for the IAD airport and shown in Table 16.

**Table 16:** Average Runway Occupancy Times in seconds

| Operation / Aircraft Type | Heavy | B757 | Large | Small |
|---|---|---|---|---|
| Arrival | 40 | 40 | 35 | 30 |
| Departure | 50 | 45 | 40 | 30 |

Average transit ground speeds between nodes in knots are shown in Table 17. Since this data is not available in any of the existing databases, ground speeds between each node in arrival and departure network are determined by experimenting the TMA of IAD airport with the MITRE Corporation runwaySimulator.

**Table 17:** Average Transit Ground Speeds between Nodes in knots

| Operation | Segment | Heavy | B757 | Large | Small |
|---|---|---|---|---|---|
| Arrivals | Entry point - Meter fix | 185 | 185 | 190 | 191 |
| | Meter fix - IAF | 185 | 185 | 190 | 191 |
| | IAF - FAF | 170 | 170 | 174 | 170 |
| | FAF - SAF | 165 | 163 | 165 | 160 |
| | SAF - Runway | 135 | 133 | 134 | 120 |
| | Runway - Runway Exit | 35 | 35 | 34 | 25 |
| Departures | Take-off - Initial Climb | 173 | 166 | 154 | 126 |
| | Initial Climb - En-route Climb | 184 | 177 | 175 | 141 |
| | En-route Climb - Departure fix | 251 | 243 | 231 | 182 |

### 6.6 Simulation-based Optimization Experiments

The simulation-based optimization (SbO) approach proposed in this dissertation is investigated in more details with three scenarios at Washington Dulles International (IAD) airport in order to conduct a proof-of-concept (validation) of the approach by analyzing the benefit of the approach over FCFS and deterministic approaches. For each scenario, these three optimization approaches (FCFS, deterministic, and SbO) are compared each other with regard to previously mentioned performance measures.

For the FCFS approach, sequencing and scheduling is calculated by using a first-come, first-served order such that aircraft land or take-off in the same order they arrive in the entry points or the holding area for landing and take-off, respectively. Since FCFS sequence is considered as the fairest runway operations schedule, average sequence changes are not considered as a performance metric. For the deterministic approach, deterministic version of the proposed approach is employed, where performance measures are calculated by using the objective function values instead of obtaining them from the simulation component. In this deterministic approach, an internal fitness function is used to account for the objective function, which is minimizing the total cost for each aircraft linearly dependent on deviation from target time, i.e. minimizing the total deviation from the target time (earliness and tardiness). Both in FCSF and deterministic approaches, the time windows, minimum separation requirements and other constraints are considered as the same in the SbO approach.

For the simulation-based optimization (SbO) approach, since it produces the best known Pareto set of solutions, i.e. a set of trade-off solutions, the best solution is determined by the following weights: runway utilization objective ($f_1$) 3/4, and fairness objective ($f_2$) 1/4. The main reason for giving runway utilization objective much higher priority than fairness objective is that from air traffic controllers' (decision-makers') perspective runway utilization is typically much more important than fairness in actual runway operations.

After finding solutions (runway operations schedules) from both FCSF and deterministic approaches, these solutions are evaluated using our discrete-event simulation model in order to determine previously explained performance metrics associated with each solution. As illustrated in Figure 47, solutions obtained from both optimization approaches are subject to solution evaluation using the simulation model to evaluate their actual performance in real-life like environment, where there are several sources of uncertainties. Solution evaluation using the simulation model is done by inputting the schedule to the model and running it 50 times. The results are recorded by averaging the performance metric obtained over the simulation runs. It is worth to mention that the all previously mentioned performance metric are obtained from the simulation model. However, the additional metric for computational time is encompass only the optimization phase, the time for solution evaluation with the simulation model is not included in this metric.

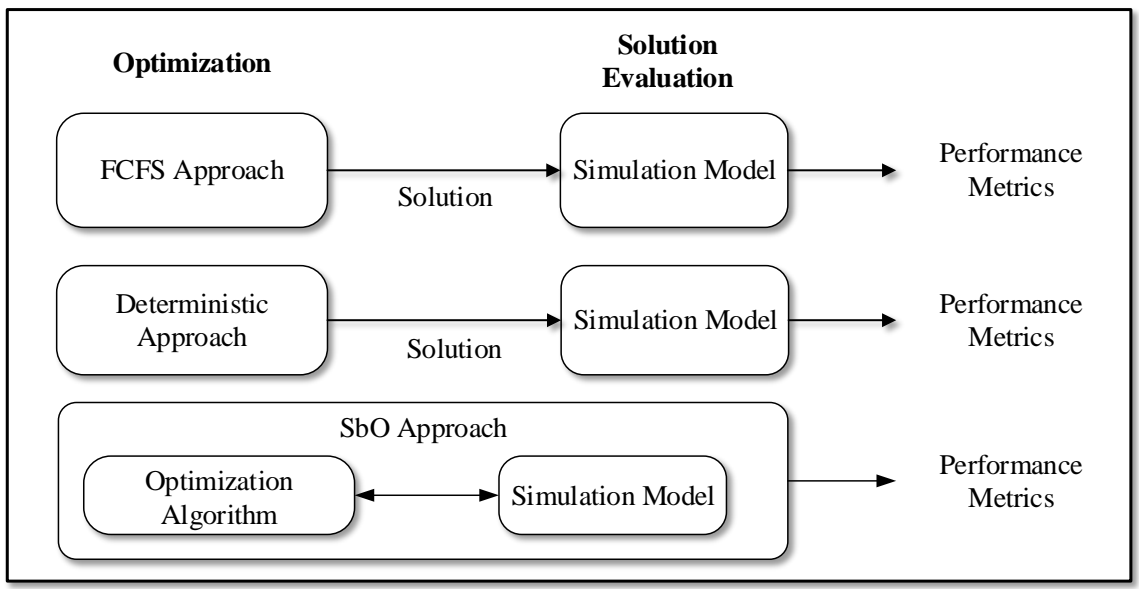

**Figure 47:** Optimization Approaches and Solution Evaluation

In order to account for the real-life operational conditions, the simulation period in the experiments consisted of half an hour before and after the one-hour solution period. Within the scenarios the runway operations scheduling problem is dealt with periodically updating the previous schedule based on a rolling horizon approach, where a large one-hour instance is divided into several sub-problems (time windows) that are solved independently. After all sub-problems are solved, the obtained solutions that pertain to different time windows are combined together to find the one-hour solution.

The three scenarios that are created for the experimental study are based on airport operation priority configuration, runway configuration and hourly air traffic demand rate, as shown in Table 18. The data required for the Scenario 1 and 2 are extracted from the dataset according to scenario configuration in order to reflect the current operation environment of IAD airport in arrival and departure priority configuration, respectively. On the other hand, the data required for the Scenario 3 is obtained from the MITRE Corporation runwaySimulator tool by simulating the future operating environment of IAD as given in the scenario data.

**Table 18:** Scenarios for the Experimental Study

| Scenario | | Arrival Runways | Departure Runways | Hourly Rate |
|---|---|---|---|---|
| Scenario 1 | Current Operations in Arrival Priority | 19C, 19L | 30 | 103 |
| Scenario 2 | Current Operations in Departure Priority | 19C, 19L | 19L, 30 | 108 |
| Scenario 3 | Future Operations | 19C, 19L, 19R | 19L, 30 | 168 |

### 6.6.1 Scenario 1 - Current Operations in Arrival Priority

This scenario reflects the current operation environment of IAD airport operating in arrival priority. IAD operates in the arrival priority configuration approximately 36 percent of the time in IFR conditions, which equates to less than 3 percent annually.

The extracted data for this scenario from the dataset has a fleet mix consists of 9.7 percent, 3.9 percent, 73 percent, and 14 percent of heavy, B757, large, and small aircraft, respectively. There are total of 103 flights in the data with 56 arrivals and 47 departures, which belongs to the busiest period of time between 16:00 and 17:00 in IAD. This hourly rate is a high traffic demand situation for IAD and close to its declared capacity in instrument conditions, namely 109 operations per hour, for south flow arrival priority configuration. Therefore, Scenario 1 is a realistic traffic demand for busy hours on a completely utilized runway system for IAD operating in arrival priority.

Table 19 summarizes the aggregated computational results for Scenario 1. We solved 3 problem instances each has 20-minute planning horizon with both FCFS, deterministic and simulation-based optimization (SbO) approaches and previously determined performance metrics are reported. We also report the average computational time, which is the CPU time consumed to find the solution.

**Table 19:** Aggregated Computational Results for Scenario 1

| Metric | FCFS | Deterministic | SbO |
|---|---|---|---|
| Average runway utilization (seconds) | 3471 | 3286 | 3189 |
| Average take-off delay (seconds) | 87 | 81 | 62 |
| Longest take-off delay (seconds) | 272 | 246 | 144 |
| Average landing delay (seconds) | 63 | 48 | 36 |
| Longest landing delay (seconds) | 241 | 217 | 127 |
| Average sequence change | 0 | 24.8 | 11.2 |
| Average computation time (seconds) | 12 | 357 | 1200 |

As shown in the computational results, in terms of runway utilization and average and longest runway operations delay, the SbO approach outperforms the FCFS and deterministic approaches. Since take-offs are cheaper to delay than landings, the average take-off delay is higher than the average landing delay. The number of average aircraft sequence change is higher in deterministic approach since this algorithm is not trying to minimize position shifts.

For Scenario 1, the average delay from three different runway operations scheduling approaches are illustrated and compared each other in Figure 48. The solutions obtained from 3 problem instances (each 20 minutes) aggregated into one graph. According to the graph, the delay savings from the SbO approach steadily increases as scheduling time progresses. These additional savings are obtained by explicitly considering the uncertainties. The box-and-whisker plots for runway utilization, sequence change and computation time are shown in Figure 49-51, and normalized Pareto-frontier for Scenario 1 is given in Figure 52.

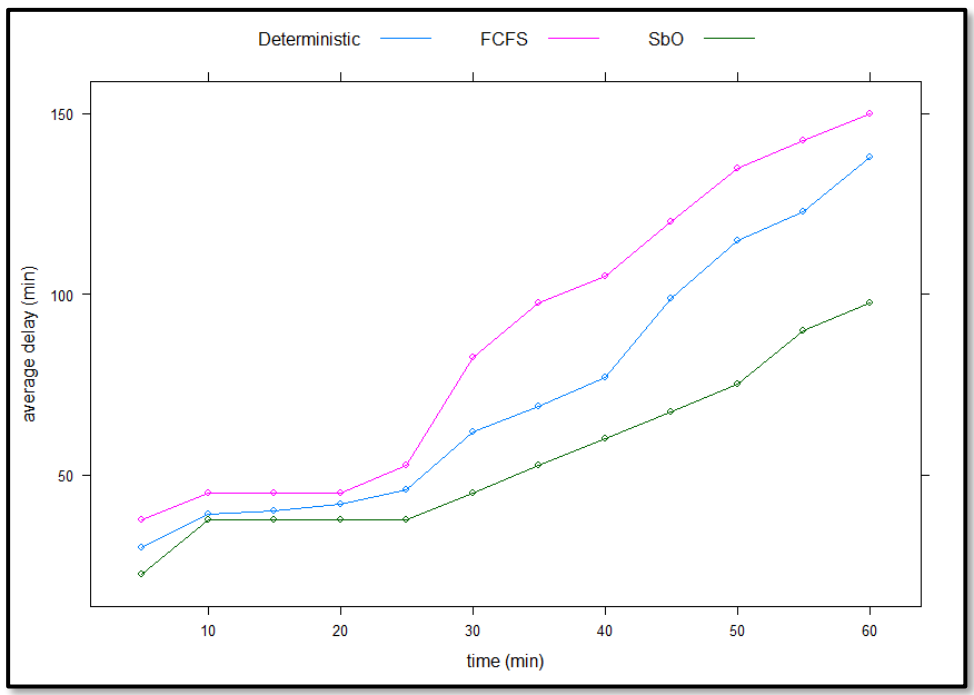

**Figure 48:** Average Delay Plot for Scenario 1

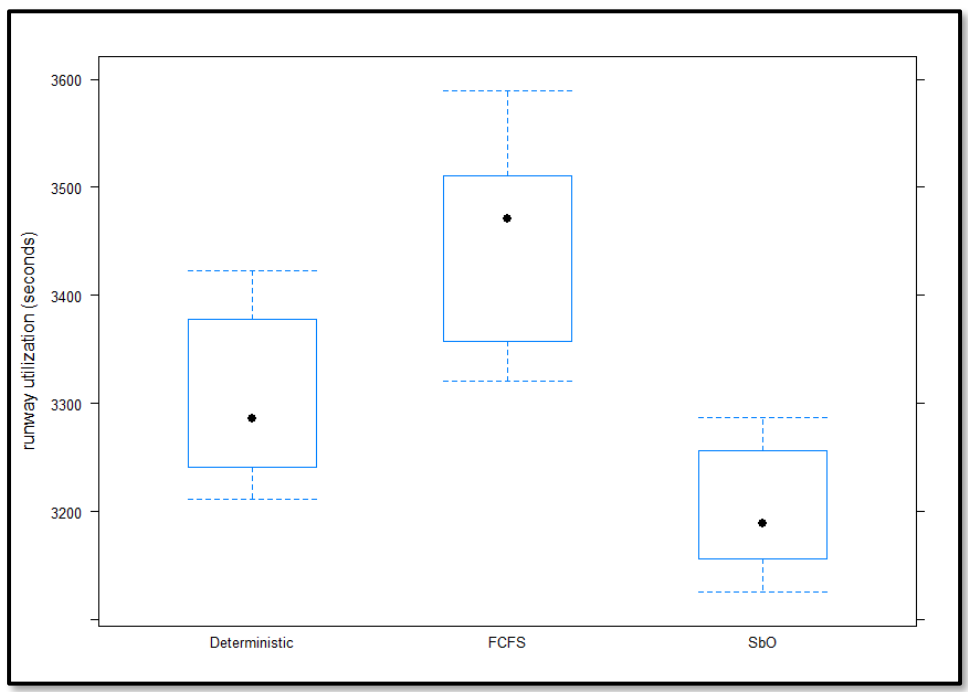

**Figure 49:** Box-and-Whisker Plot for Runway Utilization **-** Scenario 1

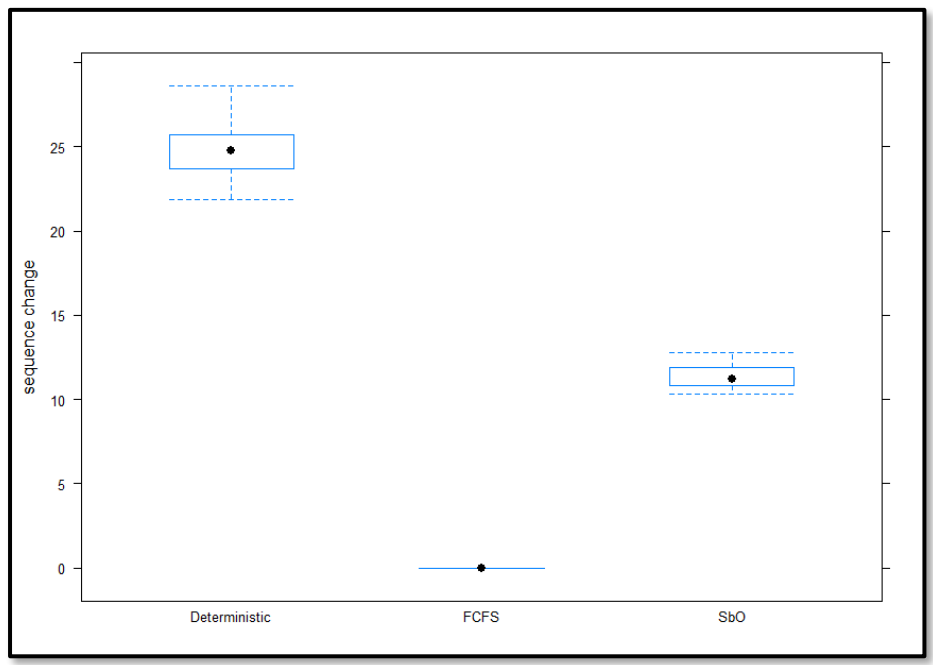

**Figure 50:** Box-and-Whisker Plot for Sequence Change **-** Scenario 1

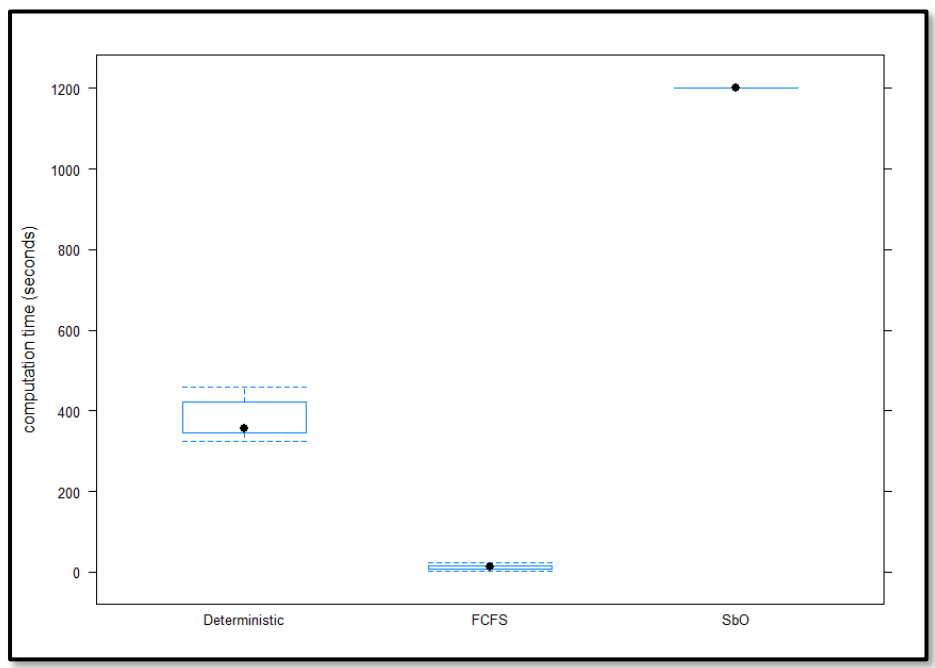

**Figure 51:** Box-and-Whisker Plot for Computation Time - Scenario 1

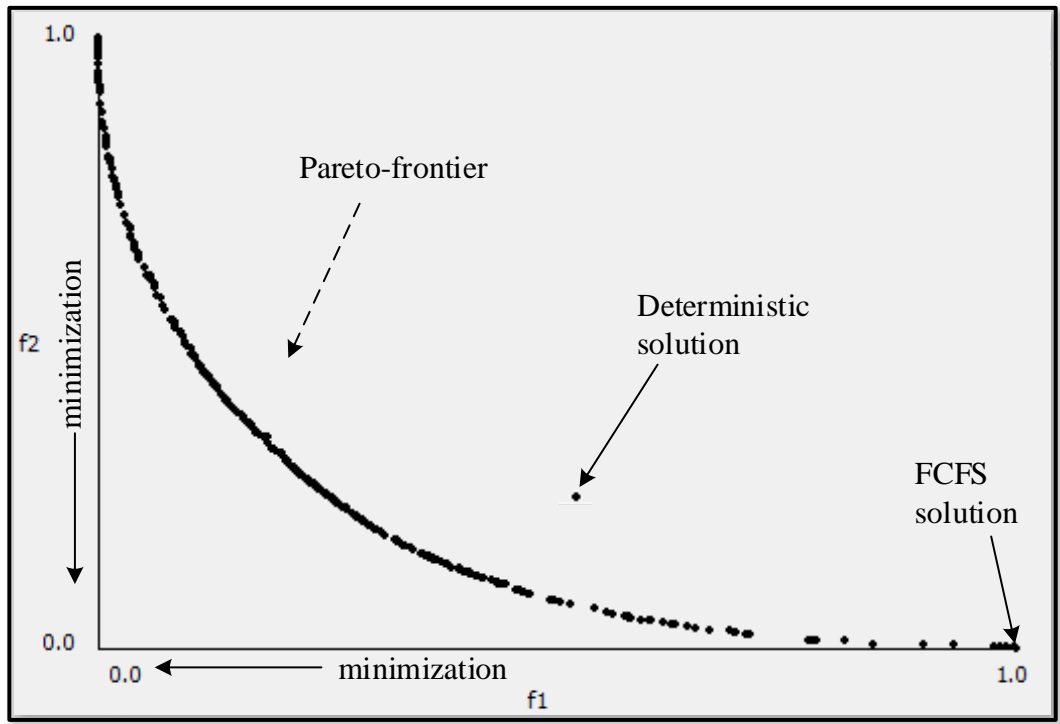

**Figure 52:** Normalized Pareto-frontier for Scenario 1

### 6.6.2 Scenario 2 - Current Operations in Departure Priority

This scenario reflects the current operation environment of IAD airport operating in departure priority. IAD operates in the departure priority configuration approximately 10 percent of the time in IFR conditions, which equates to less than 1 percent annually.

The extracted data for this scenario from the dataset has a fleet mix consists of 10.1 percent, 5.6 percent, 73.2 percent, and 11.1 percent of heavy, B757, large, and small aircraft, respectively. There were total of 108 flights in the data with 46 arrivals and 62 departures, which belongs to the busiest period of time between 17:00 and 18:00 in IAD. This hourly rate is a high traffic demand situation for IAD and close to its declared capacity in instrument conditions, namely 110 operations per hour, for south flow arrival priority configuration. Therefore, Scenario 2 is a realistic traffic demand for busy hours on a completely utilized runway system for IAD operating in departure priority.

Table 20 summarizes the aggregated computational results for Scenario 2. We solved 3 problem instances each has 20 minutes planning horizon with both FCFS, deterministic and simulation-based optimization (SbO) approaches and previously determined performance metrics are reported. We also report the average computational time, which is the CPU time consumed to find the solution.

**Table 20:** Aggregated Computational Results for Scenario 2

| Metric | FCFS | Deterministic | SbO |
|---|---|---|---|
| Average runway utilization (seconds) | 3473 | 3321 | 3273 |
| Average take-off delay (seconds) | 97 | 82 | 67 |
| Longest take-off delay (seconds) | 295 | 211 | 178 |
| Average landing delay (seconds) | 89 | 76 | 60 |
| Longest landing delay (seconds) | 277 | 246 | 182 |
| Average sequence change | 0 | 15.8 | 9.7 |
| Average computation time (seconds) | 17 | 731 | 1200 |

As shown in the computational results, in terms of runway utilization and average and longest runway operations delay, the SbO approach dominates the FCFS and deterministic approaches. Compared to Scenario 1, where runway system is operating in arrival priority configuration, average runway utilization and average runway operation delays increased. This is mainly due to complex structure of departure operations as well as the existence of more uncertainty elements in departures.

For Scenario 2, the average delay from three different runway operations scheduling approaches are illustrated and compared each other in Figure 53. The solutions obtained from 3 problem instances (each 20 minutes) aggregated into one graph. The result presented in the graph validates that the SbO approach is able to generate schedules with significantly better average delay values than deterministic and FCFS approaches. The box-and-whisker plots for runway utilization, sequence change and computation time are shown in Figure 54-56, and normalized Pareto-frontier for Scenario 2 is given in Figure 57.

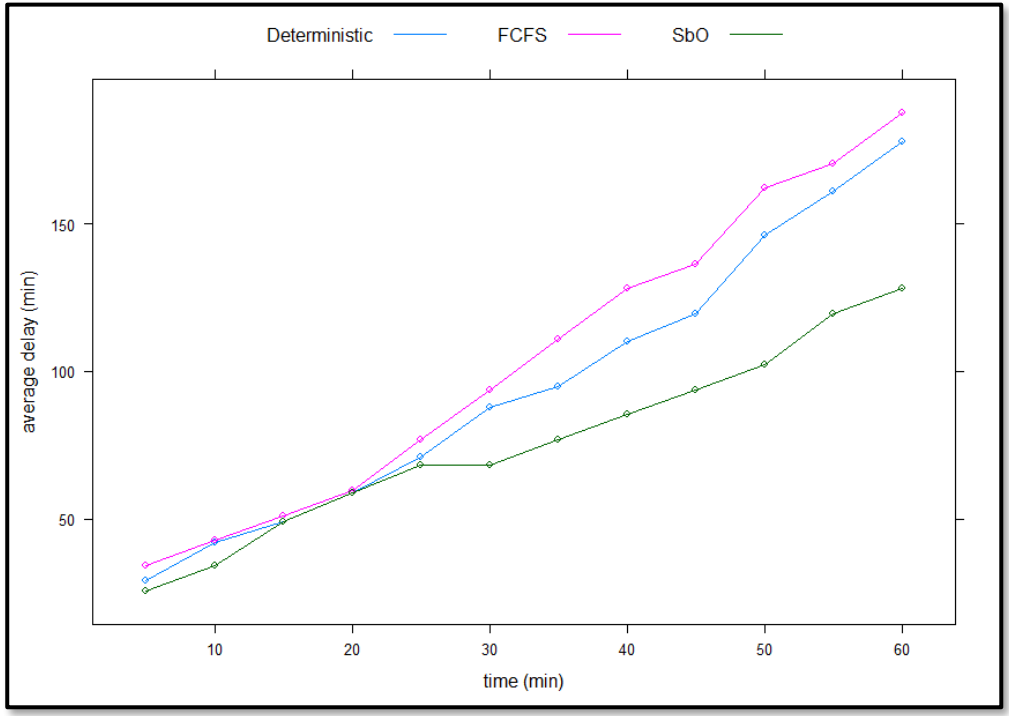

**Figure 53:** Average Delay for Scenario 2

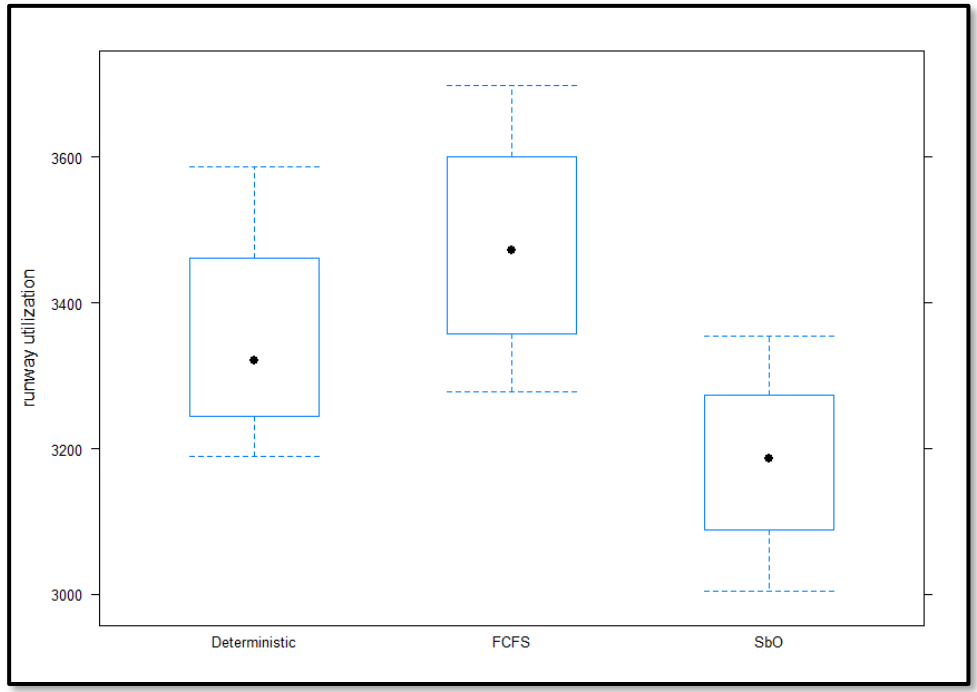

**Figure 54:** Box-and-Whisker Plot for Runway Utilization **-** Scenario 2

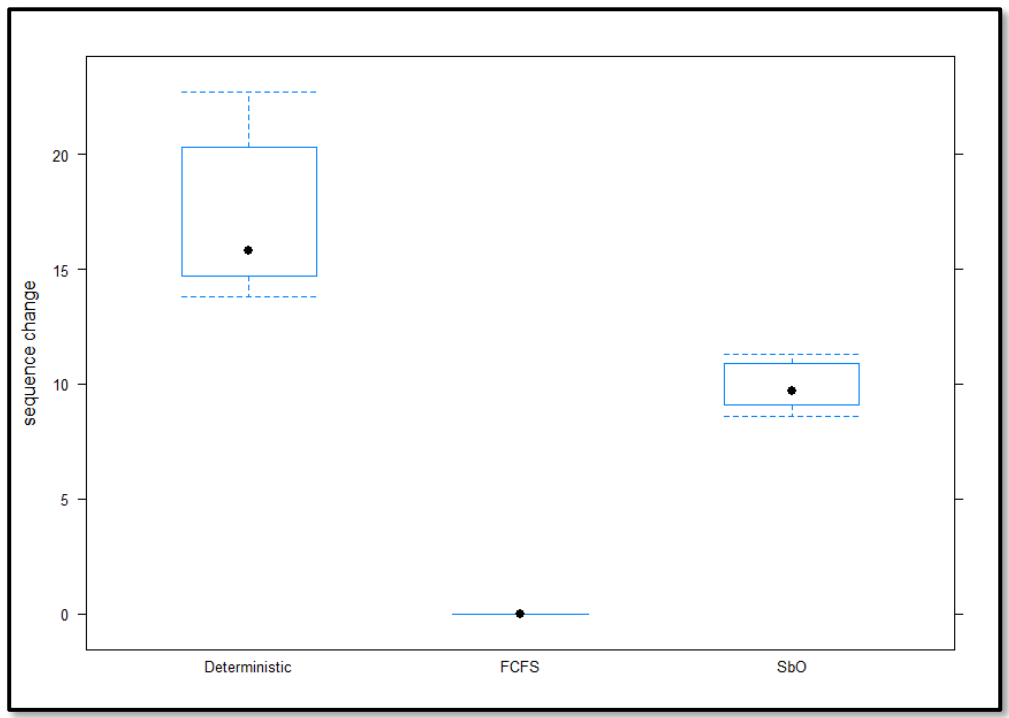

**Figure 55:** Box-and-Whisker Plot for Sequence Change **-** Scenario 2

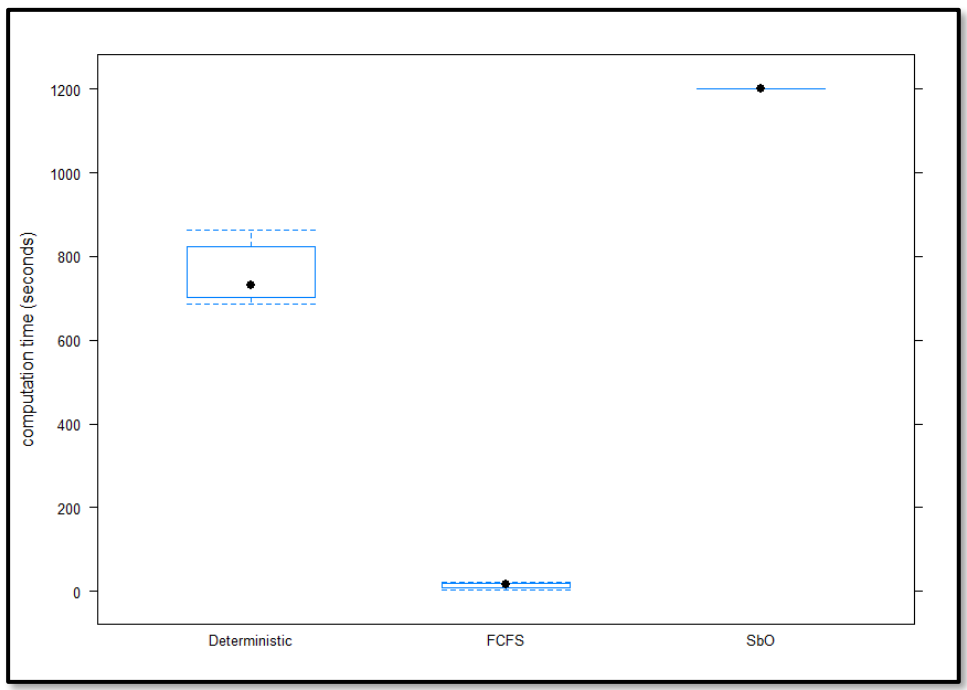

**Figure 56:** Box-and-Whisker Plot for Computation Time - Scenario 2

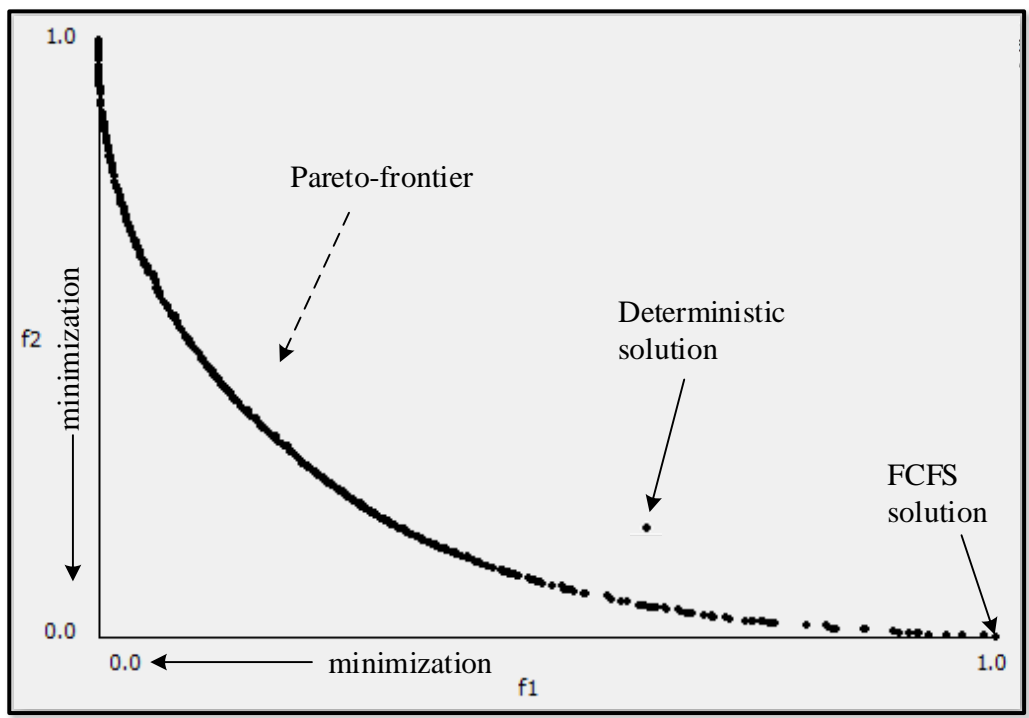

**Figure 57:** Normalized Pareto-frontier for Scenario 2

6.6.3 Scenario 3 - Future Operations

This scenario attempts to reflect the future operation environment of IAD airport. The data for the Scenario 3 is obtained from a validated and FAA approved simulation tool (the MITRE Corporation runwaySimulator ).

The data generated from the simulation tool for this scenario has a fleet mix consists of 10.1 percent, 4.2 percent, 75 percent, and 10.7 percent of heavy, B757, large, and small aircraft, respectively. There were total of 168 flights in the data with 88 arrivals and 80 departures. This hourly rate is a high traffic demand situation for IAD and close to twice its declared current capacity in instrument conditions for south flow arrival priority configuration. Therefore, Scenario 3 is a realistic traffic demand for busy hours on a completely utilized runway system in the future for IAD.

Table 21 summarizes the aggregated computational results for Scenario 3. We solved 3 problem instances each has 20 minutes planning horizon with both FCFS, deterministic and simulation-based optimization (SbO) approaches and previously determined performance metrics are reported. We also report the average computational time, which is the CPU time consumed to find the solution.

**Table 21:** Aggregated Computational Results for Scenario 3

| Metric | FCFS | Deterministic | SbO |
|---|---|---|---|
| Average runway utilization (seconds) | 3589 | 3473 | 3385 |
| Average take-off delay (seconds) | 125 | 97 | 82 |
| Longest take-off delay (seconds) | 373 | 321 | 235 |
| Average landing delay (seconds) | 98 | 88 | 70 |
| Longest landing delay (seconds) | 312 | 227 | 180 |
| Average sequence change | 0 | 28.1 | 17.8 |
| Average computation time (seconds) | 32 | 843 | 1200 |

As shown in the computational results, in terms of runway utilization and average and longest runway operations delay, the SbO approach performs better than the FCFS and deterministic approaches. It is worth to mention that in future operations there exist one more runway dedicated to arrivals compare to current operations. However, this additional runway is not able to handle additional air traffic demand completely. As a result, both runway utilization and air traffic delays are increased.

For Scenario 3, the average delay from three different runway operations scheduling approaches are illustrated and compared each other in Figure 58. The solutions obtained from 3 problem instances (each 20 minutes) aggregated into one graph. According to the graph, SbO approach is able to generate schedules with significantly better average delay values than deterministic and FCFS approaches. The box-and-whisker plots for runway utilization, sequence change and computation time are shown in Figure 59-61, and normalized Pareto-frontier for Scenario 3 is given in Figure 62.

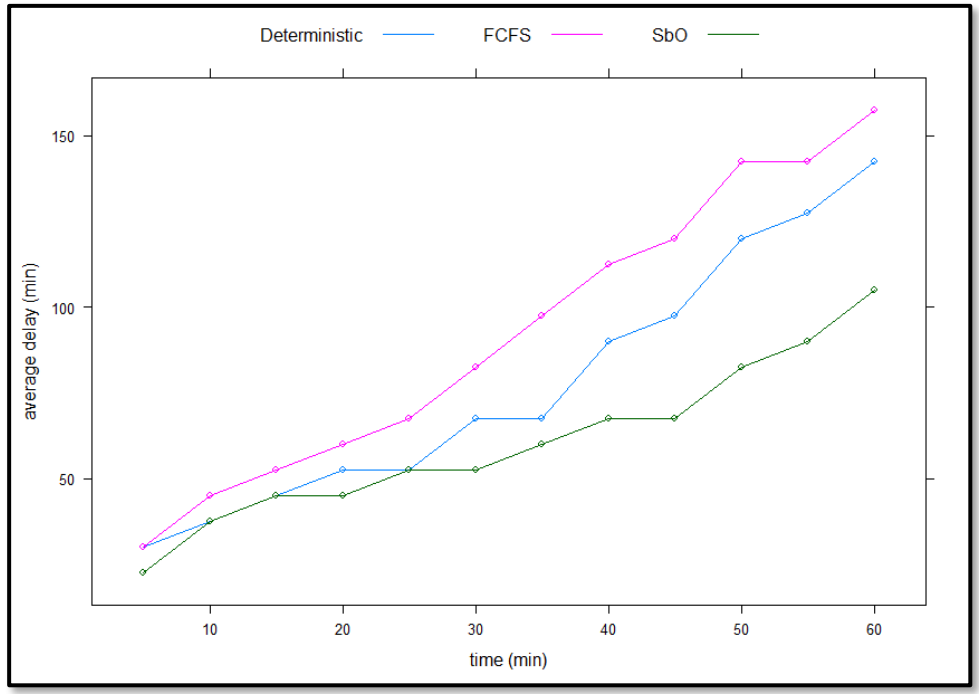

**Figure 58:** Average Delay for Scenario 3

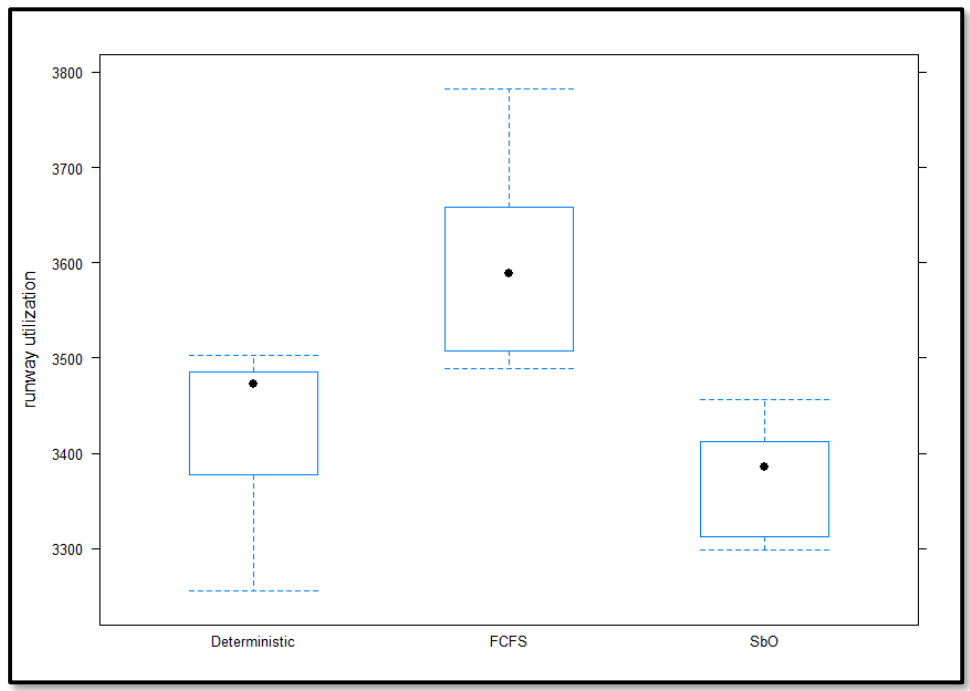

**Figure 59:** Box-and-Whisker Plot for Runway Utilization - Scenario 3

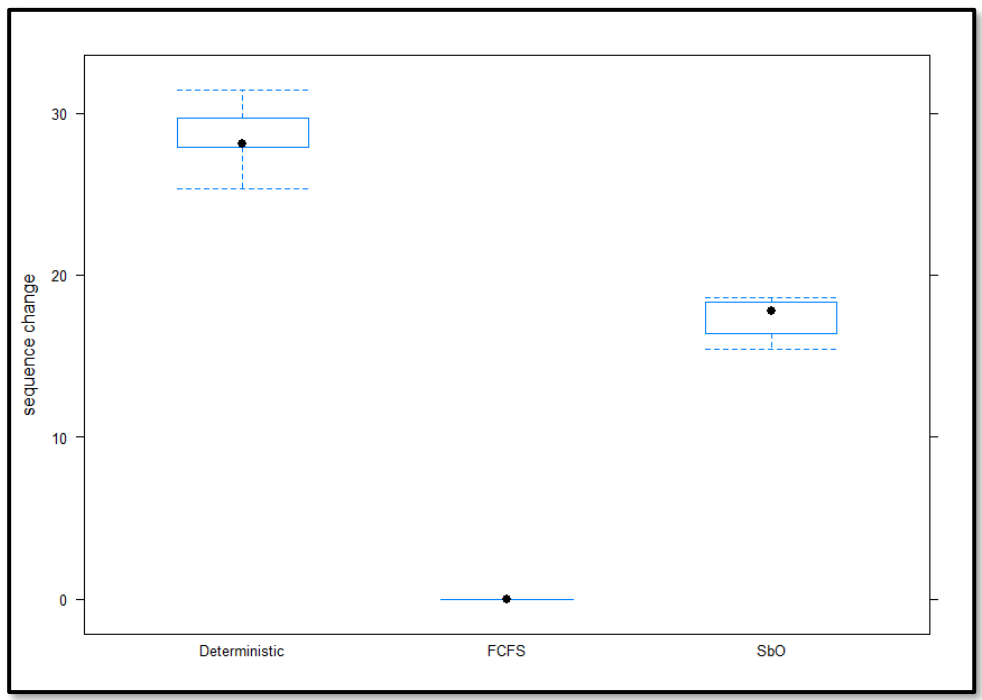

**Figure 60:** Box-and-Whisker Plot for Sequence Change **-** Scenario 3

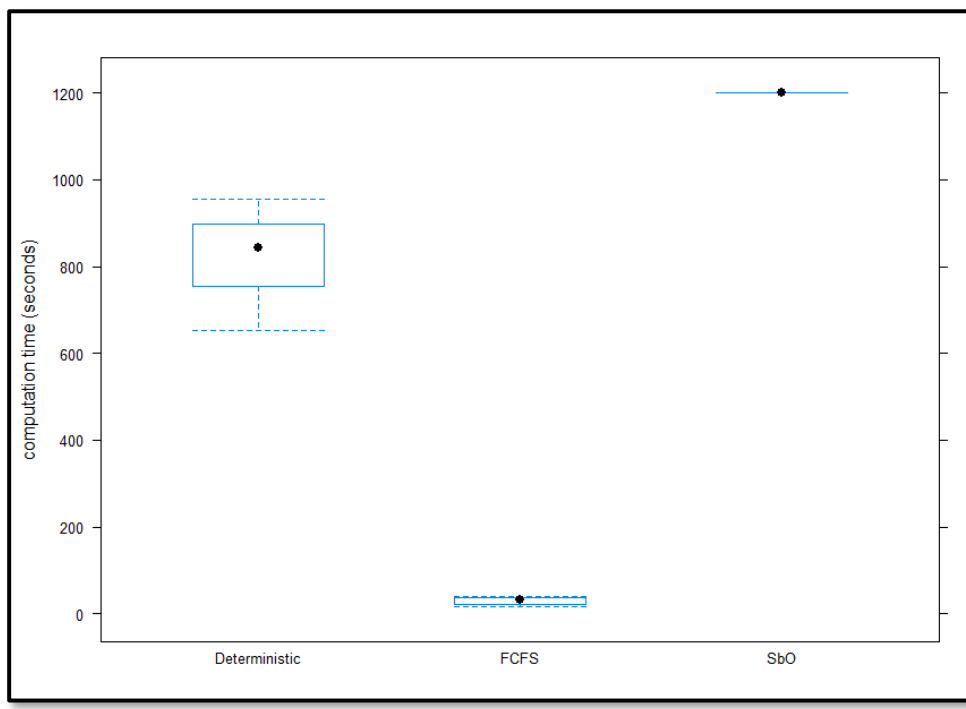

**Figure 61:** Box-and-Whisker Plot for Computation Time - Scenario 3

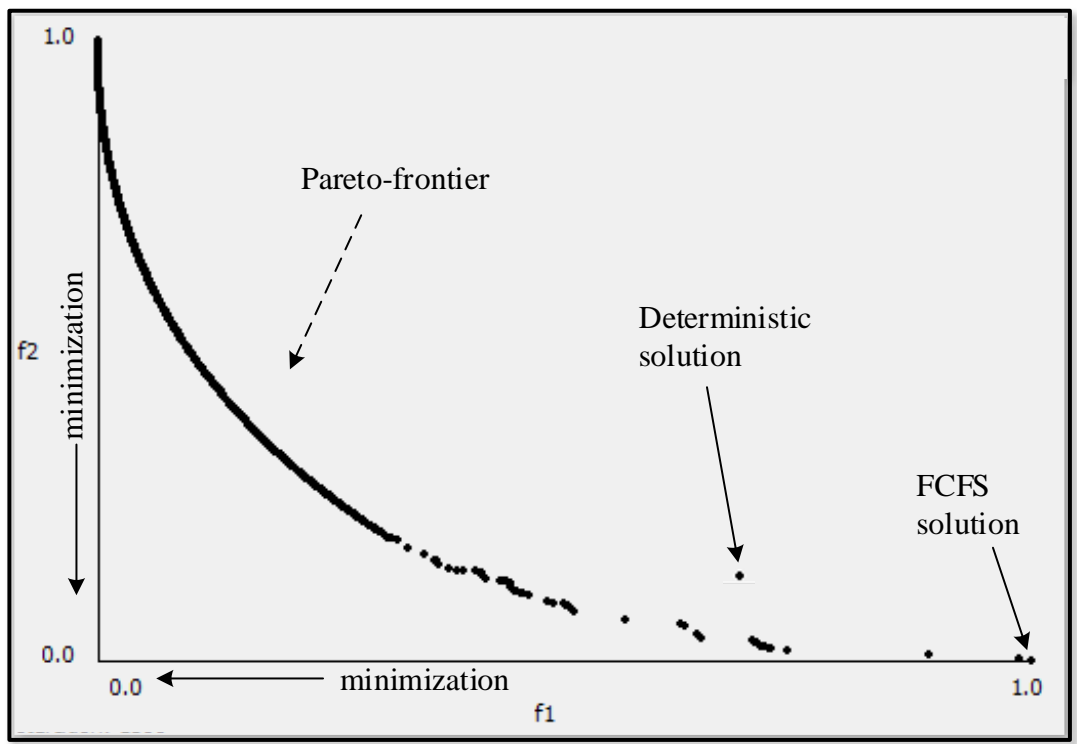

**Figure 62:** Normalized Pareto-frontier for Scenario 3

### 6.6.4 Sensitivity Analysis

Since the main purpose of this research study is to develop an optimization algorithm and propose a SbO approach rather than to conduct an actual analysis of a particular airport, sensitivity analysis is not done in an attempt to perform an extensive sensitivity analysis of all parameters. However, a sensitivity analysis is conducted in order to identify necessary features and to derive general conclusions regarding the relative sensitivity of the results to different planning horizons.

We analyze the effect of planning horizon on the runway operations' performance by comparing all scenarios. In order to evaluate the impact of the planning horizon and, in turn, the effect of computational times on runway operations, we considered five different planning horizons of 20, 25, 30, 35 and 40 minutes. Runway utilization (makespan) under different planning horizons is shown in Figure 63. As expected, when the planning horizon increased, the quality of the solution improved. SbO approach produced schedules with increased runway utilization compared to FCFS approach. Also, regarding fairness among aircraft, the maximum position shift was calculated as 3. This indicates that better objective functions can be achieved if the allocated solution time increases.

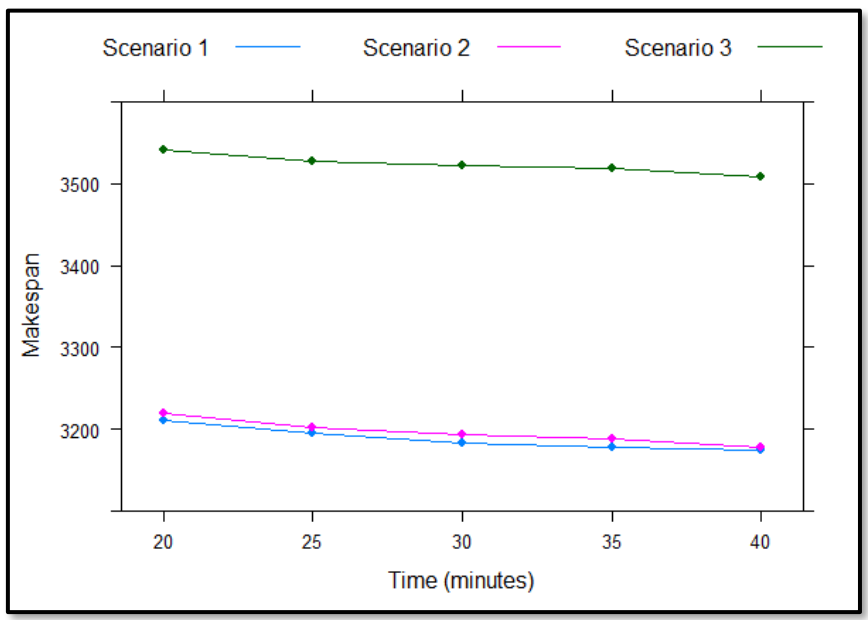

**Figure 63:** Runway Utilization under Different Planning Horizons

## 6.7 Key Findings and Analysis

The above computational simulation-based optimization experiments by using actual operations data from Washington Dulles International (IAD) airport led us to the following findings:

      (a)    Realistic scheduling of runway operations with considering the uncertainties can be of valuable use for air traffic controllers. It can enhance controllers'

tactical decision-making, and in turn, increase efficiency of the runway system and ensure safe flow of air traffic. The ad hoc decision-making based on FCFS, which is prevalent in many airports, and deterministic approaches that do not consider uncertainties can be improved with simulation-based approaches that provide optimized schedule recommendations to air traffic controllers.

(b)     The computational results show that the proposed SbO approach results in less average delay compared to both FCFS and deterministic approaches. This result provides evidence that these deterministic approaches for runway operations scheduling are essentially suboptimal in a stochastic environment. The runway utilization can be increased and runway operation delays can be reduced by using the proposed approach. However, the extent of benefits is strongly influenced by the dense of air traffic.

(c)     Since the primary computational bottleneck in the SbO framework is the simulation model, i.e. solution evaluation, utilization of problem-specific knowledge helped to keep the search effort confined within relatively reasonable limit of the solution space. As a result, approximate Pareto-optimal solutions were obtained within considerable low computation times. The computation time is an important metric since the scheduling algorithm should be able to support the air traffic controller's decision-making within a practical planning horizon, which is typically 20 minutes. This computational time make the proposed approach suitable for practical use.

(d)     Despite various models for fairness and CDM aspects of the runway operations scheduling was proposed in the academic research literature, no schedule optimization model has been deployed as part of decision support tools for air traffic controllers. The computational results indicate that trade-off solutions can be found with the price of fairness for reducing runway utilization within reasonable computational times.

(e)     Experimental results show that the operational benefits can be achieved by considering uncertainties during schedule generation. These results justify the utilization of simulation-based frameworks as part of operational planning models and decision support tools for air traffic (local) controllers, without increasing their workload.

(f)    Also, during the experiments it is observed that the computational times increased with increasing runway load. Since take-off delays are less costly compared to landing delays, the average take-off delay resulted higher than the average landing delay. Considering the fairness objective, it is also observed that the number of position shifts and the maximum position shifts to earlier or later positions are increased with increasing runway load.

(g)    Not all the airports are likely to benefit from these advanced runway scheduling optimization approaches compared to FCFS and deterministic approaches, but if the air traffic is dense, then there is potential to achieve benefits. However, considering the fact that airport infrastructures have been failing to keep pace with the growth in air traffic, and as a result, air traffic will be much denser in the future, developing and using simulation-based approaches that can produce much robust solutions seem to be a viable option for future operations.

(h)    According to the Bureau of Transportation Statistics (BTS) data between 2010 and 2015, more than 30 percent of air traffic delays are caused by the volume of air traffic. This motivated this research of effective runway operations scheduling that can reduce air traffic delays, which is mainly caused by air traffic congestion. The computational experiments provided evidence that explicit consideration of uncertainties allows air traffic controllers to produce more robust runway assignments, aircraft sequences and time schedules to maximize the runway throughput and fairness among aircraft. The approach presented in this dissertation depart from the traditional ways of controlling flows of air traffic, in particular ground delay program (GDP) and miles-in-trail (MIT) restrictions. GDP is a measure taken by the FAA in order to decrease the arrival rate to a level that can be safely handled by air traffic controllers by delaying flights on the ground before their departure, which run in lengths of 4-6 hours. In a similar manner, MIT restrictions enforce greater separations, speed limits and prevent aircraft from passing each other in the airways leading to a congested airport. These traditional approaches try to ensure that the air traffic demand is kept at a manageable level. On the other hand, the proposed scheduling approach try to increase the capacity of the runways by explicitly considering the uncertainties that may happen during runway operations and fairness among aircraft during optimization process.

(i)     Finally, the proposed SbO approach to runway operations schedule optimization is a proof-of-concept. In order to completely evaluate and operationalize the proposed SbO approach further improvements are required, and also, it should be validated by human-in-the-loop simulations before deployment.

## 6.8 Safety Risk Assessment

Since safety is the highest priority requirement in the airport industry, change in the existing procedures and process must be assessed in terms of safety risks according to the regulations enforced by the authorities. Therefore, an assessment of safety risk associated with our proposed SbO approach is conducted and results are presented below:

The FAA Air Traffic Organization (ATO) Safety Management System Manual is one of the main safety documents that provides the procedures and guidance to manage safety risk, and tries to establish a mature and integrated Safety Management System (SMS). Considering the SMS detailed in this document, the proposed approach poses a low-level of safety risk at the same time mitigates the safety risk to some extent.

There is a potential trade-off between capacity and safety. As a result of applying this approach, the number of aircraft operations per unit time will increase, and therefore, air traffic controller workload can potentially increase with a chance that the actual separation between aircraft to be violated compared to current practices. However, in computational experiments, the proposed design did not lead to changes in the air traffic control decision-making process and operational procedures. If any application of the proposed design results in changes in the air traffic control decision-making process and operational procedures, this application should be accompanied by a safety risk assessment documented in accordance with the policy outlined in the ATO SMS Manual.

In addition, we have identified a few other low-level safety risks associated with the proposed approach that can easily be mitigated. One such risk is a poor runway operations schedule generated by the proposed approach, which could cause long delays in landing and take-off air traffic at the airport. In visual meteorological conditions (VMC), the main risk transfer strategy is to change the responsibility of maintaining the minimum separation requirements from air traffic controller to pilot. In instrument meteorological conditions (IMC), the air traffic controller can mitigate the risk by analyzing the performance measures generated by the simulation model.

On the other hand, consideration of uncertainties in runway operation scheduling mitigates a level of safety risk, since any schedule that does not respect minimum separation requirements as a result of an unexpected event are considered as infeasible during the optimization process, which is commonly considered as high risk in practice due to simultaneous runway occupancy and collision risk. Therefore, the proposed approach has a potential to reduce safety risks to some extent, since it is capable of generating near-optimal schedules to reduce landing and take-off delays.

# CHAPTER 7

# CONCLUSIONS AND FUTURE RESEARCH

The focus of this dissertation has been on solving multi-objective runway operations scheduling problem with the integration of simulation with optimization to deal better with uncertainties inherent in runway operations, and developing an efficient and effective metaheuristic algorithm that can deal with two conflicting objectives. The main contribution of this dissertation research is in the field of developing a hybrid Tabu/Scatter Search algorithm and applying it to a real-life scheduling problem, where solution methodology is based on a Simulation-based Optimization (SbO) framework. The main reason for employing a SbO framework is to capture the relevant intricacies of the practical operating environment, particularly the stochastic nature of this real-life problem as well as fairness among aircraft waiting for runway operations.

This chapter presents the conclusions of this research including a brief summary of the dissertation and its main findings. Also, several indications and suggestions for future work are highlighted based on the research conducted in this dissertation. Lastly, the final conclusions that are drawn from this research effort are presented.

## 7.1 Summary

The steady increase in air traffic is expected to continue in the coming years; however, this growth has not been supported by the increase in the capacity of major airports. As a result, air traffic delays are becoming inevitable in major airports when the demand for services exceeds its capacity. Because runways are typically the primary bottleneck in air transportation system, the overall airport capacity is determined by the runways in use. In general, adding more runways to increase the capacity is often not feasible due to a number of reasons including physical limitations and environmental restrictions. The only practical option for enhancing the capacity of airports is to utilize the currently available runway capacity more efficiently. For that reason, it is significantly important to effectively utilize resources, especially runways, to achieve significant improvements in air traffic delay and, in turn, to smooth the flow of air traffic. To accomplish this, effective and efficient algorithms are required for scheduling runway operations as part of operational planning models and decision support systems used by air traffic controllers.

The runway operations scheduling problem is a decision-making process that deals with the allocation of runways to take-offs and landings over given time periods while considering operational constraints. The main challenges in solving this real-life problem are the pre-specified time windows, and the minimum separation requirements related to wake vortex which render the problem combinatorial in nature. In multiple runways case, these separation requirements are asymmetric and do not follow the triangle inequality. Therefore, appropriate modification of the aircraft sequence can greatly affect the throughput of the runway and the total flight delay. This practical problem is a computationally challenging problem faced by the local controllers, where decision support systems need to produce good quality solutions promptly and poorly utilizing runways might have severe economic and environmental consequences.

This practical scheduling problem has been researched extensively in the past several decades, and there are numerous algorithms available for the deterministic version of the problem. However, the real-life scheduling problem that includes uncertainties remains challenging from a modeling and computational tractability standpoint. It is hardly ever the case that an aircraft schedule is executed in isolation. There are many sources of uncertainty that need to be considered during scheduling, such as inclement weather, airport congestion, equipment failure, ground speed variations caused by the wind, piloting indecisions, unexpected delays in pushback or taxiing, etc. In such cases, the quasi-optimal schedules become far from optimal in practice because of challenges posed by uncertainty impacts. Also, the viewpoints of the various stakeholders who affect or be affected by the scheduling of aircraft over runways differ substantially, which needs to be considered.

The literature review on runway operations scheduling identified the knowledge gap in the literature on scalable methods that address the challenges of runway operations scheduling problem under uncertainty and consider different stakeholders' interests. Complex, dynamic and stochastic nature of the actual runway operations renders simulation-based optimization (SbO) as the only viable alternative approach for explicitly modeling uncertainty and considering multiple objectives with a computationally tractable manner. This dissertation proposes a SbO approach that tackles practical challenges of runway operations scheduling by incorporating simulation into optimization as an external fitness (objective function) evaluator. The primary advantage of this approach is in its robustness at incorporating complexity of the system to the required level of detail by application of simulation, and employing an optimization algorithm to find high-quality trade-off solutions promptly.

This research can be differentiated from the previous works mainly in following ways. First, to the best of our knowledge, this is the first SbO approach that explicitly considers uncertainties in the development of schedules for runway operations as well as considers fairness as a secondary objective. This SbO approach provides more realistic and robust solutions that can be applied to practical runway operations scheduling. Second, this approach takes advantage of the flexibility of simulation to model complexities of runway operations and integrates with optimization methods to find reasonably good quality and computationally tractable solutions. In order to accomplish this, a trade-off made between the levels of abstraction that the model reflects the real system and the requirement to keep the model as simple as possible to solve it efficiently. Third, this approach involves an optimization component (metaheuristic algorithm) that generates the (near) Pareto-optimal set of schedules, which shows the trade-offs between considered objectives, within the practical planning horizon.

Due to the problem's large, complex and unstructured search space, a hybrid Tabu/Scatter Search algorithm is developed to find best trade-off solutions by using an elitist strategy to preserve non-dominated solutions, a dynamic update mechanism to produce high-quality solutions and a rebuilding strategy to promote solution diversity. The proposed algorithm is applied to bi-objective (i.e., maximizing runway utilization and fairness) runway operations schedule optimization as the optimization component of the SbO framework, where the developed simulation model acts as an external function evaluator. Both the proposed algorithm and the discrete-event simulation model are developed by using an object-oriented architecture, which rendered the design and implementation of the models easier and more flexible. Also, a greedy heuristic algorithm is proposed to reinforce the approach by delivering promising initial solutions obtained from solving the deterministic version of the problem.

Finally, computational experiments are conducted to quantitatively evaluate the quality and efficiency of the solutions generated by the proposed hybrid Tabu/Scatter Search algorithm incorporated in the SbO, and perform a proof-of-concept (validation) of the whole SbO framework. In experimental design, design of experiments methods are employed to analyze the impacts of parameters on the simulation as well as the optimization component's performance, and to identify the appropriate parameter levels. In experimental study, first, the proposed hybrid Tabu/Scatter Search algorithm's performance is evaluated based on multi-objective benchmark problems. Then, the applicability of the proposed SbO approach is investigated by using real-life data obtained from a major international US airport. As the experimental results have shown, the proposed algorithm is capable of finding the best known and diverse Pareto set of solutions in a relatively short time, and appropriate consideration of problem-specific knowledge is highly relevant for efficiency. The main quantifiable benefits of using the proposed SbO approach include an increase in the runway utilization, delay savings and improvement in fairness, which in turn, potentially lead to economic and environmental benefits, and increased on-time performance for airlines.

## 7.2 Future Research Directions

This dissertation provides several directions for future work based on the results obtained from the conducted research. These future research directions can be divided into three main areas: (1) considering further practical aspects of runway operations scheduling problem, (2) developing high-fidelity simulation models and robust SbO approaches, and (3) extending the current multi-objective optimization algorithms. The following subsections discuss each of these main areas.

### 7.2.1 Further Considerations on Runway Operations Scheduling Problem

First of all, runway operations scheduling problem is still an active research field with numerous unexplored areas, and several research efforts are underway to enhance the effectiveness of runway operations in many ways. Some of these promising areas are detailed below:

In this dissertation, the runway operations scheduling problem is considered with the assumption that all the problem data, probability distributions, etc. are known in advance (i.e. static or offline version of the problem is considered). However, in practice, usually limited amount of information is available in advance of actual operations. As previously mentioned, the version of the problem in which related information is available only when aircraft is released to the system is referred to as dynamic, online or online-over-time scheduling version. Therefore, it would be interesting to explore some of the dynamic behavior aspects of the problem, specifically rescheduling or recovering the schedules when local disturbances, such as mechanical problems, occur during the runway operations, especially considering the future operating environment envisaged by the NextGen.

The runway operations' efficiency in multiple runways case, especially in closely spaced parallel runways, is highly restricted by the interference of wake turbulences. The current FAA enforced minimum wake turbulence separation requirements are mainly based on aircraft weight. However, recent research to re-categorize these separation requirements indicates that along with aircraft weight, other aircraft characteristics, such as speed, wingspan, also affect the power of the wake turbulence it creates as well as its sensitivity to the wake turbulence created by other aircraft. Furthermore, another area where the researchers have been focusing more is the dynamic separation standards, which allows reduced separations in favorable weather conditions when wake turbulence dissipate and actual hazardous distance becomes shorter. These new separation requirements could be considered as an extension to this dissertation research by utilizing dynamic separation standards instead of static ones.

Also, the runway operations scheduling problem might be integrated with other airside operations, such as taxi routing, gate assignment, which have an explicit impact on the input factors for runway operations. For example, if both landing and take-off aircraft are using the same taxiways, the presence of landing aircraft could affect the taxi times for take-off. In the literature, there exists several analytical models that have been developed based on time-space analysis and queuing models or sequential approaches which try to integrate taxi routing and gate assignment problems into the runway operations scheduling problem, but these models and approaches are inadequate for practical use due to their low computational performance. Utilizing a simulation-based approach has a huge potential to handle this integrated airside operations problem, which considers these interrelated problems simultaneously, in a computationally efficient way.

Another line of future research on runway operations scheduling could be to develop an optimization model that account for Collaborative Decision Making (CDM) aspects by incorporating both fairness and airline collaboration. Although fairness among aircraft is considered in our proposed SbO approach, there still remains opportunities for more complex models which consider different notions of fairness and airline collaboration, and allow for more sophisticated airline input.

As pointed out throughout the dissertation, runway operations scheduling in the presence of uncertainty is still a challenge and, in turn, has received less attention both in academic research and practice. The approach proposed in this research that considers uncertainties can be generalizable to any airport. However, each airport has local rules, regulations, and procedures that need to be considered, such as different strategies on configuring landing and take-off aircraft on the same runway or intersecting runways. Therefore, these airport-specific procedures could be added to better represent the actual runway operational environment for particular airports in future research extensions. Furthermore, weather related data could be included in these research extensions that can capture the effect of strong wind or other inclement weather conditions on scheduling runway operations.

In practice, it is commonly agreed that official FAA documents and airport standard operating procedures are more descriptive than algorithmic. These regulatory documents do not describe the whole business process in detail for scheduling aircraft in TMA. Considering that there exist numerous procedures for this complex scheduling process, another potential research area could be simulating and modeling all these procedures in an algorithmic way in order to evaluate utilization of air traffic controllers and to optimize controller workload distribution for managing the future air traffic flow and runway operations, particularly for complicated situations, such as intersecting runways, runway configurations involving more than two runways simultaneously.

Finally, another direction of future research could be to leverage detailed surface surveillance data from the Airport Surface Detection Equipment, Model-X (ASDE-X) surveillance system to manage uncertainties and CDM aspects in scheduling runway operations. ASDE-X system, which has been recently installed most of the major US airports, continuously track each aircraft on the surface of the airport. Although the main purpose of ASDE-X system is to enhance safety, the historical and live data could be used for identifying the bottlenecks in the runway system, developing metrics to evaluate the operational performance and improving the efficiency of runway operations. Even though the FAA and commercial databases, such as Operations & Performance database that we utilized during the research provide OOOI and airport-level aggregate data, the level of detail in these databases are not enough to identify the interactions between different phases of the airside operations and derive insights regarding the characteristics of these operations.

### 7.2.2 Develop Simulation Models and Simulation-based Approaches

Future research could also explore different simulation models by incorporating human performance models of air traffic controllers. For example, neural network modeling methods could be a potential area of research to model cognitive and decision-making behavior of air traffic controllers for high-fidelity. Since in this case, simulation models will most likely become more complicated and computationally expensive, these simulation models might be replaced with a computationally efficient surrogate models in order to avoid time-consuming calls of the high-fidelity models.

Simulation-based optimization (SbO) is an emerging field that tackles the increasingly complex real-life problems that are currently considered unsolvable with analytical methods as well as that will appear in the near future. The simulation technique is capable of modeling complex interrelations and practical constraints conveniently. However, in a SbO framework, utilizing simulation as an objective function evaluator brings some challenges which include the following: (1) various levels of noise stem from stochastic nature of the simulation, and (2) consumption of relatively large amounts of computation time. In order to mitigate the difficulty that simulation runs often require large amounts of computation time, distributed and parallel computing infrastructures could be employed.

Additionally, algorithm analysis and design for obtaining the Pareto-optimal solution set from a SbO framework in a computationally effective way is a relatively new field that also present a promising area of future research. The primary challenge in a SbO framework entailed by the noise stems from stochastic simulations. Therefore, different noise-handling features that mitigate the detrimental impacts of noise, such as a decrease in convergence rate, could also be investigated.

Most of the research in the literature consider a single objective in a SbO model. However, most real-world problems involve multiple objectives. Simultaneously optimizing more than one conflicting objective is natural in many practical SbO problems, which make the problem harder to tackle. Because no one solution can be considered as "optimum" to multiple objectives, the resulting simulation multi-objective optimization problem resorts to a set of trade-off solutions. It is clear that in these situations the practice of linking an optimization method with a simulation model is not an easy task. Because stochastic nature of the simulation model makes it difficult on top of the ordinary deterministic optimization setting and the computation of the objective function values in each iteration is another challenge that makes the process more daunting and time-consuming. These challenges could be partially overcome by integrating statistical factor screening techniques into the SbO framework. These techniques could be used to eliminate poor quality solutions, and in turn, the search space of the optimization could be reduced. Also, these approaches could be applied to practical problems in other domains such as production, logistics, defense, etc.

### 7.2.3 Extensions on Multi-Objective Optimization

Although most of the research on metaheuristic algorithms for MOO is dedicated to GA-based algorithms, aforementioned capabilities of Scatter Search justify further work in this field. This dissertation illustrates the employment of elitism, rebuilding and dynamic update strategy to improve the capability of the SS algorithm template in MOO. By the same token, innovative mechanisms to incorporate learning ability of neural networks and the knowledge compression ability of fuzzy logic have a potential to generate more effective algorithms in a SbO context as well as other complex problem domains.

Currently, there exist many open research lines on multi-objective evolutionary algorithms (MOEAs), which include developing more efficient algorithms, defining new performance measures, and integrating with simulation models. One of the possible solutions is to utilize common characteristics and subdomains of the search space, which are commonly referred as species and niches, respectively. By encouraging speciation and niching, an SS-based algorithm can facilitate simultaneous convergence to more than one solution in MOO.

Also, in the field of multi-objective evolutionary algorithms, there are still improvements to be made especially in terms of hybridization. One of the promising hybridization areas is the "matheuristic" algorithms that combine metaheuristic algorithms with classical exact (mathematical programming) approaches, such as B&B, in order to improve solution quality and/or computation time. In this context, exact methods may be used to solve sub-problems within a heuristic framework or metaheuristic algorithms may be used to increase the performance of exact methods.

Lastly, integrating multi-objective evolutionary algorithms into simulation-based multi-objective optimization frameworks as the optimizer component seems to be an encouraging direction for both academic research and practice. Hence, further research in this area may explore the extent and ways in which the proposed hybrid Tabu/Scatter Search algorithm can be employed as the optimization engine in other potential real-life scheduling or planning problems.

## 7.3 Conclusions

The ATC system has been becoming gradually more complex and prone to disruptions. Hence, effective and efficient methods are required to increase runway utilization, improve safety, and reduce operating and environmental costs by addressing uncertainties. The conducted literature review has shown that although a great deal of work has been done and various novel modeling approaches have been proposed in runway operations scheduling field, only a couple work exist that explicitly consider stochastic nature of the problem and different stakeholders' interests, and it is still an active research area.

In addition, combinatorial nature of the problem, uncertainties, and multiple objectives render the practical runway operations problem intractable with analytical methods. Due to this complexity and based on the literature review of potential approaches, a simulation-based optimization (SbO) approach employed for finding robust solutions to this real-life problem. The SbO is formulated as a multi-objective optimization in order to consider uncertainties as well as ensure fairness among aircraft.

The experimental results indicated that the proposed hybrid Tabu/Scatter Search algorithm has a potential to exploit knowledge obtained from the search space, utilize strategic designs, and construct search paths from a reference set of solutions to approximate the Pareto-frontier efficiently. Since computational time is the limiting factor in SbO frameworks due to costly multiple simulation evaluations, SS-based hybrid metaheuristic algorithms seem to be promising.

As a conclusion, this dissertation has shown that explicitly considering the uncertainties by utilizing a SbO approach has the potential to increase the effectiveness of runway operations. The evidence obtained from the experiments illustrates that potential operational benefits can be achieved in runway operations scheduling by building realistic models that utilize available operational data and employing these models to design and implement optimization algorithms to enhance the effectiveness of a runway system. Also, the proposed hybrid Tabu/Scatter Search algorithm seems to be a promising research direction due to its efficiency in finding a set of non-dominated solutions in a SbO framework with multiple objectives. Finally, outlined future research directions could further reduce the gap between practice and academic research in runway operations scheduling, and explore open research lines in simulation-based multi-objective optimization field.

# APPENDICES

## APPENDIX A: Additional Terminology

This appendix is to present the additional terms and acronyms used throughout the dissertation.

| | |
|---|---|
| *ASDE-X* | Airport Surface Detection Equipment Model X (ASDE-X) is a system that uses a combination of triangulation of aircraft transponder signals, also termed as multi-lateration (a technique to accurately locate aircraft), aircraft Automatic Dependent Surveillance-Broadcast (ADS-B) broadcasts, and primary radar reflections to present an airport and surrounding airspace display of the position of all aircraft and includes individual data tags indicating flight identification, aircraft tail number, and other associated flight data. It provides a position in the active movement area (not ramp) at 1-second updates. ASDE-X is primarily a safety tool designed to mitigate the risk of runway collisions. |
| *GDP* | Ground Delay Program (GDP) is a traffic management procedure where aircraft are delayed at their departure airport in order to manage demand and capacity at their arrival airport. Flights are assigned departure times, which in turn regulates their arrival time at the impacted airport. |
| *FAF* | Final Approach Fix (FAF) is the point from which the final approach to an airport is executed and which identifies the beginning of the final approach segment. The glide slope/path starts at the FAF. |
| *IAF* | Initial Approach Fix (IAF) is the point where the initial approach segment of an instrument approach begins. An instrument approach procedure may have more than one Initial approach fix and initial approach segment. The initial approach fix is usually a designated intersection. The IAF may be collocated with the intermediate fix of the instrument approach. |
| *RNAV* | Area Navigation (RNAV) is a method of navigation which permits aircraft operation on any desired flight path within the coverage of ground or space-based navigation aids or within the limits of the capability of self-contained aids, or a combination of these. |
| *OOOI* | Out of the gate, Off the ground, On the ground, and Into the Gate (OOOI) is a capability that provides four data points to measure and measure the efficiency of aircraft ground movements. Avionics equipment on many aircraft automatically reports these times to the operator via avionics, which in turn provides this report to the FAA. |
| *TMA* | Terminal Maneuvering Area (TMA), in Europe it is commonly referred as traffic control area, is a general term used to describe airspace in which approach control service or airport traffic control service is provided. |

*Weather Conditions* — Air traffic rules are traditionally applied based on prevailing weather conditions. Visibility and cloud ceiling are the two primary factors in determining the weather category for an airport. Three specific weather scenarios - visual, instrument, and marginal will be considered, which are the fundamental scenarios utilized for airport capacity profiles (Jennifer Gentry et al., 2014). These scenarios are detailed below:

*(a)*     *Visual meteorological conditions (VMC)*: Ceiling and visibility allow for visual approaches and *visual flight rules (VFR)* apply. These rules depend on pilots to visually maintain adequate separation.

*(b)*     *Instrument meteorological conditions (IMC)*: Ceiling less than 1,000 feet or visibility less than 3 statute miles. *Instrument flight rules (IFR)* apply and radar separation between aircraft is required.

*(c)*     *Marginal meteorological conditions (MMC)*: Ceiling and visibility better than instrument conditions but without meeting criteria for visual approaches. It is basically an IFR environment, but with a better visibility so the 2 NM departure/arrival separation is superseded by visual separations.

## APPENDIX B: Various Stakeholders and their Desired Interests

This appendix is to provide some example objectives of various stakeholders in mathematical terms. Before presenting the objective functions that pertain to each stakeholder, the notation used throughout these objective functions are shown below:

$N$     : set of $n$ aircraft, $N = \{1, 2, ..., n\}$

$i, j$     : aircraft indices

$r_j$     : ready time for aircraft $j$

$\delta_j$     : target time for aircraft $j$

$d_j$     : due time for aircraft $j$

$w_j$     : weight value assigned to aircraft $j$ based on its operation type and class

$T_j$     : piecewise tardiness of aircraft $j$ with respect to target time for aircraft $j$

$T_{max}$     : maximum tardiness

$C_j$     : landing/take-off time of aircraft $j$

$C_{max}$     : makespan (the completion time of the last scheduled aircraft)

$I_j$     : air traffic controllers' intervention counts to aircraft $j$

$c_j(t)$     : landing/take-off cost of aircraft $j$ at time $t$

$\alpha_j$     : penalty cost for aircraft $j$ when it lands or takes-off after its target time

$\beta_j$     : penalty cost for aircraft $j$ when it lands or takes-off before its target time

$k_j$     : penalty for aircraft $j$ if its place in land/take-off sequence changes due to delays

$v_j$     : fuel burn cost of aircraft $j$ depends on its operation type and class

$h_j$     : arrival time of aircraft $j$ to the holding area for take-off

$p_j$     : passenger capacity of aircraft $j$

$n_j(t)$     : $CO_2$ emission per unit time associated with tardiness of aircraft $j$

$t_j$     : start landing/take-off time of aircraft $j$

$x_j$     : 1, if aircraft $j$'s place in land/take-off sequence changes due to delays, 0, otherwise.

**ANSPs or air traffic controllers**:

- minimizing the arrival/departure delay (total weighted tardiness)

$$min. \quad \sum_{j \in N} w_j T_j$$

- minimizing the sum of the costs of deviation from the target times

$$min. \quad \sum_{j \in N} (\alpha_j T_j + \beta_j E_j)$$

- minimizing total prioritized land/take-off time (total weighted completion time)

$$min. \quad \sum_{j \in N} w_j C_j$$

- maximizing the runway throughput

$$min. \quad \sum_{j \in N} C_{max}$$

- minimizing the maximum arrival/departure delay

$$min. \quad \sum_{j \in N} T_{max}$$

- minimizing air traffic controllers' workload (air traffic controllers' intervention) (A heuristic model can be built to simulate air traffic controllers' intervention behavior. In each simulation run, this heuristic model can be called to check if there is any air traffic controllers' intervention and count ($I$) will be accumulated.)

$$min. \quad \sum_{j \in N} I_j$$

- maximizing fairness among the aircraft operators (The Constrained Position Shifting (CPS) concept, first proposed by (Roger George Dear, 1976), helps maintaining fairness among the aircraft operators by preventing a specific flight from waiting longer relative to FCFS order.)

$$min. \quad \sum_{j \in N} c_j(t)$$

- minimizing the aircraft waiting time in holding area for take-off

$$min. \quad \sum_{j \in N} t_j - h_j$$

**Airlines**:

- minimizing operating costs (minimizing the total fuel cost resulting from the deviation of aircraft start times from their respective ready times)

$$min. \sum_{j \in N} v_j (t_j - r_j)$$

- minimizing total passenger delays (passenger capacity of each aircraft is considered as a weight factor)

$$min. \sum_{j \in N} p_j T_j$$

**Airport management**:

- minimizing the aircraft sequence changes due to delays

$$min. \sum_{j \in N} k_j x_j$$

**Government agencies**:

- minimizing environmental effects (air pollution) (minimizing the $CO_2$ emitted to achieve an aircraft schedule.)

$$min. \sum_{j \in N} n_j(t) T_j$$

**APPENDIX C: High-level Block Diagram of the Simulation Model**

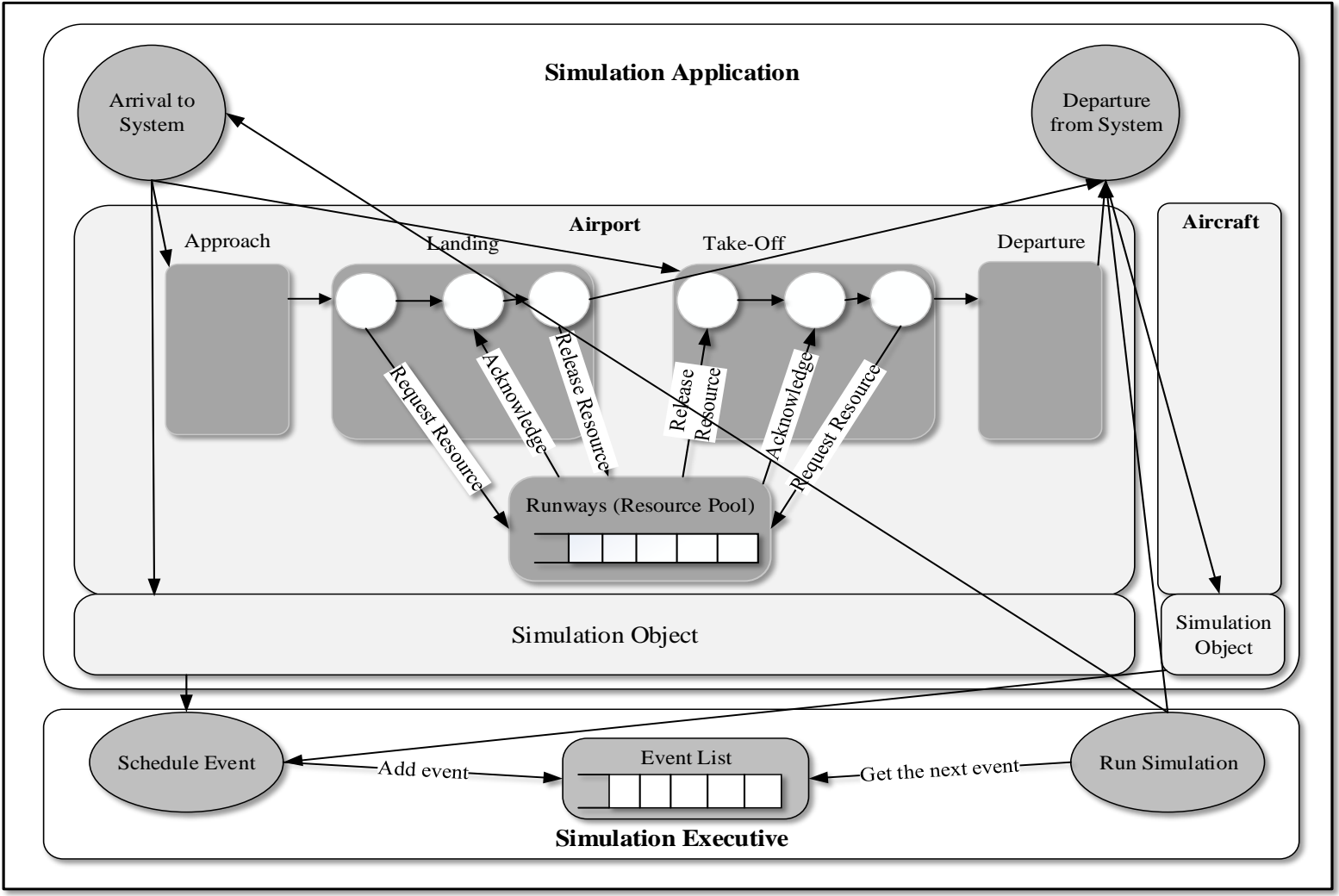

# APPENDIX D: Arrival and Departure Flow Chart Diagrams

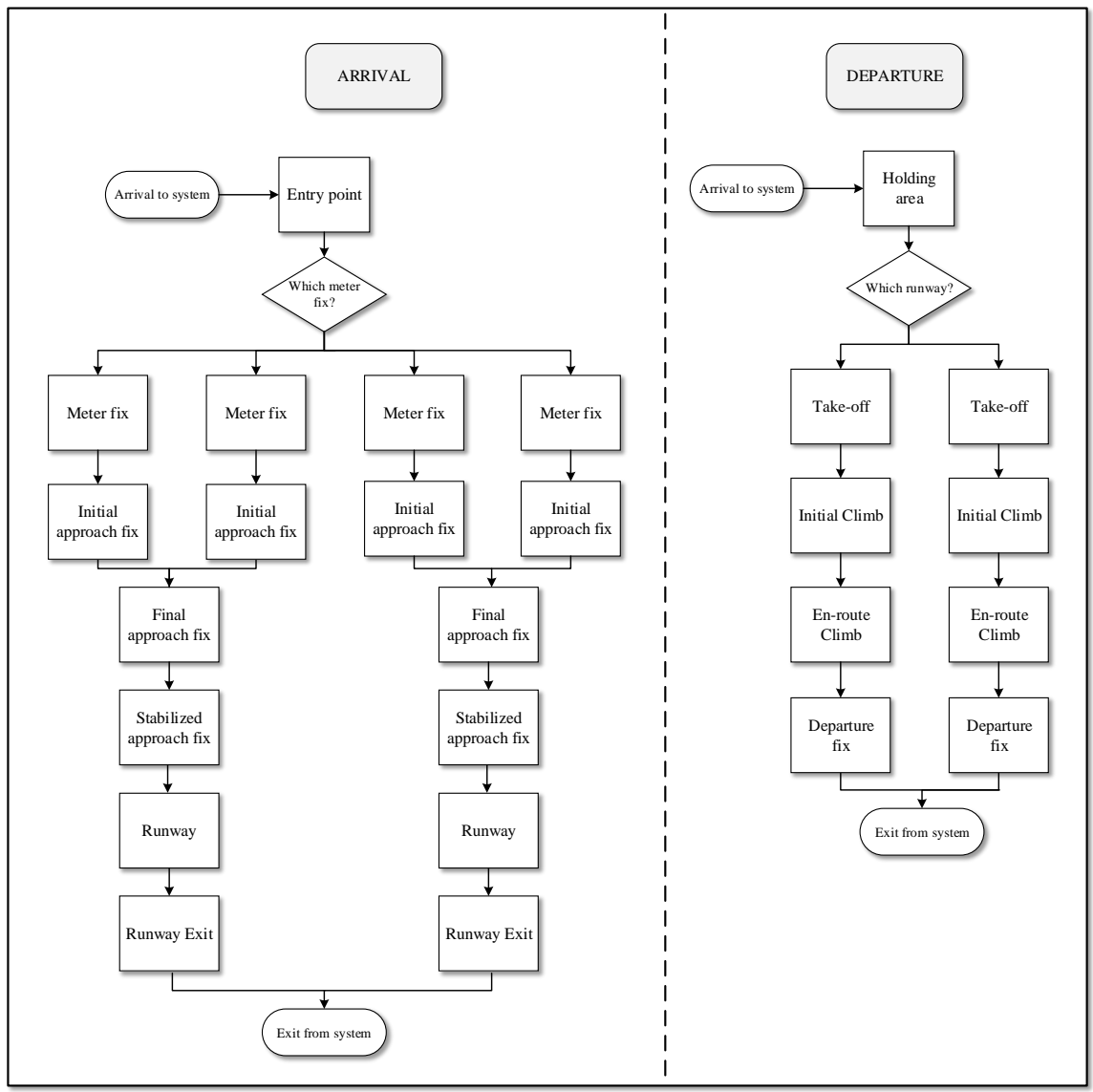

**APPENDIX E: Results for Multi-Objective Optimization Experiments**

    This appendix is to provide the results for multi-objective optimization experiments, which include Pareto-frontiers after 20, 40, 60, 80, 100 and 120 Iterations for ZDT3 and F&F, respectively.

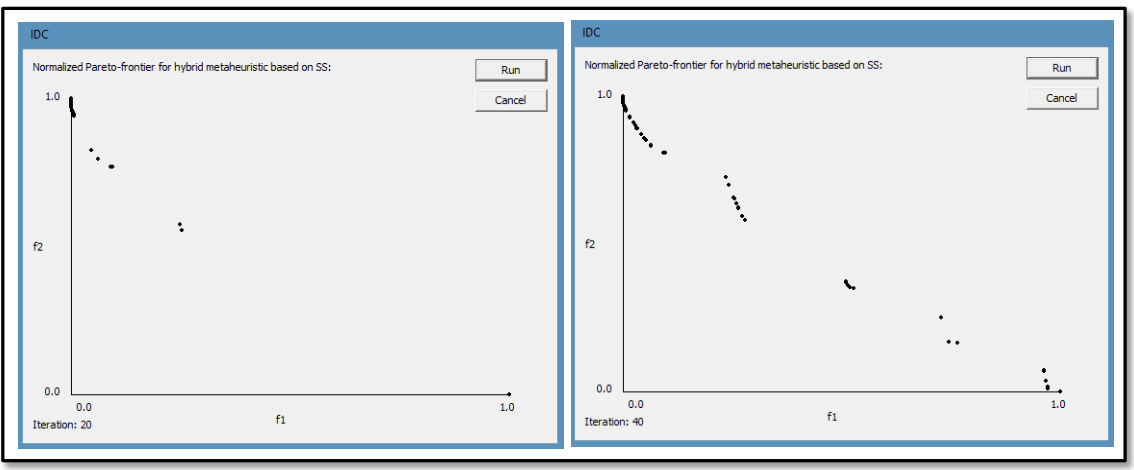

Pareto-frontier after 20 & 40 Iterations for ZDT3

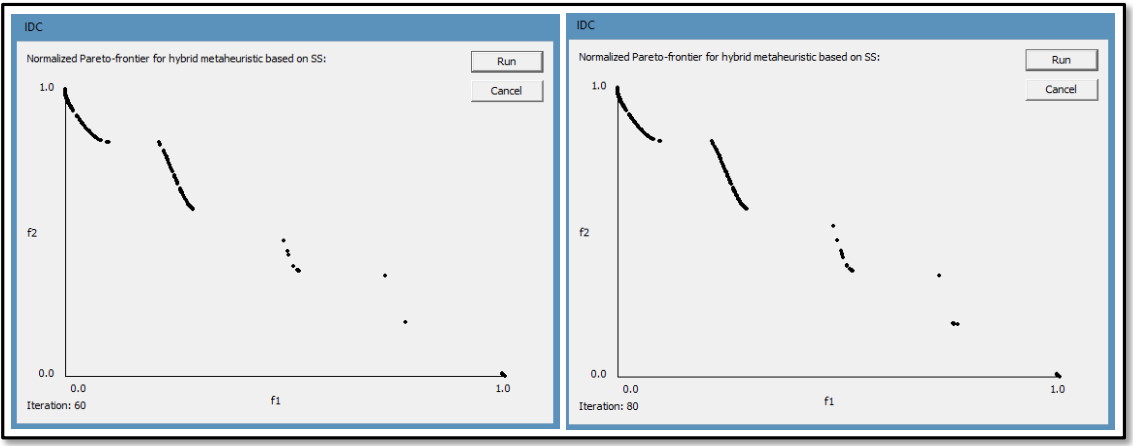

Pareto-frontier after 60 & 80 Iterations for ZDT3

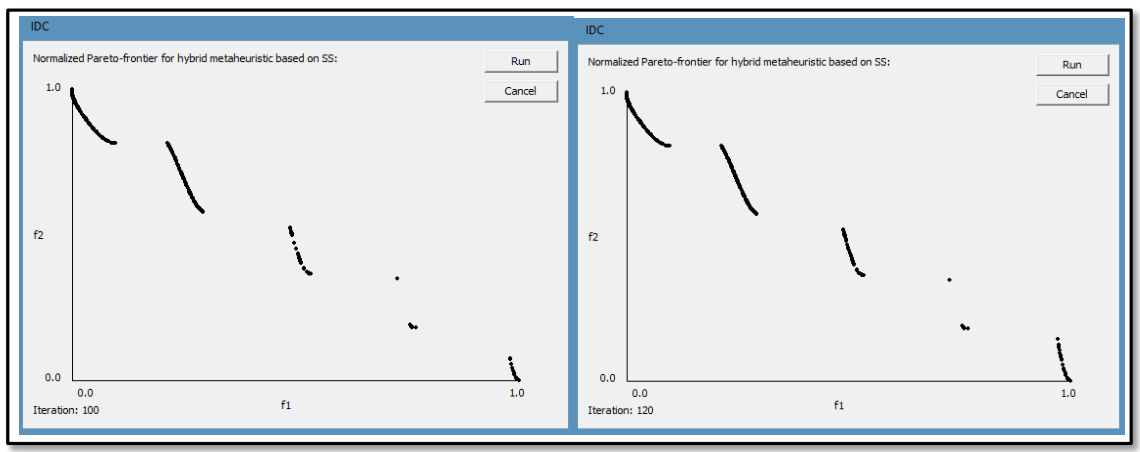

Pareto-frontier after 100 & 120 Iterations for ZDT3

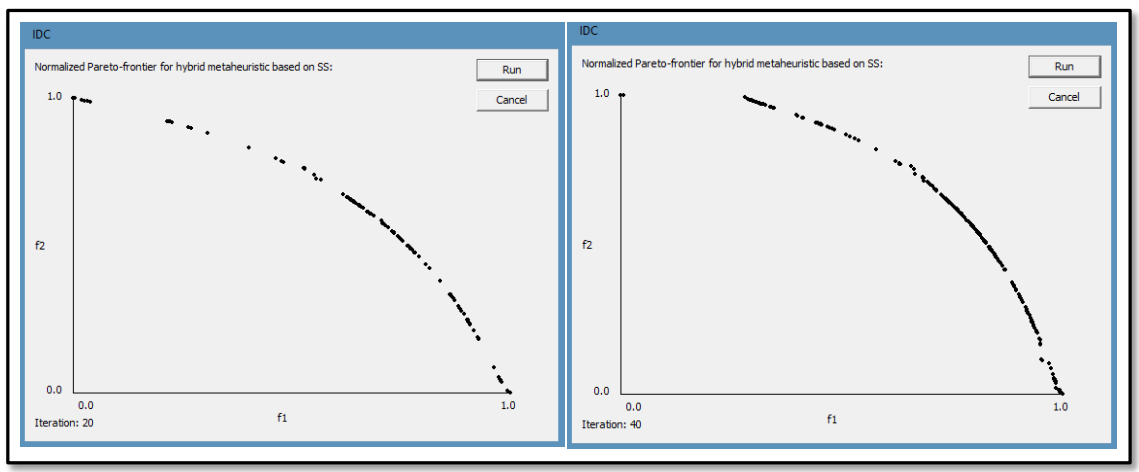

Pareto-frontier after 20 & 40 Iterations for F&F

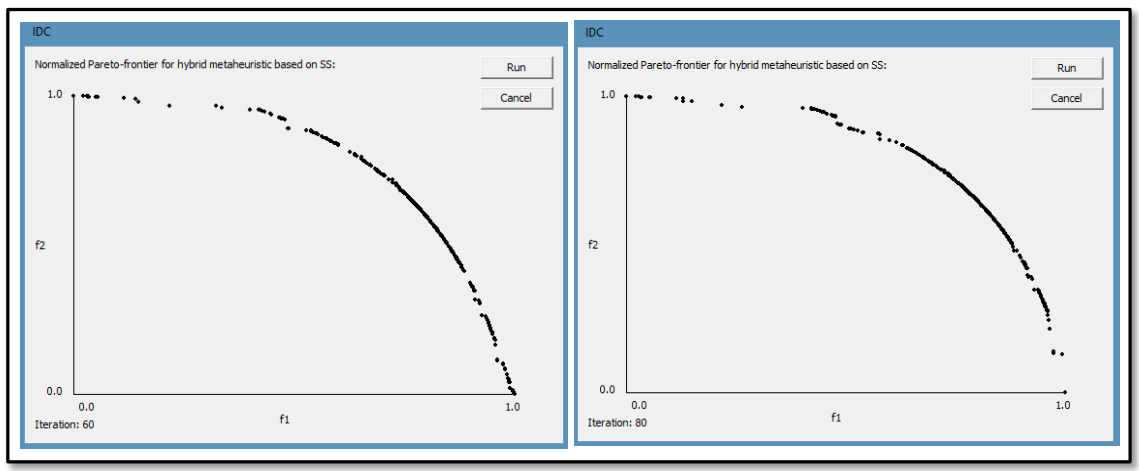

Pareto-frontier after 60 & 80 Iterations for F&F

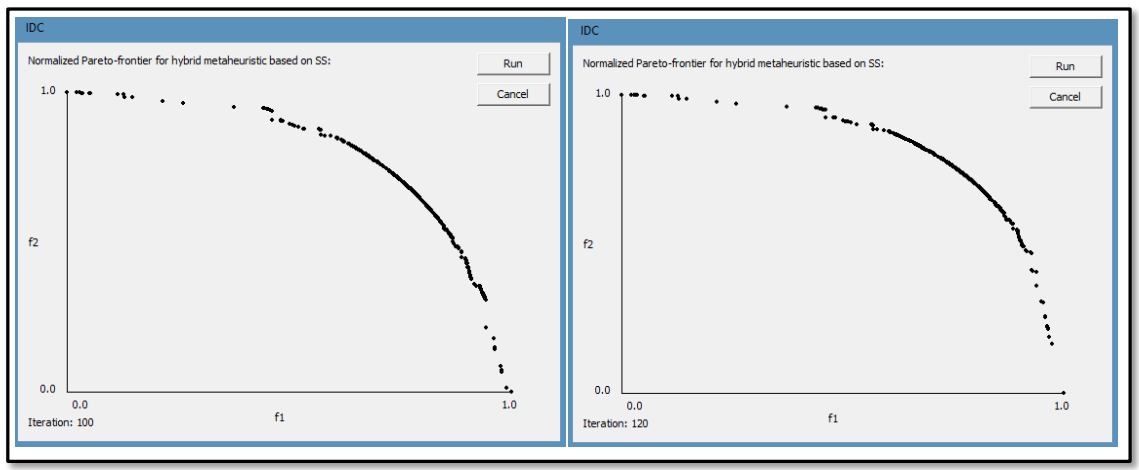

Pareto-frontier after 100 & 120 Iterations for F&F

**APPENDIX F: Results for Simulation-based Optimization Experiments**

This appendix is to provide the results for simulation-based optimization experiments, which include Pareto-frontiers after 20, 40, 60, 80, 100 and 120 iterations for Scenario 1, Scenario 2 and Scenario 3, respectively. Also, Pareto-frontiers after 300 and 500 iterations for Scenario 3 are included.

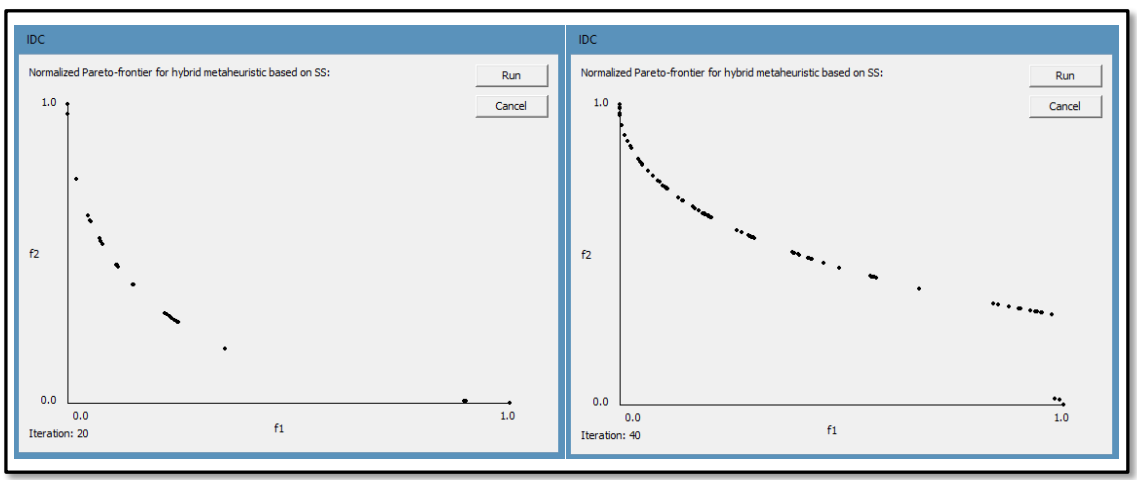

Pareto-frontier after 20 & 40 Iterations for Scenario 1
(*f1*: runway utilization, *f2*: fairness)

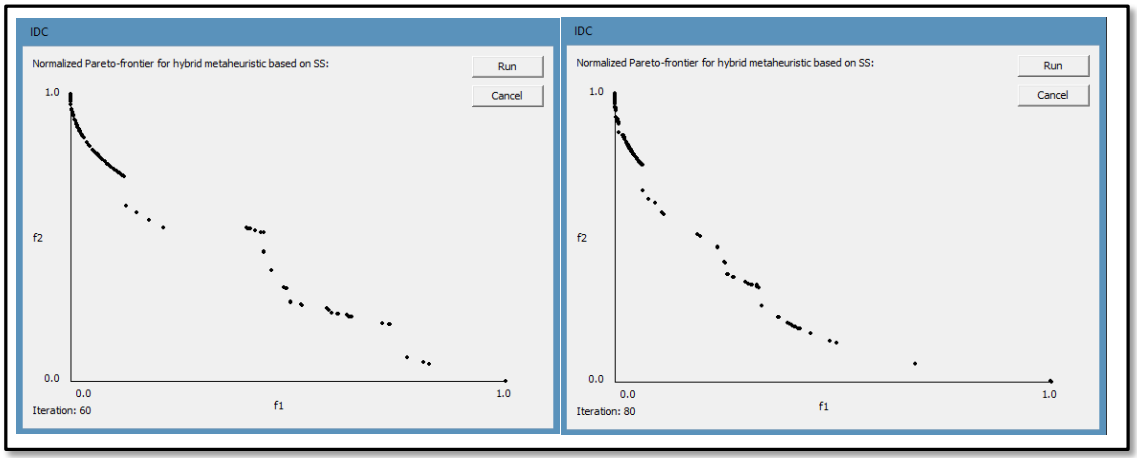

Pareto-frontier after 60 & 80 Iterations for Scenario 1
(*f1*: runway utilization, *f2*: fairness)

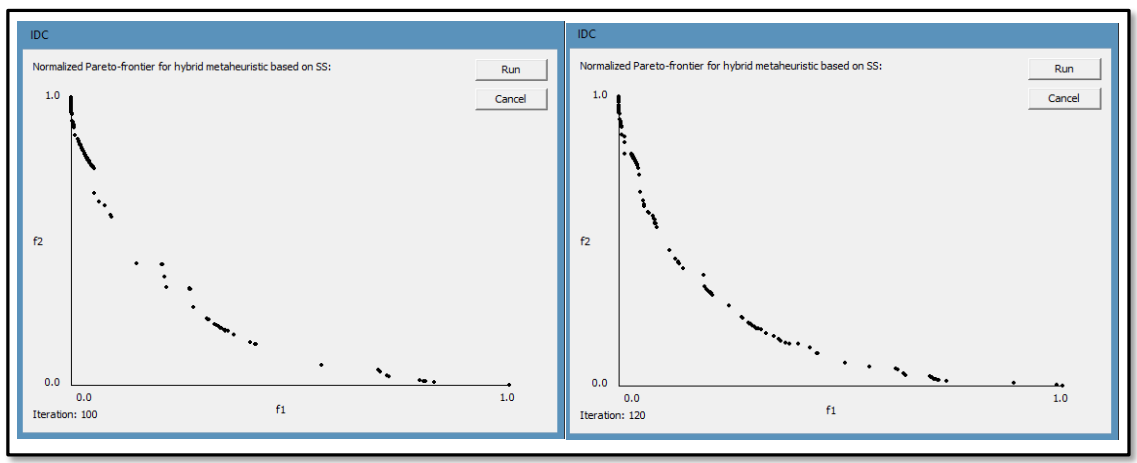

Pareto-frontier after 100 & 120 Iterations for Scenario 1
(*f1*: runway utilization, *f2*: fairness)

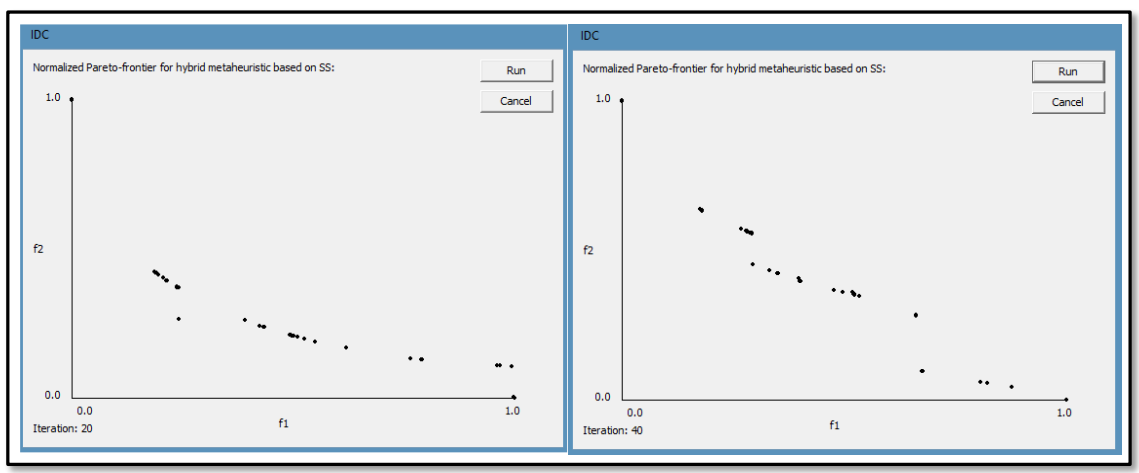

Pareto-frontier after 20 & 40 Iterations for Scenario 2
(*f1*: runway utilization, *f2*: fairness)

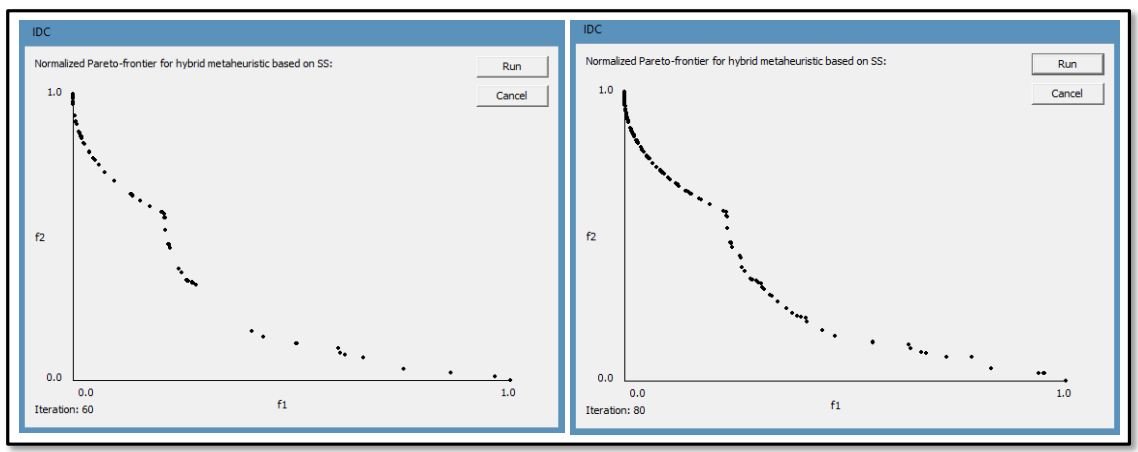

Pareto-frontier after 60 & 80 Iterations for Scenario 2
(*f1*: runway utilization, *f2*: fairness)

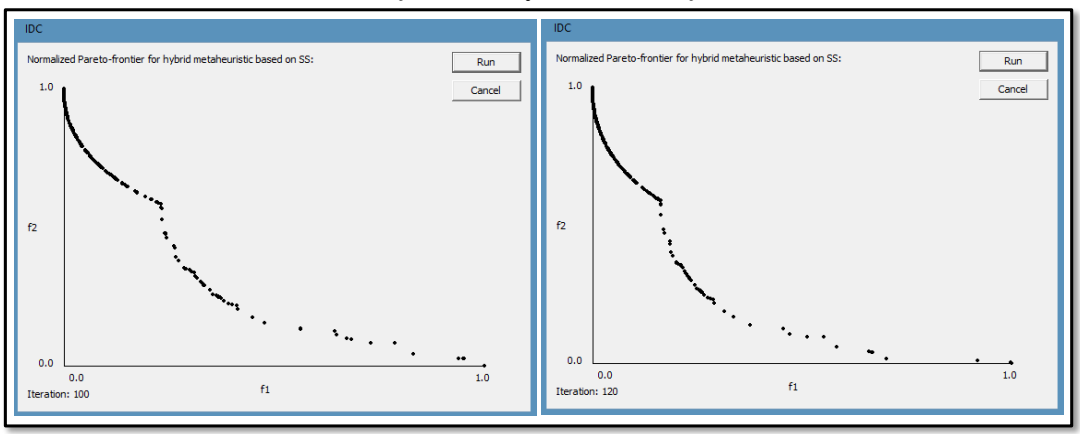

Pareto-frontier after 100 & 120 Iterations for Scenario 2
(*f1*: runway utilization, *f2*: fairness)

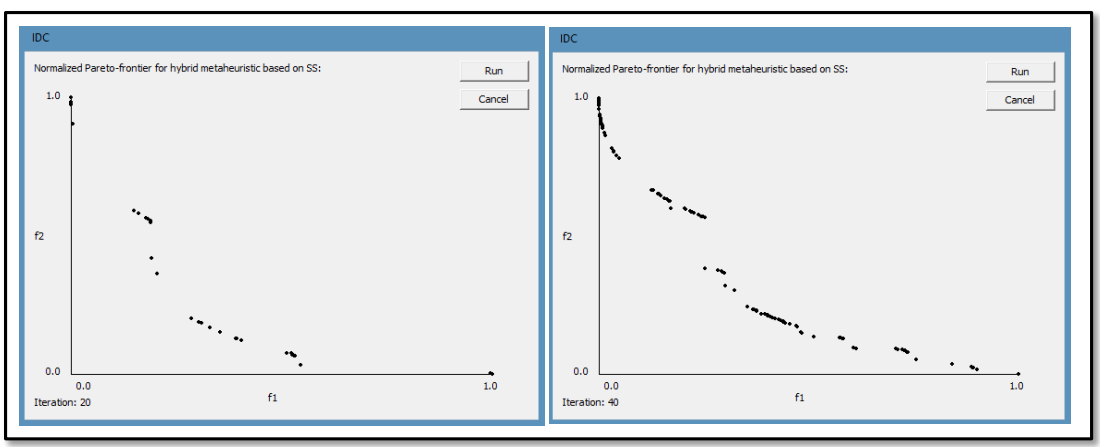

Pareto-frontier after 20 & 40 Iterations for Scenario 3
(*f1*: runway utilization, *f2*: fairness)

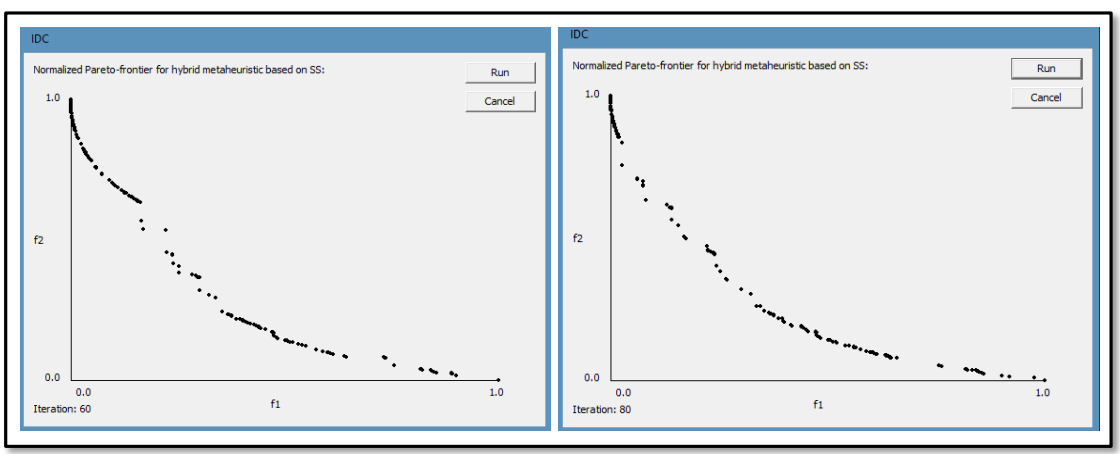

Pareto-frontier after 60 & 80 Iterations for Scenario 3

(*f1*: runway utilization, *f2*: fairness)

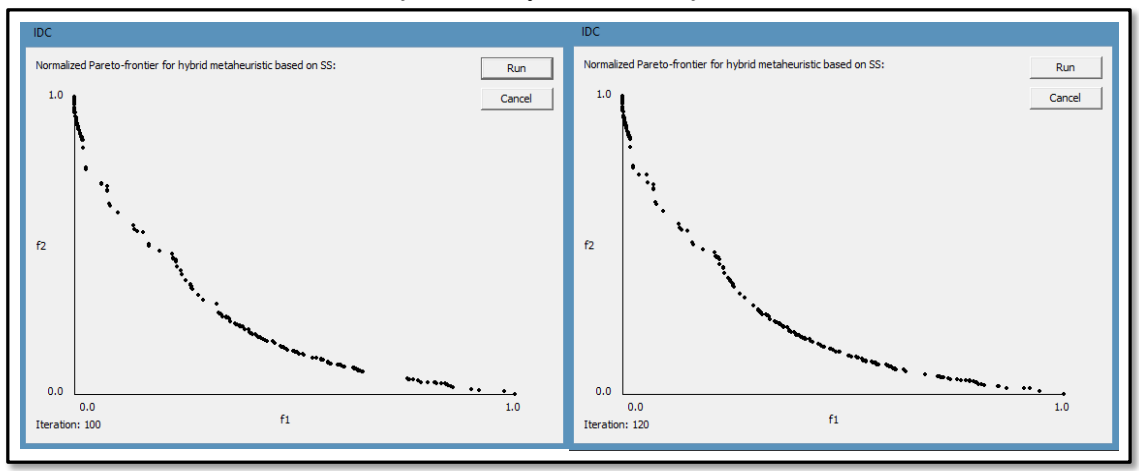

Pareto-frontier after 100 & 120 Iterations for Scenario 3
(*f1*: runway utilization, *f2*: fairness)

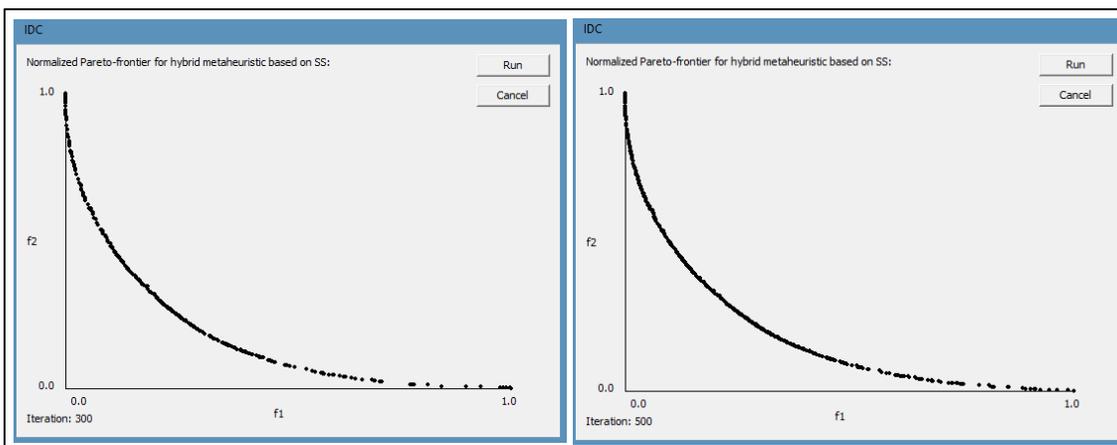

Pareto-frontier after 300 & 500 Iterations for Scenario 3
(*f1*: runway utilization, *f2*: fairness)

# VITA

Bulent Soykan

Department of Engineering Management and Systems Engineering
Old Dominion University, Norfolk, VA
e-mail: bsoyk001@odu.edu , soykanb@gmail.com

EDUCATION:

| | |
|---|---|
| Old Dominion University, Norfolk, VA | 2016 |
| Ph.D., Engineering Management | |
| Turkish Military Academy, Defense Sciences Institute, Ankara, Turkey | 2007 |
| M.S., Technology Management (Project Management) | |
| Turkish Military Academy, Ankara, Turkey | 1997 |
| B.S., Systems Engineering | |

RESEARCH AREAS:

Decision-Making under Uncertainty, Mathematical Programming, Multi-objective Optimization, Simulation Analysis & Modelling, Metaheuristic Algorithms, Evolutionary & Genetic Algorithms, Object-Oriented Design & Implementation, Simulation-based Optimization, Engineering Management, Project & Program Management, Command & Control

PUBLICATIONS RESULTING FROM THIS DISSERTATION RESEARCH:

Soykan, B. & Rabadi, G. (2016). A Tabu Search Algorithm for the Multiple Runway Aircraft Scheduling Problem. In *Heuristics, Metaheuristics and Approximate Methods in Planning and Scheduling* (pp. 165-186). Springer International Publishing.

Soykan, B. & Rabadi, G. (2016). A Hybrid Metaheuristic Algorithm for Multi-Objective Runway Scheduling Problem in Simulation-based Optimization. In *MODSIM World 2016 Conference.*

Soykan, B. & Rabadi, G. (2016). Multi-Objective Simulation-based Optimization of Runway Operations Scheduling Using a Hybrid Metaheuristic Algorithm. Technical Report for *Airport Cooperative Research Program, University Design Competition for Addressing Airport Needs (2015-2016 Academic Year).*

The word processor for this dissertation was the author.